\documentclass[lot, lof]{puthesis}
\usepackage{graphicx}
\usepackage{subfigure}
\usepackage{amsmath}
\usepackage{pslatex}
\usepackage{nomencl}
\usepackage{losymbol}
\usepackage{arabtex}
\usepackage{utf8}
\makeindex
\newcommand{\printmode}{}

\title{Novel Quadrotor Manipulation System}
\submitted{\textnormal{September 2013}}  % degree conferral date (January, April, June, September, or November)
\copyrightyear{2013}  % year in which the copyright is secured by publication of the dissertation.
\school{Innovative Design Engineering}
\department{Mechatronics and Robotics Engineering}
\degree{\textnormal{Master of Science}}
\author{Ahmed Mohammed Elsayed Khalifa}
\department{Mechatronics and Robotics Engineering}

\usepackage{amsfonts}
\usepackage{graphicx}
\usepackage{verbatim}
\usepackage{multirow}
\usepackage{longtable}
\usepackage{amsthm}
\usepackage{setspace}
\ifdefined\printmode

\usepackage{url}

\else

\ifdefined\proquestmode
\usepackage{hyperref}
\hypersetup{bookmarksnumbered}

\makeatletter
\hypersetup{pdftitle=\@title,pdfauthor=\@author}
\makeatother

\else
\usepackage{hyperref}
\hypersetup{colorlinks,bookmarksnumbered}

\makeatletter
\hypersetup{pdftitle=\@title,pdfauthor=\@author}
\makeatother

\fi % proquest or online formatting
\fi % printed or online formatting

\ifodd 2
\renewcommand{\maketitlepage}{}
\renewcommand*{\makeabstract}{}

\else

\abstract{
This thesis introduces a novel quadrotor manipulation system that consists of 2-link manipulator attached to the bottom of a quadrotor. This new system presents a solution for the drawbacks found in the current quadrotor manipulation system which uses a gripper fixed to a quadrotor. Unlike the current system, the proposed system enables the end-effector to achieve any arbitrary orientation and thus increases its degrees of freedom from 4 to 6. Also, it provides enough distance between the quadrotor and the object to be manipulated. System kinematics and dynamics are derived. Picking and placing a payload using the end-effector of the manipulator are also modeled.

To study the feasibility of the proposed system, a quadrotor with high enough payload to add the 2-link manipulator is constructed. Its parameters are identified to be used in the simulation and controller design of the proposed system. A CAD model is developed to calculate the mass and moments of inertia in an accurate way. Direct relationships between Pulse Width Modulation (PWM) and each of the angular speeds, thrust forces, and drag moments of the rotors are identified. A Direction Cosine Matrix (DCM) complementary filter is used to estimate the attitude of the quadrotor based on the IMU measurements. Attitude stabilization controller is designed based on feedback linearization technique to test the identified parameters and the attitude estimation. The results of the experiments show satisfactory accuracy of the identified structure parameters, the identified rotor assembly parameters and the attitude estimation algorithm.

The controller of the proposed quadrotor manipulation system is designed based on three control techniques: feedback linearization based PID control, direct fuzzy logic control, and fuzzy model reference learning control. These controllers are tested to provide system stability and trajectory tracking under the effect of picking and placing a payload and the effect of changing the operating region. Simulation results show that the fuzzy model reference learning control technique has superior performance. The results indicate the feasibility of the proposed system.

}

\acknowledgements{
First of all, I would like to thank Prof. Ahmed Abo-Ismail, the founder of the Mechatronics and Robotics Engineering Department for his great efforts in establishing all facilities in the department.

I would like to thank my supervisors, Prof. Ahmed Abo-Ismail, Dr. Mohamed Fanni and Dr. Ahmed Ramadan, for their advice, support, availability to speak with me, and their help and fruitful discussions throughout the period of doing this research.

A special thanks to Dr. Mohamed Fanni who provides me with this novel idea that is the topic of this thesis. I learned a lot from his experience in life, science, and research. He used to exert his best efforts in helping me to solve the problems that I faced during my research.

I must not forget to thank my colleague Eng. Mahmoud Elsamanty, Ph.D. candidate at E-JUST, who shared with me, as a team work, the problem of parameter identification, attitude estimation and control of the qaudrotor, in addition to, for offering his aid and knowledge on 3D modelling using SOLIDWORKS.

The author is supported by a scholarship from the Mission Department, Ministry of Higher Education of the Government of Egypt which is gratefully acknowledged.

Last but not least, I want to give special thanks to my family, my father, my mother and my wife, for all their love, support, and for encouraging me to work my way through college, a path that made me gain precious life experiences, which I would have missed otherwise.
\\
\\
\\
\hfil Ahmed M. Khalifa
\\
\hfill September 2013
}

\fi  % disable frontmatter

\renewcommand{\contentsname}{\uppercase{Table of Contents}}

\usepackage{titlesec}
\titleformat{\chapter}[display]
  {\normalfont\huge\bfseries\centering}
  {\chaptertitlename\ \thechapter}{20pt}{\Huge}
\begin{document}
\makefrontmatter
\addsymbol{$(.)_d$}{Desired value of a variable}
\addsymbol{$\{B\}, O_{B}-x y z$}{Body-fixed frame}
\addsymbol{$\{E\},O_{I}-X Y Z$}{Reference frame that it is supposed to be earth-fixed and inertial}
\addsymbol{$\dot{\eta}_{1}$}{time derivative of body position expressed in the earth-fixed frame}
\addsymbol{$\dot{\eta}_{2}$}{Angular velocities expressed in the inertial frame}
\addsymbol{$\dot{\omega}_{i}^{i}$}{Angular acceleration of frame $i$}
\addsymbol{$\dot{v}_{i}^{c_i}$}{Linear acceleration of the center of mass of link $i$}
\addsymbol{$\eta_{1}$}{Vector of the body position coordinates in the earth-fixed reference frame}
\addsymbol{$\eta_{2}$}{Vector of body Euler-angle coordinates in an earth-fixed reference frame}
\addsymbol{$\eta_{ee_1}$}{Vector of the end effector position in the earth-fixed reference frame}
\addsymbol{$\eta_{ee_2}$}{Vector of the end effector orientation in the earth-fixed reference frame}
\addsymbol{$\mu_i$}{Membership value}
\addsymbol{$\nu_{1}$}{Linear velocity of the origin of the body-fixed frame with respect to the origin of the earth-fixed frame expressed in the body-fixed frame}
\addsymbol{$\nu_{2}$}{Angular velocity of the body-fixed frame with respect to the earth-fixed frame expressed in the body-fixed frame}
\addsymbol{$\omega_{i}^{i}$}{Angular velocity of frame $i$ expressed in frame $i$}
\addsymbol{$\Omega_j$}{Propeller/Rotor($j$)'s angular speed}
\addsymbol{$\rho$}{Density of air}
\addsymbol{$\tau_{a_1}$}{Rolling moment}
\addsymbol{$\tau_{a_2}$}{Pitching moment}
\addsymbol{$\tau_{a_3}$}{Yawing moment}
\addsymbol{$\tau_{c_i}$}{Time constant of the reference model}
\addsymbol{$\theta{1} \; and \; \theta{2}$}{Manipulator joints' angles}
\addsymbol{${\phi}$}{Roll angle; Rotation around x-axis}
\addsymbol{${\psi}$}{Yaw angle; Rotation around z-axis}
\addsymbol{${\theta}$}{Pitch angle; Rotation around y-axis}
\addsymbol{$A^{0}_{1}$}{Transformation matrix from manipulator frame \{0\} to manipulator frame \{1\}}
\addsymbol{$A^{1}_{2}$}{Transformation matrix from manipulator frame \{1\} to manipulator frame \{2\}}
\addsymbol{$A^{B}_{0}$}{Transformation matrix from body frame(\{B\}) to manipulator frame \{0\}}
\addsymbol{$a_j\; and \; b_j$}{Parameters of the linear relationship between PWM and the rotor($j$)'s angular velocity}
\addsymbol{$b_1\; and \; b_2$}{Friction coefficients of manipulator's joints}
\addsymbol{$b_m$}{Center  of the membership  function associated with $U^m$}
\addsymbol{$c$}{Change/Rate of error signal ($e$)}
\addsymbol{$C(\alpha)$}{Cosine of angle $\alpha$}
\addsymbol{$c_j \; and \; d_j$}{Parameters of the linear relationship between PWM and the rotor($j$)'s thrust force}
\addsymbol{$c_P$}{Power coefficient}
\addsymbol{$c_T$}{Thrust coefficient}
\addsymbol{$CAD$}{Computer Aided Design}
\addsymbol{$CG_{i}$}{Point of center of gravity of link ($i$)}
\addsymbol{$CM$}{Center of Mass}
\addsymbol{$d$}{Distance between the quadrotor center of mass and the rotation axis of the propeller}
\addsymbol{$d_{i} \; ,a_{i},\; \alpha_{i}, \; and \; \theta_{i}$}{DH Parameters of the manipulator}
\addsymbol{$DCM$}{Direction cosine matrix}
\addsymbol{$DFLC$}{Direct Fuzzy Logic Control}
\addsymbol{$DOF$}{Degrees Of Freedom}
\addsymbol{$e/\tilde{.}$}{Error signal between desired input and actual output of plant}
\addsymbol{$E^i$ \; and\; $CE^j$}{$i^{th}$ ($j^{th}$) linguistic value associated with $e (c)$, respectively}
\addsymbol{$e_j \; and \; h_j$}{Parameters of the linear relationship between PWM and the rotor($j$)'s drag moment}
\addsymbol{$ESC$}{Electronic Speed Controller}
\addsymbol{$f_{(i,i-1)}^{i} / n_{(i,i-1)}^{i}$}{Resulting force/moment exerted on link $i$ by link $(i -1)$ at point $O_{(i-1)}$}
\addsymbol{$F_{i}^{i}/N_{i}^{i}$}{Inertial forces/moments acting on link $i$}
\addsymbol{$F_{m,q_x}^I,\; F_{m,q_y}^I,\; and\; F_{m,q_z}^I$}{Interaction forces from the manipulator to the quadrotor in $X$,$Y$, and $Z$ directions defined in the inertial frame}
\addsymbol{$F_{m,q}^{B} \; and \; M_{m,q}^{B}$}{The interaction forces and moments of the manipulator acting on the quadrotor expressed in body frame}
\addsymbol{$F_j$}{Propeller/Rotor($j$)'s thrust force, $j=1-4$}
\addsymbol{$FLC_\phi,\; FLC_\theta,\; and  \; FLC_\psi$}{Fuzzy logic controllers designed to control the quadrotor's roll($\phi$), pitch($\theta$) and yaw($\psi$) angles.}
\addsymbol{$FLC_{\theta_1} \; and \; FLC_{\theta_2}$}{Fuzzy logic controllers designed to control the two joints' angles of the manipulator.}
\addsymbol{$FLC_x,\; FLC_y, \; and \; FLC_z$}{Fuzzy logic controllers designed  to control the quadrotor's linear position}
\addsymbol{$FMRLC$}{Fuzzy Model Reference Learning Controller}
\addsymbol{$g$}{Value of acceleration due to gravity ($9.81m/s^2$)}
\addsymbol{$g^{I}$}{Vector of gravity expressed in inertial frame $I$}
\addsymbol{$g_{y_e},\; g_{y_c}\; and\; g_p$}{scaling gains for normalization of its universe of discourses of $y_e$, $y_c$ and $p$ respectively}
\addsymbol{$g_e,\; g_c,\; and\; g_u$}{Scaling controller gains for the error, $e$, change in error, $c$, and controller output, $u$ of the FMRLC}
\addsymbol{$GC$}{Geometrical Center}
\addsymbol{$GPS$}{Global Positioning System}
\addsymbol{$I_{i}^{i}$}{Inertia matrix of link $i$ about its center of mass coordinate frame}
\addsymbol{$I_f$}{Inertia matrix of the vehicle around its body-frame}
\addsymbol{$I_r$}{Rotor inertia}
\addsymbol{$ICSP$}{In-Circuit Serial Programming}
\addsymbol{$IED$}{Improvised Explosive Device}
\addsymbol{$IMU$}{Inertial Measurement Unit}
\addsymbol{$J_{v}$}{Jacobian matrix that relates the body-fixed and earth fixed angular velocities}
\addsymbol{$K_{e_i},\; K_{c_i},\; and\; K_{u_i}$}{Input and Output scaling factors for the error, change of error, and fuzzy output for DFLC,$i$ = $x$, $y$, $z$, $\phi$, $\theta$, $\psi$, $\theta_1$, $\theta_2$}
\addsymbol{$K_F$}{Constant relates a propeller thrust with the angular speed}
\addsymbol{$K_M$}{Constant relates propeller moment with the angular speed}
\addsymbol{$K_p,\; K_d \; and \; K_i$}{PID parameters}
\addsymbol{$L_i$}{Length of manipulator's link($i$)}
\addsymbol{$m$}{Mass of quadrotor}
\addsymbol{$M_{m,q_\phi}^B,\; M_{m,q_\theta}^B,\; and \; M_{m,q_\psi}^B$ }{Interaction moments from the manipulator to the quadrotor around $X$, $Y$, and $Z$  directions defined in the inertial frame.}
\addsymbol{$m_{p}$}{Mass of payload}
\addsymbol{$m_i$}{Mass of manipulator's link($i$)}
\addsymbol{$M_j$}{Propeller/Rotor($j$)'s drag moment}
\addsymbol{$m_p$}{Mass of payload placed on end effector}
\addsymbol{$MCU$}{MicroContoller Unit}
\addsymbol{$MEMS$}{MicroElectroMechanical System}
\addsymbol{$mtr_i$}{Manipulator's motor ($i$)}
\addsymbol{$NMEA$}{National Marine Electronics Association}
\addsymbol{$p$}{Output of fuzzy inverse model}
\addsymbol{$P^m$}{Consequent linguistic value associated with $p$}
\addsymbol{$P_{B0}^{B}$}{Position vector of the origin $O_{0}$ relative to frame $B$}
\addsymbol{$P_j$}{Propeller($j$)'s consumed power}
\addsymbol{$PWM$}{Pulse Width Modulation}
\addsymbol{$r$}{Propeller radius}
\addsymbol{$r(s)$}{Desired value of the plant}
\addsymbol{$R^{B}_{I}$}{Rotation matrix expressing the transformation from the inertial frame to the body-fixed frame}
\addsymbol{$r_{c_i}^{i}$}{Vector from the origin of frame $(i – 1)$ to the center of mass of link $i$}
\addsymbol{$R_{i}^{(i-1)}$}{Rotation matrix from frame $i$ to frame $(i -1)$}
\addsymbol{$r_{i}^{i}$}{Vector from the origin of frame $(i – 1)$ to the origin of link $i$}
\addsymbol{$S(\alpha)$}{Sine of angle $\alpha$}
\addsymbol{$skew (.)$}{Skew symmetric matrix}
\addsymbol{$SSC$}{ Serial servo controller}
\addsymbol{$T$}{Total thrust applied to the quadrotor from all four rotors}
\addsymbol{$T^{I}_{2}$ }{The total transformation matrix that relates the end effector frame to the inertial frame}
\addsymbol{$T_{ee}$ }{General form for a transformation matrix}
\addsymbol{$T_{m_1}\; and \; T_{m_2}$}{Torques acting on manipulator's joints }
\addsymbol{$T_{mtr_i}$}{Torque Loaded on manipulator's motor($i$); $i=0-2$}
\addsymbol{$T_a$}{Time interval for the auto-tuning mechanism}
\addsymbol{$T_p$}{Period of time between each passage of the propeller blades}
\addsymbol{$U^m$}{Consequent linguistic value associated with $u$}
\addsymbol{$u_{\phi}$}{PID control signal for quadrotor's roll stabilization}
\addsymbol{$u_{\psi}$}{PID control signal for quadrotor's yaw stabilization}
\addsymbol{$u_{\theta_1}$}{PID control signal for manipulator's joint1 stabilization}
\addsymbol{$u_{\theta_2}$}{PID control signal for manipulator's joint2 stabilization}
\addsymbol{$u_{\theta}$}{PID control signal for quadrotor's pitch stabilization}
\addsymbol{$u_j$}{PWM signal applied to rotor $j$}
\addsymbol{$u_z$}{PID control signal for quadrotor's altitude stabilization}
\addsymbol{$UART$}{Universal Asynchronous Receiver/Transmitter}
\addsymbol{$UAV$}{Unmanned Aerial Vehicle}
\addsymbol{$v_{i}^{i}$}{Linear velocity of the origin of frame $i$}
\addsymbol{$VTOL$}{Vertical Take-Off and Landing}
\addsymbol{$y_{c}$}{Change/Rate of error signal ($y_{e}$)}
\addsymbol{$y_{e}$}{Error signal between actual output of plant and reference system}
\addsymbol{$y_{m}(s)$}{Output response of the reference system}
\addsymbol{$Y_e^k \; and\; Y_c^s$}{$k^{th}$ ($s^{th}$) linguistic value associated with $y_e (y_c)$, respectively}
\addsymbol{$z_{(i-1)}^{(i-1)}$}{Unit vector pointing along the $i^{th}$ joint axis and expressed in the $(i-1)^{th}$ link coordinate system}

\chapter{\uppercase{Introduction}\label{ch:intro}}

\section{Definition and Purpose of  Quadrotor System}
\begin{figure}[!h]
      \centering
      \includegraphics[width=0.5\columnwidth, height= 5cm]{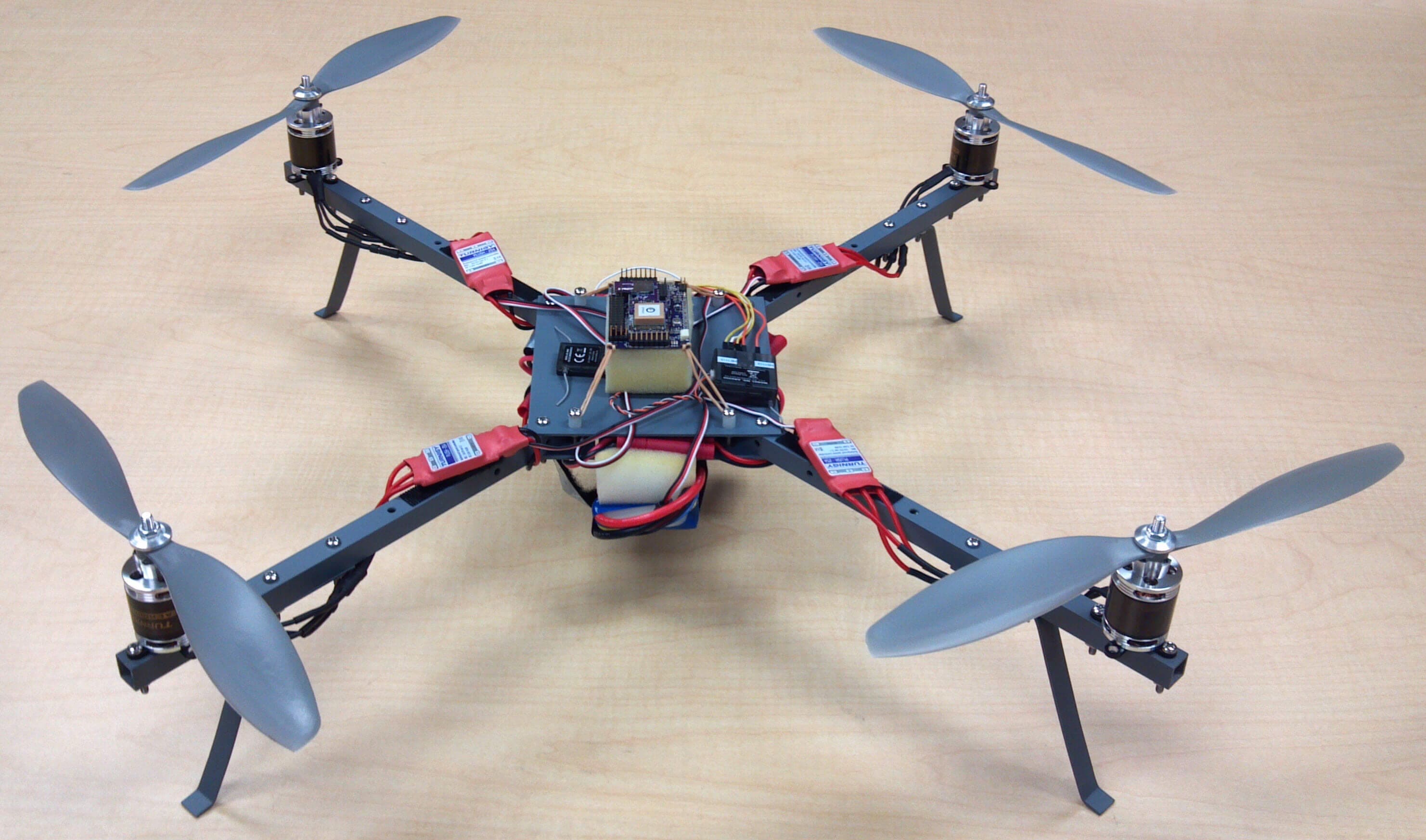}
      \caption{Typical Quadrotor.}
      \label{quad_fig}
   \end{figure}

A quadrotor,  also called a quadrotor helicopter, quadrocopter or quadcopter, is an aircraft that becomes airborne due to the lift force provided by four rotors usually mounted in cross configuration, hence its name (see Figure \ref{quad_fig}).
%=============================================
% ========================================

Early in the history of flight, quadrotor configurations were seen as possible solutions to some of the persistent problems in vertical flight; torque-induced control issues (as well as efficiency issues originating from the tail rotor, which generates no useful lift) can be eliminated by counter-rotation and the relatively short blades are much easier to construct \cite{aerodynamics}.

A number of manned designs appeared in the 1920s and 1930s. These vehicles were among the first successful heavier-than-air vertical take off and landing (VTOL) vehicles. However, early prototypes suffered from poor performance, and latter prototypes required too much pilot work load, due to poor stability augmentation and limited control authority. More recently quadrotor designs have become popular in Unmanned Aerial Vehicle (UAV) research.

These vehicles use an electronic control system and electronic sensors to stabilize the aircraft. With their small size and agile maneuverability, these quadrotors can be flown indoors as well as outdoors.

There are several advantages to quadrocopters over comparably-scaled helicopters. First, quadrotors do not require mechanical linkages to vary the rotor blade pitch angle as they spin. This simplifies the design and maintenance of the vehicle. Second, the use of four rotors allows each individual rotor to have a smaller diameter than the equivalent helicopter rotor, allowing them to possess less kinetic energy during flight. This reduces the damage caused when the rotors hit anything. For small-scale UAVs, this makes the vehicles safer for close interaction. Some small-scale quadrotors have frames that enclose the rotors, permitting flights through more challenging environments, with lower risk of damaging the vehicle or its surroundings. Due to their ease of both construction and control, quadrotor aircraft are frequently used as amateur model aircraft projects \cite{quad_history, quad_prototype}.

At present, there are three main areas of quadrotor development: military, transportation (of goods and people) and UAVs  \cite{quad_history}. UAVs are subdivided into two general categories, fixed wing UAVs and rotary wing UAVs. Rotary wing UAVs like quadrotors have several advantages over fixed-wing airplanes. They can move in any direction and are capable of hovering and fly at low speeds. In addition, the VTOL capability allows deployment in almost any terrain while fixed-wing aircraft require a prepared airstrip for takeoff and landing.

Given these characteristics, quadrotors can be used in search and rescue missions, meteorology, penetration of hazardous environments and other applications suited for such an aircraft. Also, they are playing an important role in research areas like control engineering, where they serve as prototypes for real life applications.

Furthermore, quadrotor  vehicles  possess  certain  essential  characteristics,  which  highlight their potential for use in search and rescue applications. Characteristics that provide a clear advantage over other flying UAVs include their (VTOL) and hovering capability, as well as their ability to make slow precise movements. There are also  definite  advantages  to  having  a  four  rotor  based  propulsion  system,  such  as  a  higher payload  capacity,  and  impressive  maneuverability,  particularly  in  traversing  through  an environment with many obstacles, or landing in small areas \cite{Fuzzy_Quad_only}.
\section{Quadrotor Operation}
\begin{figure}[!h]
      \centering
      \includegraphics[width=0.8\columnwidth, height=8cm]{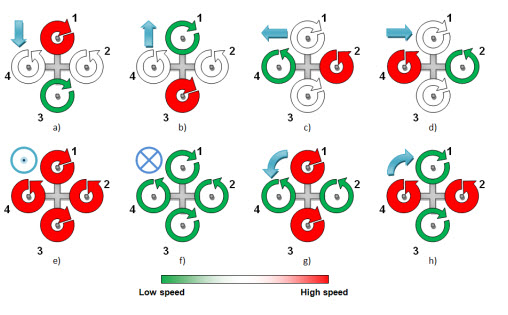}
      \caption{Illustration of the various movements of a quadrotor:
      a) Motion in x-direction(Forward)-Pitching, b) Motion in x-direction(Backward)-Pitching, c) Motion in y-direction(Left)-Rolling, d) Motion in y-direction(Right)-Rolling e) Motion in z-direction(Up), f) Motion in z-direction(Down), g) Rotation around z-Axis(Heading Left)-Yawing, and h) Rotation around z-Axis(Heading Right)-Yawing.}
      \label{quad_operation}
   \end{figure}

\begin{figure}[!h]
      \centering
      \includegraphics[width=0.8\columnwidth, height= 6.5cm]{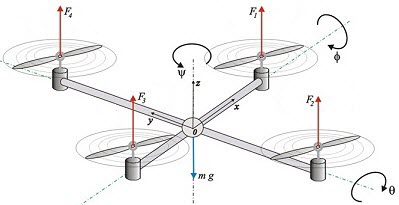}
      \caption{Yaw, Pitch and Roll Rotations of Quadrotor.}
      \label{quad_frames}
   \end{figure}

Possible movements of a quadrotor are shown in Figures \ref{quad_operation} and \ref{quad_frames}. By varying the speeds of each rotor, the flight of the quadrotor is  controlled. Each rotor in the quadrotor has a role in producing a certain amount of thrust and torque about its center of rotation, as well as for a drag force opposite to the rotorcraft's direction of flight. The quadrotor's propellers are divided in two pairs, two pusher and two puller blades, that work in opposite directions.  As a consequence, the resulting net torque can be null if all propellers turn with the same angular velocity, thus allowing for the aircraft to remain still around its center of gravity.
% ===========================================
% ==================================================

In order to define an aircraft's orientation (or attitude) around its center of mass, aerospace engineers usually define three dynamic parameters, the angles of yaw ${\psi}$(rotation around z-axis), pitch ${\theta}$ (rotation around y-axis) and roll ${\phi}$ (rotation around x-axis). This is very useful because the forces used to control the aircraft act around its center of mass, causing it to pitch, roll or yaw (see Figure \ref{quad_frames}).
Changes in the pitch angle are induced by contrary variation of speeds in propellers 1 and 3 , resulting in forward or backwards translation (Movement in x-direction). If we do this same action for propellers 2 and 4, we can produce a change in the roll angle and we will get lateral translation (Movement in y-direction). Yaw is induced by mismatching the balance in aerodynamic torques (i.e. by offsetting the cumulative thrust between the counter-rotating blade pairs). So, by changing these three angles in a quadrotor, we are able to make it maneuver in any direction \cite{quad_prototype}.
%====================================================
\section{Literature Review}
\begin{figure}[!h]
      \centering
      \includegraphics[width=0.7\columnwidth, height=5cm]{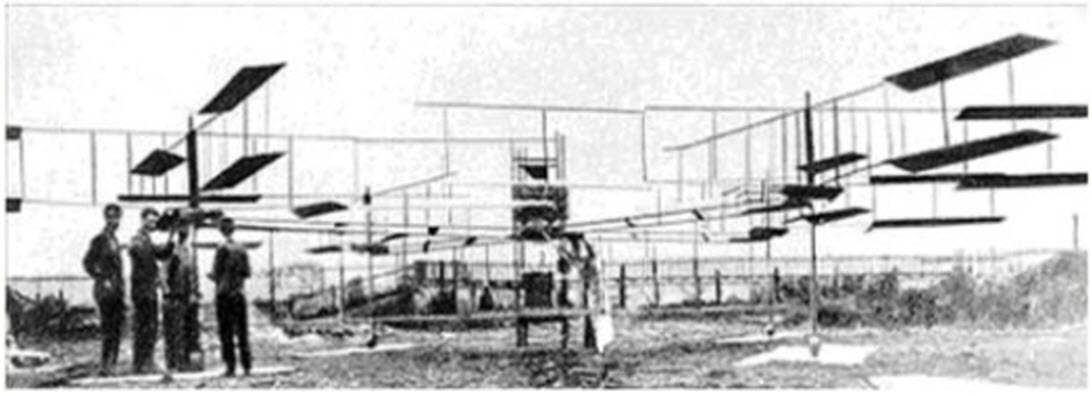}
      \caption{Bréguet Richet Gyroplane No. 1.}
      \label{Gyroplane}
   \end{figure}

Louis Bréguet and Jacques Bréguet were the first to  construct a quadrotor, which they named Bréguet  Richet Gyroplane No. 1.  The  first  flight  demonstration  of  Gyroplane  No. 1 with no control surfaces was achieved  on  29 September 1907. Figure  \ref{Gyroplane} shows the huge quadrotor with double layered propellers being prepared for its first manned flight \cite{quad_history, Fuzzy_Quad_only}.
% ===========================================
% ==================================================

\begin{figure}[!h]
      \centering
      \includegraphics[width=0.7\columnwidth, height=5cm]{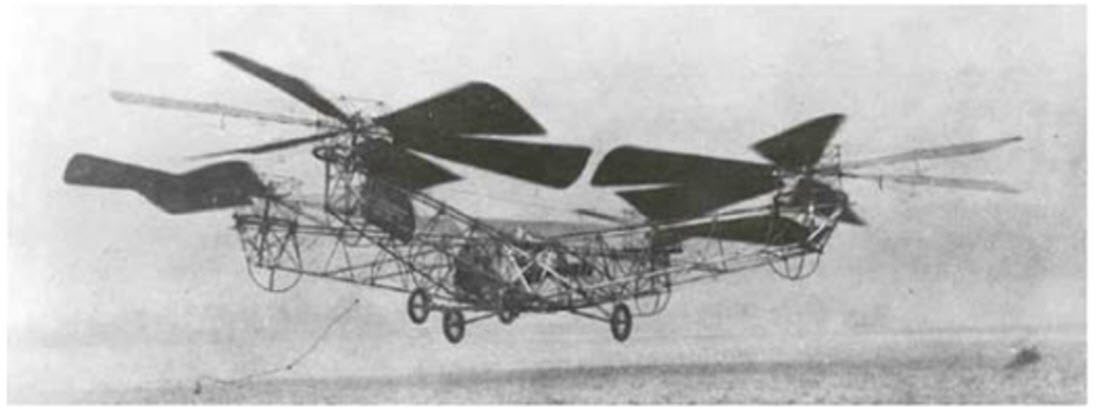}
      \caption{Quadrotor designed by George De Bothezat, in 1923. }
      \label{designed_by_George}
   \end{figure}

\begin{figure}[!h]
      \centering
      \includegraphics[width=0.7\columnwidth, height=5cm]{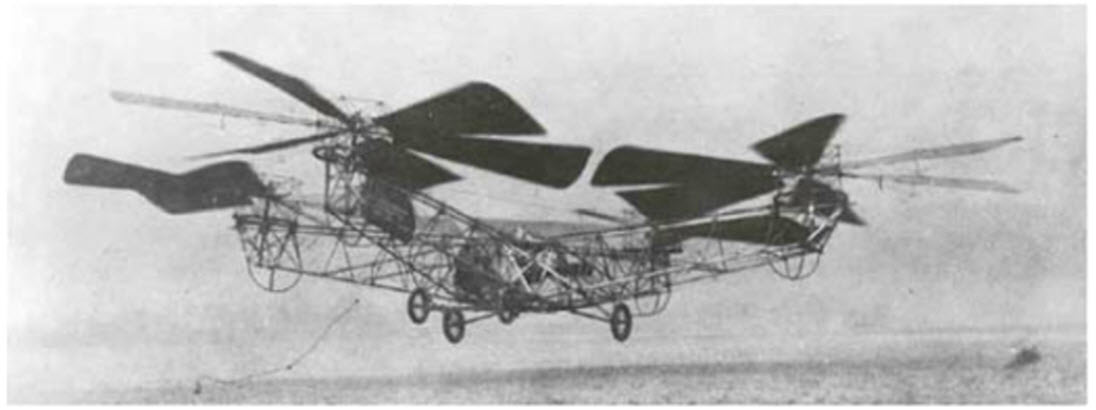}
      \caption{OE hmichen quadrotor designed in 1924.}
      \label{OE_hmichen}
   \end{figure}

Later, two additional designs were developed and experimental flights were conducted. The first, by Georges de Bothezat and Ivan Jerome in 1922, had six-bladed rotors placed at each end of an X-shaped truss structure, as shown in Figure \ref{designed_by_George}. The second, shown in Figure \ref{OE_hmichen}, was built by Étienne OE hmichen in 1924, and set distance records, including achieving the first kilometer long helicopter flight \cite{quad_history, Fuzzy_Quad_only}.
% ===========================================
% ==================================================
% ===========================================
% ==================================================

\begin{figure}[!h]
      \centering
      \includegraphics[width=0.5\columnwidth, height=5cm]{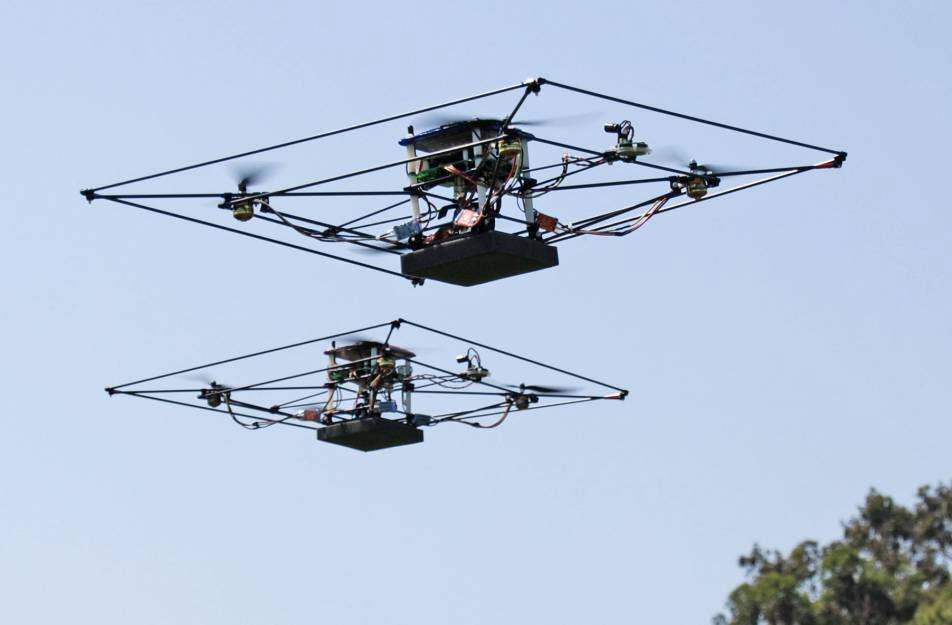}
      \caption{The STARMAC of Stanford University (typical modern quadrotor design).}
      \label{STARMAC}
   \end{figure}

The Stanford test bed of Autonomous Rotorcraft for Multi Agent Control (STARMAC) project at Stanford University performed some of the initial work on making small-scale quadrotors autonomous. Stanford was able to modify commercially available quadrotors (Dragonfly X4s) to follow a series of GPS way points. After achieving this system, Stanford created the STARMAC II series of quadrotors (see Figure \ref{STARMAC}) with the goal of improving stability and control to make quadrotors super stable \cite{quad_history, Fuzzy_Quad_only, quad_commercial}.
% ===========================================
% ==================================================

Recently, quadrotor is used as a research platform. Quadrotors are a useful tool for university researchers to test and evaluate new ideas in a number of different fields, including flight control theory, navigation, real time systems, and robotics. In recent years many universities have shown quadrotors performing increasingly complex aerial manoeuvres. Swarms of quadrotors can hover in mid-air, in formation, autonomously perform complex flying routines such as flips, darting through hula-hoops and organise themselves to fly through windows as a group \cite{quad_popular, quad_swarm}.

Because they are so manoeuvrable, quadrotors could be useful in all kinds of situations and environments. Quadrotors capable of autonomous flight could help remove the need for people to put themselves in any number of dangerous positions. This is a prime reason that research interest has been increasing over the years \cite{quad_popular}.

Moreover, Quadrotors are used for surveillance and reconnaissance by military and law enforcement agencies, as well as search and rescue missions in urban environments. One such example is the Aeryon Scout, created by Canadian company Aeryon Labs, which is a small UAV that can quietly hover in place and use a camera to observe people and objects on the ground. The company claims that the machine played a key role in a drug bust in Central America by providing visual surveillance of a drug trafficker's compound deep in the jungle \cite{quad_armed, quad_guys}.

In addition, The largest use of quadrotors has been in the field of aerial imagery. Quadrotor UAVs are suitable for this job because of their autonomous nature and huge cost savings. Capturing aerial imagery with a quadrotor is as simple as programming GPS coordinates and hitting the go button. Using on-board cameras, users have the option of being streamed live to the ground. Many companies have used this for real estate photography to industrial systems inspection. Various organizations are taking advantage of the quadrotor’s closed-circuit television capabilities and ability to provide an eye in the sky to the action below \cite{quad_commercial}.

From the above discussion, we can notice that quadrotor UAVs offer promises of speed and access to regions that are otherwise inaccessible to ground robotic vehicles. However,  most research on UAVs has typically been limited to  monitoring and surveillance applications where the objectives are limited to "look" and "search" but "do not touch".

\begin{figure}[!h]
      \centering
      \includegraphics[width=0.8\columnwidth, height=7cm]{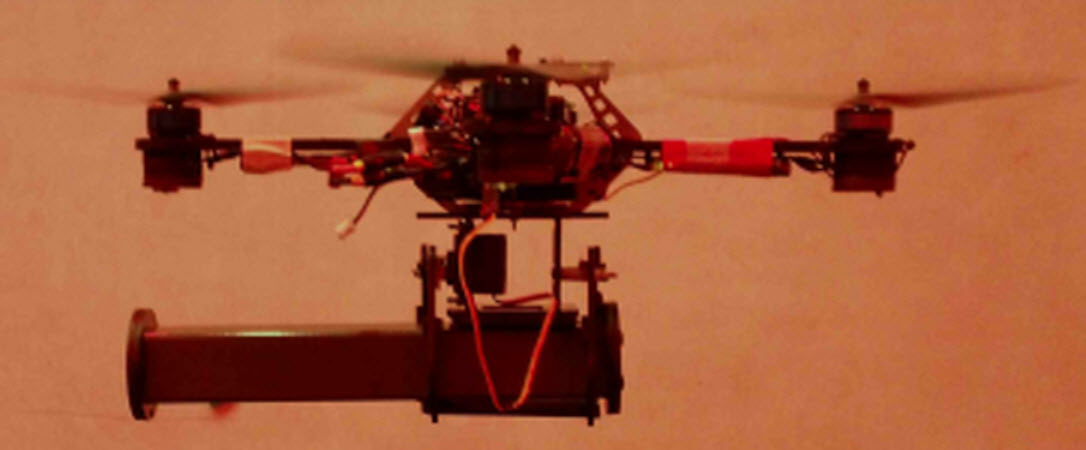}
      \caption{Quadrotor Gripper System by D. Mellinger (2011).}
      \label{quad_gripper}
   \end{figure}

\begin{figure}[!h]
      \centering
      \includegraphics[width=0.5\columnwidth, height=4cm]{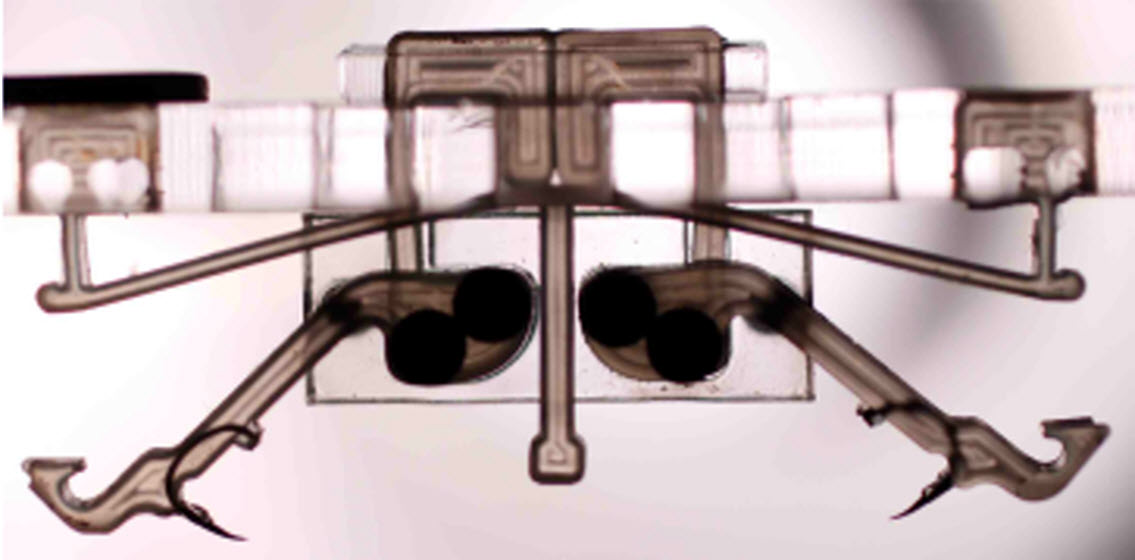}
      \caption{Gripper Design for the Quadrotor Gripper System  by D. Mellinger (2011).}
      \label{gripper_design}
   \end{figure}

In \cite{Aerial_Grasping}, the author presents the design and control of an aerial grasping. In this work \cite{Aerial_Grasping}, a gripper was built and attached to the bottom of the quadrotor as shown in Figures \ref{quad_gripper} and \ref{gripper_design}. By allowing quadrotors, to interact with the environment, we get an entire new set of applications. First, allowing robots to fly and perch on rods or beams allows  them  to increase  the  endurance  of  their  missions. Indeed, if perches are equipped with charging stations, robots can recharge their batteries extending their lives substantially. Second, the ability to grasp objects allows robots to access payloads that are unavailable to ground robots. There are fewer workspace constraints for aerial robots. Third, they are able to assemble structures and scaffolds of arbitrary height in three dimensions without requiring such special purpose structures as tower cranes.
% ===========================================
% ==================================================
% ===========================================
% ==================================================
\section{The Proposed System}
From the quadrotor operation, we can notice that we have to make a pitch rotation in order to make quadrotor to move in X-direction, while the movement in Y-direction is caused by changing roll angle. Furthermore, if a gripper is attached to the bottom of quadrotor, such that a quadrotor grasping system can be built, then the resulting aerial grasping system has 4 Degrees Of Freedom (DOF). Three translations DOF and one rotational DOF (Yaw). In addition, the gripper cannot posses pitch or roll rotation without moving along x- or y- direction.

The proposed system is a new quadrotor manipulation system. This system consists of two-link manipulator, with two revolute joints, is attached to the bottom of a quadrotor. The two axes of the revolute joints are perpendicular to each other. With this new system, the capability of manipulating objects with arbitrary location and orientation is achieved because the DOF are increased from 4 to 6. In addition, the manipulator provides sufficient distance between the quadrotor and the object location. Moreover, this configuration is the best one because it achieves minimum manipulator weight for arbitrary aerial manipulation.

There are two main challenges that face the aerial manipulation. First, lack of a stable platform. This is because the arial platform is different from ground vehicle that can remain stationary and provide a very stable base during the manipulation. Second, dynamics of the robot are significantly altered by the addition and manipulation of payloads. Therefore, Adaptive and robust controller will be needed in order to overcome these challenges.
\section{Motivation}
The proposed system has a lot of potential applications such as:
\begin{itemize}
\item Demining applications/Improvised Explosive Device (IED) Disposal.
\item Accessing payloads that are inaccessible to ground robots.
\item Performing maintenance for a bridge or building.
\item Hazardous material handling and removal.
\item Removing obstacles that are blocking the view of a target or perch to conserve power.
\end{itemize}
\section{Contributions of this Thesis}
During the development of this thesis, set of contributions were made that are worth mentioning:
\begin{itemize}
\item Introducing and describing a novel quadrotor manipulation system.
\item Detailed mathematical modeling including kinematics and dynamics of this system taking into account the effect of payload.
\item Building a quadrotor system as an initial step to build the complete proposed system.
\item Building experimental test rigs in order to determine the quadrotor parameters in order to be used in system simulation and controller design.
\item Using a sensor fusion technique to estimate the quadrotor's altitude and attitude using a 10 DOF Inertial Measurement Unit (IMU).
\item Implementing an onboard algorithm for attitude estimation and attitude stabilization.
\item Building experimental test rigs in order to test the estimated parameters of quadrotor and the estimation algorithm for attitude.
\item Design three different types of control techniques in order to achieve system stability and trajectory tracking under the effects of changing system operating region and picking/placing a payload.
\item Simulation of the system's equations of motion and the control laws in MATLAB/SIMULINK program.

\end{itemize}
\section{Thesis Structure}
In Chapter 1, we begin by defining what a quadrotor is, and what its main applications are. After defining the maneuvering capabilities of this kind of air vehicle, its historical evolution is presented, from the beginning of the twentieth century to the present day. Next, brief description of the quadrotor grasping system is presented. Finally, the reasons that led to the development of proposed system are explained, and a brief description of the main areas of research with this new system is made.

Chapter 2 focuses on the proposed system design. It describes the quadrotor manipulation system. In addition, the selected components for developing the system is presented.

Chapter 3 presents the kinematic and dynamic analysis of the quadrotor. Next, we define the system parameters, and present experiments that are done to determine the system parameters. Finally, it provides a technique for attitude estimation using sensor fusion on a 10 DOF IMU. Also, It presents the online implementation of the attitude estimation and stabilization. Finally, the experimental results are shown.

Chapter 4 focuses on the kinematic(forward and inverse) and dynamic analysis of the new quadrotor manipulation system. Moreover, it presents the effect of adding a payload to the manipulator's end effector

Chapter 5 describes the controller design process of the system. It starts with the definition of the objectives to be fulfilled by the system.  Next, controller design based on three different control techniques are described with their simulation results and observations. First, feedback linearization technique, followed by, Direct Fuzzy Logic control, and finally, Adaptive fuzzy logic control.

This thesis ends in Chapter 6 with the final conclusions and suggestions for future improvements.

\chapter{\uppercase{The Proposed System}\label{ch:proposedsystem}}

\begin{figure}[!h]
      \centering
      \includegraphics[width=1\columnwidth, height= 11cm]{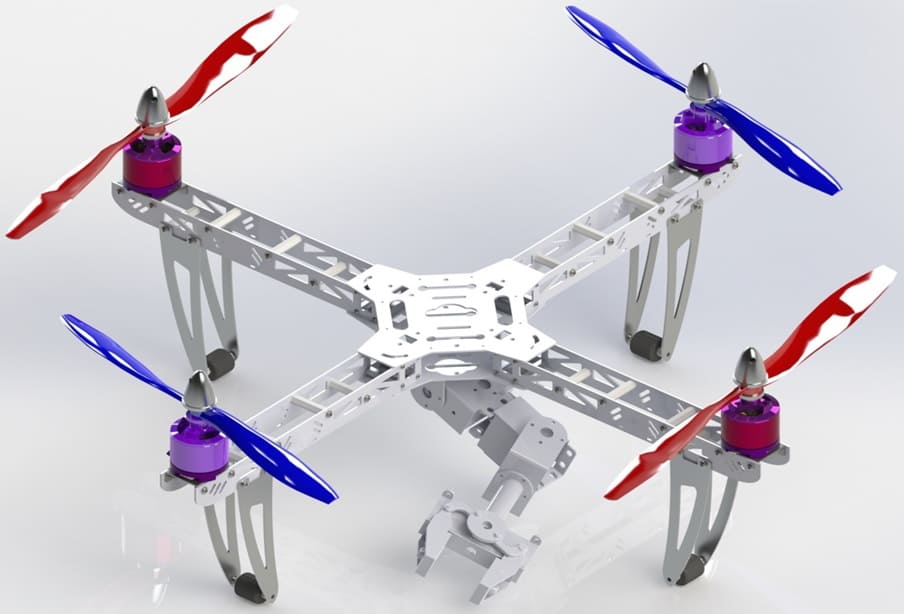}
      \caption{3D CAD Model of the New Quadrotor Manipulation System}
      \label{3D-CAD-MODEL}
\end{figure}

The structure of the proposed system is shown in Figure \ref{3D-CAD-MODEL} using a 3D CAD model. The proposed system is a new quadrotor manipulation system. This system consists of two-link manipulator, with two revolute joints, attached to the bottom of a quadrotor. The two axes of the revolute joints are perpendicular to each other. The axis of the first revolute joint that is fixed with respect to the quadrotor is parallel to the body x-axis of the quadrotor (see Figure \ref{quad_frames}). The axis of the second joint will be parallel to the body y-axis of quadrotor when the first link is vertical. So, the pitching and rolling rotation of the end effector is now possible independently on the horizontal motion of the quadrotor. With this new system, the capability of manipulating objects with arbitrary location and orientation is achieved because the DOF are increased from 4 to 6. In addition, the manipulator provides sufficient distance between the quadrotor and the object location. Furthermore, this configuration is the best one because it achieves the minimum manipulator weight for arbitrary aerial manipulation. The light weight is important especially for aerial vehicle.
% =======================================
% ===========================================

The proposed quadrotor manipulation system consists mainly from two parts; the quadrotor and the manipulator. During the reset of this chapter we will describe each of them such that they can combined, and then, perform the required task.

Our target is to design a light and simple 2 DOF manipulator that can carry as much as possible of a payload. One of the available and famous company to sell the components of such type of manipulator is \uppercase{"lynxmotion"} \cite{lynxmotion}.

The main components of the the proposed system should be:
\begin{itemize}
\item  4 Electric Motors and 4 Electronic Speed Controllers (ESC).
\item 4 Propellers.
\item 1 On-board Processing and Control Unit.
\item Wireless Communication unit between the base station and the quadrotor.
\item Altitude and Attitude Measurement Units.
\item 1 Airframe for Supporting all the Aircraft's Components.
\item  Gripper Mechanism (End Effector).
\item  2 Motors and their Accessories for the Manipulator Construction (two links with two revolute joints).
\item  Motors's Drive Unit.
\item On-board Power Supplies (battery) for the Motors and the Electronics.
\end{itemize}
% ===========================================
\section{The Two-Link Manipulator}
Figure \ref{3D-CAD-MODEL} shows 3D CAD model for the proposed manipulator. The design of this manipulator is based on light weight and enough workspace under the quadrotor.
%% ===========================================
%\begin{figure}[!h]
%      \centering
%      \includegraphics[width=0.7\columnwidth, height= 7cm]{figures/3dmodel_manp}
%      \caption{3D CAD Model for the proposed Two-Link Manipulator.}
%      \label{3dmodel_manp}
%\end{figure}
%% ===========================================

\begin{figure}[!h]
      \centering
      \includegraphics[width=0.9\columnwidth, height=5cm]{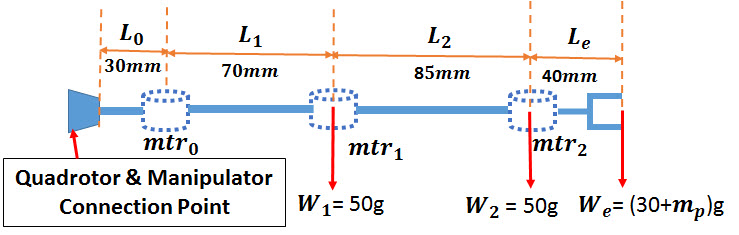}
      \caption{Schematic Diagram of the Manipulator Used to Select the Joints Motors.}
      \label{2d_manp_motorsel}
\end{figure}

Starting with design of the gripper mechanism, a little gripper, from \uppercase{"lynxmotion"}, is used. It is made from injection molded ABS. This gripper can open to 1.3 in. It has a driving motor ($mtr_2$), see Figure \ref{2d_manp_motorsel}, of type servo motor HS-422. This gripper can carry a payload of 200 g \cite{gripper_payload}. Thus, a payload of 150 g is chosen as a maximum allowable value for the payload carried by the gripper (for safety purpose).
%Figure \ref{gripper_mech} presents a schematic diagram for the gripper with the generated forces due to the driving motor and the payload.
%
%Let start from the payload and build the calculation based on a payload of 200 g (for safety purpose).
%
%Applying Newton laws to the forces in Figure \ref{gripper_mech}:
%\begin{equation}
% F_{mtr_2}= \frac{m_{p}*g}{2\mu cos(67.16^{o})}
% \label{F_mtr2}
%\end{equation}
%
%where, $F_{mtr_2}$ is the force generated from the rotation of driving motor, $\mu$ is the friction coefficient (has values from 0.15 to 0.3).
%
%Let make our design based on the worst case i.e. $\mu$ = 0.15, then from (\ref{F_mtr2}) the required force is 16.8316 N. This force is generated from the motor torque ($T_{mtr_2}$) through the gripper mechanism of diameter ($d_{mtr_2}$) equal to 18.43 mm, thus the required torque is:
%\begin{equation}
% T_{mtr_2} = F_{mtr_2} * d_{mtr_2} = 0.3102 N.m
% \label{t_mtr2}
%\end{equation}

%Therefore, a servo motor (HS-422), Voltage = 4.8 - 6.0 vdc, Torque = 0.324 N.m, and Weight = 45 g, is sufficient \cite{lynxmotion}.
% =======================================================
%\Begin{Figure}[!h]
%      \Centering
%      \Includegraphics[Width=1\Columnwidth, Height=12Cm]{Figures/Gripper_Mech}
%      \Caption{Schematic Diagram For The Gripper With The Generated Forces}
%      \Label{Gripper_Mech}
%\End{Figure}
% ===========================================

The total length of the arm is chosen such that it can provide enough work space. If the length is increased , then a motor with larger output torque is required, and thus larger weight and more consumed power are resulted.

Based on this design, the required specifications of the selected joints motors are set from Figure \ref{2d_manp_motorsel}.

The torques, $T_{mtr_0}$ and $T_{mtr_1}$, of motors ($mtr_0$) and ($mtr_1$), see Figure \ref{2d_manp_motorsel}, can be calculated from(\ref{torque_mtr0} and \ref{torque_mtr1}):
\begin{equation}
T_{mtr_0} = L_1 W_1 + (L_1 + L_2 ) W_2 + (L_1 + L_2 + L_e) W_e
\label{torque_mtr0}
\end{equation}
\begin{equation}
T_{mtr_1} = L_2 W_2 + (L_2 + L_e) W_e
\label{torque_mtr1}
\end{equation}

where $W_e$ is the weight from the end effector and the payload that it has mass of $m_p$.

The average weight of the available motors in the light categories is 50 g (including the arm accessories that will be described next).

Therefore, from the values of lengths and masses shown in Figure \ref{2d_manp_motorsel}, the value of $T_{mtr_0}$ is 0.45 N.m and $T_{mtr_1}$ is 0.26 N.m (@ payload; $m_p$ = 150 g). Multiplying this value by factor of 1.2 for safety, the selected motors should have output torque at least of 0.55 N.m for $mtr_0$ and 0.31 N.m for $mtr_1$.
% ===========================================
% ===========================================

\begin{figure}
      \centering
      \includegraphics[width=0.3\columnwidth, height=4cm]{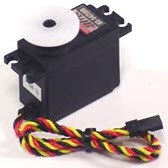}
      \caption{HS-5485HB (89 oz.in) Digital Standard Servo.}
      \label{servomotor}
\end{figure}

 A digital standard servo (HS-5485HB), see Figure \ref{servomotor}, is used to provide the rotational motion (Revolute Joint) for joint 1. It has a rotational range of $180^{o}$, weight of 45 g, speed of $350^{o}/s$, and operating voltage of 6 v. It is capable of producing a torque of 0.63 N.m. Another  digital standard servo (HS-422) is used for joint 2. It has a rotational range of $180^{o}$, weight of 45 g, speed of $350^{o}/s$, operating voltage of 6 v, and torque of 0.32 N.m \cite{lynxmotion}.
  % ===========================================
% ===========================================

\begin{figure}[!h]
    \centering
    \begin{tabular}{cc}
    \subfigure[Aluminum Tubing - 1.50 in]{\includegraphics[width=0.4\columnwidth,height=6cm]{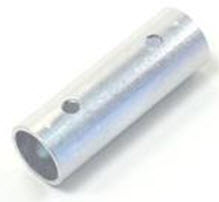}}&
    \subfigure[Aluminum Multi-Purpose Servo Bracket Two Pack]{\includegraphics[width=0.5\columnwidth,height=6cm]{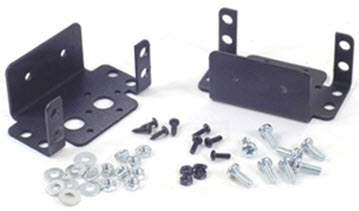}}\\
    \subfigure[Aluminum Tubing Connector Hub]{\includegraphics[width=0.4\columnwidth,height=6cm]{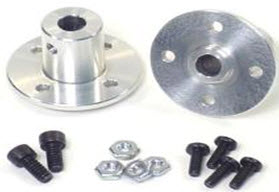}}&
    \subfigure[Aluminum Long "C" Servo Bracket with Ball Bearings Two Pack]{\includegraphics[width=0.5\columnwidth,height=6cm]{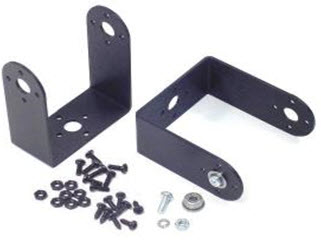}}
    \end{tabular}
    \caption{Accessories for Building the Manipulator: a) Aluminum Tubing, b) Servo Bracket, c) Connector Hub, and d) Long "C" Servo Bracket.}
    \label{accessories}
\end{figure}

\begin{figure}[!h]
    \centering
    \begin{tabular}{cc}
    \subfigure[Aluminum Tubing]{\includegraphics[width=0.4\columnwidth,height=7cm]{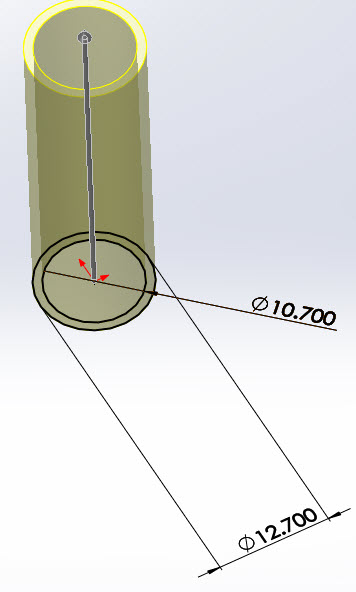}}&
    \subfigure[Servo Bracket]{\includegraphics[width=0.5\columnwidth,height=7cm]{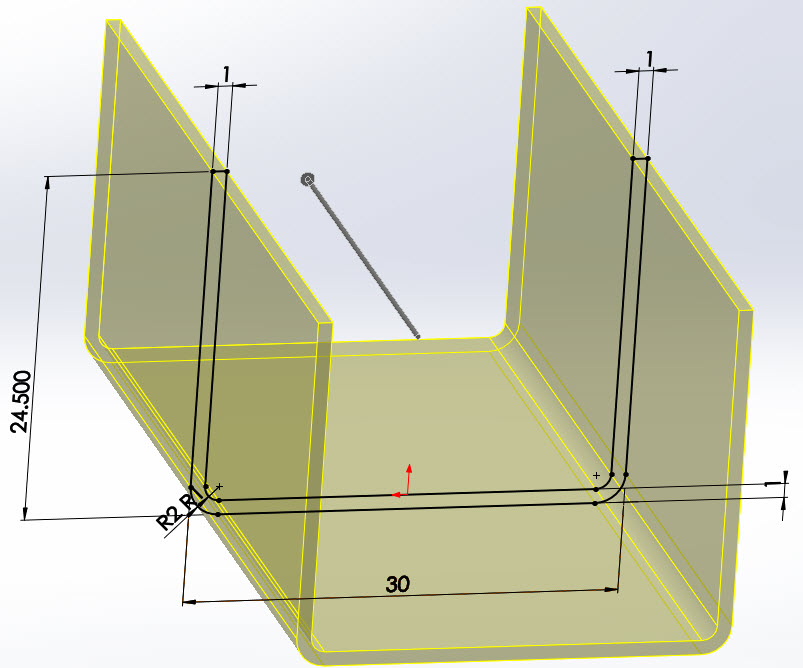}} \\
    \subfigure[Connector Hub]{\includegraphics[width=0.5\columnwidth,height=8cm]{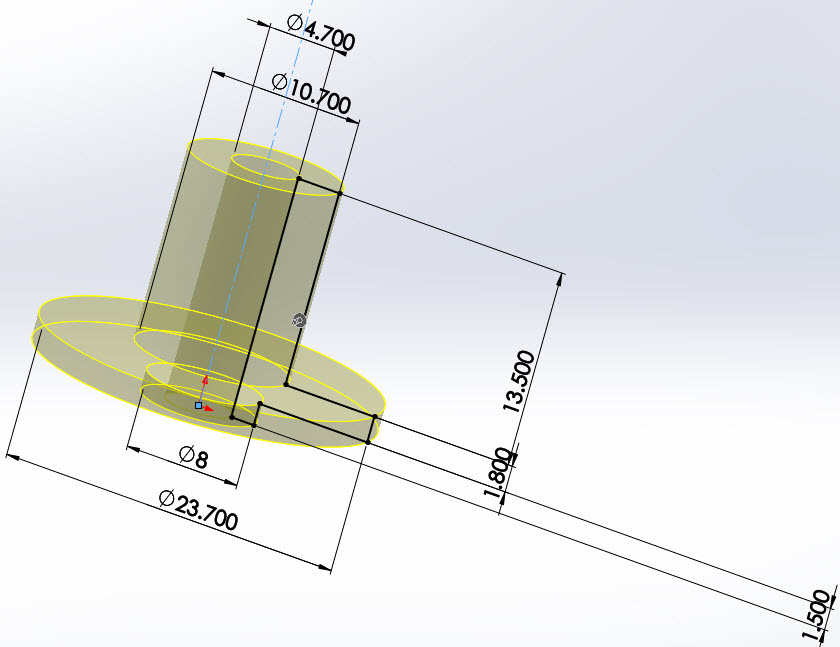}}&
    \subfigure[Aluminum Long "C" Servo Bracket]{\includegraphics[width=0.5\columnwidth,height=7cm]{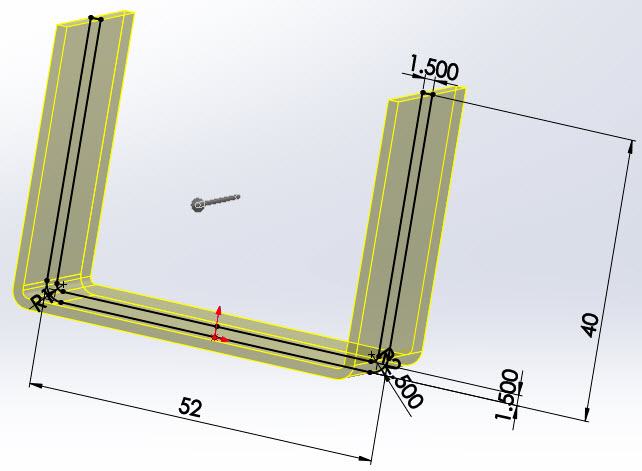}}
    \end{tabular}
    \caption{Dimensions of the Accessories for Building the Manipulator in (mm): a) Aluminum Tubing, b) Servo Bracket, c) Connector Hub, and d) Long "C" Servo Bracket.}
    \label{accessories_dim}
\end{figure}

Figure \ref{accessories} presents the required accessories to build the the two arms of the manipulator. Also, the dimensions of these accessories are given in Figure \ref{accessories_dim}.
% ===========================================
% ===========================================

\begin{figure}[!h]
\centering
\begin{tabular}{cc}
 \subfigure[Total Deformation]{\includegraphics[width=8cm]{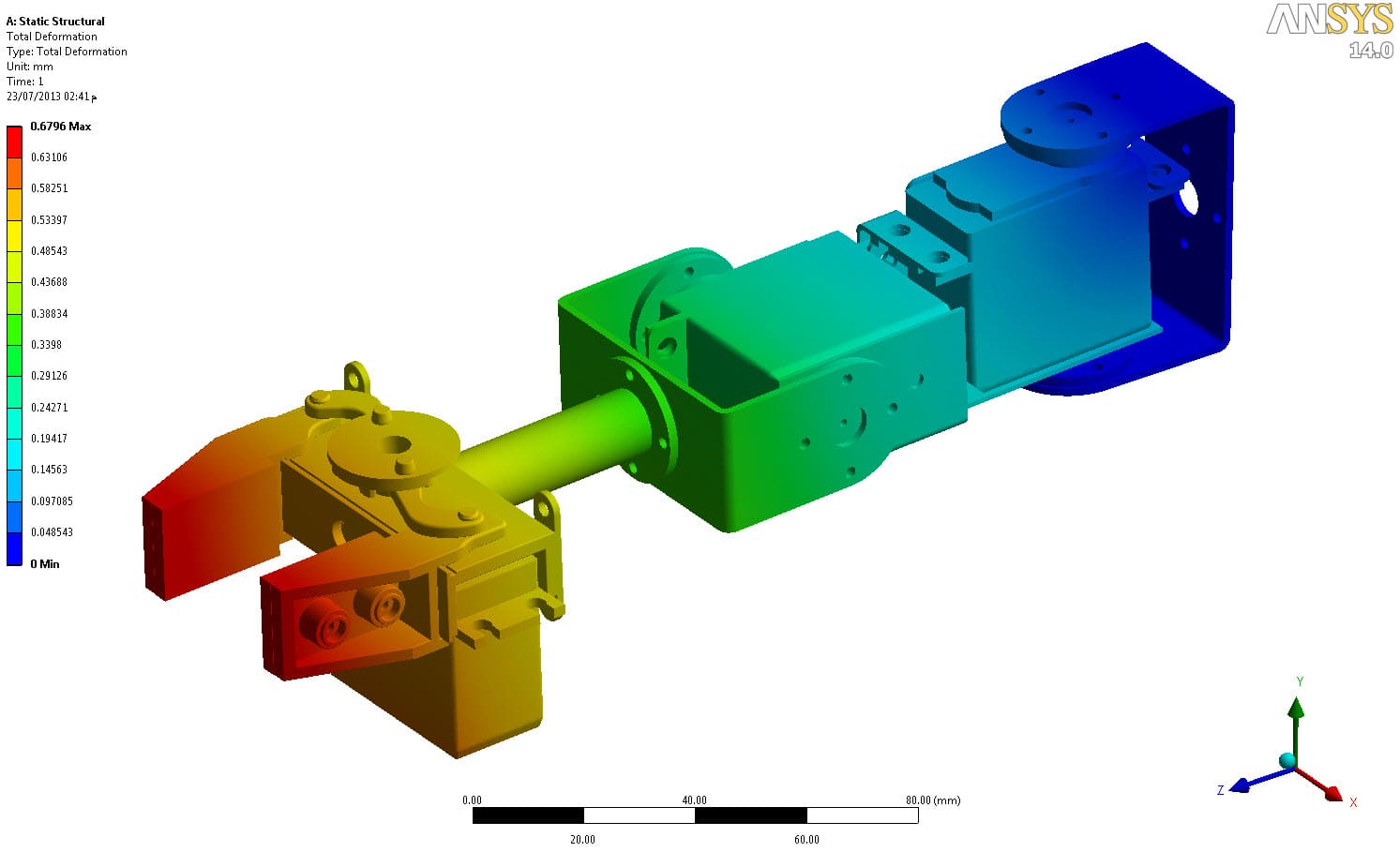}}&
 \subfigure[Deformation in X-axis]{\includegraphics [width=8cm]{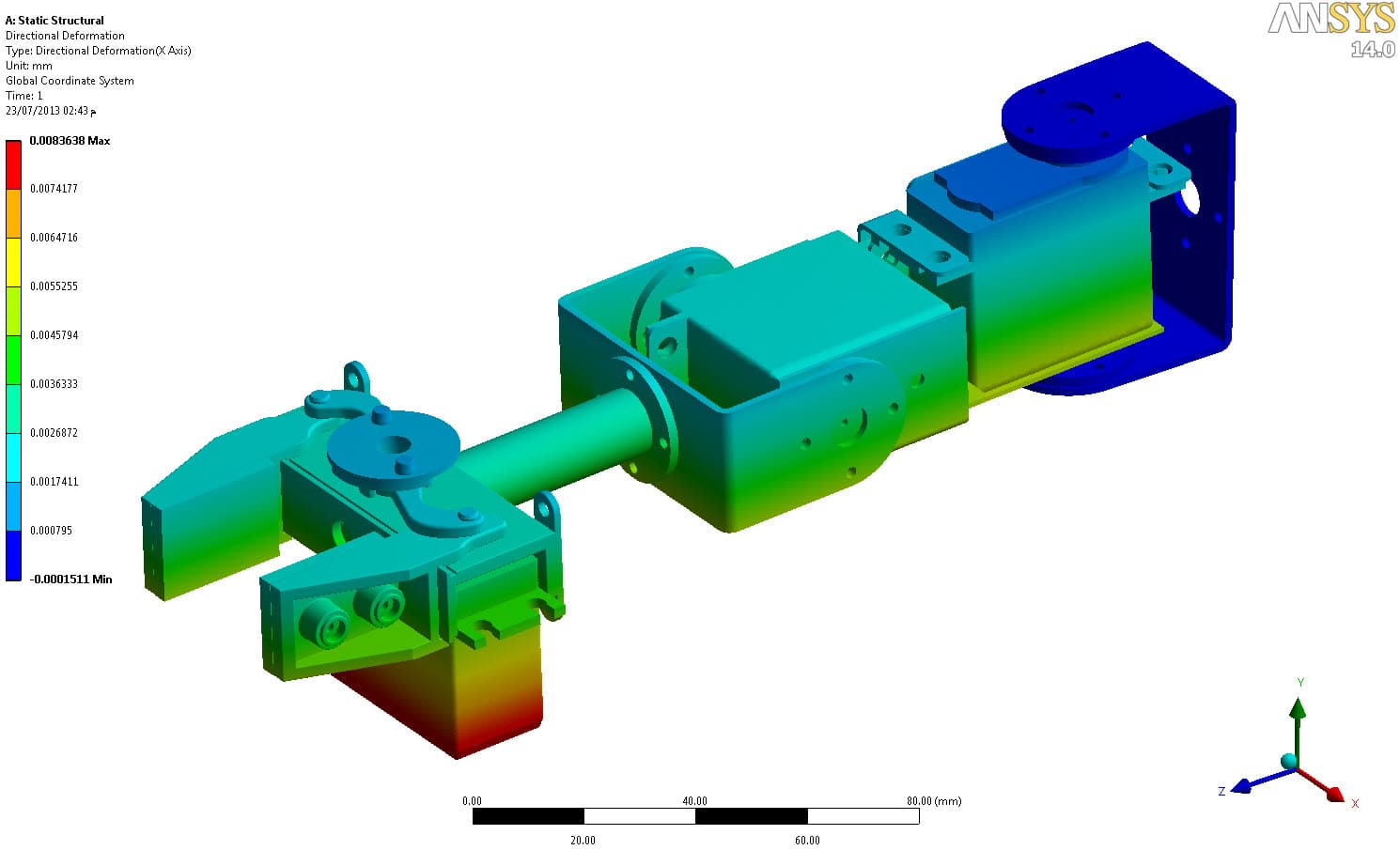}} \\
 \subfigure[Deformation in Y-axis]{\includegraphics [width=8cm]{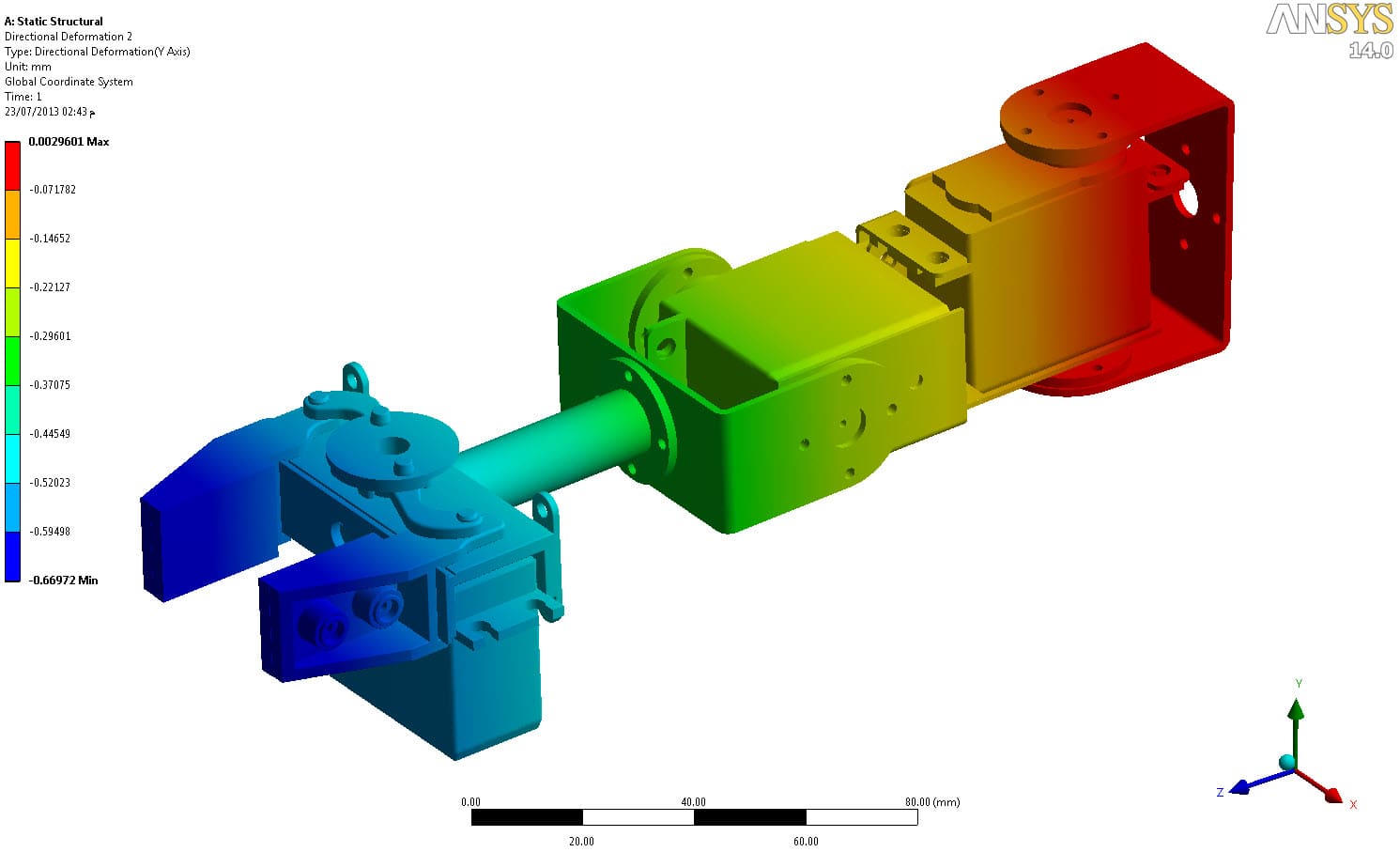}} &
 \subfigure[Deformation in Z-axis]{\includegraphics [width=8cm]{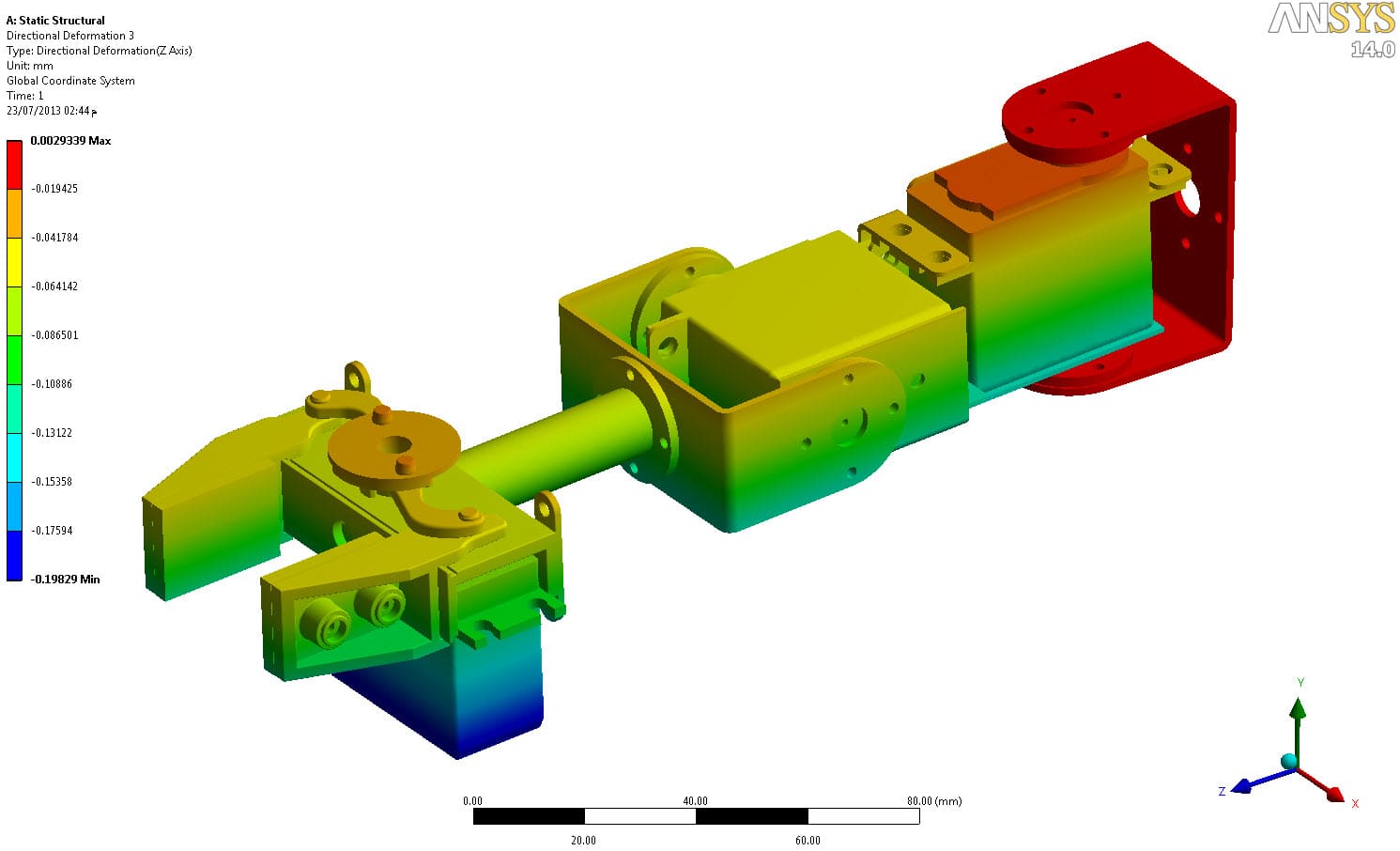}}
\end{tabular}
\caption{Manipulator's Structure Deflections Using ANSYS: a) Total Deformation, b) Deformation in X-axis, c) Deformation in Y-axis, and d) Deformation in Z-axis.}
\label {FEA_ANSYS_deflec}
\end{figure}

\begin{figure}[!h]
      \centering
      \includegraphics[width=0.8\columnwidth, height=7cm]{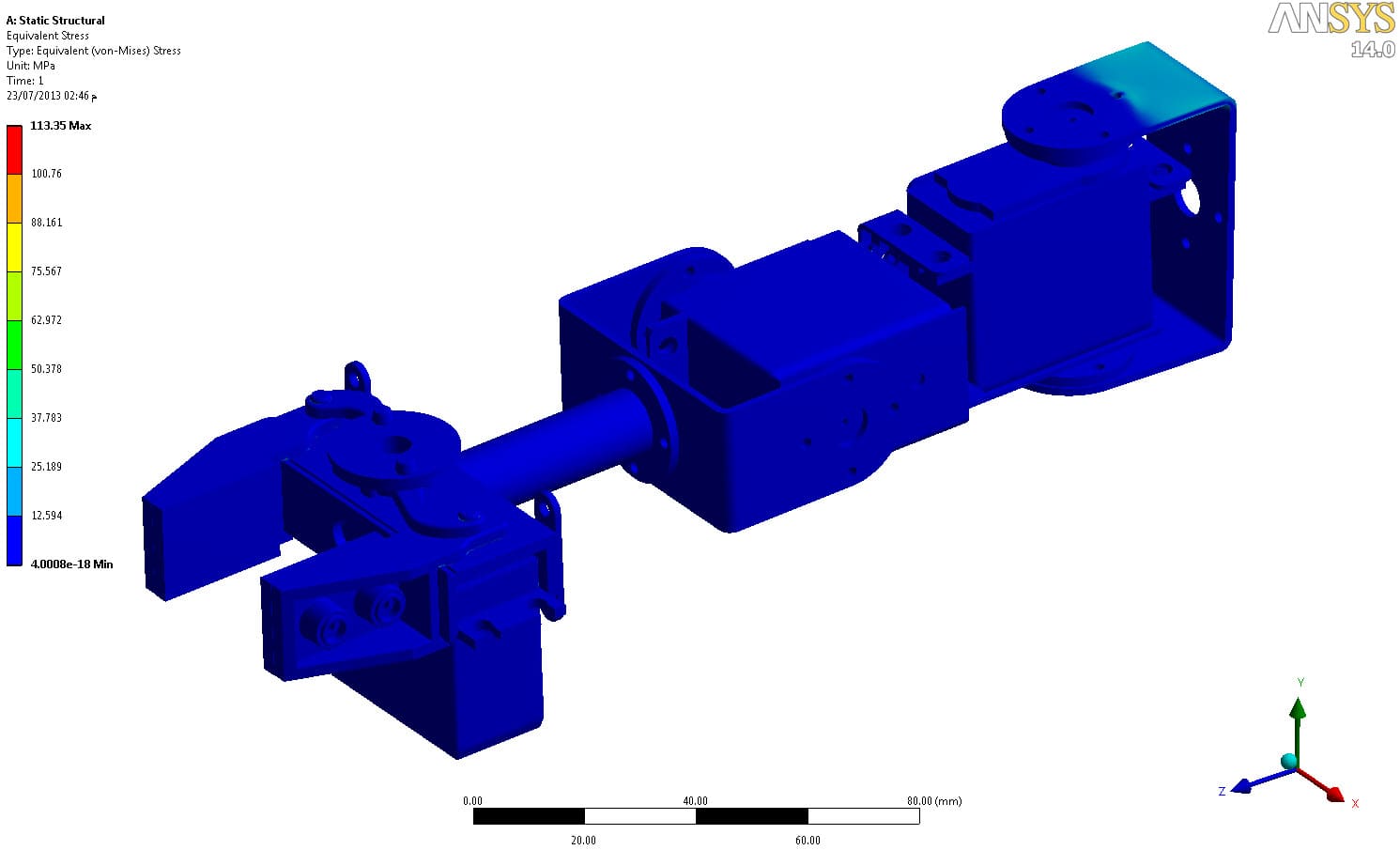}
      \caption{Manipulator's Structure Stress Analysis using ANSYS}
      \label{FEA_ANSYS_stress}
\end{figure}

The safety of this design and structure, with respect to the deflections and stress, is tested and checked through finite element analysis using ANSYS software (see Figures \ref{FEA_ANSYS_deflec} and \ref{FEA_ANSYS_stress}). From this figures, the maximum deflection is about 0.6 mm which is small compared to the chosen allowable is 1mm. In addition, the maximum stress of the structure is 113 MPa which is smaller than the yield stress of aluminum alloy that is 270 MPa \cite{al_stress}. Therefore, this design is safe.
%==============================================
% ===========================================
%==============================================
\subsection{Driver Unit}
\begin{figure}[!h]
      \centering
      \includegraphics[width=0.4\columnwidth, height=3cm]{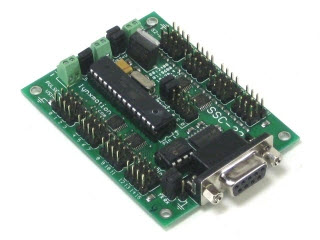}
      \caption{SSC-32 Servo Controller from Lynxmotion}
      \label{servocontroller}
\end{figure}

Serial servo controller (SSC-32) from \uppercase{Lynxmotion}, see Figure \ref{servocontroller}, is a small preassembled servo controller with some big features. It has high resolution (1 $\mu$s) for accurate positioning, and extremely smooth moves. The range is 0.50 ms to 2.50 ms for a range of about $180^{o}$. This board contains a MCU of Atmel ATMEGA168-20PU as well as driver interface between the controller unit and the motors. This driver unit operates at voltage of 12 v \cite{lynxmotion}. This unit will take its commands from the on-Board controller unit (Arduino MEGA 2560).
% ===========================================

In the next sections, the construction of the quadrotor is presented.
\section{Airframe}

The airframe is the mechanical structure of an aircraft that supports all the components.

\begin{figure}[!h]
      \centering
      \includegraphics[width=0.5\columnwidth, height=5cm]{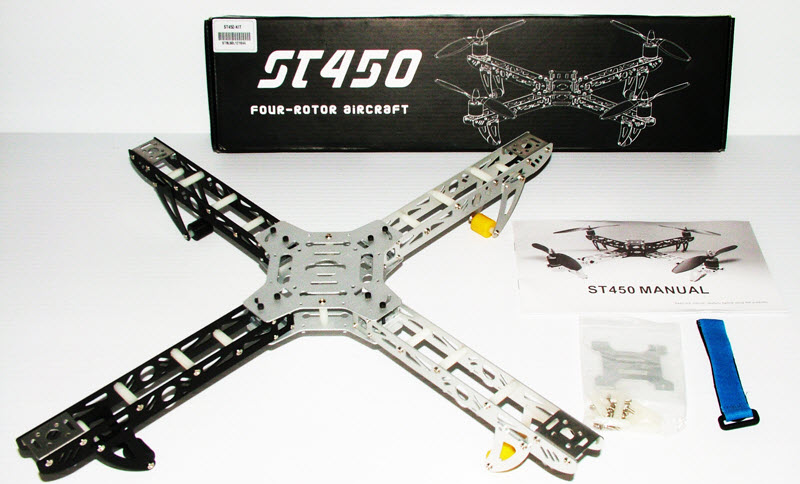}
      \caption{ST450 Quadrotor Airframe.}
      \label{airframe}
\end{figure}

The chosen airframe for the quadrotor was the "ST450 metal folding" model (see Figure \ref{airframe}), with 280 g of mass and made of Aluminum alloy material. The legs of the frame are supported by a piece of rubber to reduce the effect of landing \cite{airframe_ref}.
% =======================================
% ===========================================

\section{Rotor Assembly}
The rotor-assembly consists of three components; the propeller, electric motor, and the Electronic Speed Controller (ESC).

Considering the quadrotor-manipulator system has a maximum weight of 1.5 kg and that we have four rotors, it is mandatory that each rotor-assembly is able to provide at least 375 g in order to achieve lift off.

The typical behavior of a propeller can be defined by three parameters \cite{quad_prototype}:
\begin{itemize}
\item  Thrust coefficient $c_T$.
\item  Power coefficient $c_P$.
\item  Propeller radius $r$.
\end{itemize}
These parameters allow the calculation of a propeller's thrust force $F$ :
\begin{equation}
\label{F_thrust}
F = c_{T} \frac{4\rho r^{4}}{\pi^{2}} \Omega^{2}
\end{equation}
and power $P$:
\begin{equation}
\label{P_power}
P = c_{P} \frac{4\rho r^{5}}{\pi^{3}} \Omega^{3}
\end{equation}
where $\Omega$ is the propeller angular speed and $\rho$ the density of air.
These formulas show that both thrust and power increase greatly with propeller's diameter. If the diameter is big enough, then it should be possible to get sufficient thrust while demanding low rotational speed of the propeller. Consequently, the motor driving the propeller will have lower power consumption, giving the quadrotor higher flight autonomy.

\begin{figure}[!h]
      \centering
      \includegraphics[width=0.3\columnwidth, height=3cm]{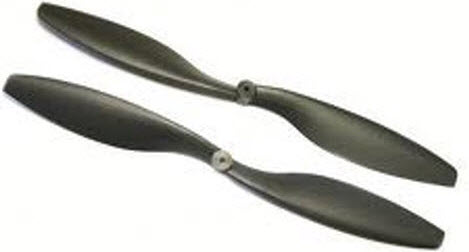}
      \caption{EPP1045 Propellers.}
      \label{propeller}
\end{figure}

\begin{figure}[!h]
      \centering
      \includegraphics[width=0.7\columnwidth, height=8cm]{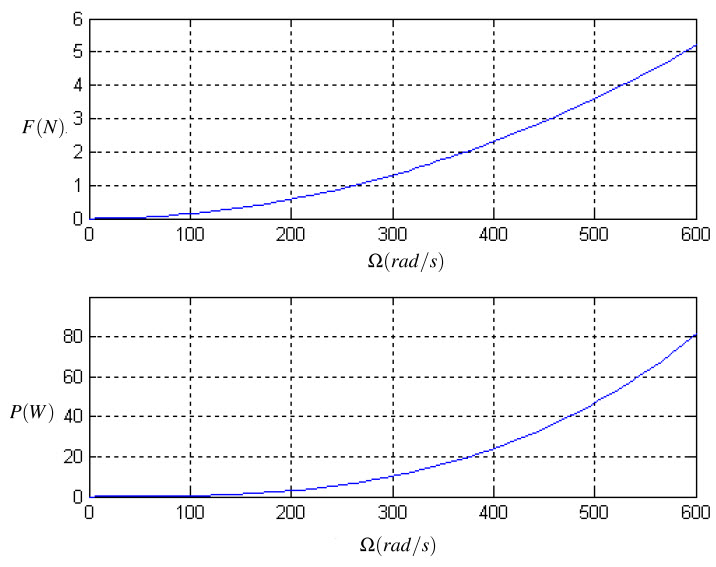}
      \caption{Theoretical Thrust and Power of EPP1045 Propeller \cite{quad_prototype}.}
      \label{propeller_data}
\end{figure}

The "EPP1045" propeller, (see Figure \ref{propeller}) has a diameter of 10" (25.4 cm) and weighting 23 g, is selected. Figure \ref{propeller_data} gives the theoretical thrust/power and angular velocity relationship of this propeller, from which we can see that a propeller will have to achieve approximately 490 rad/s  to provide the minimum 375 g required for lift-off and will need a power of 43 W .
% =======================================
% ===========================================
% ===========================================

The motors usually implemented in this kind of application are electric Direct Current (DC) motors. They are lighter than combustion engines and do not need a combustible fuel, which, among other benefits, decreases the risk of explosion.

DC motors available in the radio control hobby market are either brushed or brushless. Brushless motors are expensive but have higher efficiency, power, and do not need regular maintenance. Brushed motors are cheap but have a shorter lifetime and their brushes need regular replacements.  For these reasons it is preferable to use brushless motors, because loss of structural integrity of the quadrotor due to motor failure should be avoided by using more reliable equipment.

\begin{figure}[!h]
      \centering
      \includegraphics[width=0.4\columnwidth, height=4cm]{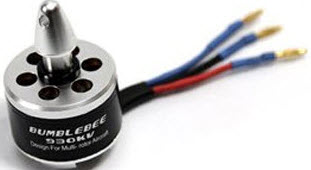}
      \caption{Electric DC Brushless motor BumbleBee 930 KV ST2812.}
      \label{motor}
\end{figure}

The selected motor was the "930KV ST2812" model from the manufacturer BumbleBee (see Figure \ref{motor}). This motor is able to rotate at 8800 rpm, which is larger than the required speed of the propeller for taking off,  at current of 14.5 A , weights 80 g and has a maximum current of 19 A /11.1 V (maximum power is 211 W), and a KV rating of 930 rpm/V \cite{motor_ref}. Thus, this motor is suitable with the selected propeller, because it can provide rotational speed and power that is more than the required.
% ===========================================
% ===========================================

\begin{figure}[!h]
      \centering
      \includegraphics[width=0.3\columnwidth, height=3cm]{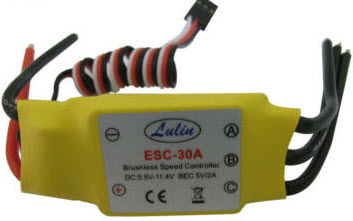}
      \caption{Lulin 30A Brushless ESC.}
      \label{ESC}
\end{figure}

The speed of a brushless motor is controlled by an Electronic Speed Controller (ESC). This hardware receives the power from the battery and drives it to the motor according to a PWM signal that is provided by the controller unit. The "Lulin 30 A" ESC is well suited for the job at hand (Figure \ref{ESC}). It has a mass of 9 g and is capable of providing up to 30 A of current (which is also larger than the maximum allowable current of the BumbleBee 930KV ST2812 motor) \cite{ESC_ref}.
% ===========================================
% ===========================================
\section{Microcontroller Unit}
A stabilization system is necessary to drive the quadrotor because it is a naturally unstable vehicle. The implementation of stabilization control algorithms can be easily implemented by small size microcontrollers.

\begin{figure}[!h]
      \centering
      \includegraphics[width=0.4\columnwidth, height= 3cm]{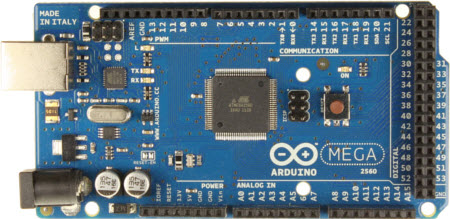}
      \caption{Arduino Mega 2560.}
      \label{arduino2560}
\end{figure}

One microcontroller that has gotten special attention from the robotics community world-wide is the Arduino.  This microcontroller platform is quite inexpensive and has a C-based language development environment that is very intuitive to use.  From the different versions of the Arduino, the selected one for this project was the Arduino Mega 2560 (see Figure \ref{arduino2560}). It has 54 digital input/output pins of which 15 can be used as PWM outputs, 16 analog inputs, 4 UARTs (hardware serial ports), 16 MHz crystal oscillator, USB connection, power jack, an ICSP header, and 256 KB flash memory for storing code \cite{arduino_ref}.
% ===========================================
% ===========================================
\section{Wireless Communication}
\begin{figure}[!h]
      \centering
      \includegraphics[width=0.3\columnwidth, height=2.5cm]{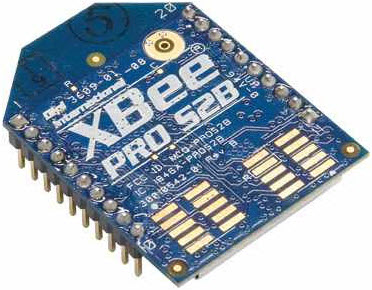}
      \caption{Zigbee Pro-63 mw PCB Antenna Series2 wireless module.}
      \label{zigbee}
\end{figure}

Wireless communications are always a challenge. One has to weight important factors like power consumption, weight, transmission speed and reliability. Fortunately, it is possible to use hardware with the Arduino that allows to satisfy all the previous conditions, e.g.  the Zigbee Pro- 63 mW PCB Antenna Series2 (Figure \ref{zigbee}). This module is a Zigbee Pro module with power of 63 mW and PCB Antenna Series 2 version.  Its communication range is 100 m indoors and 1500 m outdoors. Also, it can be wrapped into a serial command set this is useful because the Arduino can use serial communication. Two XBee modules are going to be used: one for the quadrotor and another in the ground station computer that will handle all telemetry for system identification and control purposes \cite{zigbee_ref}.
% ===========================================
% ===========================================
\section{Sensors}
The sensors of a rotorcraft are a key element of the control loop.  They are responsible for providing information like aircraft attitude, acceleration, altitude, global position, and other relevant data.

In this system we use 3 sensor units; Inertial Measurement Unit (IMU), Sonar unit, and GPS unit.

\begin{figure}[!h]
      \centering
      \includegraphics[width=0.6\columnwidth, height=6cm]{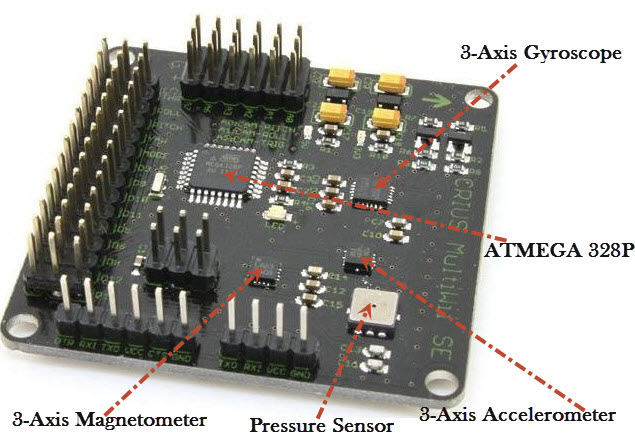}
      \caption{Multiwii ZMR board type used as an IMU.}
      \label{IMU}
\end{figure}

The Inertial Measurement Unit is 10 DOF unit (see Figure \ref{IMU}) that is a Multiwii ZMR board type which  consists  of  high  quality  MEMS  sensors. It contains triple axis accelerometer (BMA180) providing three accelerations (one for each axis of a Cartesian coordinate system), triple axis gyroscope (ITG3205) providing three angular rates,  triple axis magnetometer (HMC5883L) providing the magnetic north direction and good estimation of yaw angle, and pressure sensor (BMP085) providing the aircraft altitude.

All these sensors are connected to ATMEGA328P microcontroller  on  the  same  board. This microcontroller  handles  all  the  readings  from  all  sensors through an $I^{2}C$ bus. Multiwii   ZMR  board  in connected via serial port to Arduino Mega2560 MCU, which contains the main control algorithm.

Readings from the accelerometer, gyroscope, and magnetometer, are fused to provide a high accurate estimation for the quadrotor attitude ($\phi$, $\theta$, and $\psi$).

\begin{figure}[!h]
      \centering
      \includegraphics[width=0.3\columnwidth, height=2.5cm]{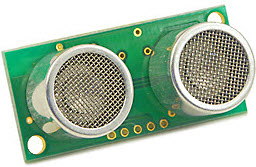}
      \caption{Devantech SRF04 Ultrasonic Range Finder.}
      \label{sonar}
\end{figure}

To calculate the altitude, the pressure sensor will be used at distance lager than 3 m, while if the distance is lower than 3 m the sonar sensor (SRF04) \cite{sonar_ref}, see Figure \ref{sonar}, will be used because it is more sensitive to small distances in rage of 1 cm, and thus, it is very effective during landing.

\begin{figure}[!h]
      \centering
      \includegraphics[width=0.3\columnwidth, height=2cm]{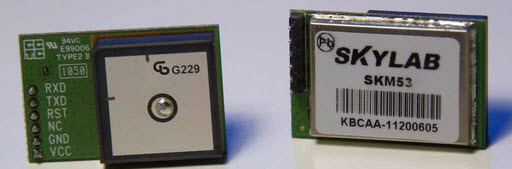}
      \caption{Skylab UART GPS Module.}
      \label{GPS}
\end{figure}

The SKM53 GPS from Skylab (see Figure \ref{GPS}) is used to determine the system's global lateral and longitudinal  position. It is a small size low weight module that comes with embedded antenna. This GPS is easy to use and connect and can be integrated with Arduino. It has output format of NMEA-0183 standard \cite{GPS_ref}.
% ===========================================
% ===========================================
% ===========================================
\section{Battery}
A recent advance in power storage technology, provide high capacity, light and robust power source that has a large market spectrum of applications, including RC aircraft. A Lithium Polymer Battery
% (see Figure \ref{battery})
is used to power both the electronics components and the motors. It is capable of providing 5200 mAh of current at 11.1 v and it has a weight of 413 g. Thus, it is capable of providing the required power to the electronic components and the motors. Moreover, IMAX B6 Balance Charger is used to charge this type of battery
 %(see Figure \ref{charger})
 \cite{battery_ref}.
% ===========================================
%\begin{figure}[!h]
%      \centering
%      \includegraphics[width=0.3\columnwidth, height=2cm]{figures/battery}
%      \caption{Lithium Polymer Battery (11.1 V, 5200 mAH),BAT-LIPO-002-1.}
%      \label{battery}
%\end{figure}
% ===========================================
%% ===========================================
%\begin{figure}[!h]
%      \centering
%      \includegraphics[width=0.3\columnwidth, height=2cm]{figures/charger}
%      \caption{IMAX B6 Balance Charger50 W, 5 A, CH-LIPO-003.}
%      \label{charger}
%\end{figure}
%% ===========================================

\chapter{\uppercase{Modeling, Identification and Control of Quadrotor}\label{ch:qudmodelidentresults}}

This chapter describes a methodology to identify all the parameters of the quadrotor constructed in the previous chapter. The identified parameters include the structure parameters and rotor assembly parameters. These parameters will be used in the system simulation and controller design later. A CAD model is developed using SOLIDWORKS to calculate the mass moments of inertia and all the missing geometrical parameters. Three simple test rigs are built and used to identify the relationship between the motor input Pulse Width Modulation (PWM) signal and the angular velocities, the thrust forces, and drag moments of the rotors. A simple algorithm is implemented to an IMU for estimating the attitude and altitude of the quadrotor. Experimental set up is built to verify and test the accuracy of these estimation and identification techniques. This is achieved by testing a controller designed based on feedback linearization method to stabilize the quadrotor attitude.

\section{Quadrotor Modeling}

The quadrotor dynamic model is presented in this section to emphasis the importance of the quadrotor parameters identification and to be used as a basis for modeling and  control synthesis of the proposed system presented latter. Figure \ref{quad_frames} presents a schematic diagram of the quadrotor system with the relevant frames, in addition to, the corresponding drag moment, force, and angular velocity of each propeller. There are some assumptions for the dynamic model. First the quadrotor structure is symmetrical and rigid. Second the propeller has a fixed pitch angle. Finally, the center of mass (CM) is coincident with the geometrical center (GC).
The equations of motion of the quadrotor were driven in \cite{Accurate_Modeling, Non_linear_Observer_design, Modelling_Ident_msc_thesis}.

The rotational kinematics of the quadrotor is represented through Euler Angles. A rigid body is completely described by its position and orientation with respect to reference frame $\{E\}$, $O_{I}$-X Y Z, that it is supposed to be earth-fixed and inertial. Let define $\eta_{1}$ as
\begin{equation}
  \eta_{1} = [X, Y, Z]^T
\end{equation}
the vector of the body position coordinates in the earth-fixed reference frame. The vector $\dot{\eta}_{1}$ is the corresponding time derivative (expressed in the earth-fixed frame). If one defines
\begin{equation}
  \nu_{1} = [u, v, w]^T
\end{equation}
as the linear velocity of the origin of the body-fixed frame $\{B\}$, $O_{B}$-x y z, whose origin is coincident with the center of mass (CM), with respect to the origin of the earth-fixed frame expressed in the body-fixed frame (from now on: body-fixed linear velocity) the following relation between the defined linear velocities holds:
\begin{equation}
\label{linear_vels}
\nu_{1} = R^{B}_{I} \dot{\eta}_{1}
\end{equation}
where $R^{B}_{I}$ is the rotation matrix expressing the transformation from the inertial frame to the body-fixed frame.

Let define $\eta_{2}$ as
\begin{equation}
  \eta_{2} =  [\phi, \theta, \psi]^T
\end{equation}
the vector of body Euler-angle coordinates in an earth-fixed reference frame. Those are commonly named roll, pitch and yaw angles and corresponds to the elementary rotation around $X$, $Y$ and $Z$ in fixed frame. The vector $\dot{\eta}_{2}$ is the corresponding time derivative (expressed in the inertial frame). Let define
\begin{equation}
  \nu_{2} =  [p, q, r]^T
\end{equation}
as the angular velocity of the body-fixed frame with respect to the earth-fixed frame expressed in the body-fixed frame (from now on: body-fixed angular velocity). The vector $\dot{\eta}_{2}$ is related to the body-fixed angular velocity by a proper Jacobian matrix:
\begin{equation}
\label{angular_vels}
\nu_{2} = J_{v} \dot{\eta}_{2}
\end{equation}
The matrix $J_{v}$ can be expressed in terms of Euler angles as:
\begin{equation}
J_{v}= \begin{bmatrix}
             1 & 0 & -S(\theta) \\
             0& C(\phi)& C(\theta) S(\phi) \\
             0& -S(\theta) & C(\theta) C(\phi) \\
           \end{bmatrix}
\label{jac}
\end{equation}
where $C(\alpha)$ and $S(\alpha)$ are short notations for $cos(\alpha)$ and $sin(\alpha)$.
The rotation $R^{B}_{I}$ matrix needed to transform the linear velocities, is expressed in terms of Euler angles by the following:
\begin{equation}
R^{B}_{I}= \begin{bmatrix}
             C(\psi) C(\theta) & S(\psi) C(\theta) & -S(\theta) \\
             -S(\psi) C(\phi)+S(\phi) S(\theta) C(\psi) & C(\psi) C(\phi)+ S(\psi) S(\theta) S(\phi) & C(\theta) S(\phi) \\
             S(\psi) S(\phi)+ C(\psi) S(\theta) C(\phi) & -C(\psi) S(\phi)+ S(\psi) S(\theta) C(\phi) & C(\theta) C(\phi) \\
           \end{bmatrix}
\label{rot_mat}
\end{equation}

From Figure\ref{quad_frames}, the angular velocity of rotor $j$, denoted with $\Omega_{j}$, creates a thrust force $F_{j}$ and drag moment $M_{j}$. Based on the momentum theory, both thrust force and drag moment are proportional to the square of the angular speed of the propeller. The consumed power of rotor $j$, $P_{j}$, is the drag moment times the angular velocity as stated in the following equations:
\begin{equation}
\label{Pi}
P_{j} = \Omega_{j} M_{j}
\end{equation}
Replacing this equation (\ref{Pi}) in (\ref{P_power}) leads to:
\begin{equation}
\label{M_moment}
M_{j} = c_{P} \frac{4\rho r^{5}}{\pi^{3}} \Omega^{2}_{j}
\end{equation}
Noting that all variables in equations (\ref{F_thrust}) and (\ref{M_moment}) are constant with exception for angular speed, propeller moment and thrust, we can rewrite these equations:
\begin{equation}
\label{Fi}
F_{j} = K_{F_{j}} \Omega_{j}^{2}
\end{equation}
%=============================================
\begin{equation}
\label{Mi}
M_{j} = K_{M_{j}} \Omega_{j}^{2}
\end{equation}
%=============================================
where $K_{M_j}$ and $K_{F_j}$ are constants that respectively relate a propeller moment and thrust with the angular speed.

The equation of motion of the quadrotor is obtained using Newton-Euler formalism.
%==================================================
\begin{equation}
\label{quad_Xdd}
m\ddot{X} = T(C(\psi) S(\theta) C(\phi) + S(\psi) S(\phi))
\end{equation}
\begin{equation}
\label{quad_Ydd}
m\ddot{Y} = T(S(\psi) S(\theta) C(\phi) - C(\psi) S(\phi))
\end{equation}
\begin{equation}
\label{quad_Zdd}
m\ddot{Z} = -mg + T C(\theta) C(\phi)
\end{equation}
\begin{equation}
\label{quad_Phdd}
I_{x}\ddot{\phi} = \dot{\theta}\dot{\phi}(I_{y}-I_{z}) - I_{r}\dot{\theta}\overline{\Omega} + \tau_{a_1}
\end{equation}
\begin{equation}
\label{quad_thdd}
I_{y}\ddot{\theta} = \dot{\psi}\dot{\phi}(I_{z}-I_{x}) + I_{r}\dot{\phi}\overline{\Omega} + \tau_{a_2}
\end{equation}
\begin{equation}
\label{quad_epdd}
I_{z}\ddot{\psi} = \dot{\theta}\dot{\phi}(I_{x}-I_{y}) + \tau_{a_3}
\end{equation}
The last three equations are derived, assuming that there are small variations in the three angles $\phi$, $\theta$ and $\psi$ such that the corresponding time derivatives of Euler angles are equivalent to the body-fixed angular velocities, i.e $J_v = I_{(3x3)}$ such that equation (\ref{angular_vels})becomes
\begin{equation}
\label{angular_vels_wout_J}
\nu_{2} = \dot{\eta}_{2} ,
\end{equation}

The variables in the above equations are defined as follows:
$m$ is the mass of the quadrotor.
$T$ is the total thrust applied to the quadrotor from all four rotors, and is given by:
\begin{equation}
T = \sum\limits_{j=1}^{4}(F_j) = \sum\limits_{j=1}^{4}(K_{F_j} \Omega_j^2)
\label{thrust_sum}
\end{equation}
$\tau_{a_1}$, $\tau_{a_2}$, and $\tau_{a_3}$  are the three input moments about the three body axes,These moments are the rolling, pitching, yawing moment about x-, y-, and z-axis of the body frame respectively, and they are given as:
\begin{equation}
\tau_{a_1} = d(F_4 - F_2) = d(K_{F_4} \Omega_4^2 - K_{F_2} \Omega_2^2)
\label{Ta1}
\end{equation}
\begin{equation}
\tau_{a_2} = d(F_3 - F_1) = d(K_{F_3} \Omega_3^2 - K_{F_1} \Omega_1^2)
\label{Ta2}
\end{equation}
\begin{equation}
\tau_{a_3} = -M_1 + M_2 - M_3 + M_4  = -K_{M_1}\Omega_1^2 + K_{M_2}\Omega_2^2 - K_{M_3}\Omega_3^2 + K_{M_4}\Omega_4^2
\label{Ta3}
\end{equation}
$d$ is the distance between the quadrotor center of mass and the rotation axis of the propeller.
\begin{equation}
\overline{\Omega} = \Omega_1 - \Omega_2 + \Omega_3 - \Omega_4
\label{omega_bar}
\end{equation}
$I_r$ is the rotor inertia. $I_f$ is the inertia matrix of the vehicle around its body-frame assuming that the vehicle is symmetric about x-, y- and z-axis and it is given by:
\begin{equation}
I_f= \begin{bmatrix}
             I_x & 0 & 0 \\
             0 & I_y & 0 \\
             0& 0 & I_z \\
           \end{bmatrix}
\label{quad_inertia}
\end{equation}

The parameters presented in the previous equations are anonymous. They must be identified.

\section{Quadrotor Structure Parameters}
\begin{figure}[!h]
      \centering
      \includegraphics[width=0.8\columnwidth, height=8cm]{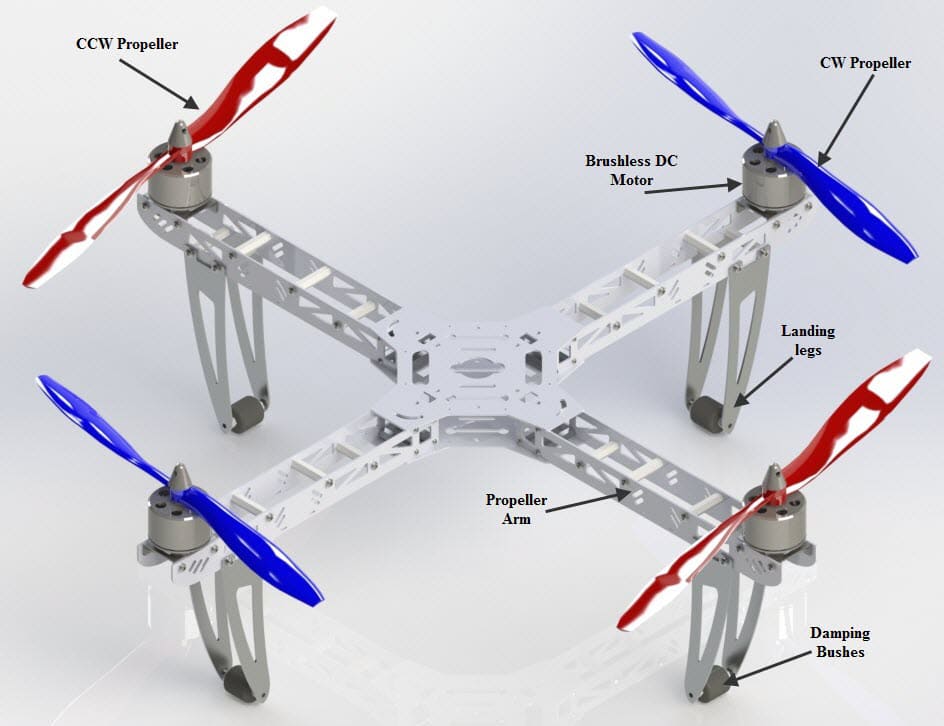}
      \caption{Quadrotor CAD Model.}
      \label{quad_CAD}
\end{figure}

A CAD model, as shown in Figure \ref{quad_CAD}, is developed using SOLIDWORKS by modeling all the parts of the quadrotor. The modeled parts are motors, electronic parts, battery and the aluminum frames, which are assembled together. The rotor arms are manufactured from aluminum sheet 1 mm thickness. These arms are engraved to lighten the total weight and decrease the resistance of aerodynamics during flying. All these holes complicate the calculations of the mass moment of inertia using the principal laws. So, the mass moments of inertia of the quadrotor structure and rotors are extracted directly from the CAD model.

The inertia matrix obtained from the CAD model is diagonal and positive definite. Table \ref{quad_struct_par} presents the mass moment of inertia about x-, y- and z-axis of the body frame, total mass, the mass moment of inertia of the rotor $I_r$ and center distance $d$ between the rotor axis and the quadrotor center. The mass of the quadrotor is also measured using a scale and found to be enough close to the man obtained from the CAD model.
% =======================================
% ===========================================
\begin{table}[!h]
\caption{Quadrotor Structure Parameters}
\label{quad_struct_par}
\begin{center}
\setlength{\tabcolsep}{4pt}
\begin{tabular}{|c||c||c|}
\hline
\hline
Parameter	&Value	&	Unit \\
\hline
$I_x$ & $13.215 X 10^{-3}$ & $kg.m^2$\\
\hline
$I_y$ & $12.522 X 10^{-3}$ & $kg.m^2$\\
\hline
$I_z$ & $23.527 X 10^{-3}$ & $kg.m^2$\\
\hline
$I_r$ & $33.216 X 10^{-6}$ & $kg.m^2$\\
\hline
$d$ & $223.5 X 10^{-3}$ & $m$\\
\hline
$m$ & $0.952$ & $kg$\\
\hline
\hline
\end{tabular}
\end{center}
\end{table}
\section{Rotor Assembly Identification}
The rotor assembly (ESC, motor, and propeller) is the most important part which delivers the lifting force that permits the quadrotor to fly. This assembly is consists of an ESC, brushless DC motor and propeller that has two blades. To identify the rotor assembly, it is needed to find the relationship between input and output of each motor. Motor voltage is the input of the rotor assembly. However, PWM is used as input for its simplicity and it can be directly programmed using  Microcontroller Unit (MCU). Outputs of rotor assembly are angular velocity, thrust force and drag moment.
\subsection{Angular Speed Identification}
The problem is that no direct relationship between the motor input signal and the propeller angular speed exists. The input signal is PWM which is generated precisely using Arduino Mega 2560 MCU. PMW signal in this case has a limited boundary from $1000\mu s$ to $2000\mu s$. Motors have no response in case of PWM values smaller than the lower limit. On the other hand, a saturation phenomenon occurs for values larger than the upper limit of PWM. An optical tachometer device is required to measure the output angular velocity of the propeller in order to establish a direct relationship between PWM, and the angular velocity. However, this instrument is expensive. In addition, It use may decrease the accuracy due to its interaction with the light structure of the rotor assembly.

\begin{figure}[!h]
      \centering
      \includegraphics[width=0.8\columnwidth, height=7cm]{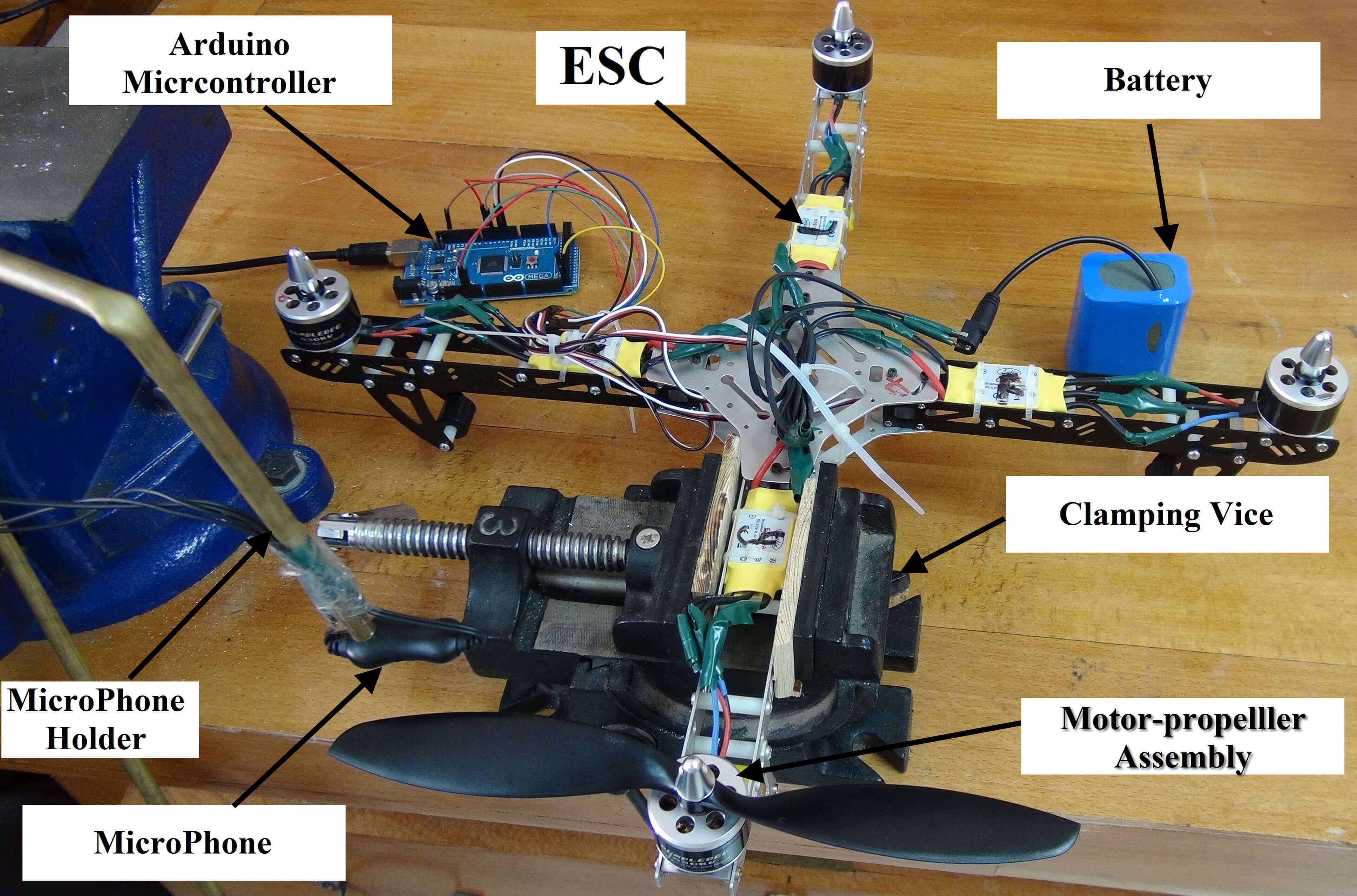}
      \caption{Test Rig for Identifying Rotor Angular Speed.}
      \label{rotorangvel}
\end{figure}

A solution for this problem is developed by using a simple microphone placed 2cm over the tip of the propeller to work as a rudimentary tachometer. Figure \ref{rotorangvel} shows a clamping vice used to clamp the quadrotor during the test. Microphone holder is used to ensure that the microphone position over the propeller tip is high enough to avoid hitting by the propeller.
% =======================================
% ===========================================

The concept of using microphone as a rudimentary tachometer is simple. At each time the propeller pass under the microphone the air between the propeller and microphone is sucked downward. In this case the microphone records the suction pulses. The microphone readings can be captured in real time using MATLAB. This data is processed to obtain the number of pulse-pairs per unit time which indicate the angular velocity accurately. The experiments are made as follows. First, the readings are captured after 30 s from starting the motor rotation, which make the motor reach to the steady state condition. Then suction pulses are recorded for 5 s. This process is repeated three times for each PWM signal. Finally; these values are averaged to determine the angular velocity for each motor at different PWM inputs.

\begin{figure}[!h]
      \centering
      \includegraphics[width=0.8\columnwidth, height=7cm]{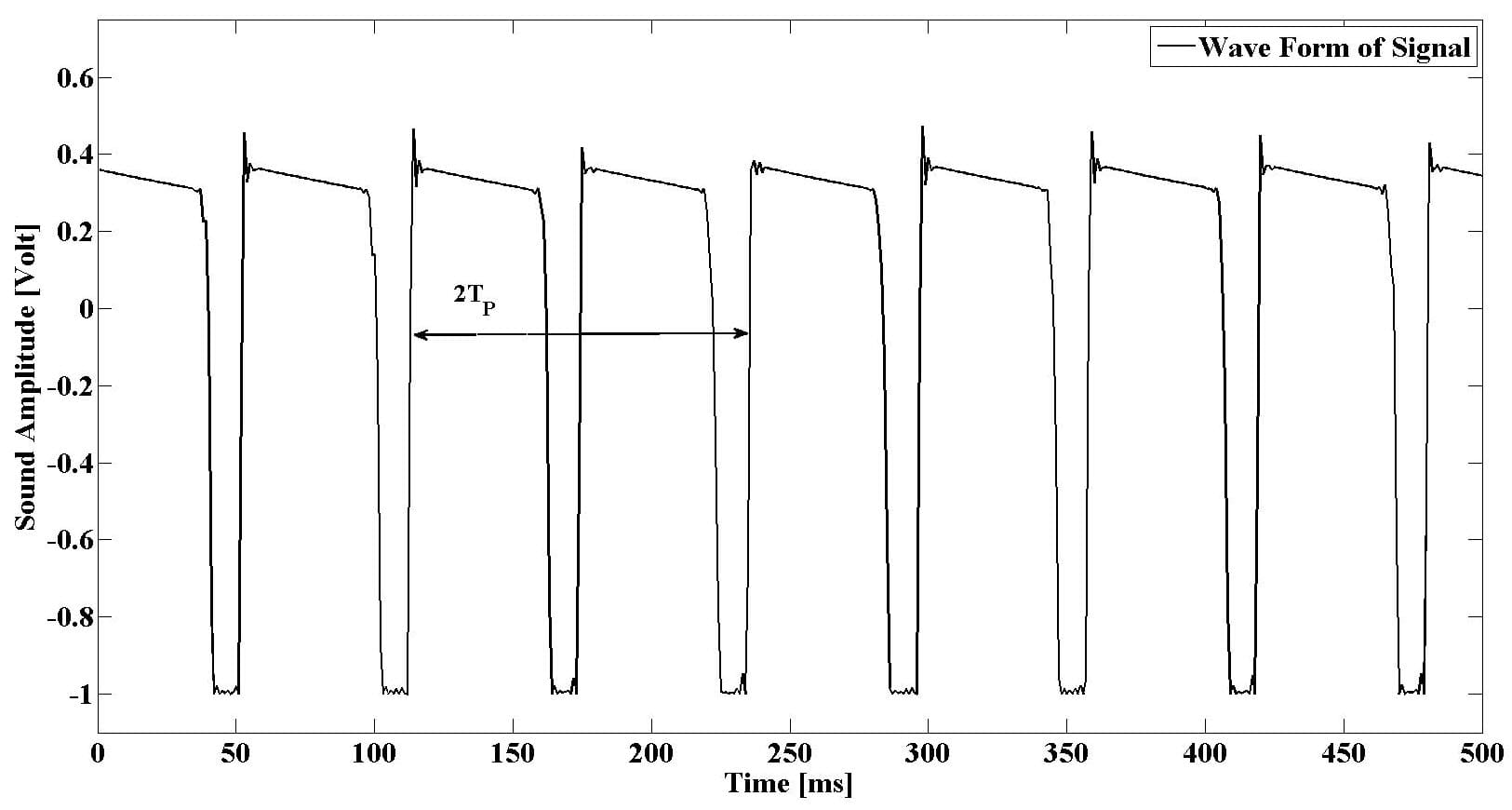}
      \caption{Propeller Sound  Signal.}
      \label{soundsignal}
\end{figure}

\begin{figure}[!h]
      \centering
      \includegraphics[width=0.8\columnwidth, height=7cm]{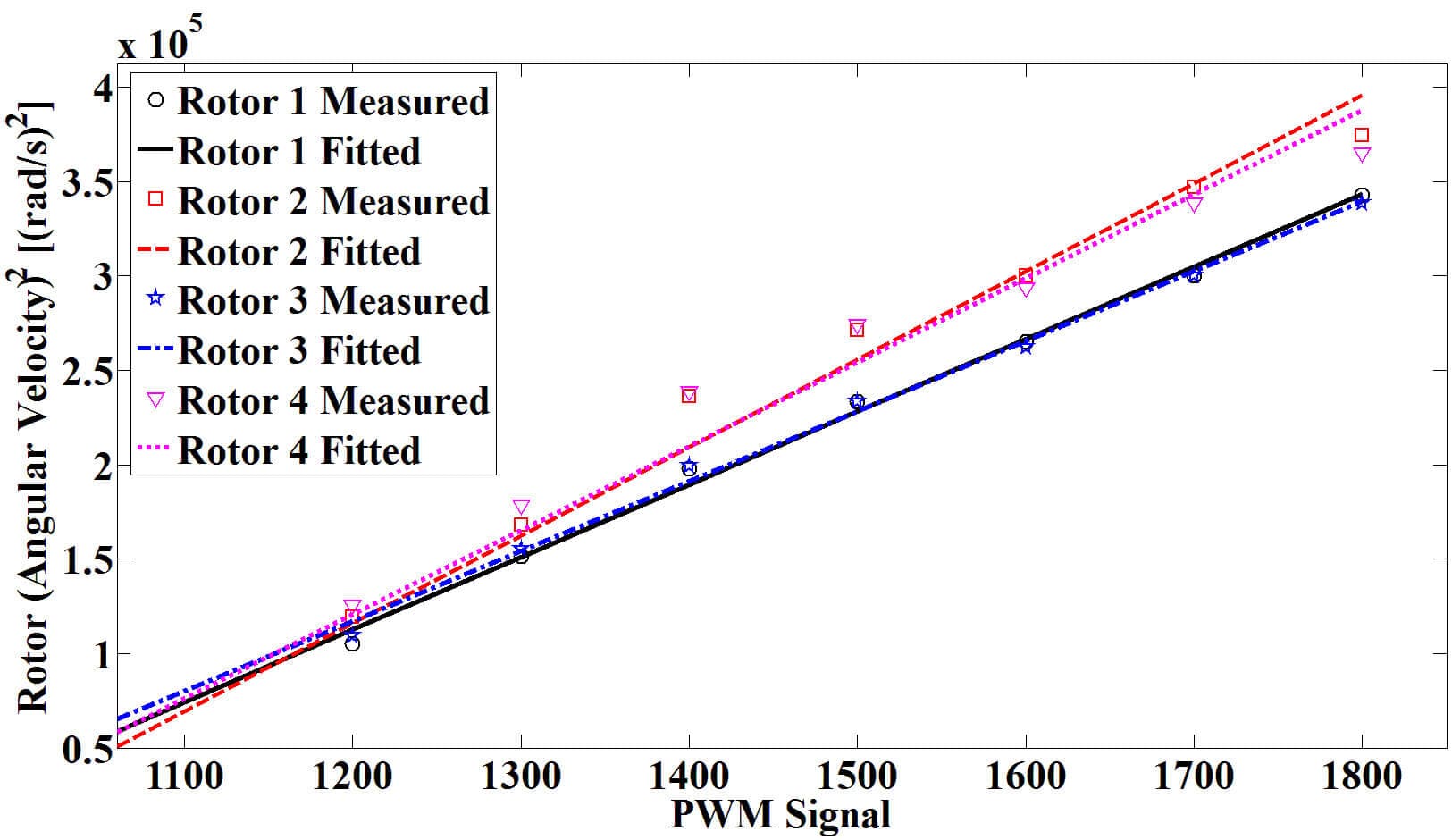}
      \caption{Relationship Between motor PWM and rotor angular velocity.}
      \label{pwmvsw}
\end{figure}

The rotor angular speed is calculated by measuring the period of time between each passage of the propeller blades $T_p$:
\begin{equation}
\Omega = \frac{\pi}{T_p}
\label{w}
\end{equation}
Figure \ref{soundsignal} shows a sample signal from the propeller sound. It can be noticed that each two pulses can indicate a complete turn for the propeller. One can notice from Figure \ref{soundsignal} that the ratio between noise and pressure signal is very low. So, it is possible to identify the sound pressure waves clearly due to the rotation of the propeller. The input/output data is gathered for one motor. This data is used to determine the relationship between PWM and rotor's angular speed. All the rotors' obtained data for the 4 rotors are plotted in Figure \ref{pwmvsw}. It is found that there is a linear relationship between the PWM, $u$, and the squared angular velocity as stated in (\ref{pwm_wi}). It is observed from Figure \ref{pwmvsw} that the four rotors are not identical in their angular velocities at the same value of PWM.
% =======================================
% ===========================================
% =======================================
% ===========================================
\begin{equation}
\Omega_j^2 = a_j u_j + b_j
\label{pwm_wi}
\end{equation}

After finishing the identification; constants for the four motors are obtained. Table \ref{pwmvsangvel} presents the values of these constants which satisfy the linear relationship (\ref{pwm_wi}). It is easy now to obtain precise values for angular velocity to help in building a robust controller.
% ===========================================
\begin{table}
\caption{Curve Fitting Parameters for PWM V.S. Angular Velocity}
\label{pwmvsangvel}
\begin{center}
\setlength{\tabcolsep}{4pt}
\begin{tabular}{|c||c||c||c||c|}
\hline
\hline
Parameter &Rotor 1& Rotor 2 &Rotor 3 &Rotor 4 \\
\hline
$a_j$ & $420.5$ & $466$ & $411.4$ & $445$\\
\hline
$b_j$ & $-4.06 X 10^{5}$ & $-4.43 X 10^{5}$ & $-3.92 X 10^{5}$ & $-4.13 X 10^{5}$ \\
\hline
\hline
\end{tabular}
\end{center}
\end{table}
\subsection{Thrust Force and Drag Moment Identifications}

In this section, a new test rig is developed to measure the thrust force and drag moment simultaneously. The test rig consists of a lever arm of 50 cm length and 25X25X1 mm hollow square cross section. The arm is pivoted at its center point by steel pin of 10 mm diameter. Rotor is mounted on one end of the lever arm using 4 screw bolts, M2.5. On the other end, electronic balance is mounted under the lever arm. This lever mechanism is clamped from the pivot pin to a clamping vice to prevent it from flying.

\begin{figure}[!h]
      \centering
      \includegraphics[width=0.8\columnwidth, height=7cm]{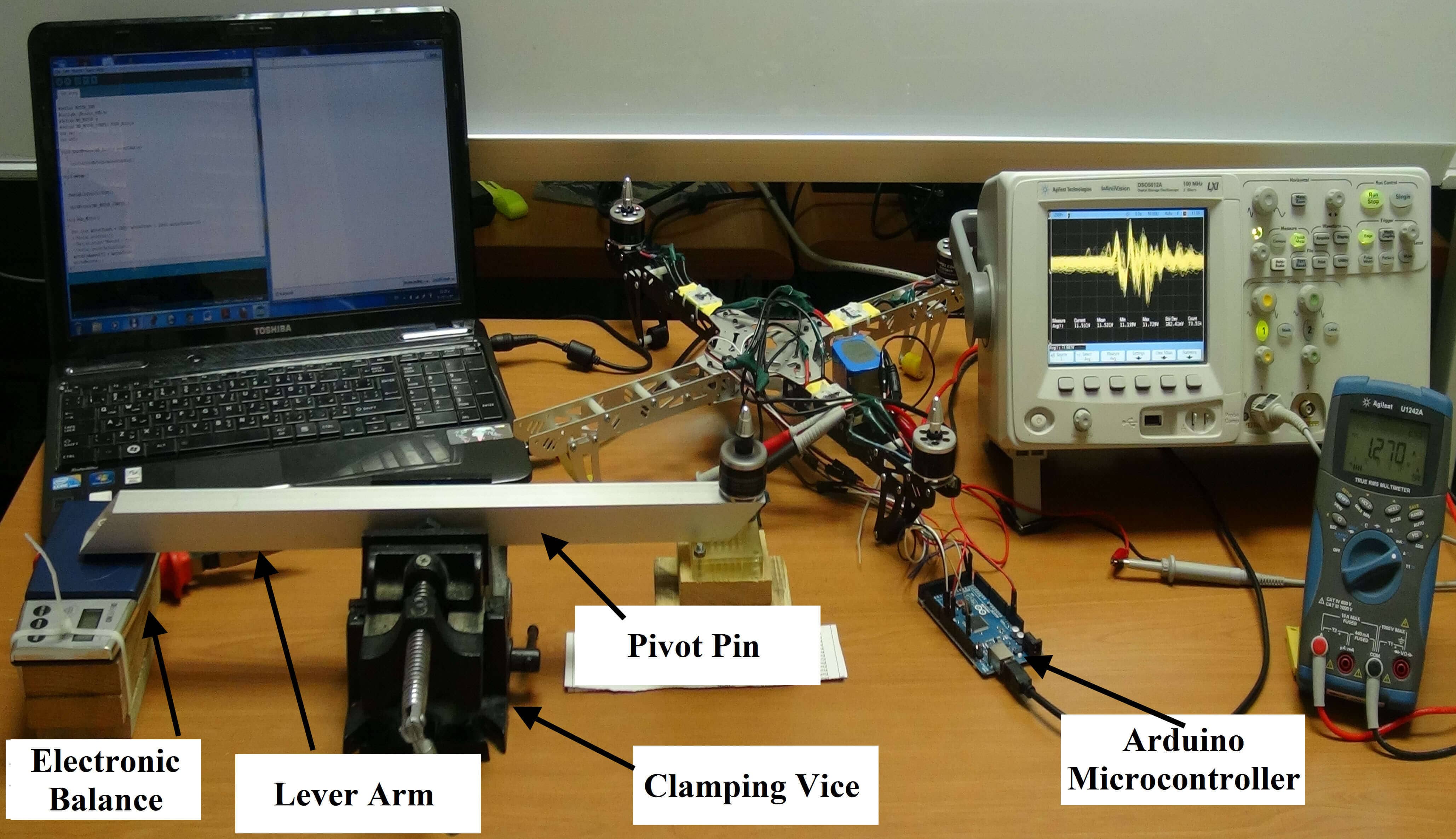}
      \caption{Test Rig for Identifying Thrust Force and Drag Moment.}
      \label{thrustdragexper}
\end{figure}

\begin{figure}[!h]
      \centering
      \includegraphics[width=0.7\columnwidth, height=7cm]{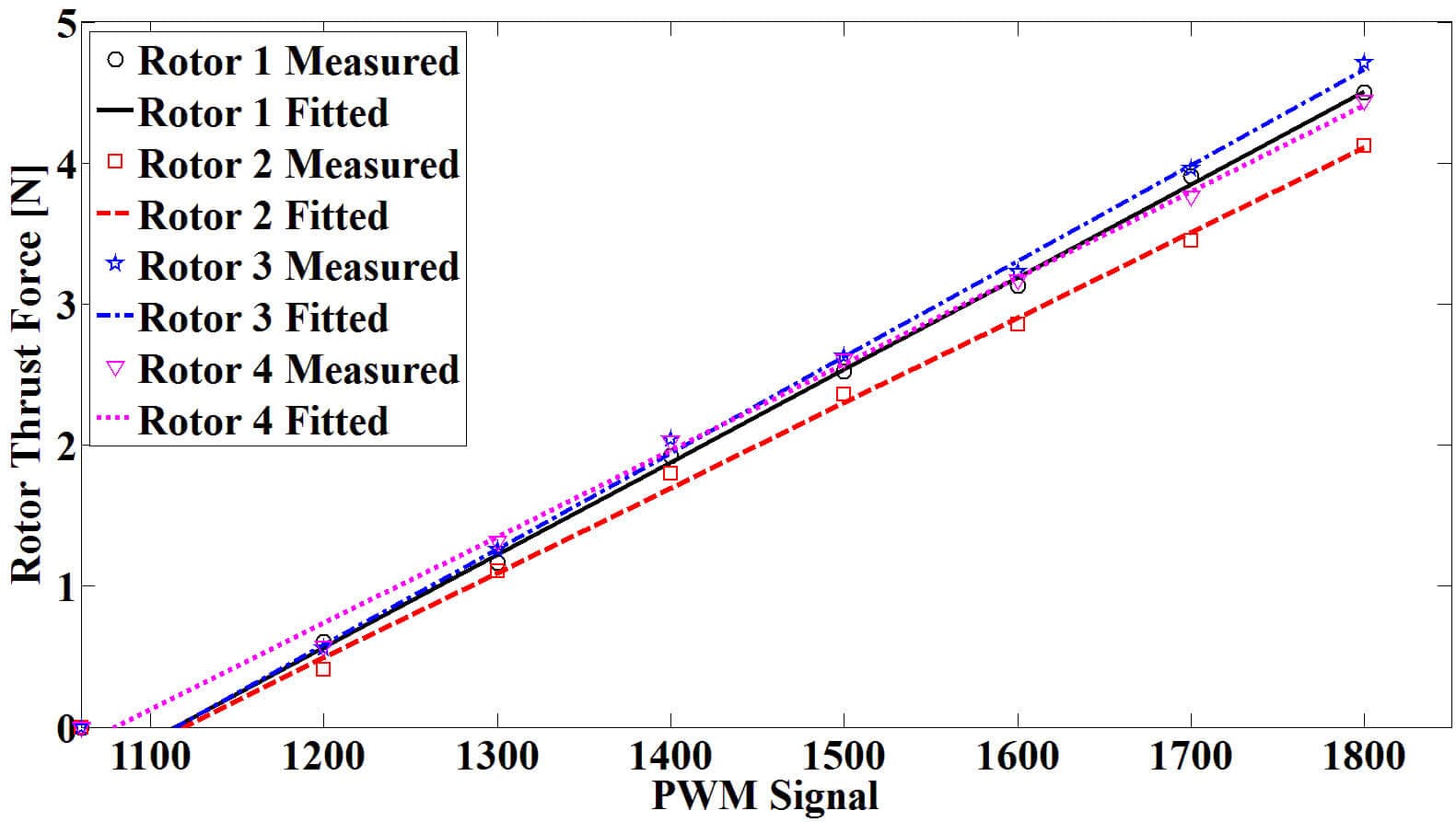}
      \caption{Relationship between Motor PWM and Rotor Thrust Force.}
      \label{pwmvsthrust}
\end{figure}

\begin{figure}[!h]
      \centering
      \includegraphics[width=0.7\columnwidth, height=7cm]{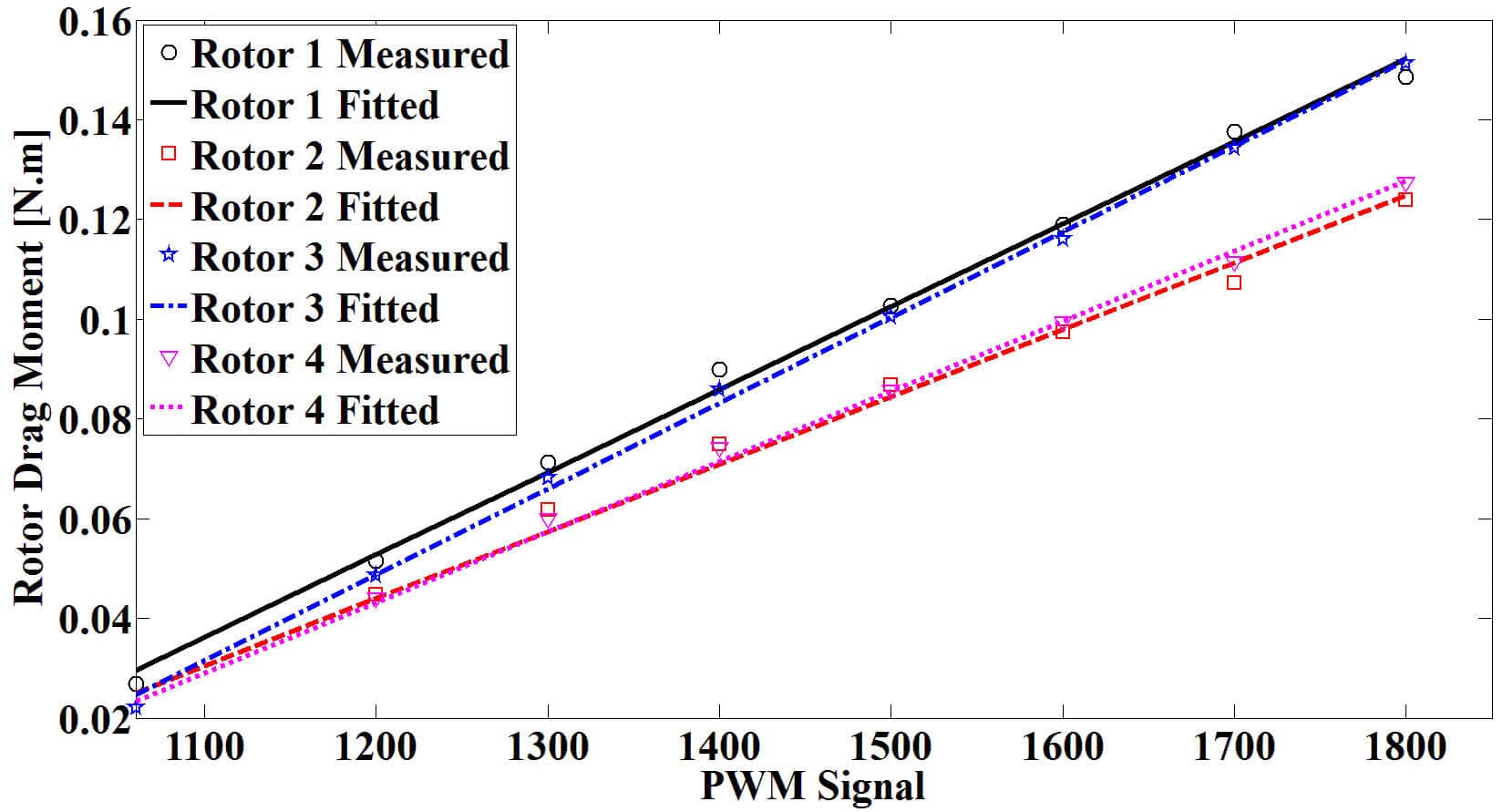}
      \caption{Relationship between Motor PWM and Rotor Drag Moment.}
      \label{pwmvsmoment}
\end{figure}

The concept of this lever mechanism is that the propeller rotation produces a vertical upward thrust force. This force tends to rotate the lever arm about its pivot pin, and hence generates an equivalent downward force at the other end of the arm. This force can be measured using an electronic balance. The thrust force at certain motor speed is equal to the difference between the balance reading at this motor speed and the balance reading at zero motor speed. On the same time, the drawn current and voltage are measured as shown in Figure \ref{thrustdragexper}. The consumed power by the motor is then calculated, and from (\ref{Pi}) the drag moment is determined.
% =======================================
% ===========================================
The tests are carried out by varying the PWM from 1060 $\mu s$ to 1800 $\mu s$, to obtain the measurements of thrust, drawn current and voltage. Gathering data for all motors is made by repeating the previous procedures. This data is plotted then fitted. It is observed that there are linear relationships between PWM and both thrust force and drag moment as shown in Figure \ref{pwmvsthrust} and Figure \ref{pwmvsmoment} respectively. TABLE \ref{pwmvsthrustmoment} presents the values of the constants in equations (\ref{F_j}, \ref{M_j}) which express these linear relationships. It can be noticed from Figure \ref{pwmvsthrust}, and Figure \ref{pwmvsmoment} that the motors are not identical, and having variance in the thrust force and the drag moment coefficients.
% ===========================================
\begin{equation}
F_j = c_j u_j + d_j
\label{F_j}
\end{equation}
% ===========================================
\begin{equation}
M_j = e_j u_j + h_j
\label{M_j}
\end{equation}
% =======================================
% ===========================================
% ===========================================
\begin{table}[!h]
\caption{Curve Fitting Parameters for PWM V.S. Thrust Force and Drag Moment}
\label{pwmvsthrustmoment}
\begin{center}
\setlength{\tabcolsep}{4pt}
\begin{tabular}{|c||c||c||c||c|}
\hline
\hline
Parameter &Rotor 1& Rotor 2 &Rotor 3 &Rotor 4 \\
\hline
$c_j$ & $0.6566$ & $0.6029$ & $0.6805$ & $0.6119$\\
\hline
$d_j$ & $-731.4$ & $-674.4$ & $-758.3$ & $-660.5$ \\
\hline
$e_j$ & $0.0001658$ & $0.0001348$ & $0.000172$ & $0.000141$\\
\hline
$h_j$ & $-0.1462$ & $-0.1178$ & $-0.1577$ & $-0.126$\\
\hline
\hline
\end{tabular}
\end{center}
\end{table}
% ===========================================

Transformation from the motors signals $u_1$, $u_2$, $u_3$ and $u_4$ to system thrust and moments $T$, $\tau_{a_1}$, $\tau_{a_2}$ and $\tau_{a_3}$ is derived by substituting in (\ref{thrust_sum},\ref{Ta1},\ref{Ta2} and \ref{Ta3}) by (\ref{F_j}, \ref{M_j}). It is as following:
%========================================
\begin{equation}
T = c_1 u_1 + c_2 u_2 + c_3 u_3 + c_4 u_4 + d_1 + d_2 + d_3 + d_4
\label{T_pwm}
\end{equation}
% =======================================
\begin{equation}
\tau_{a_1} = d c_4 u_4 - d c_2 u_2 + d(d_4 - d_2)
\label{Ta1_pwm}
\end{equation}
% =======================================
\begin{equation}
\tau_{a_2} = d c_3 u_3 - d c_1 u_1 + d(d_3 - d_1)
\label{Ta2_pwm}
\end{equation}
%========================================
\begin{equation}
\tau_{a_3} = -e_1 u_1 + e_2 u_2 - e_3 u_3 + e_4 u_4 - h_1 + h_2 - h_3 + h_4
\label{Ta3_pwm}
\end{equation}
% =======================================

Equations (\ref{T_pwm}-\ref{Ta3_pwm}) can be put in a matrix form as following:
\begin{equation}
\underbrace{\begin{bmatrix}
  T \\
  \tau_{a_1} \\
  \tau_{a_2}  \\
  \tau_{a_3}  \\
\end{bmatrix}
}_{U}
=
\underbrace{ \begin{bmatrix}
  c_1 & c_2 & c_3 & c_4 \\
  0 & -dc_2 & 0 & dc_4 \\
  -dc_1 & 0 & dc_3 & 0 \\
  -e_1 & e_2 & -e_3 & e_4 \\
\end{bmatrix}
}_{G}
\underbrace{\begin{bmatrix}
  u_{1} \\
  u_{2} \\
  u_{3} \\
  u_{4} \\
\end{bmatrix}
}_{u_{j}}
+
\underbrace{\begin{bmatrix}
  d_1 + d_2 + d_3 + d_4 \\
  d(d_4 - d_2) \\
  d(d_3 - d_1) \\
  -h_1 + h_2 - h_3 + h_4 \\
\end{bmatrix}
}_{A}
\label{pwm_controlsignal}
\end{equation}
and thus,
%========================================
\begin{equation}
[u_i] = G^{-1}\{[U] - [A]\}
\label{ui_consig}
\end{equation}
% =======================================

Now, after the identification, the direct relationships between the motors input signals, and the quadrotor input thrust and moments are clearly derived. The previous equations make the rotor control and its implementation into an embedded system easier which helps in achieving robustness.

For system simulation purposes, determination of $K_{F_j}$ and $K_{M_j}$ parameters must be done. From the data of $F_j$, $M_j$ and $\Omega_{j}^{2}$, these parameters can be estimated and they are given in Table \ref{vales_KF_KM}.
% ===========================================
\begin{table}[!h]
\caption{Rotor Assembly Parameters ($K_{F_j}$ and $K_{M_j}$)}
\label{vales_KF_KM}
\begin{center}
\setlength{\tabcolsep}{4pt}
\begin{tabular}{|c||c||c||c||c||c|}
\hline
\hline
Parameter &Rotor 1& Rotor 2 &Rotor 3 &Rotor 4& Unit \\
\hline
$K_{F_j}$ & $1.667x10^{-5}$ & $1.285x10^{-5}$ & $1.711x10^{-5}$ & $1.556x10^{-5}$ & $kg.m.rad^{-2}$\\
\hline
$K_{M_j}$ & $3.965x10^{-7}$ & $2.847x10^{-7}$ & $4.404x10^{-7}$ & $3.170x10^{-7}$ & $kg.m^{2}.rad^{-2}$ \\
\hline
\hline
\end{tabular}
\end{center}
\end{table}
% ===========================================
\section{IMU Sensor Fusion}
Sensor fusion \cite{sensor_fuse} is a term used to combine the data of different types of sensors to enhance the accuracy of the measurements, and decrease the level of noise as much as possible.

\begin{figure}[!h]
      \centering
      \includegraphics[width=0.6\columnwidth, height=5cm]{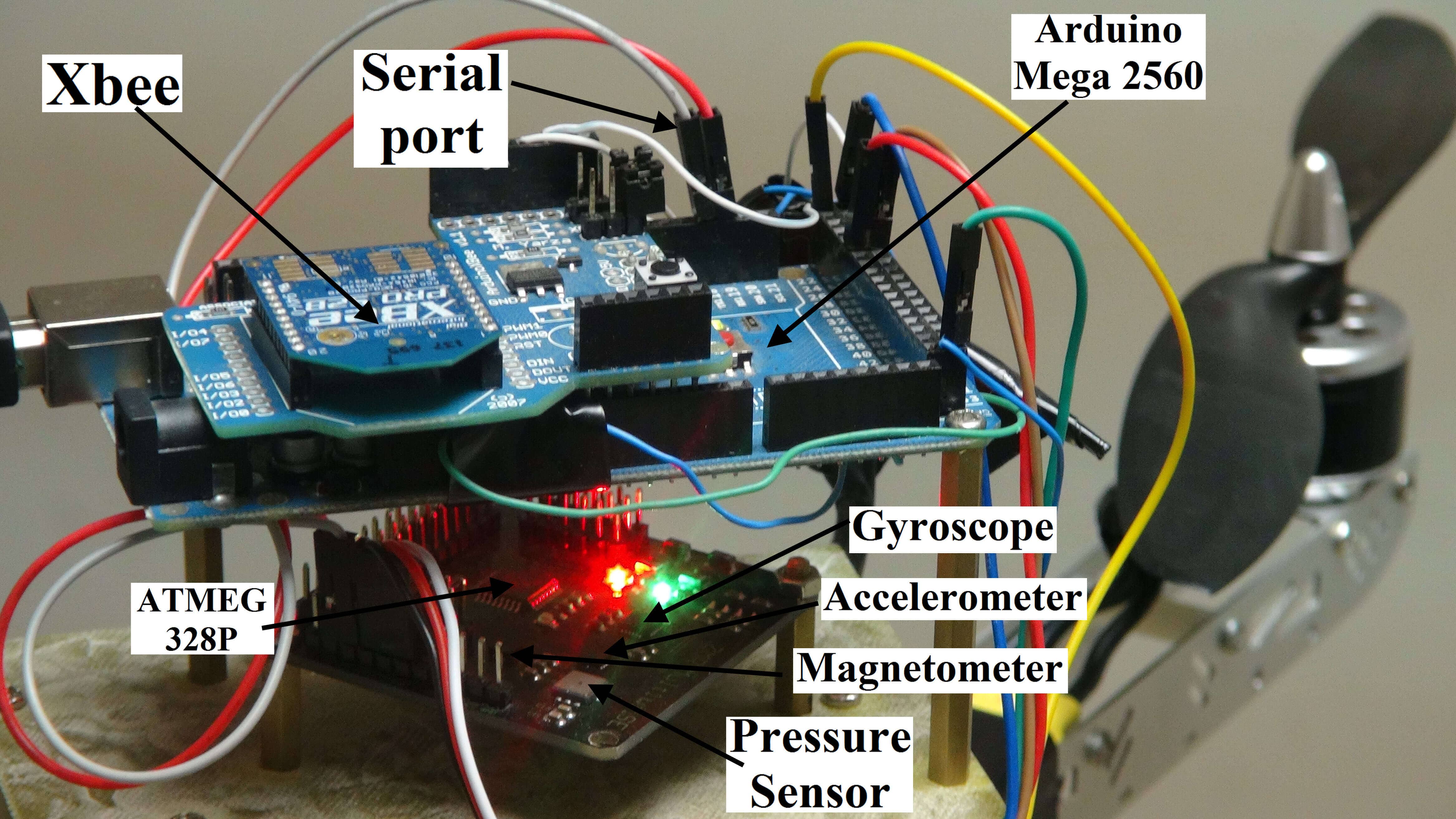}
      \caption{IMU and Arduino board connection.}
      \label{practiimu}
\end{figure}

A Multiwii ZMR board type is used as an IMU which consists of high quality MEMS sensors. The board containing a three axes gyroscope (ITG3205), a three axes accelerometer (BMA180), a three axes magnetometer (HMC5883L), and embedded pressure sensor (BMP085). All these sensors are connected to ATMEGA 328P microcontroller on the same board as shown in Figure \ref{practiimu}. This MCU handles all the readings from all sensors through an $I^{2}C$ bus. IMU sensor fusion is done to combine all the sensors readings to estimate the orientation of the quadrotor namely roll, pitch, and yaw angles as well as their rate. Multiwii ZMR board in connected via serial port to Arduino Mega 2560 MCU, which has the main control algorithm.

\begin{figure}[!h]
      \centering
      \includegraphics[width=0.7\columnwidth, height=19cm]{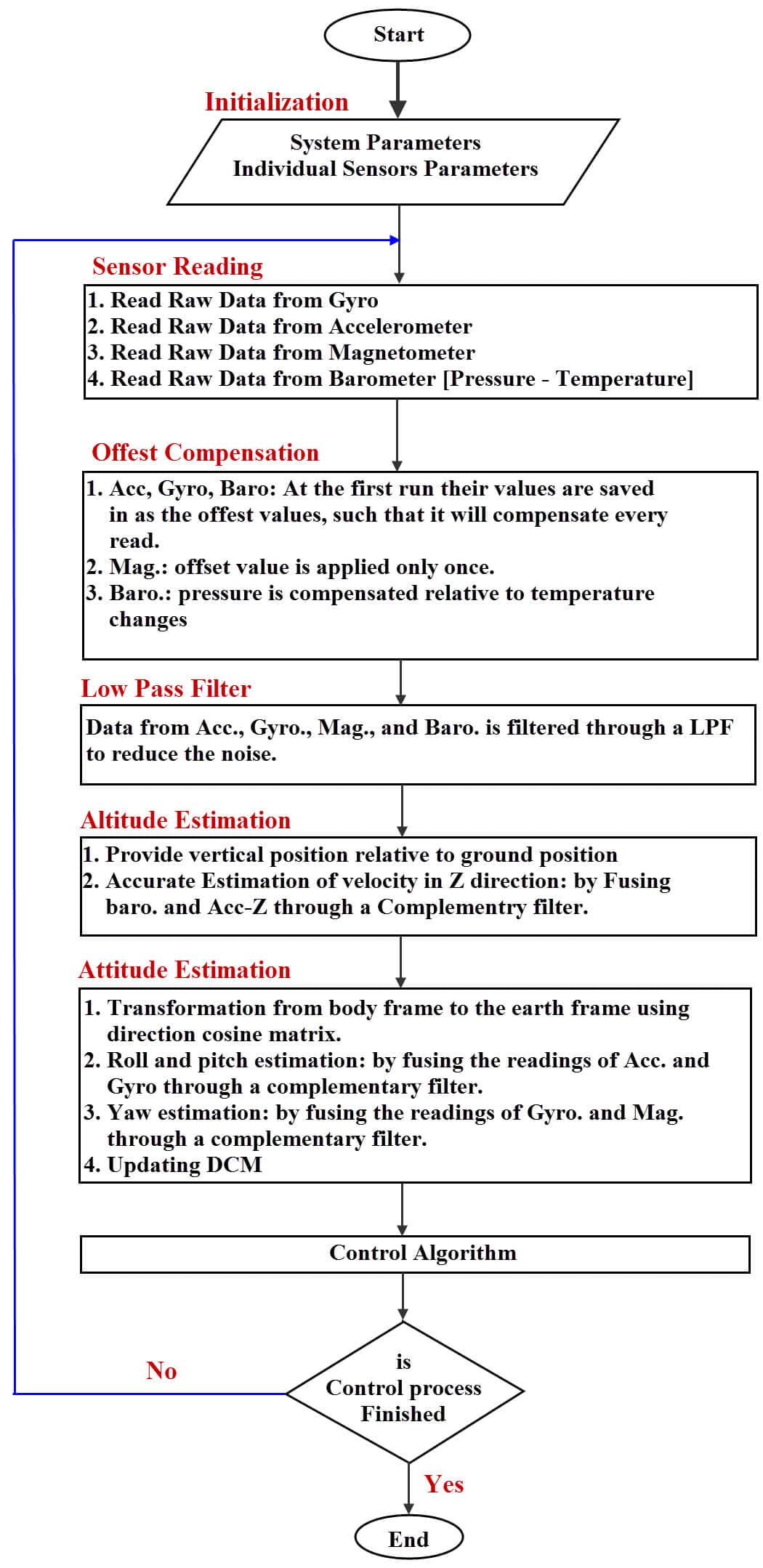}
      \caption{Flowchart for Estimating Attitude of Quadrotor.}
      \label{sensorfusionflowchart}
\end{figure}

The main objective is to estimate the Euler angles of the quadrotor, in order to use these angles as feedback signals for quadrotor attitude stabilization. These angles cannot be estimated using a single sensor, because each sensor has its own problem. The disadvantage of accelerometer is its mechanical vibration, while drift is the disadvantage of gyroscope. Direction cosine matrix (DCM) complimentary filter algorithm is used to estimate the quadrotor attitude \cite{planar_low_cost_IMU}. DCM \cite{DCM_IMU} is used to transform from the body frame to the earth frame. Then a complimentary filter \cite{comp_filter} is used to estimate the roll, and pitch angles based on fusing gyroscope and accelerometer data together. The usage of gyroscope is to make fine tuning for the DCM matrix which is returned by the accelerometer. Another complementary filter is used to estimate the yaw angle by fusing the gyroscope and magnetometer data as shown in Figure \ref{sensorfusionflowchart}. Gyroscope is used with magnetometer to enhance the yaw angle determination relative to the earth's magnetic north. Finally the altitude is estimated using the pressure sensor \cite{applied_robotics, imu_theory, imu_guide, quad_sensorplatform,DCM_Comp_Filt}.
% =======================================
% ===========================================
% ===========================================
\section{Testing and Results}
After identifying all the quadrotor parameters and constructing the attitude estimation algorithm, their verification and testing are desired. So, this section describes a control system design based on the technique of feedback linearization which is used to test the attitude stabilization of the quadrotor using the identified parameters.

Feedback linearization \cite{applied_nonlinear} is used to transform the nonlinear system dynamics into a linear system. The control laws are chosen so that we can obtain a good tracking performance. The controllers' laws for the attitude namely roll, pitch, and yaw angles are stated as follows:
%========================================
\begin{equation}
\tau_{a_1} = I_x u_{\phi} - \dot{\theta} \dot{\psi} (I_y - I_z) + I_r \dot{\theta} \overline{\Omega}
\label{FBL_Ta1}
\end{equation}
% =======================================
\begin{equation}
u_{\phi} = \ddot{\phi}_{d} + K_{p_\phi} e_{\phi} + K_{d_\phi} \dot{e}_{\phi}+ K_{i_\phi}\int\limits_0^t{e_{\phi} dt}
\label{FBL_uph}
\end{equation}
% =======================================
\begin{equation}
\tau_{a_2} = I_y u_{\theta} - \dot{\phi} \dot{\psi} (I_z - I_x) - I_r \dot{\phi} \overline{\Omega}
\label{FBL_Ta2}
\end{equation}
% =======================================
\begin{equation}
u_{\theta} = \ddot{\theta}_{d} + K_{p_\theta} e_{\theta} + K_{d_\theta} \dot{e}_{\theta}+ K_{i_\theta}\int\limits_0^t{e_{\theta} dt}
\label{FBL_uth}
\end{equation}
% =======================================
\begin{equation}
\tau_{a_3} = I_z u_{\psi} - \dot{\phi} \dot{\theta} (I_x - I_y)
\label{FBL_Ta3}
\end{equation}
% =======================================
\begin{equation}
u_{\psi} = \ddot{\psi}_{d} + K_{p_\psi} e_{\psi} + K_{d_\psi} \dot{e}_{\psi}+ K_{i_\psi}\int\limits_0^t{e_{\psi} dt}
\label{FBL_uep}
\end{equation}
% =======================================
Where $K_p$, $K_d$ and $K_i$ are the proportional, differential and integral gains respectively. These gains are tuned manually in order to obtain a satisfied controller performance. Table \ref{PIDparquad} presents the used values of these parameters.
% ===========================================
\begin{table}[!h]
\caption{PID Controllers' Parameters for quadrotor }
\label{PIDparquad}
\begin{center}
\setlength{\tabcolsep}{4pt}
\begin{tabular}{|c||c||c||c|}
\hline
\hline
Parameter &$\phi$& $\theta$ &$\psi$ \\
\hline
$K_p$ & $100$ & $85$ & $0.01$\\
\hline
$K_i$ & $10$ & $10$ & $250$\\
\hline
$K_d$ & $1$ & $1$ & $260$\\
\hline
\hline
\end{tabular}
\end{center}
\end{table}
% ===========================================

\begin{figure}[!h]
      \centering
      \includegraphics[width=0.6\columnwidth, height=6cm]{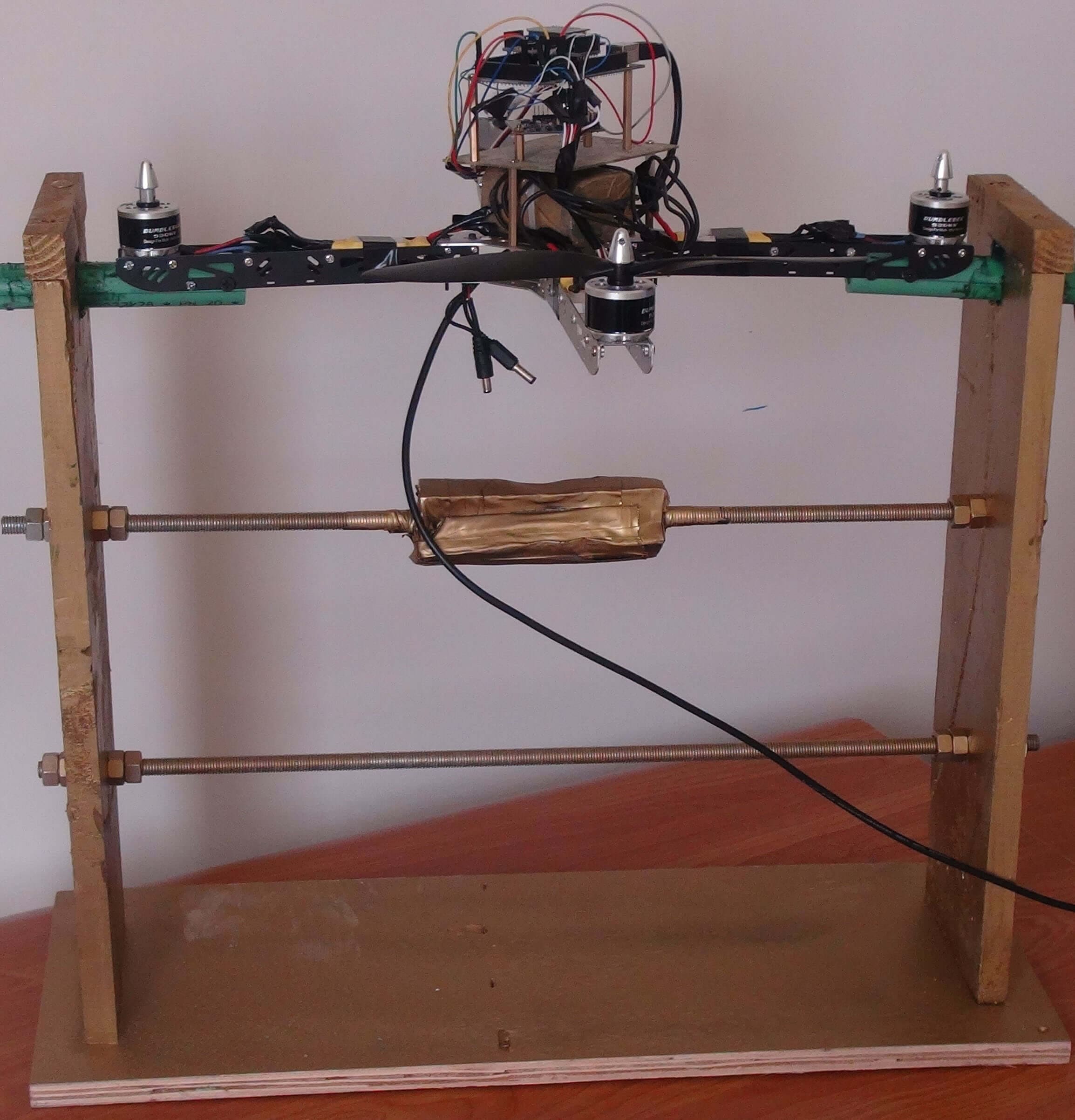}
      \caption{Roll and Pitch Angle Test Rig.}
      \label{phthtestrig}
\end{figure}

\begin{figure}[!h]
      \centering
      \subfigure[$\phi$]{\centering \label{quadphexp}\includegraphics[width=0.65\columnwidth, height=6cm]
      {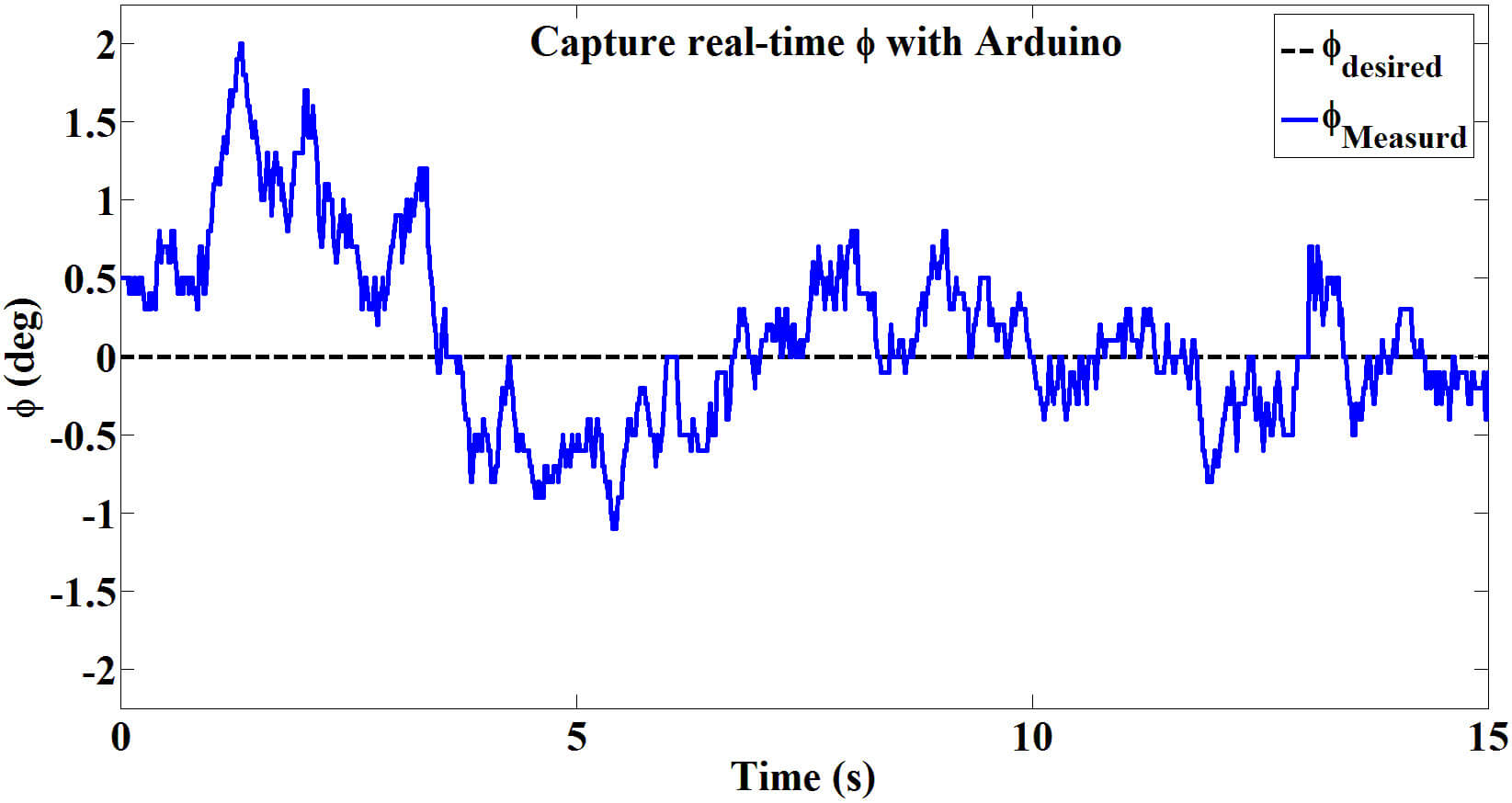}}\hspace{2cm}
      \subfigure[$\theta$]{\centering \label{quadthetaexp} \includegraphics[width=0.65\columnwidth, height=6cm]{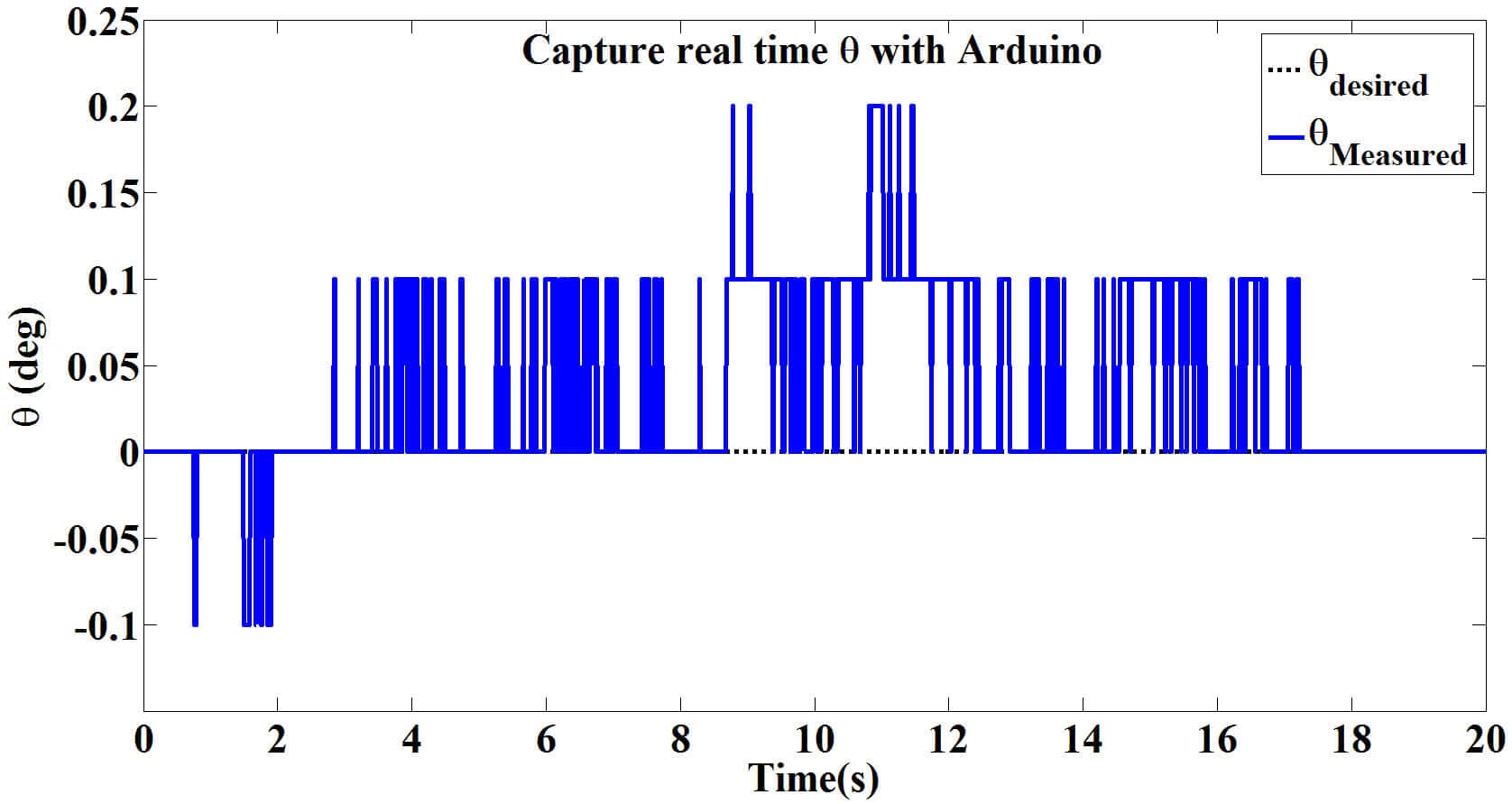}}
      \caption{Experimental Results for Testing Roll and Pitch Angles: a) $\phi$ and b) $\theta$.}
      \label{quadrollpitch}
\end{figure}

\begin{figure}[!h]
      \centering
      \includegraphics[width=0.65\columnwidth, height=6cm]{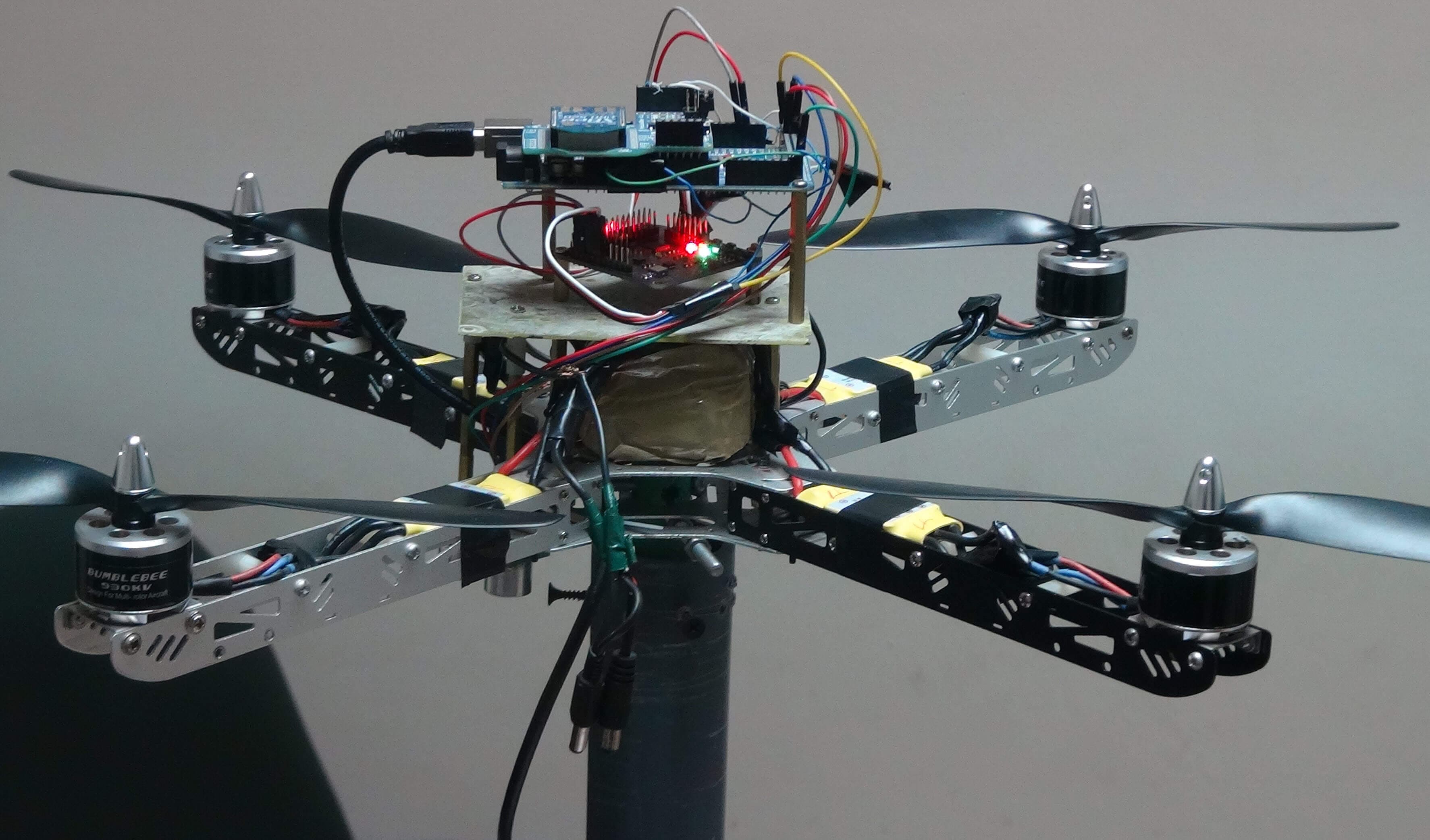}
      \caption{Yaw Angle Test Rig.}
      \label{eptestrig}
\end{figure}

\begin{figure}[!h]
      \centering
      \includegraphics[width=0.65\columnwidth, height=6cm]{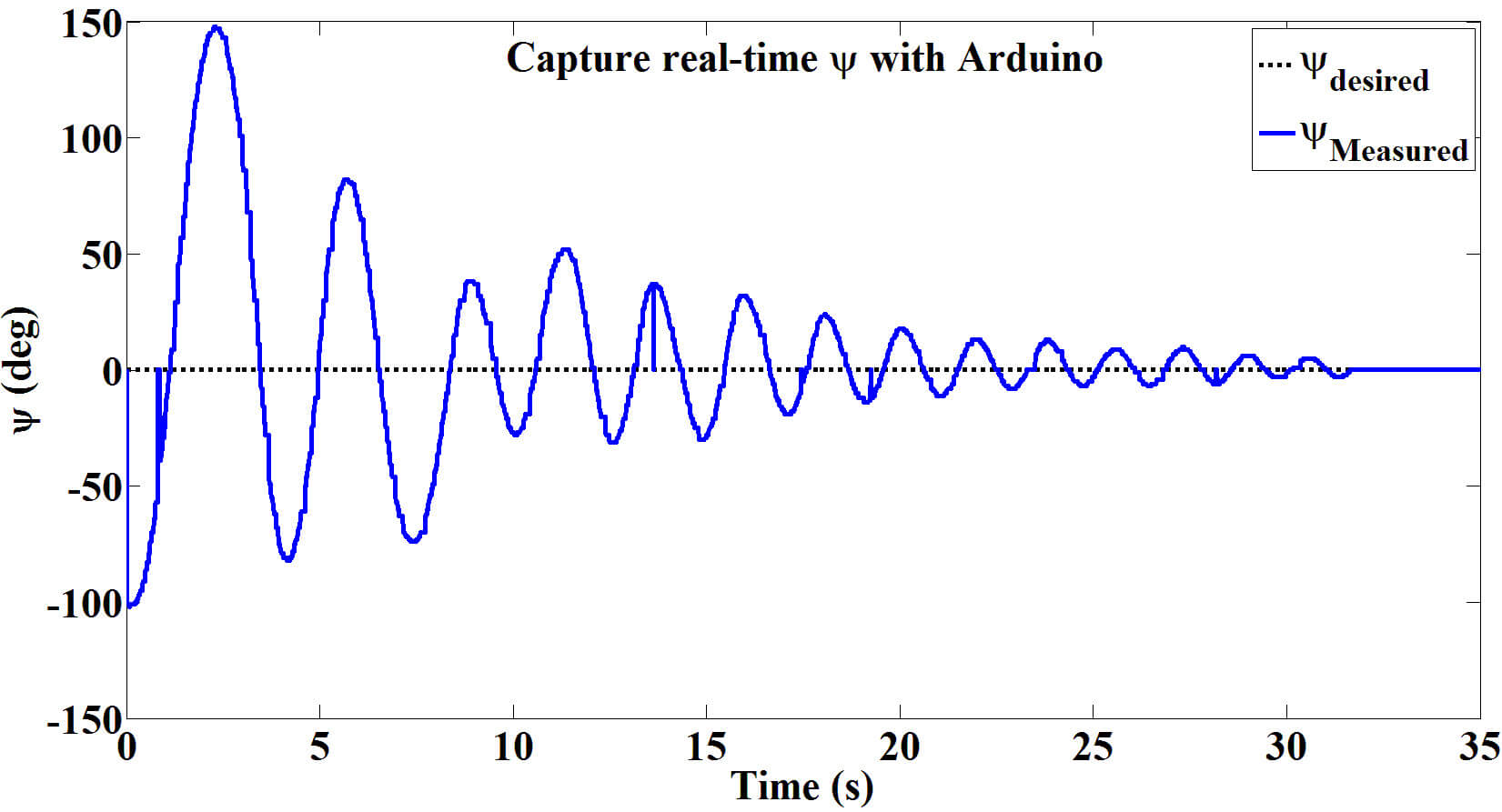}
      \caption{Experimental Result for Testing Yaw Angle.}
      \label{quadepexp}
\end{figure}

Two different test rigs are constructed to check the identified parameters of the quadrotor and attitude estimation. Figure \ref{phthtestrig} presents the test rig which is used to perform stabilizing control of roll, and pitch angles. This rig is a wooden bracket used to support the quadrotor. It contains two holes to mount two cylindrical supports. Then the quadrotor is fixed to these supports. These supports allow the rotation about x-axis or y-axis. The controller which presented in (\ref{FBL_Ta1} - \ref{FBL_uep}) has been implemented. These controllers are executed on Arduino Mega 2560 MCU, with a sampling time of 1.8ms. Arduino Mega 2560 is connected to MATLAB in order to record the measured data from the test rig.
% ===========================================
% ===========================================
Figure \ref{quadrollpitch} presents roll, and pitch controllers' responses. These controllers achieve the stability of the quadrotor, and rotate the quadrotor successfully to the desired roll and pitch angles. For the roll angle a small oscillation happens that ranges from $0.5^{o}$ to $-0.5^{o}$ as shown in Figure \ref{quadphexp}. This oscillation is satisfactory in the flying and it can be reduced by fine tuning of the controller parameters. Figure \ref{quadthetaexp} shows the response of the quadrotor to the pitch angle controller. One can notice that a very small oscillation occurs. The range of this oscillation is between $0.2^{o}$ and $-0.1^{o}$. It is accepted for flying and hovering.
% ===========================================
% ===========================================
The second test rig is designed to control the yaw angle. This rig consists of two cylindrical tubes. One of them is fixed to the ground using clamping vice. The other one is fixed to the quadrotor. A ball bearing is fitted between these two tubes to facilitate the rotational motion about z-axis as shown in Figure \ref{eptestrig}. The experimental result for the yaw angle is presented in Figure \ref{quadepexp}. In this figure, one can see that the controller achieves the stability of the yaw angle although the system starts from large initial conditions.

Finally the experiments show that the quadrotor identified parameters are accurate in such way that the feedback linearization algorithm stabilizes the quadrotor in an efficient manner.
% ===========================================
% ===========================================
% ===========================================
% ===========================================
%\section{conclusion}
%In this paper a complete methodology for identifying the quadrotor parameters is presented in details. Quadrotor dynamic model is described. A CAD model was developed to calculate the mass moment of inertia in an accurate way. Direct relationships between PWM and each of the angular speed, thrust force, and drag moment of the rotors are identified. A new relationship that directly linking the quadrotor control input (forces and moments) with the motor control PWM signal is achieved and tested. A DCM complementary filter is used to estimate the attitude of the quadrotor based on the IMU measurements. Controller is designed based on the feedback linearization technique to test the identified parameters and the attitude estimation. Finally, the result of the experiments shows a satisfied accuracy of the proposed techniques of the identified structure parameters, the identified rotor assembly parameters, the used IMU with attitude estimation algorithm, and the control design for quadrotor attitude stabilization.

\chapter{\uppercase{Quadrotor manipulation system Modeling}\label{ch:modeling}}
\begin{figure}[!h]
      \centering
      \includegraphics[width=0.8\columnwidth, height=8cm]{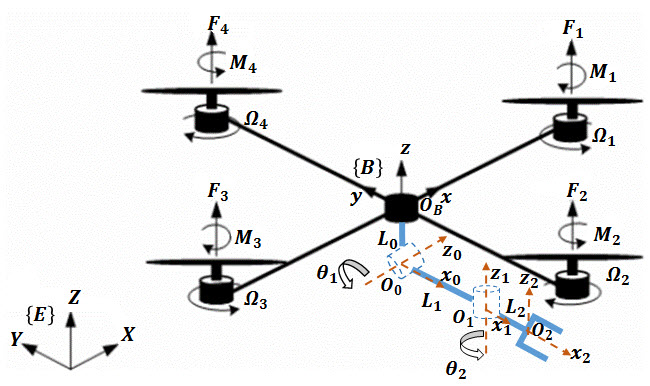}
      %\framebox{\parbox{1cm}{(1.a)}}
      \caption{Schematic of Quadrotor Manipulation System Frames}
      \label{frames}
\end{figure}

The proposed system consists of  two-link manipulator attached to the bottom of a quadrotor. The manipulator has two revolute joints. The axis of joint 1 ($z_{0}$ in Figure \ref{frames}) is parallel to one in-plane axis of the quadrotor ($x$ in Figure \ref{frames}) and perpendicular to the axis of joint 2. Also, the axis of joint 2 ($z_{1}$ in Figure \ref{frames}) is parallel to the other in-plane axis of the quadrotor ($y$ in Figure \ref{frames}) at extended configuration. So, the pitch and roll orientation of the end effector is now possible independently on the longitudinal and lateral motion of the quadrotor. Therefore, the end effector can perform any arbitrary position and orientation, and hence, a 6–DOF aerial manipulator is obtained.
% =======================================
% ===========================================

The kinematic and dynamic analysis of the quadrotor were presented in Chapter \ref{ch:qudmodelidentresults}.
\section{System Kinematics}
Figure \ref{frames} presents a sketch of the Quadrotor-Manipulator System with the relevant frames. The frames are assumed to satisfy the Denavit-Hartenberg(DH) convention \cite{Spong}. Table \ref{D_H} presents DH parameters for the 2-Link Manipulator.
%================================================
\begin{table}[!h]
\caption{DH Parameters for the Manipulator}
\label{D_H}
\begin{center}
\begin{tabular}{|c||c||c||c||c|}
\hline
\hline
Link(i) & $d_{i}$ & $a_{i}$ & $\alpha_{i}$ & $\theta_{i}$ \\
\hline
0	& $-L_{0}$ & 0  & $-\pi/2$ & $-\pi/2$ \\
\hline
1	& 0 & $L_{1}$  & $\pi/2$ & $\theta_{1}$ \\
\hline
2	& 0 & $L_{2}$ & 0 & $\theta_{2}$ \\
\hline
\hline
\end{tabular}
\end{center}
\end{table}
%===================================================
The position and orientation of the end effector relative to the body-fixed frame is easily obtained by multiplying the following homogeneous transformation matrices $A^{B}_{0}$, $A^{0}_{1}$, $A^{1}_{2}$ :
\begin{equation}
A^{B}_{0}= \begin{bmatrix}
             0 & 0 & 1 & 0 \\
             -1 & 0 & 0 & 0 \\
             0 & -1 & 0 & -L_{0} \\
             0 & 0 & 0 & 1 \\
           \end{bmatrix}
\label{AB_0}
\end{equation}
\begin{equation}
A^{0}_{1}= \begin{bmatrix}
             C(\theta_{1}) & 0 & S(\theta_{1}) & L_{1} C(\theta_{1}) \\
             S(\theta_{1}) & 0 & -C(\theta_{1}) & L_{1} S(\theta_{1})\\
             0 & 1 & 0 & 0\\
             0 & 0 & 0 & 1 \\
           \end{bmatrix}
\label{A0_1}
\end{equation}
\begin{equation}
A^{1}_{2}= \begin{bmatrix}
             C(\theta_{2}) & -S(\theta_{2}) & 0 & L_{2} C(\theta_{2}) \\
             S(\theta_{2}) & C(\theta_{2}) & 0 & L_{2} S(\theta_{2})\\
             0 & 0 & 1 & 0\\
             0 & 0 & 0 & 1 \\
           \end{bmatrix}
\label{A0_1}
\end{equation}
where $\theta_{1}$ and $\theta_{2}$ are the manipulator joints' angles.
\subsection{Forward Kinematics}
Let define the position and orientation of the end effector expressed in the inertial frame, as $\eta_{ee_1}$ and $\eta_{ee_2}$ respectively.
\begin{equation}
\eta_{ee_1} = [x_{ee},y_{ee},z_{ee}]^T
\label{etae1}
\end{equation}
\begin{equation}
\eta_{ee_2} = [\phi_{ee},\theta_{ee},\psi_{ee}]^T
\label{etae2}
\end{equation}
The  forward  kinematics  problem  consists  of  determining the operational coordinates ($\eta_{ee_1}$ and $\eta_{ee_2}$) of the end effector, as a function of the quadrotor movements ($X$, $Y$, $Z$, and $\psi$) as well as the motion of the manipulator's joints ($\theta_{1}$ and $\theta_{2}$). This problem is solved by computing the homogeneous transformation matrix composed of relative translations and rotations.

The transformation matrix from the body frame to the inertial frame $A^{I}_{B}$ which is:
\begin{equation}
A^{I}_{B} = R^{I}_{B}*transl(X,Y,Z)
\label{AIB}
\end{equation}
where $R^{I}_{B}$ is 4x4 matrix , and $transl(X,Y,Z)$ is 4x4 matrix that describes the translation of $X$, $Y$ and $Z$ in the inertial coordinates.
%
%From (\ref{AIB}) this matrix is :
%\begin{equation}
%A^{I}_{B}= \begin{bmatrix}
%             C(\psi) & -S(\psi) & 0 & X \\
%             S(\psi) & C(\psi) & 0 & Y \\
%             0 & 0 & 1 & Z \\
%             0 & 0 & 0 & 1 \\
%           \end{bmatrix}
%\label{AIB_detail}
%\end{equation}

The total transformation matrix that relates the end effector frame to the inertial frame is $T^{I}_{2}$, which is given by:
\begin{equation}
T^{I}_{2} = A^{I}_{B} A^{B}_{0} A^{0}_{1} A^{1}_{2}
\label{T2IB}
\end{equation}
Define the general form for this transformation matrix as a function of end effector variables($\eta_{ee_1}$ and $\eta_{ee_2}$), as following:
\begin{equation}
T_{ee}= \begin{bmatrix}
             r_{11} & r_{12} & r_{13} & x_{ee}\\
             r_{21} & r_{22}& r_{23} & y_{ee} \\
             r_{31} & r_{32} & r_{33} & z_{ee} \\
             0 & 0 & 0 & 1 \\
           \end{bmatrix}
\label{T_e}
\end{equation}
Equating (\ref{T2IB}) and (\ref{T_e}), an expression for the parameters of $T_{ee}$ ($r_{ij}$, $x_{ee}$, $y_{ee}$, and $z_{ee}$; $i,j=1, 2, 3$) can be found, from which values of the end effector variables can determined.
Euler angles of the end effector ($\phi_{ee}$, $\theta_{ee}$ and $\psi_{ee}$) can be computed from the rotation matrix of $T_{ee}$ as in \cite{euler_2_rot}.
\subsection{Inverse Kinematics}
The inverse kinematics problem consists of determining the quadrotor movements ($X$, $Y$, $Z$, and $\psi$) as well as the motion of the manipulator's joints ($\theta_{1}$ and $\theta_{2}$) as function of operational coordinates ($\eta_{ee_1}$ and $\eta_{ee_2}$) of the end effector.

The inverse kinematics solution is essential for the robot's control, since it allows to  compute the required quadrotor movements and manipulator joints angles to move the end effector to a desired position and orientation.

The rotations of the end effector can be parameterized by using several methods one of them, that is chosen, is the euler angles \cite{euler_2_rot}.

Equation (\ref{T2IB}) can be expressed, after putting $\phi$ = $\theta$ = 0 since we apply inverse kinematics for reset position, as following:
\begin{equation}
\resizebox{1\hsize}{!}{$
T^{I}_{2}= \begin{bmatrix}
C(\psi)S(\theta_2)+C(\theta_1)C(\theta_2)S(\psi)&C(\psi)C(\theta_2)-C(\theta_1)S(\psi)S(\theta_2)&S(\psi)S(\theta_1)&X+L_1 C(\theta_1)S(\psi)+L_2 C(\psi)S(\theta_2)+L_2 C(\theta_1)C(\theta_2)S(\psi)\\
S(\psi)S(\theta_2)-C(\psi)C(\theta_1)C(\theta_2)&C(\theta_2)S(\psi)+C(\psi)C(\theta_1)S(\theta_2)&-C(\psi)S(\theta_1)&Y- L_1 C(\psi)C(\theta_1)+L_2 S(\psi)S(\theta_2)-L_2 C(\psi)C(\theta_1)C(\theta_2)\\
-C(\theta_2)S(\theta_1)&S(\theta_1)S(\theta_2)&C(\theta_1)&Z-L_0 -L_1 S(\theta_1)-L_2 C(\theta_2)S(\theta_1)\\
0& 0&0&  1\\
\end{bmatrix}
$}
\label{T2IB_detail}
\end{equation}
From (\ref{T2IB_detail}) and (\ref{T_e}), the inverse kinematics of the system can be derived. According to the structure of (\ref{T2IB_detail}), the inverse orientation is carried out first followed by inverse position. The inverse orientation has three cases as following:

\uppercase{\textbf{Case 1}}:

Suppose that not both of $r_{13}$ , $r_{23}$ are zero. Then from (\ref{T2IB_detail}), we deduce that $sin(\theta_1)$ $\neq$ 0 and $r_{33}$ $\neq$ $\pm$1. In the same time, $cos(\theta_1)$ = $r_{33}$ and $sin(\theta_1)$ = $\pm$ $\sqrt{1-r^{2}_{33}}$ and thus,
\begin{equation}
\theta_1 = atan2(\sqrt{1-r^{2}_{33}},r_{33})
\label{theta_1_1_inv_kin}
\end{equation}
or
\begin{equation}
\theta_1 = atan2(-\sqrt{1-r^{2}_{33}},r_{33})
\label{theta_1_2_inv_kin}
\end{equation}
where the function atan2(.) is defined as:
The function $\alpha$ = $atan2(yy, xx)$  computes the arc tangent function, where $xx$ and
$yy$ are the cosine and sine, respectively, of the angle $\alpha$. This function uses the signs of $xx$ and $yy$ to select the appropriate quadrant for the angle $\alpha$.

If we choose the value for $\theta_1$ given by Equation (\ref{theta_1_1_inv_kin}), then $sin(\theta_1)> 0$, and
\begin{equation}
\psi = atan2(r_{13},-r_{23})
\label{psi_1_inv_kin}
\end{equation}
\begin{equation}
\theta_2 = atan2(r_{32},-r_{31})
\label{theta_2_1_inv_kin}
\end{equation}

If we choose the value for $\theta_1$ given by Equation (\ref{theta_1_2_inv_kin}), then $sin(\theta_1)< 0$, and
\begin{equation}
\psi = atan2(-r_{13},r_{23})
\label{psi_2_inv_kin}
\end{equation}
\begin{equation}
\theta_2 = atan2(-r_{32},r_{31})
\label{theta_2_2_inv_kin}
\end{equation}
Thus there are two solutions depending on the sign chosen for $\theta_1$. If $r_{13}$ = $r_{23}$ = $0$, then the fact that $T_{ee}$ is orthogonal implies that $r_{33}$ = $\pm$1.

\uppercase{\textbf{Case 2}}:

If $r_{13}$ = $r_{23}$ = $0$ and $r_{33}$ = 1, then $cos(\theta_1)$ = 1 and $sin(\theta_1)$ = 0, so that $\theta_1$ = 0. In this case, the rotation matrix of (\ref{T2IB_detail})becomes
\begin{equation}
R^{I}_{2}= \begin{bmatrix}
S(\theta_2+\psi)&C(\theta_2+\psi)&0\\
-C(\theta_2+\psi)&S(\theta_2+\psi)&0\\
0&0&1\\
\end{bmatrix}
\label{R2IB_detail_theta_1_0}
\end{equation}
Thus the sum $\theta_2+\psi$ can be determined as
\begin{equation}
\theta_2+\psi = atan2(r_{11},r_{12})
\label{theta2psi_inv_kin}
\end{equation}
We can assume any value for $\psi$ and get $\theta_2$. Therefor, there are infinity of solutions.

\uppercase{\textbf{Case 3}}:

If $r_{13}$ = $r_{23}$ = $0$ and $r_{33}$ = -1, then $cos(\theta_1)$ = -1 and $sin(\theta_1)$ = 0, so that $\theta_1$ = $\pi$. In this case, the rotation matrix of (\ref{T2IB_detail}) becomes:
\begin{equation}
R^{I}_{2}= \begin{bmatrix}
S(\theta_2-\psi)&C(\theta_2-\psi)&0\\
C(\theta_2-\psi)&-S(\theta_2-\psi)&0\\
0&0&-1\\
\end{bmatrix}
\label{R2IB_detail_theta_1_1}
\end{equation}

Thus, $\theta_2-\psi$ can be determined as
\begin{equation}
\theta_2-\psi = atan2(r_{11},r_{12})
\label{theta2psi_inv_kin2}
\end{equation}

We can assume any value for $\psi$ and get $\theta_2$. Therefor, there are infinity of solutions.

In cases 2 and 3, putting $\psi$ will lead to find $\theta_2$.

Finally, the inverse position is determined from:
\begin{equation}
X = x_{ee}-(L_1 C(\theta_1)S(\psi)+L_2 C(\psi)S(\theta_2)+L_2 C(\theta_1)C(\theta_2)S(\psi))
\label{X_inv_kin2}
\end{equation}
\begin{equation}
Y = y_{ee}-(-L_1 C(\psi)C(\theta_1)+L_2 S(\psi)S(\theta_2)-L_2 C(\psi)C(\theta_1)C(\theta_2))
\label{Y_inv_kin2}
\end{equation}
\begin{equation}
Z = z_{ee}-(Z-L_0 -L_1 S(\theta_1)-L_2 C(\theta_2)S(\theta_1))
\label{Z_inv_kin2}
\end{equation}

\section{System Dynamics}

\begin{figure}[!h]
      \centering
      \includegraphics[width=0.7\columnwidth, height=8cm]{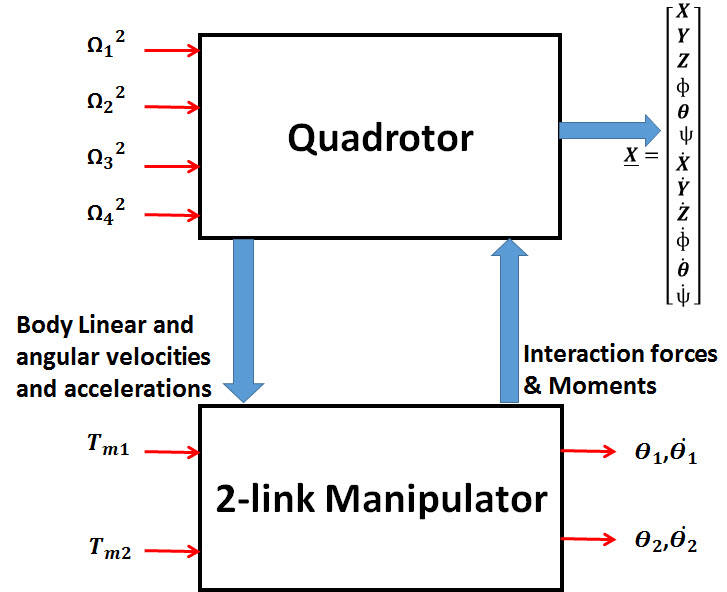}
      \caption{Effects of Adding a Manipulator to the Quadrotor}
      \label{effects_Qaud_MAnp}
   \end{figure}

In Figure \ref{effects_Qaud_MAnp}, a block diagram that shows the effects of adding a manipulator to a quadrotor is presented.

For the manipulator dynamics, Recursive Newton Euler method \cite{Tsai} is used to derive the equations of motion. Since the quadrotor is considered to be the base of the manipulator, the initial linear and angular velocities and accelerations, used in Newton Euler algorithm, are that of the quadrotor expressed in body frame. Applying the Newton Euler algorithm to the manipulator considering that the link (with length $L_{0}$) that is fixed to the quadrotor is the base link, manipulator's equations of motion can be obtained, in addition to, the forces and moments, from manipulator, that affect the quadrotor.

Let us define for each link $i$, the following variables:

$\omega_{i}^{i}$, angular velocity of frame $i$ expressed in frame $i$, $\dot{\omega}_{i}^{i}$, angular acceleration of frame $i$, $v_{i}^{i}$, linear velocity of the origin of frame $i$, $\dot{v}^{i}_{c_i}$, linear acceleration of the center of mass of link $i$, $\dot{v}_{i}^{i}$, linear acceleration of the origin of frame $i$, $r_{i}^{i}$ , the vector from the origin of frame $(i-1)$ to the origin of link $i$, $r_{c_i}^{i}$, the vector from the origin of frame $(i-1)$ to the center of mass of link $i$, $g^{I}$, the vector of gravity expressed in inertial frame $I$, $z_{(i-1)}^{(i-1)}$, is a unit vector pointing along the $i^{th}$ joint axis and expressed in the $(i-1)^{th}$ link coordinate system, $R_{i}^{(i-1)}$, rotation matrix from frame $i$ to frame $(i-1)$, $I_{i}^{i}$, the inertia matrix of link $i$ about its center of mass coordinate frame, and $f_{(i,i-1)}^{i}$ / $n_{(i,i-1)}^{i}$ are the resulting force/moment exerted on link $i$ by link $(i-1)$ at point $O_{(i-1)}$, where $i = 1, 2$.
% ===========================================
   % ===========================================

For the link $0$:
\begin{equation}
\label{W00}
\omega_{0}^{0} = R^{0}_{I} \nu_{2}
\end{equation}
% ===============================================
\begin{equation}
\label{Wd00}
\dot{\omega}_{0}^{0} = R^{0}_{I} \dot{\nu}_{2}
\end{equation}
% ===========================================
\begin{equation}
\label{r00}
r_{0}^{0} = [0, L_{0}, 0]^{T}
\end{equation}
% ===============================================
\begin{equation}
\label{v00}
v_{0}^{0} =  R^{0}_{B} \nu_{1} + \omega_{0}^{0} x r_{0}^{0}
\end{equation}
% ===============================================
\begin{equation}
\label{vd00}
\dot{v}_{0}^{0} =  R^{0}_{B} \dot{\nu}_{1} + \dot{\omega}_{0}^{0} x r_{0}^{0} + \omega_{0}^{0} x (\omega_{0}^{0} x r_{0}^{0})
\end{equation}
% ===============================================
For link $i$ $(i = 1, 2)$ we calculate the following variables:
\begin{equation}
\label{wii}
\omega_{i}^{i} = R^{i}_{i-1} (\omega_{i-1}^{i-1} + \dot{\theta}_{i} z_{i-1}^{i-1})
\end{equation}
% ===============================================
\begin{equation}
\label{wii}
\dot{\omega}_{i}^{i} = R^{i}_{i-1} (\dot{\omega}_{i-1}^{i-1} + \ddot{\theta}_{i} z_{i-1}^{i-1} + \omega_{i-1}^{i-1} x \dot{\theta}_{i} z_{i-1}^{i-1})
\end{equation}
% ===============================================
\begin{equation}
\label{vii}
v_{i}^{i} = R^{i}_{i-1} v_{i-1}^{i-1} + \omega_{i}^{i} x r_{i}^{i}
\end{equation}
% ===============================================
\begin{equation}
\label{vdii}
\dot{v}_{i}^{i} = R^{i}_{i-1} \dot{v}_{i-1}^{i-1} + \dot{\omega}_{i}^{i} x r_{i}^{i} + \omega_{i}^{i} x (\omega_{i}^{i} x r_{i}^{i})
\end{equation}
% ===============================================
\begin{equation}
\label{vcdii}
\dot{v}_{c_i}^{i} =\dot{v}_{i}^{i} + \dot{\omega}_{i}^{i} x r_{c_i}^{i} + \omega_{i}^{i} x (\omega_{i}^{i} x r_{c_i}^{i})
\end{equation}
% ===============================================
The inertial forces and moments acting on link $i$ are given by:
% ===============================================
\begin{equation}
\label{Fii}
F_{i}^{i} = - m_{i} \dot{v}_{c_i}^{i}
\end{equation}
% ===============================================
\begin{equation}
\label{Nii}
N_{i}^{i} = - I_{i}^{i} \dot{\omega}_{i}^{i} - \omega_{i}^{i} x I_{i}^{i} \omega_{i}^{i}
\end{equation}
% ===============================================
The total forces and moments acting on link $i$ are given by:
% ===============================================
\begin{equation}
\label{fi_i-1}
f_{i,i-1}^{i} = f_{i+1,i}^{i} - m_{i} g_{i} - F_{i}^{i}
\end{equation}
% ===============================================
\begin{equation}
\label{ni_i-1}
n_{i,i-1}^{i} = n_{i+1,i}^{i} +(f_{i}^{i} + r_{c_i}^{i}) x f_{i,i-1}^{i} - r_{c_i}^{i} x f_{i+1,i}^{i} - N_{i}^{i}
\end{equation}
% ===============================================
\begin{equation}
\label{fi-1_i-1}
f_{i,i-1}^{i-1} = R_{i}^{i-1} f_{i,i-1}^{i}
\end{equation}
% ===============================================
\begin{equation}
\label{ni-1_i-1}
n_{i,i-1}^{i-1} = R_{i}^{i-1} n_{i,i-1}^{i}
\end{equation}
% ===============================================
where,
% ===============================================
\begin{equation}
z_{0}^{0} = [0,0,1]^T
\end{equation}
% ===============================================
\begin{equation}
r_{1}^{1} = [L_{1},0,0]^T
\end{equation}
% ===============================================
\begin{equation}
r_{c_1}^{1} = [-L_{1}/2,0,0]^T
\end{equation}
% ===============================================
\begin{equation}
z_{1}^{1} = [0,0,1]^T
\end{equation}
% ===============================================
\begin{equation}
r_{2}^{2} = [L_{2},0,0]^T
\end{equation}
% ===============================================
\begin{equation}
r_{c_2}^{2} = [-L_{2}/2,0,0]^T
\end{equation}
The gravity vector expressed in frames 1 and 2 are:
% ===============================================
\begin{equation}
\label{g2}
g^{2} = R_{I}^{2} g^{I}
\end{equation}
% ===============================================
\begin{equation}
\label{g1}
g^{1} = R_{I}^{1} g^{I}
\end{equation}
%=====================
where,
% ===============================================
\begin{equation}
\label{RI1}
R_{I}^{1} = R_{0}^{1} R_{B}^{0} R_{I}^{B}
\end{equation}
% ===============================================
\begin{equation}
\label{RI2}
R_{I}^{2} = R_{1}^{2} R_{I}^{1}
\end{equation}
and,
\begin{equation}
g^{I} = [0,0,-g]^T
\end{equation}
%=====================================
Let links 1 and 2 be square beams of relatively small cross-sectional area, then
 % ===============================================
\begin{equation}
\label{Iii}
I_{i}^{i} = \frac{m_{i} L_{i}^{2}}{12}  \begin{bmatrix}
                  0 & 0 & 0 \\
            0 & 1 & 0 \\
             0 & 0 & 1 \\
           \end{bmatrix}
\end{equation}
where $m_i$  and $L_i$ are the mass and length of link $i$.

The torques acting on joints 1 and 2 are finally given by:
% ===============================================
\begin{equation}
\label{Tm1details}
T_{m_1} = (n_{1,0}^{0})^{T}  z_{0}^{0} + b_{1} \dot{\theta}_{1}
\end{equation}
% ===============================================
\begin{equation}
\label{Tm2details}
T_{m_2} = (n_{2,1}^{1})^{T}  z_{1}^{1} + b_{2} \dot{\theta}_{2}
\end{equation}
where, $b_1$and $b_2$ are  friction coefficients.

The interaction forces and moments of the manipulator acting on the quadrotor expressed in body frame, $F_{m,q}^{B}$ and $M_{m,q}^{B}$ are given as follows:
\begin{equation}
\begin{bmatrix}
  F_{m,q}^{B} \\
  M_{m,q}^{B} \\
\end{bmatrix}
=
\begin{bmatrix}
  R_{0}^{B} & 0_{3x3}   \\
  skew(P_{B0}^{B})R_{0}^{B} & R_{0}^{B} \\
\end{bmatrix}
\begin{bmatrix}
  f_{1,0}^{0} \\
  n_{1,0}^{0} \\
\end{bmatrix}
\label{interact_f_M}
\end{equation}
where, $skew (.)$ is skew symmetric matrix \cite{Spong} of $P_{B0}^{B} = [0, 0, -L_{0}]^{T} $, which is the position vector of the origin $O_{0}$ relative to frame $B$.
The interaction forces expressed in the inertial frame are:
\begin{equation}
\label{FmqI}
F_{m,q}^{I} = R_{B}^{I} F_{m,q}^{B}
\end{equation}

The equations of motion of the manipulator are:

\begin{equation}
\label{arm1}
M_{1}\ddot{\theta}_{1} = T_{m_1} + N_1
\end{equation}
\begin{equation}
\label{arm2}
M_{2}\ddot{\theta}_{2} = T_{m_2} + N_2
\end{equation}
where, $T_{m_1}$ and $T_{m_2}$ are the manipulator-actuators' torques. $M_1$, $M_2$, $N_1$, and $N_2$ are  nonlinear terms and they are functions in the system states $(\eta_{2}, \nu_{2}, \dot{\nu}_{2}, \dot{\nu}_{1}, \theta_{1}, \theta_{2}, \dot{\theta}_{1}, \dot{\theta}_{2})$.

The equations of motion of the quadrotor after adding the forces/moments applied by the manipulator are:
\begin{equation}
\label{Xdd}
m\ddot{X} = T(C(\psi) S(\theta) C(\phi) + S(\psi) S(\phi)) + F_{m,q_x}^I
\end{equation}
\begin{equation}
\label{Ydd}
m\ddot{Y} = T(S(\psi) S(\theta) C(\phi) - C(\psi) S(\phi)) + F_{m,q_y}^I
\end{equation}
\begin{equation}
\label{Zdd}
m\ddot{Z} = -mg + T C(\theta) C(\phi) + F_{m,q_z}^I
\end{equation}
\begin{equation}
\label{Phdd}
I_{x}\ddot{\phi} = \dot{\theta}\dot{\phi}(I_{y}-I_{z}) - I_{r}\dot{\theta}\Omega + T_{a_1} + M_{m,q_\phi}^B
\end{equation}
\begin{equation}
\label{thdd}
I_{y}\ddot{\theta} = \dot{\psi}\dot{\phi}(I_{z}-I_{x}) + I_{r}\dot{\phi}\Omega + T_{a_2} + M_{m,q_\theta}^B
\end{equation}
\begin{equation}
\label{epdd}
I_{z}\ddot{\psi} = \dot{\theta}\dot{\phi}(I_{x}-I_{y}) + T_{a_3} + M_{m,q_\psi}^B
\end{equation}
where  $F_{m,q_x}^I$, $F_{m,q_y}^I$, and $F_{m,q_z}^I$ are the interaction forces from the manipulator to the quadrotor in $X$,$Y$, and $Z$ directions defined in the inertial frame, and  $M_{m,q_\phi}^B$, $M_{m,q_\theta}^B$, and $M_{m,q_\psi}^B$  are the interaction moments from the manipulator to the quadrotor around $X$, $Y$, and $Z$  directions defined in the inertial frame.
\subsection{Effect of adding a payload to the manipulator's end effector}
Applying a payload of mass equal to $m_{p}$ (see Figure \ref{effect_paylaod}) will change link 2's parameters such as mass moments of inertia, total mass of this link, and center of gravity of this link as following:
\begin{figure}[!h]
      \centering
      \includegraphics[width=0.7\columnwidth, height=8cm]{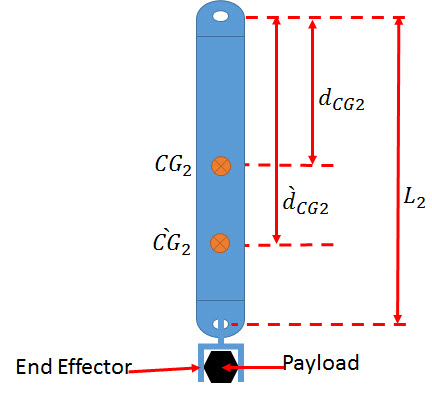}
      %\framebox{\parbox{1cm}{(1.a)}}
      \caption{Schematic Diagram of Link $2$ after Adding the Payload}
      \label{effect_paylaod}
\end{figure}
% ===========================================
\begin{equation}
\grave{I_{2}^{2}} = I_{2}^{2} + m_{2} (\grave {d_{CG_2}} - d_{CG_2})^{2}+ m_{p} (L_{2} - \grave {d_{CG_2}})^{2}
\label{I22new}
\end{equation}
%===================================================
\begin{equation}
\grave{d_{CG_2}} = \frac{m_{2} d_{CG_2} + m_{p} L_{2}}{m_{2} + m_{p}}
\label{CGnew}
\end{equation}
%===================================================
\begin{equation}
\grave{m_{2}} = m_{2} + m_{p}
\label{m2new}
\end{equation}
%====================================================
where $CG_{2}$ is the point of center of gravity of link 2, and $(\grave{.})$ refers to the value of the parameter after adding the payload.

Changing the point of center of gravity of link 2 will change the $r_{c_2}^{2}$ to be
\begin{equation}
r_{c_2}^{2} = [-(L_2 - \grave{d_{CG_2}}), 0, 0]^{T}
\label{rc22new}
\end{equation}

Substituting from the equations(\ref{I22new}, \ref{m2new} and \ref{rc22new}) in the previous dynamic equations, we can study the effect of picking/placing the payload ($m_p$).

\chapter{\uppercase{Controller Design and Simulation Results of the Proposed System}\label{ch:controldesignsimresults}}

\section{Controller Design for the Quadrotor Manipulation system\label{sec:controldesign}}

Quadrotor is an under-actuated system, because it has four inputs (angular velocities of its four rotors) and six variables to be controlled. By observing the operation of the quadrotor, one can find that the movement in $X$- direction is based on the pitch rotation, $\theta$. Also the movement in $Y$- direction is based on the roll rotation, $\phi$. Therefore, motion along $X$- and $Y$-axes will be controlled through controlling $\theta$ and $\phi$.

\begin{figure}
      \centering
      \includegraphics[width=1\columnwidth, height=7cm]{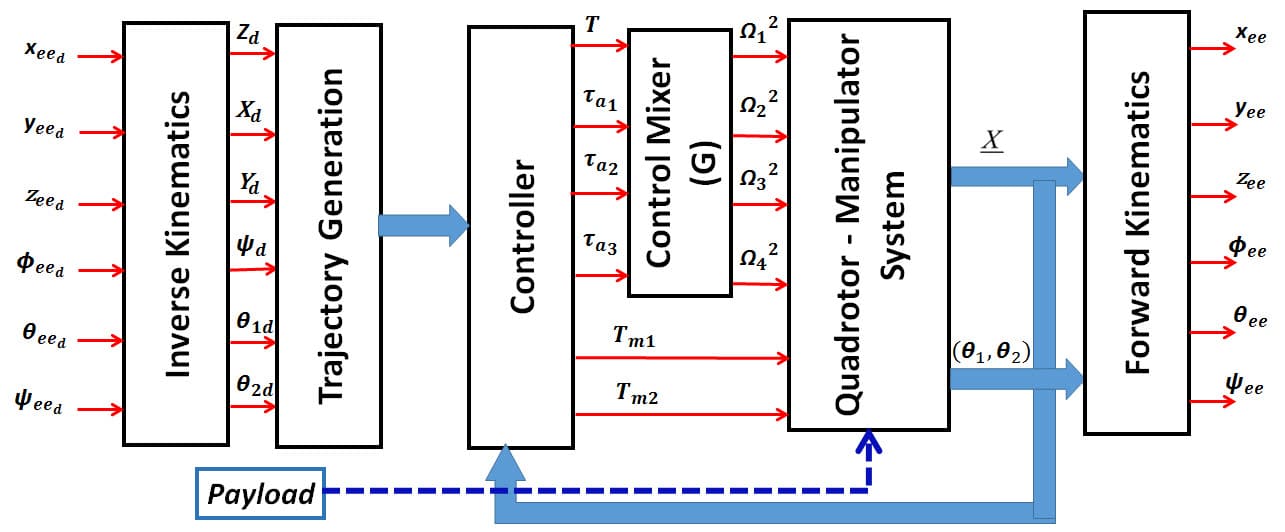}
          \caption{Block Diagram of the Control System}
      \label{General_Control_block}
   \end{figure}

Figure \ref{General_Control_block} presents a block diagram of the proposed control system. The desired values for the end effector's position ($x_{ee_d}$, $y_{ee_d}$ and $z_{ee_d}$) and orientation ($\phi_{ee_d}$, $\theta_{ee_d}$ and $\psi_{ee_d}$) are converted to the desired values of the quadrotor ($X_{d}$, $Y_{d}$, $Z_{d}$ and $\psi_{d}$) and joints variables ($\theta_{1_d}$ and $\theta_{2_d}$) through the inverse kinematics that are derived in Chapter \ref{ch:modeling}. Next, these values is applied to a trajectory generation algorithm which will be explained later. After that, the controller block receives the deired values and the feedback signals from the system and provides the control signals ($T$, $\tau_{a_1}$, $\tau_{a_2}$, $\tau_{a_3}$, $T_{m_1}$ and $T_{m_2}$). The matrix G of the control mixer, in Figure \ref{General_Control_block}, is used to transform the assigned thrust force and moments of the quadrotor (the control signals) from the controller block into assigned angular velocities of the four rotors. This matrix can be derived from (\ref{thrust_sum}-\ref{Ta3}) and presented as following:
\begin{equation}
\begin{bmatrix}
  \Omega_1^2 \\
  \Omega_2^2 \\
  \Omega_3^2 \\
  \Omega_4^2 \\
\end{bmatrix}
=
\underbrace{ \begin{bmatrix}
  K_{F_1} & K_{F_2} & K_{F_3} & K_{F_4} \\
  0 & -d K_{F_2} & 0 & d K_{F_4} \\
  -d K_{F_1} & 0 & d K_{F_3} & 0 \\
  -K_{M_1} & K_{M_2} & -K_{M_3} & K_{M_4} \\
\end{bmatrix}
^{-1}
}_{G}
\begin{bmatrix}
  T \\
  \tau_{a_1} \\
  \tau_{a_2} \\
  \tau_{a_3} \\
\end{bmatrix}
\label{control_matrix}
\end{equation}

Finally, The actual values of the quadrotor and joints are converted to the actual values of the end effector variables through the forward kinematics which are derived in Chapter \ref{ch:modeling}.

The control design criteria are to achieve system stability and zero position error, for the movements in $X$, $Y$, $Z$, and $\psi$ directions as well as for joints' angles $\theta_1$ and $\theta_2$ and consequently for the end effector variables ($\eta_{ee_1}$ and $\eta_{ee_2}$), under the effect of:
\begin{itemize}
\item Picking and placing a payload.
\item Changing the operating region of the system.
\end{itemize}

Noting that in the task space, a position tracking is implemented, and in the joint space, trajectory tracking is required.
%===================================================
%==============================================
\section{Trajectory Generation}
Quintic Polynomial Trajectories \cite{Spong} are used as the reference trajectories for $X$, $Y$, $Z$, $\psi$, $\theta_1$, and $\theta_2$. Those types of trajectories have sinusoidal acceleration which is better in order to avoid vibrational modes.

\begin{figure}
\centering
\begin{tabular}{cc}
 \subfigure[X Desired Trajectory]{\includegraphics[width=8cm]{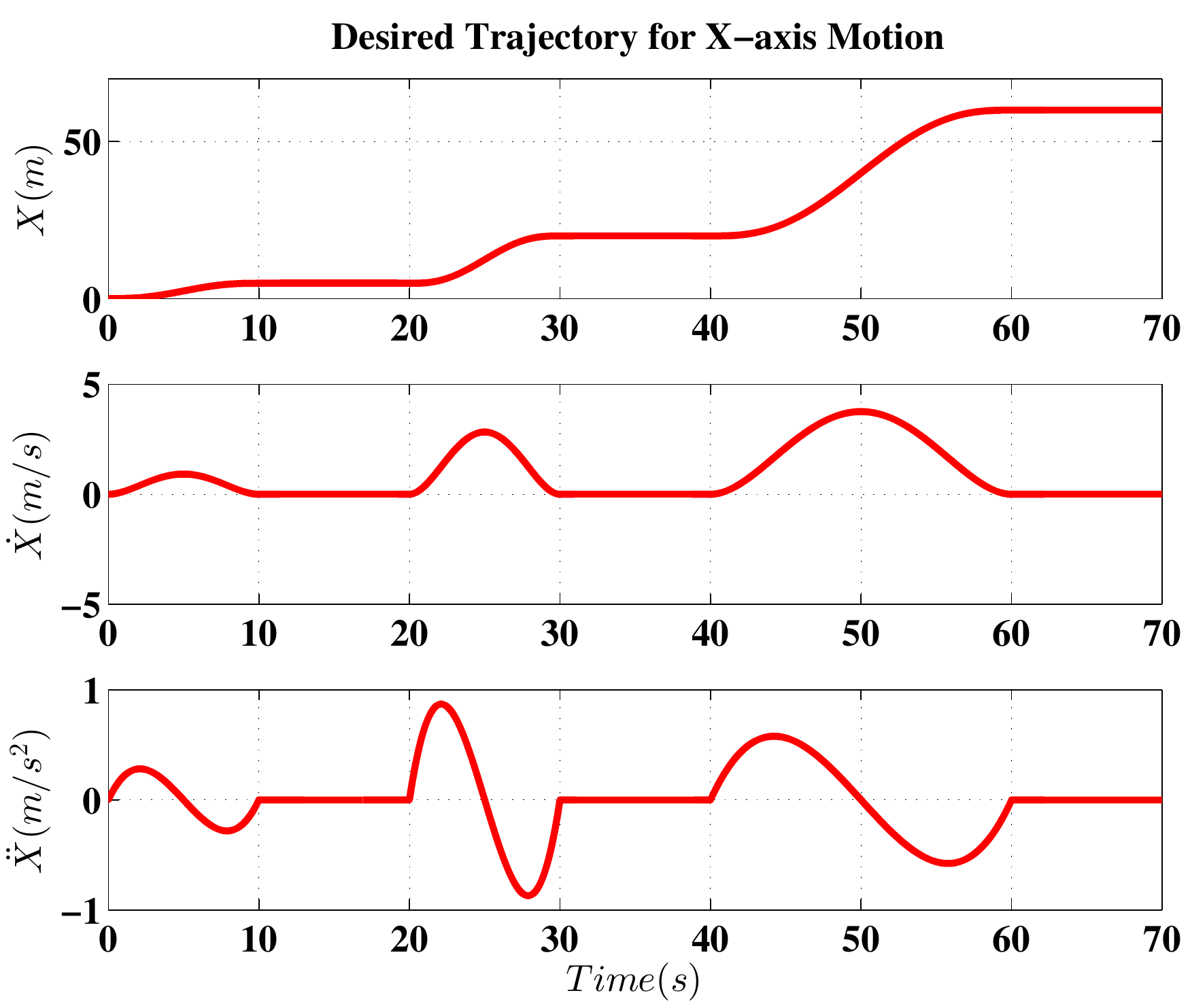}}&
 \subfigure[Y Desired Trajectory]{\includegraphics [width=8cm]{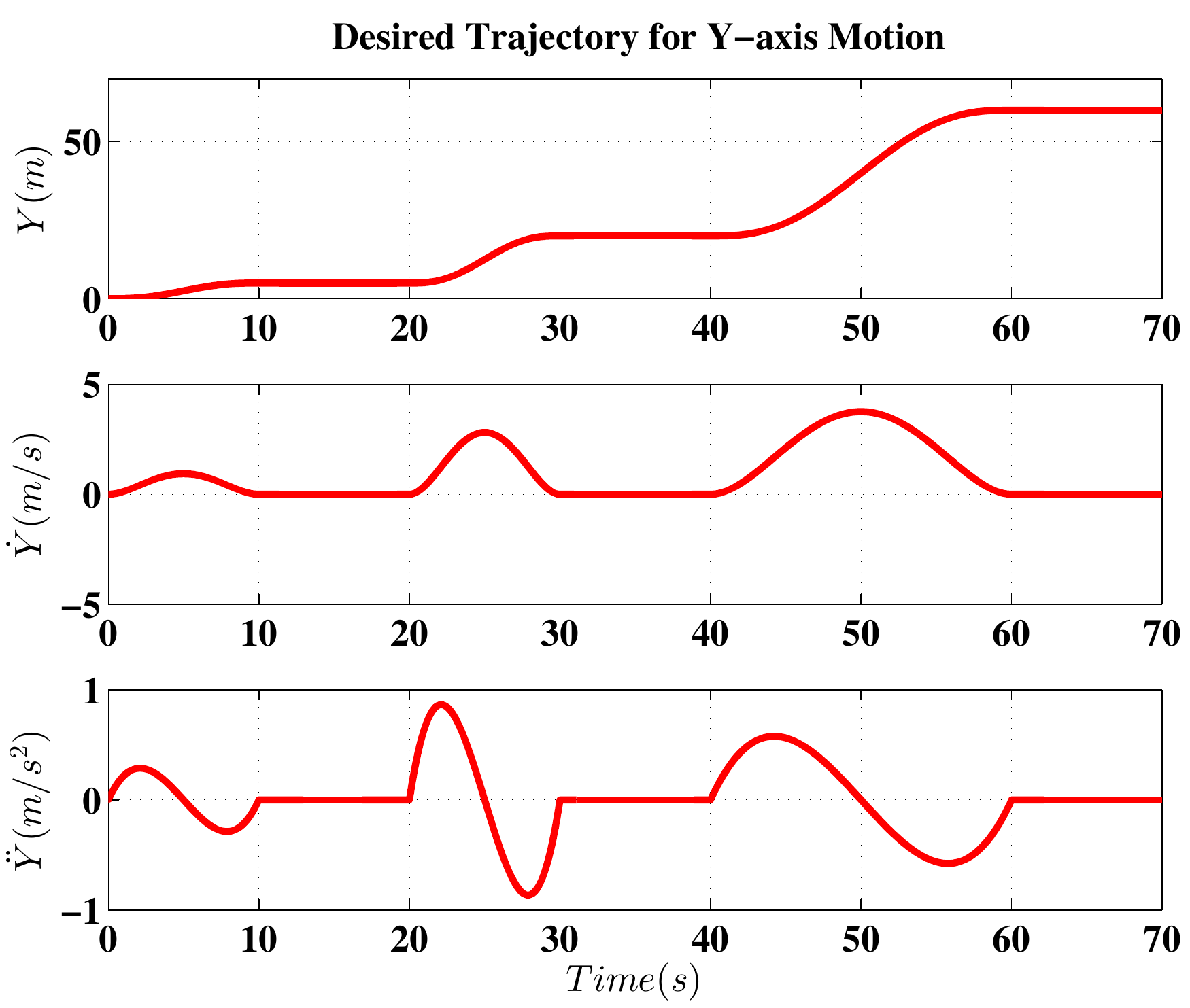}} \\
 \subfigure[Z Desired Trajectory]{\includegraphics [width=8cm]{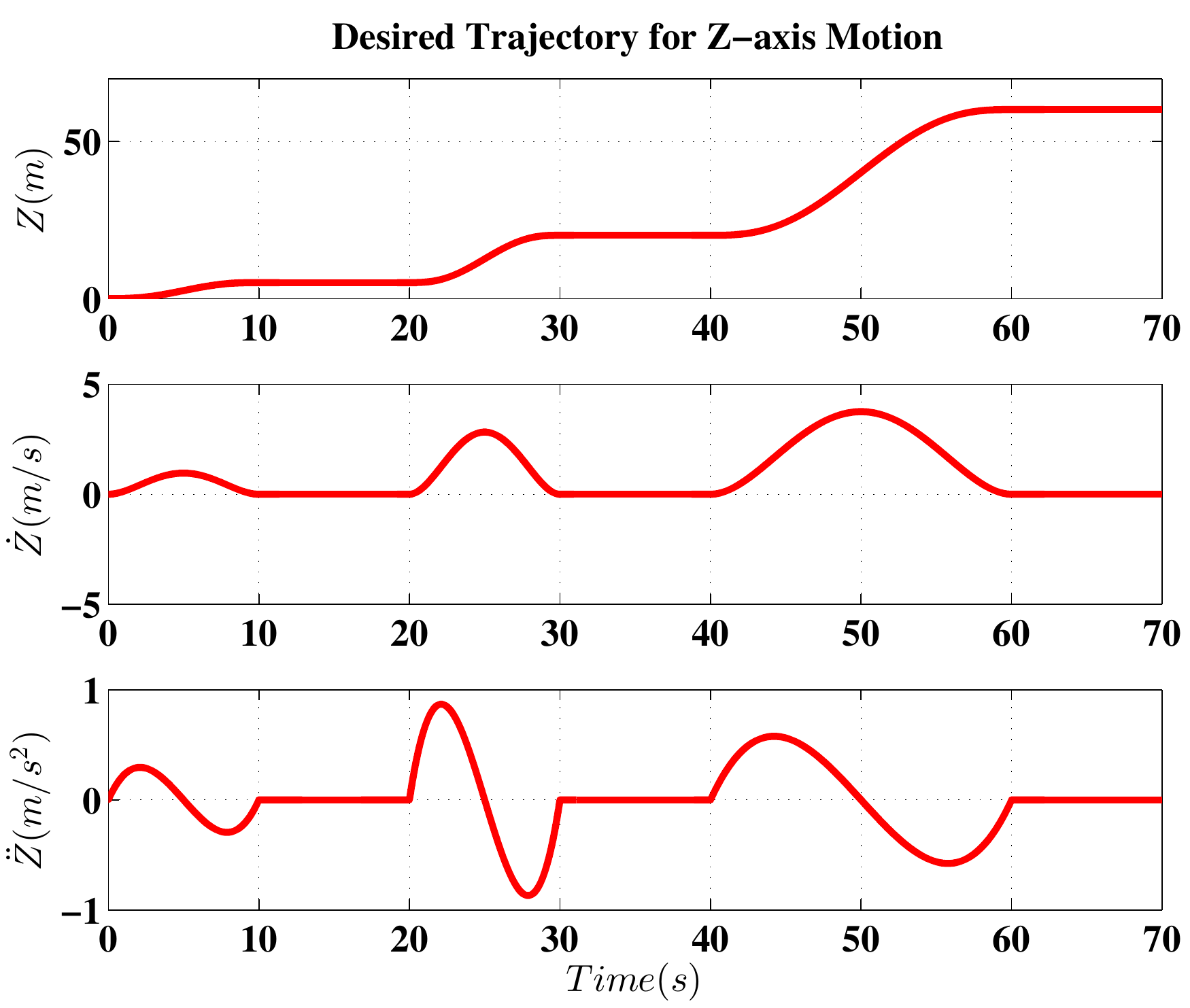}} &
 \subfigure[$\psi$ Desired Trajectory]{\includegraphics [width=8cm]{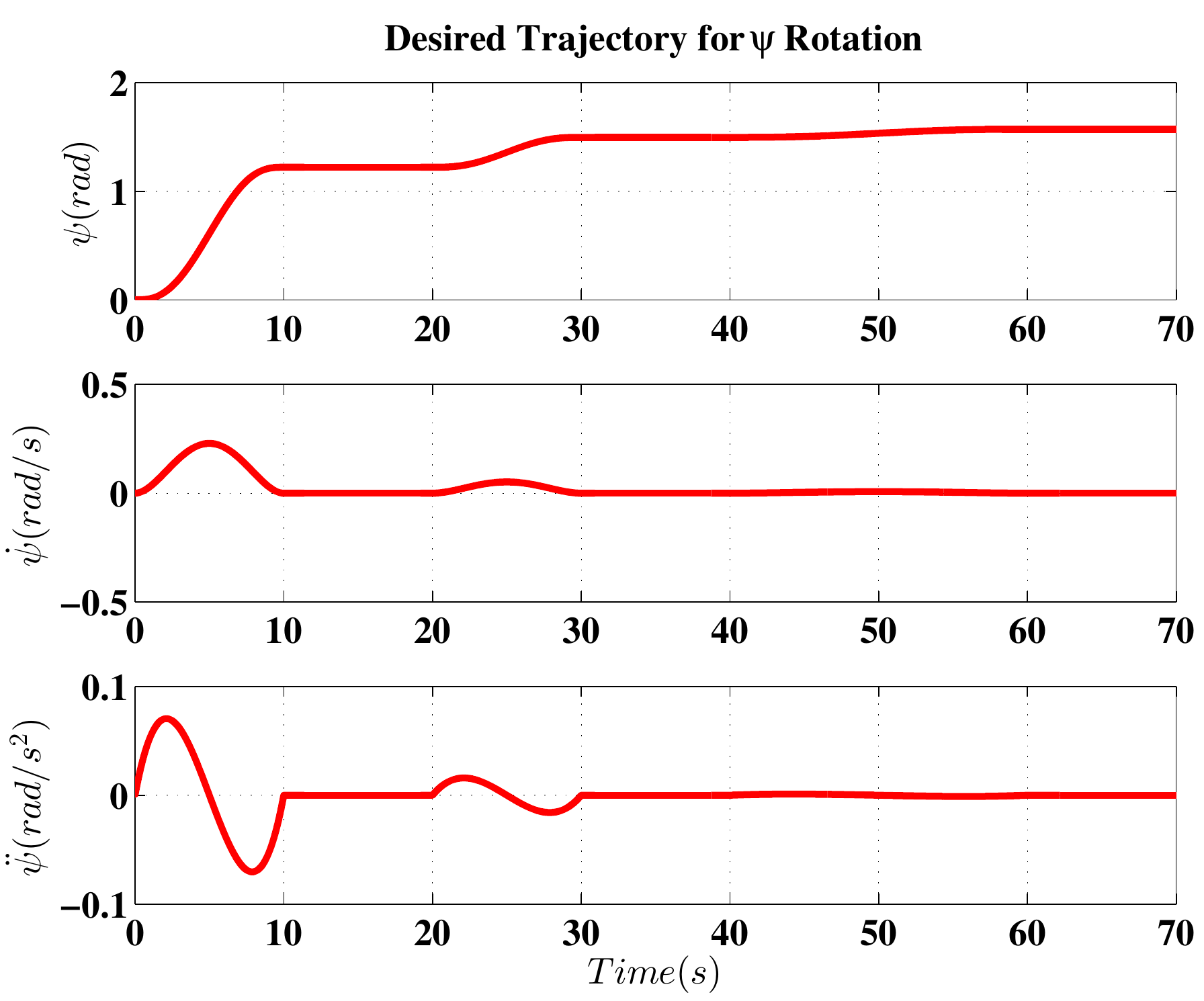}}\\
 \subfigure[$\theta_1$ Desired Trajectory]{\includegraphics [width=8cm]{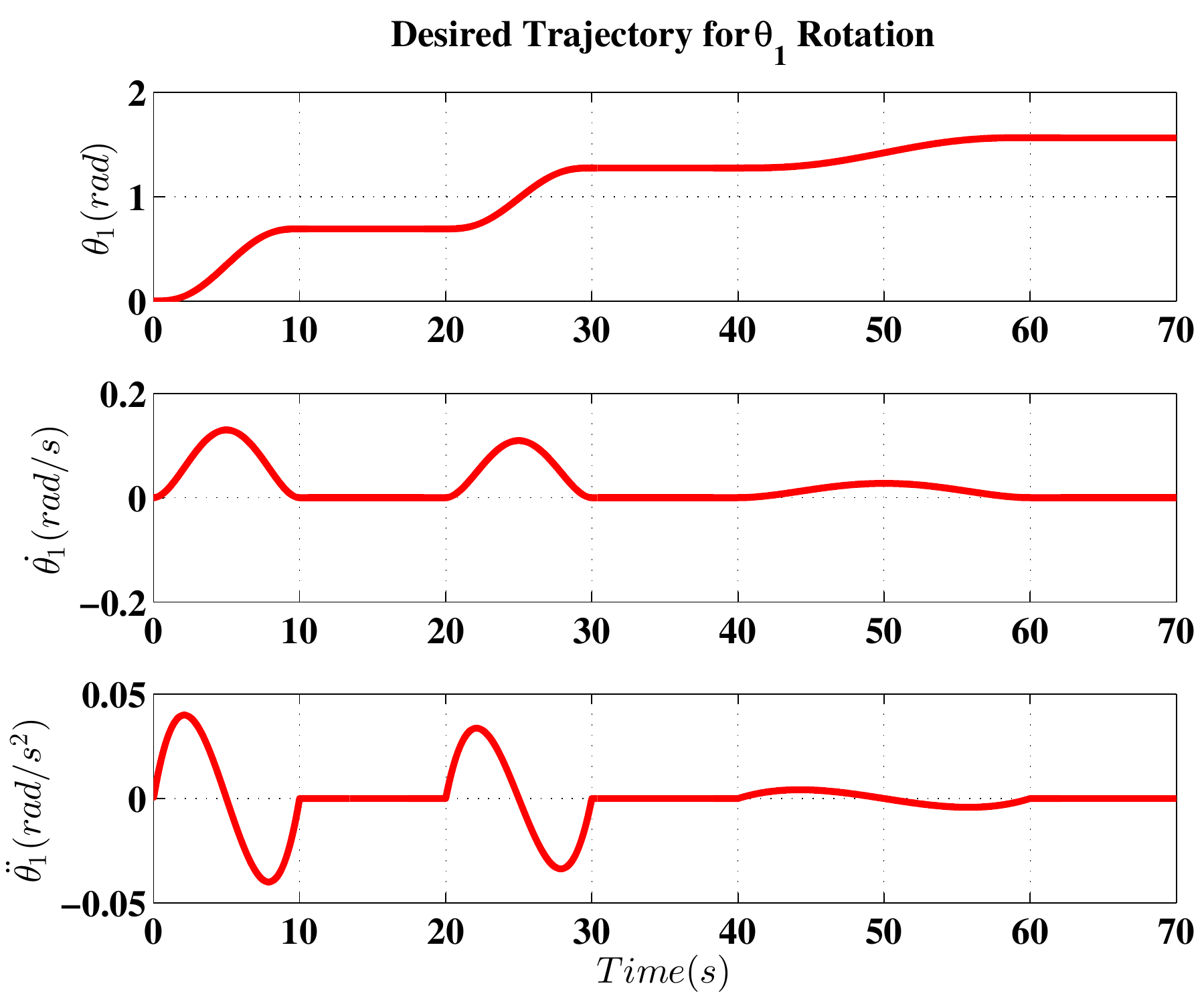}}&
 \subfigure[$\theta_2$ Desired Trajectory]{\includegraphics [width=8cm]{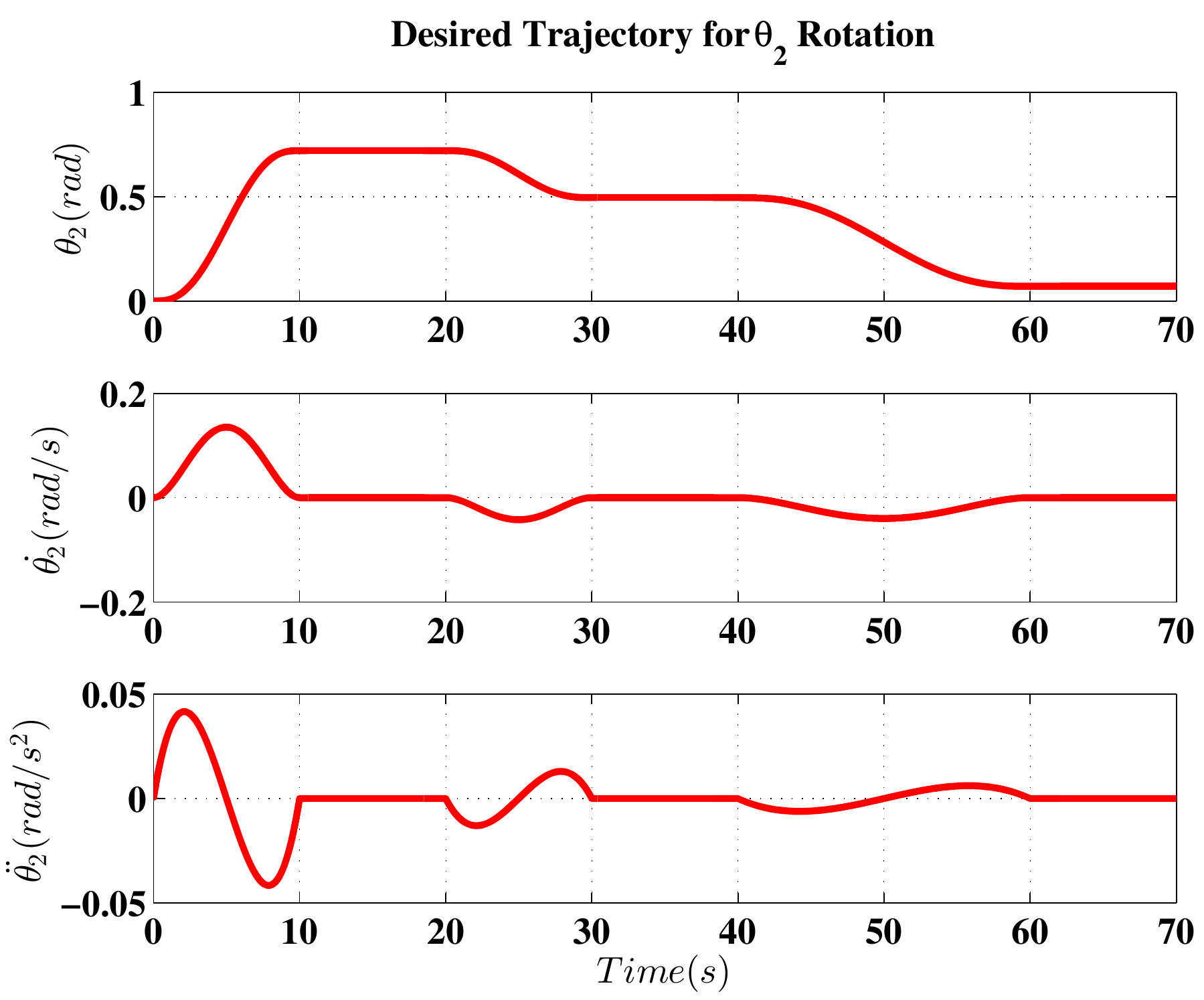}}
\end{tabular}
\caption{The Generated Desired Trajectories for the Quadrotor and Manipulator Variables: a) X, b) Y, c) Z, d) $\psi$, e) $\theta_1$, and f) $\theta_2$.}
\label {Desired_Traj}
\end{figure}

The desired values of end effector position and orientation (Multi-region of operation and point-to-point control) are with the following values:
\begin{itemize}
  \item $x_{ee_d}$ = 5 m to 20 m then 60 m.
  \item $y_{ee_d}$ = 5 m to 20 m then 60 m.
  \item $z_{ee_d}$ = 5 m to 20 m then 60 m.
  \item $\phi_{ee_d}$ = 0.5 $rad$ to 1 $rad$ then 1.5 $rad$
  \item $\theta_{ee_d}$ = 0.5 $rad$ to 1 $rad$ then 1.5$rad$
  \item $\psi_{ee_d}$ = 0.5 $rad$ to 1 $rad$ then 1.5 $rad$
\end{itemize}
Figure \ref{Desired_Traj} shows the generated desired trajectories for $X$, $Y$, $Z$, $\phi$, $\theta$ and $\psi$ using the inverse kinematics and then the algorithm for generating the trajectories.
%===================================================
%\begin{figure}[!]
%\centering
%\subfigure[X Desired Trajectory]{\includegraphics[width=15cm]{figures/x_des_traj}}\hspace{2cm}
%\subfigure[Y Desired Trajectory]{\includegraphics [width=15cm]{figures/y_des_traj}}\hspace{2cm}
%\subfigure[Z Desired Trajectory]{\includegraphics [width=15cm]{figures/z_des_traj}}\hspace{2cm}
%\subfigure[$\psi$ Desired Trajectory]{\includegraphics [width=15cm]{figures/ep_des_traj}}\hspace{2cm}
%\subfigure[$\theta_1$ Desired Trajectory]{\includegraphics [width=15cm]{figures/th1_des_traj}}\hspace{2cm}
%\subfigure[$\theta_2$ Desired Trajectory]{\includegraphics [width=15cm]{figures/th2_des_traj}}
%\caption{The Generated desired Trajectories for the quadrotor and manipulator variables . a) X, b) Y, c) Z, d) $\psi$, e) $\theta_1$, and f) $\theta_2$.}
%\label {Desired_Traj}
%\end{figure}
%==============================================
%==============================================
\section{Feedback Linearization Based PID Controller \label{FBL_Tech}}
This section discusses the control system design based on the technique of feedback linearization \cite{Spong, applied_nonlinear}. Feedback linearization transforms the nonlinear system dynamics into a linear system. Then the control laws are chosen so that zero tracking errors are achieved.

\begin{figure}
      \centering
      \includegraphics[width=1\columnwidth, height=8cm]{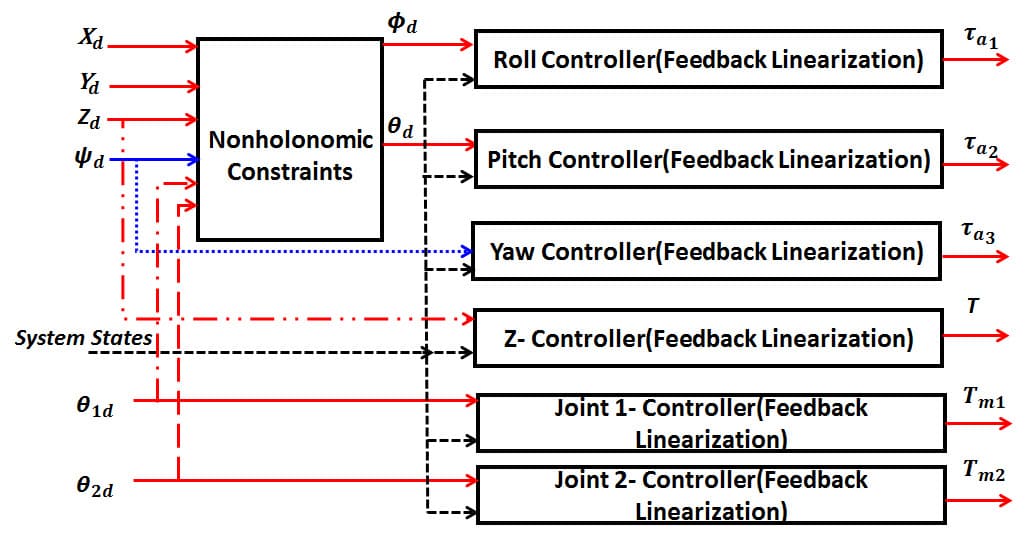}
      %\framebox{\parbox{1cm}{(1.a)}}
      \caption{Details of the Controller Block in Case of Feedback Linearization}
      \label{FBL_controller}
   \end{figure}

Figure \ref{FBL_controller} presents the block diagram of this control technique. In this Figure, the nonholonmic constraints are used to determine the desired trajectories of  $\theta$ and $\phi$ from the desired trajectories of $X$, $Y$, $Z$, $\psi$, $\theta_1$, and $\theta_2$ and their derivatives. Then feedback linearization controllers are used to obtain a zero tracking errors for $\theta$, $\phi$, Z, $\psi$, $\theta_1$ and $\theta_2$.

The nonholonmic constraints define the coupling between various states of the system. They are used to determine the desired trajectories of $\theta$ and $\phi$. From the equations of the translation dynamics (\ref{Xdd}-\ref{Zdd}), one can extract the expressions of these high order nonholonmic constraints:
\begin{equation}
\sin(\phi) = \frac{(\ddot{X}-F_{m,q_x}^I)\sin(\psi) - (\ddot{Y}-F_{m,q_y}^I)\cos(\psi)}{\sqrt{(\ddot{X}-F_{m,q_x}^I)^{2}+(\ddot{Y}-F_{m,q_y}^I)^{2}+(\ddot{Z}+g-F_{m,q_z}^I)^{2}}}
\label{sin_ph}
\end{equation}
\begin{equation}
\tan(\theta) = \frac{(\ddot{X}-F_{m,q_x}^I)\cos(\psi) + (\ddot{Y}-F_{m,q_y}^I)\sin(\psi)}{\ddot{Z}+g-F_{m,q_z}^I}
\label{tan_th}
\end{equation}
where $F_{m,q_x}^I$, $F_{m,q_y}^I$, and $F_{m,q_z}^I$ are functions of the system states and their derivatives.

Putting subscript $d$ to all variables in (\ref{sin_ph}) and (\ref{tan_th}), then $\phi_d$ and $\theta_d$ can be obtained numerically.
%=================================================

$Z$-Controller can be developed by expressing the equation of motion in $Z$-direction in the following form:
%===============================
\begin{equation}
(m\ddot{Z} + m g - F_{m,q_z}^I)/(C(\phi) C(\theta)) = T
\label{FBLquadmanp_Z}
\end{equation}
%===============================
The following control input will cancel out the nonlinearities in the system;
%===============================
\begin{equation}
T = (m u_{z} + m g - F_{m,q_z}^I)/(C(\phi) C(\theta))
\label{FBLquadmanp_T}
\end{equation}
%===============================
where,
%===============================
\begin{equation}
u_{Z} = \ddot{Z}_{d} + K_{p_z} e_{z} + K_{d_z} \dot{e}_{z}+ K_{i_z}\int\limits_0^t{e_{z} dt}
\label{FBLquadmanp_uz}
\end{equation}
%===============================
This control law leads to the exponential stable dynamics
%===============================
\begin{equation}
\ddot{e}_{z} + K_{d_z} \dot{e}_{z} + K_{p_z} e_{z} + K_{i_z}\int\limits_0^t{e_{z} dt} = 0
\label{FBLquadmanp_ezdd}
\end{equation}
%===============================
which implies that the error, $e_{z} \rightarrow 0$.

For $\phi$, $\theta$, $\psi$, $\theta_1$ and $\theta_2$ controllers, similar control laws are chosen:
%===============================
\begin{equation}
\tau_{a_1} = I_x u_{\phi} - \dot{\theta} \dot{\psi} (I_y - I_z) + I_r \dot{\theta} \overline{\Omega} - M_{m,q_\phi}^B
\label{FBLquadmanp_Ta1}
\end{equation}
%===============================
where,
%===============================
\begin{equation}
u_{\phi} = \ddot{\phi}_{d} + K_{p_\phi} e_{\phi} + K_{d_\phi} \dot{e}_{\phi}+ K_{i_\phi}\int\limits_0^t{e_{\phi} dt}
\label{FBLquadmanp_uph}
\end{equation}
%===============================
\begin{equation}
\tau_{a_2} = I_y u_{\theta} - \dot{\phi} \dot{\psi} (I_z - I_x) - I_r \dot{\phi} \overline{\Omega} - M_{m,q_\theta}^B
\label{FBLquadmanp_Ta2}
\end{equation}
%===============================
where,
%===============================
\begin{equation}
u_{\theta} = \ddot{\theta}_{d} + K_{p_\theta} e_{\theta} + K_{d_\theta} \dot{e}_{\theta}+ K_{i_\theta}\int\limits_0^t{e_{\theta} dt}
\label{FBLquadmanp_uth}
\end{equation}
%===============================
\begin{equation}
\tau_{a_3} = I_z u_{\psi} - \dot{\phi} \dot{\theta} (I_x - I_y) - M_{m,q_\psi}^B
\label{FBLquadmanp_Ta3}
\end{equation}
%===============================
where,
%===============================
\begin{equation}
u_{\psi} = \ddot{\psi}_{d} + K_{p_\psi} e_{\psi} + K_{d_\psi} \dot{e}_{\psi}+ K_{i_\psi}\int\limits_0^t{e_{\psi} dt}
\label{FBLquadmanp_uep}
\end{equation}
%===============================
\begin{equation}
T_{m_1} = M_1 u_{\theta_1} - N_1
\label{FBLquadmanp_Tm1}
\end{equation}
%===============================
where,
%===============================
\begin{equation}
u_{\theta_1} = \ddot{\theta_1}_{d} + K_{p_{\theta_1}} e_{\theta_1} + K_{d_{\theta_1}} \dot{e}_{\theta_1}+ K_{i_{\theta_1}}\int\limits_0^t{e_{\theta_1} dt}
\label{FBLquadmanp_um1}
\end{equation}
%===============================
\begin{equation}
T_{m_2} = M_2 u_{\theta_2} - N_2
\label{FBLquadmanp_Tm2}
\end{equation}
%===============================
where,
%===============================
\begin{equation}
u_{\theta_2} = \ddot{\theta_2}_{d} + K_{p_{\theta_2}} e_{\theta_2} + K_{d_{\theta_2}} \dot{e}_{\theta_2}+ K_{i_{\theta_2}}\int\limits_0^t{e_{\theta_2} dt}
\label{FBLquadmanp_um2}
\end{equation}
%===============================
where $K_p$, $K_d$ and $K_i$ are the controller parameters.
\subsection{Simulation Results}
The system equations of motion and the control laws are simulated using MATLAB/SIMULINK program.
The identified Parameters of the system obtained in Chapter \ref{ch:qudmodelidentresults} are listed in Table \ref{sys_par}. The controller parameters of the feedback linearization controller are given in Table \ref{FBL parameters}. Those parameters are tuned to get the required system performance.

\begin{figure}
\centering
\begin{tabular}{cc}
 \subfigure[$X$ Response]{\includegraphics[width=0.5\columnwidth, height=5cm]{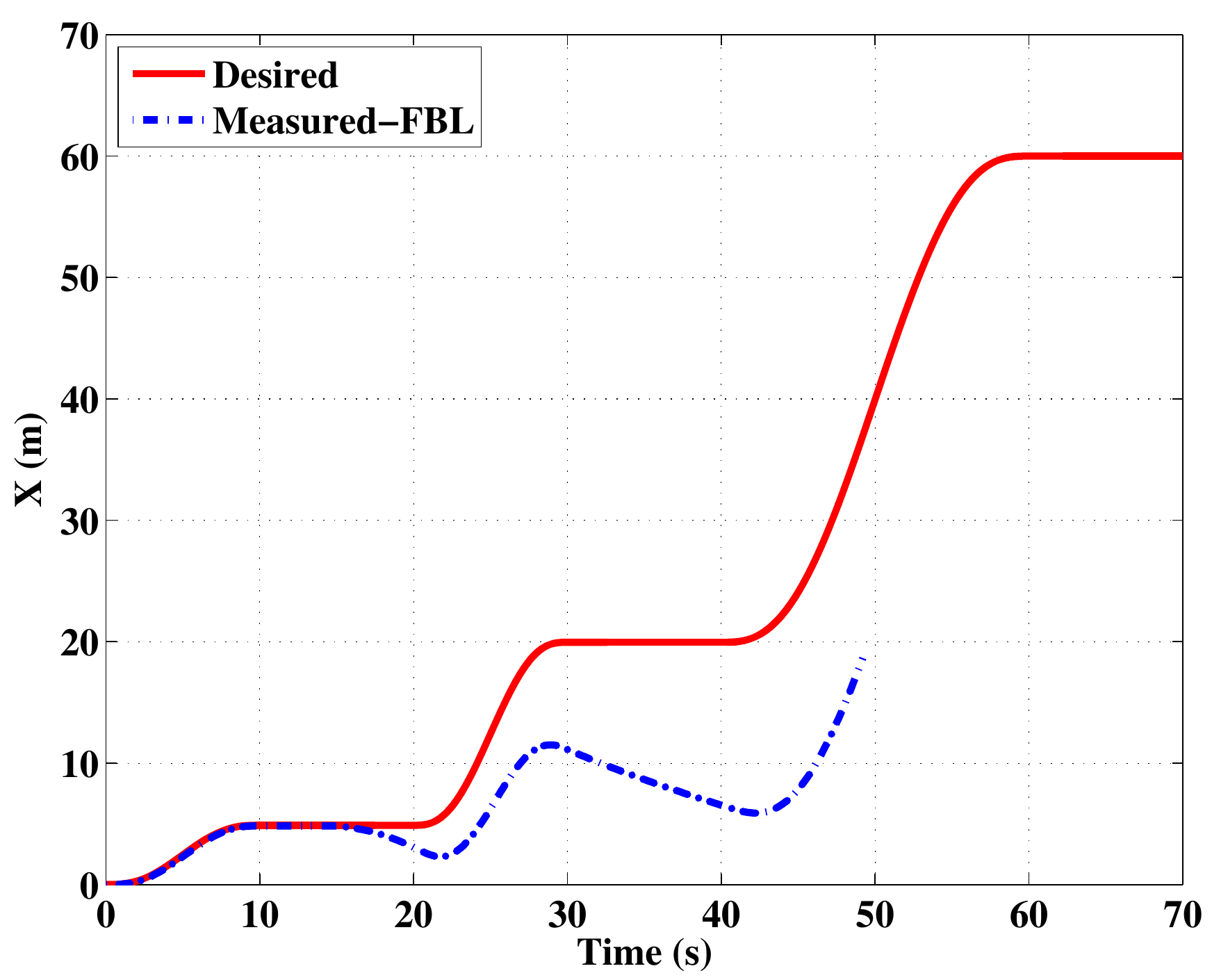}}&
 \subfigure[$Y$ Response]{\includegraphics [width=0.5\columnwidth, height=5cm]{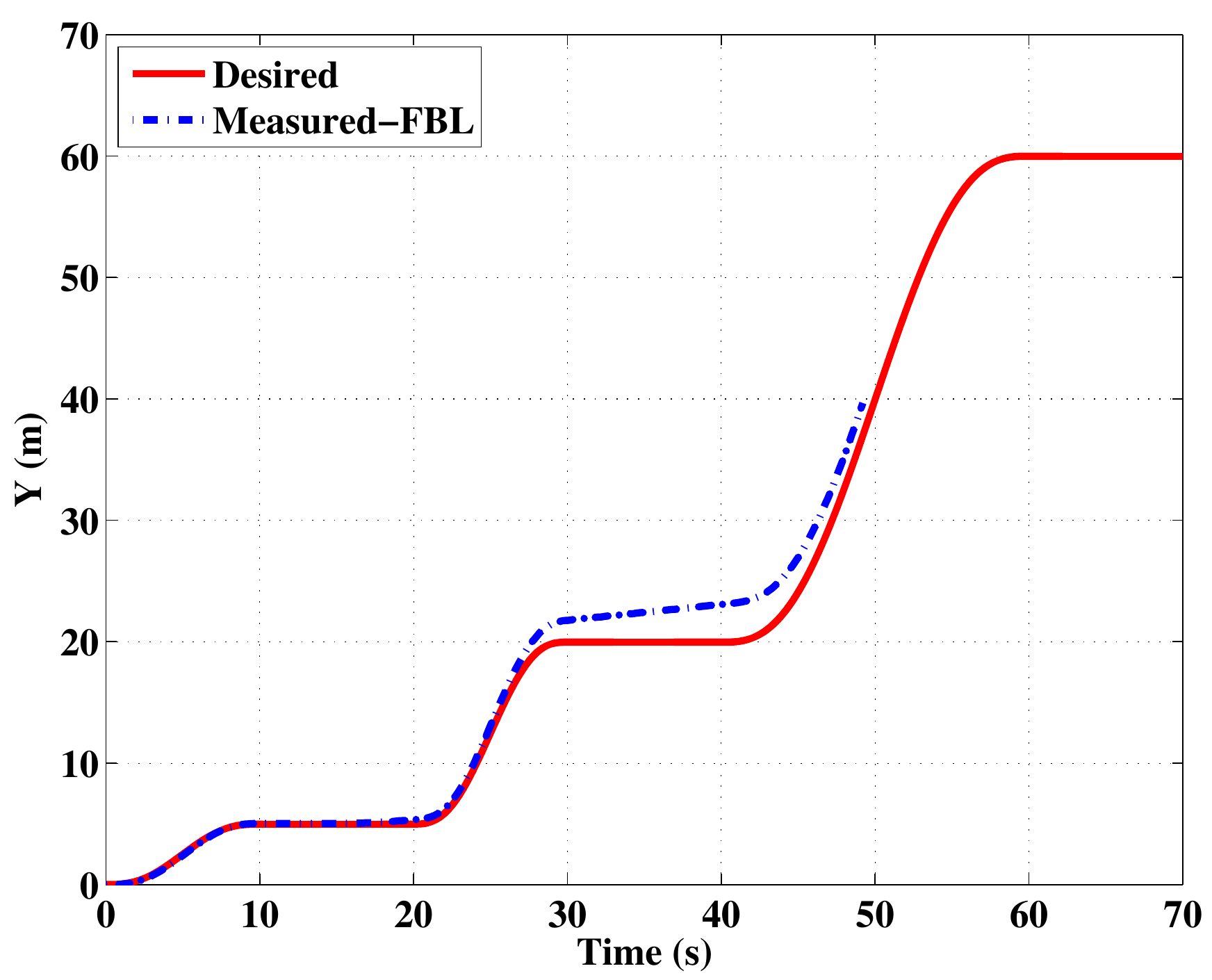}} \\
 \subfigure[$Z$ Response]{\includegraphics [width=0.5\columnwidth, height=5cm]{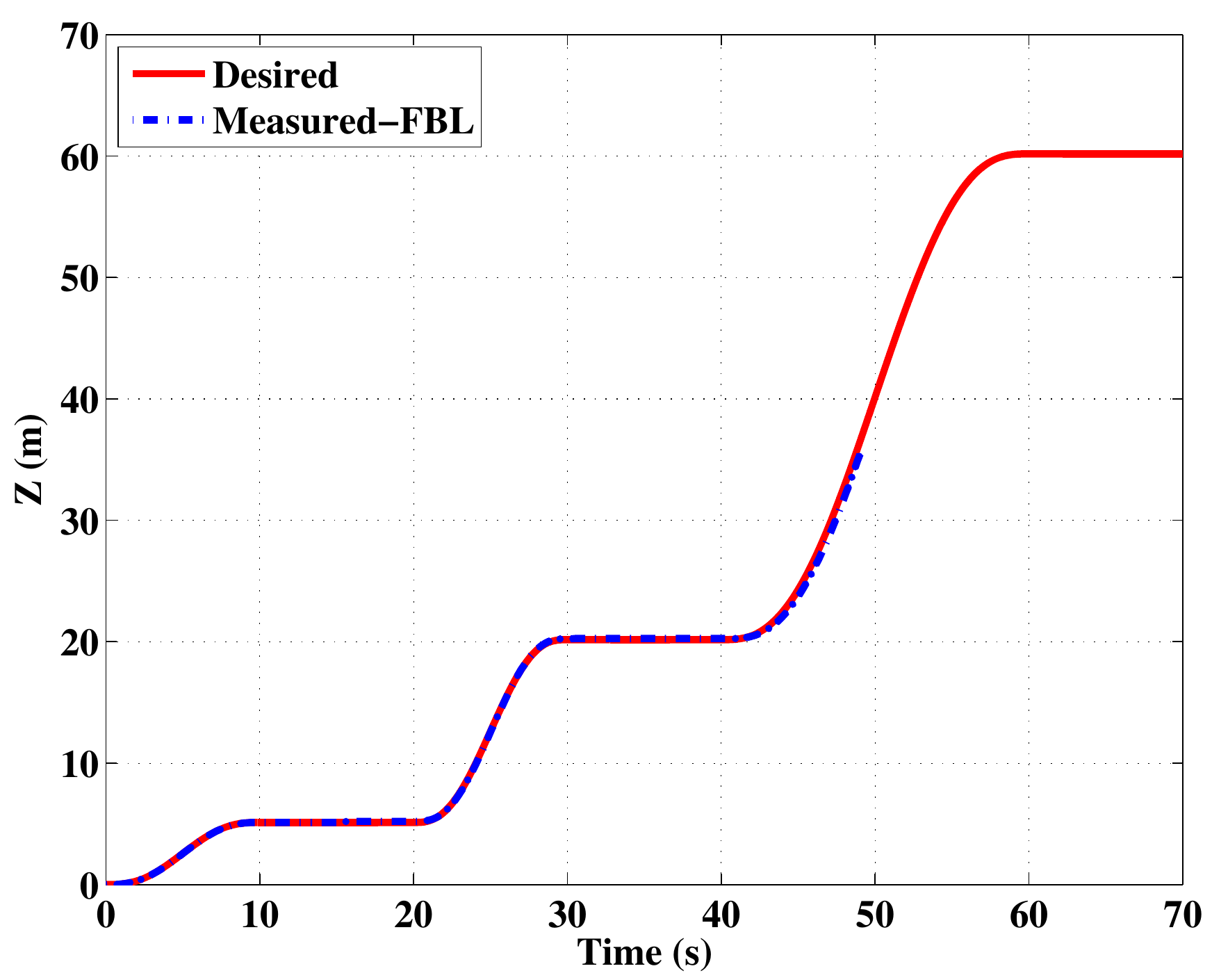}} &
 \subfigure[$\psi$ Response]{\includegraphics [width=0.5\columnwidth, height=5cm]{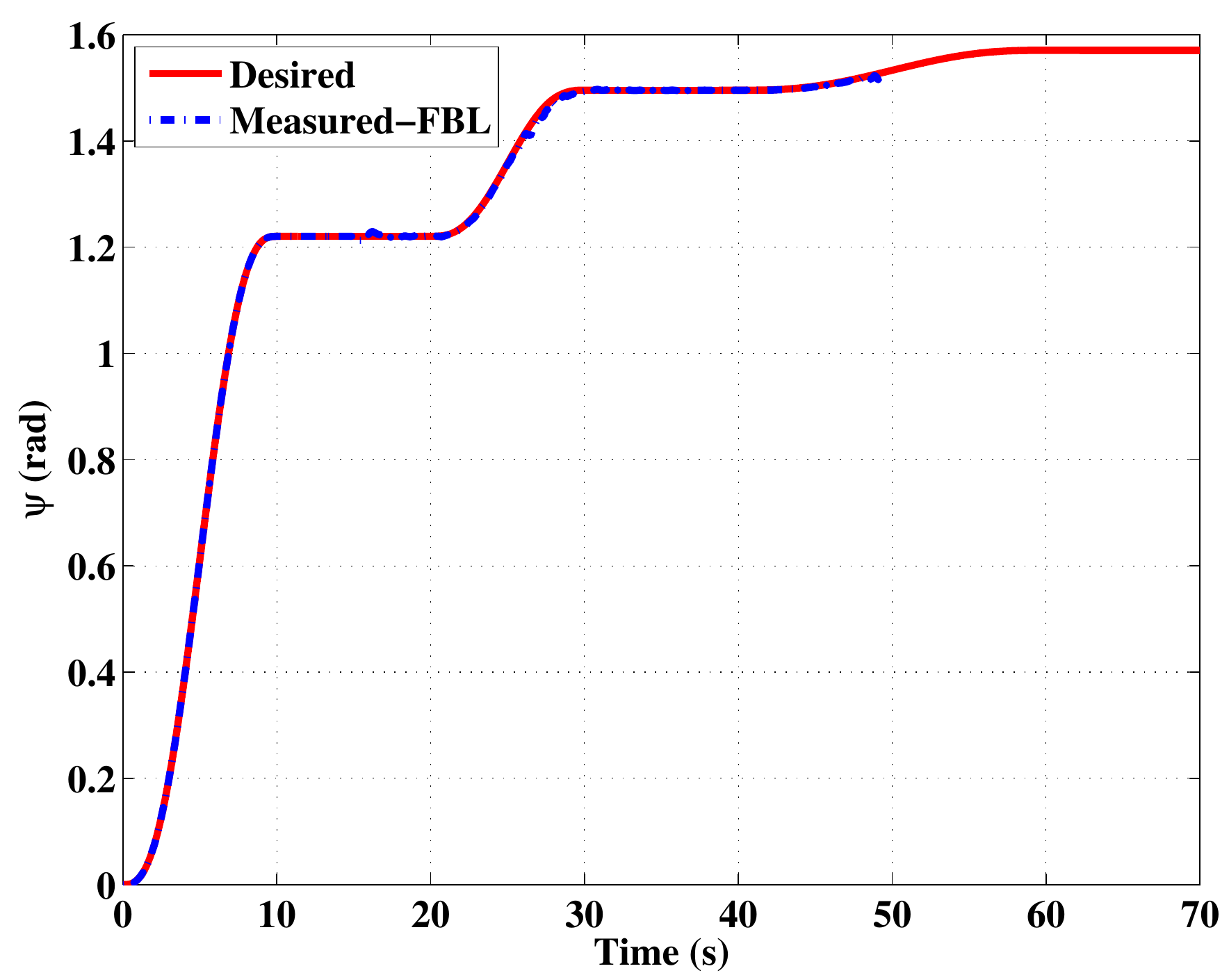}}\\
 \subfigure[$\theta_1$ Response]{\includegraphics [width=0.5\columnwidth, height=5cm]{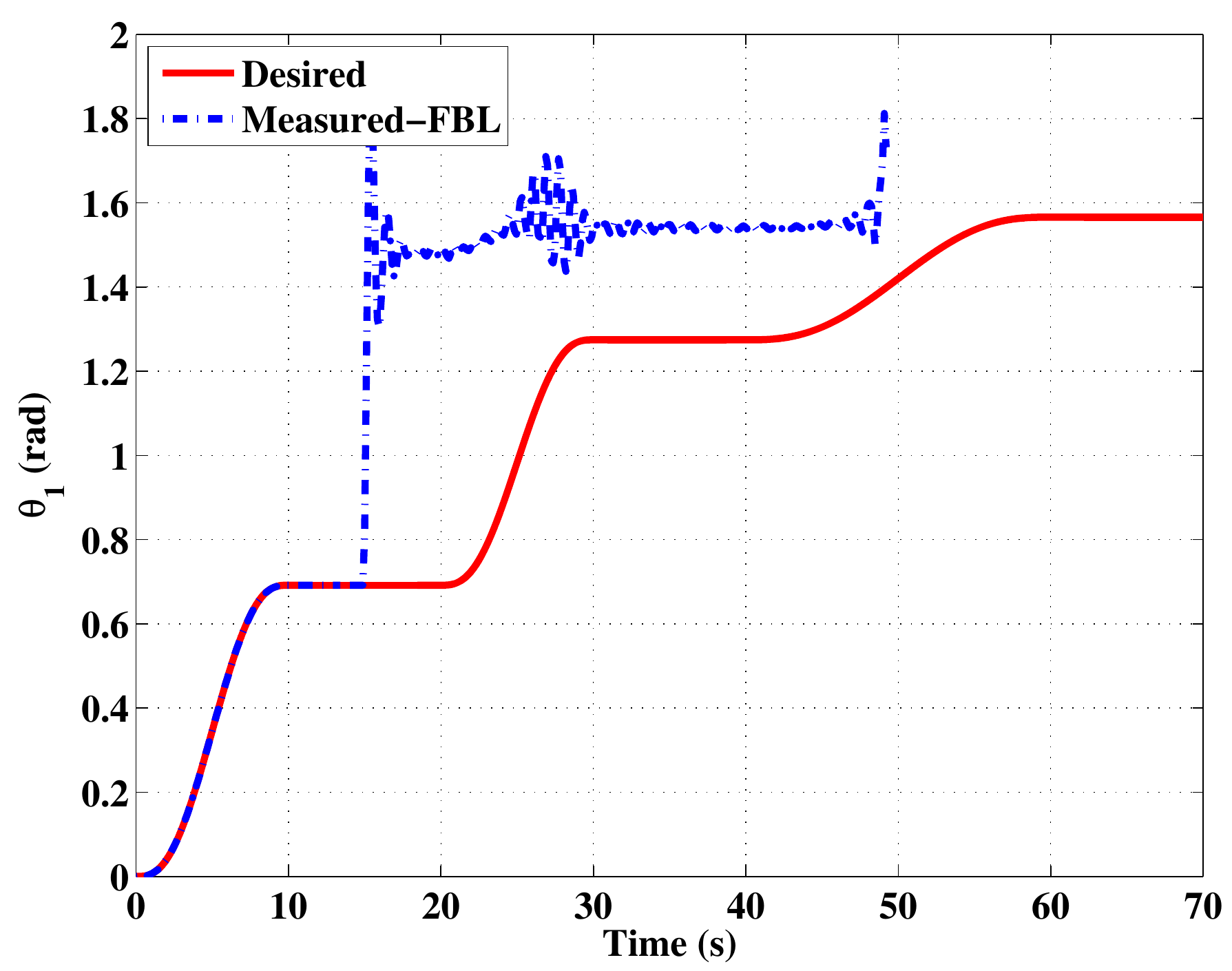}}&
 \subfigure[$\theta_2$ Response]{\includegraphics [width=0.5\columnwidth, height=5cm]{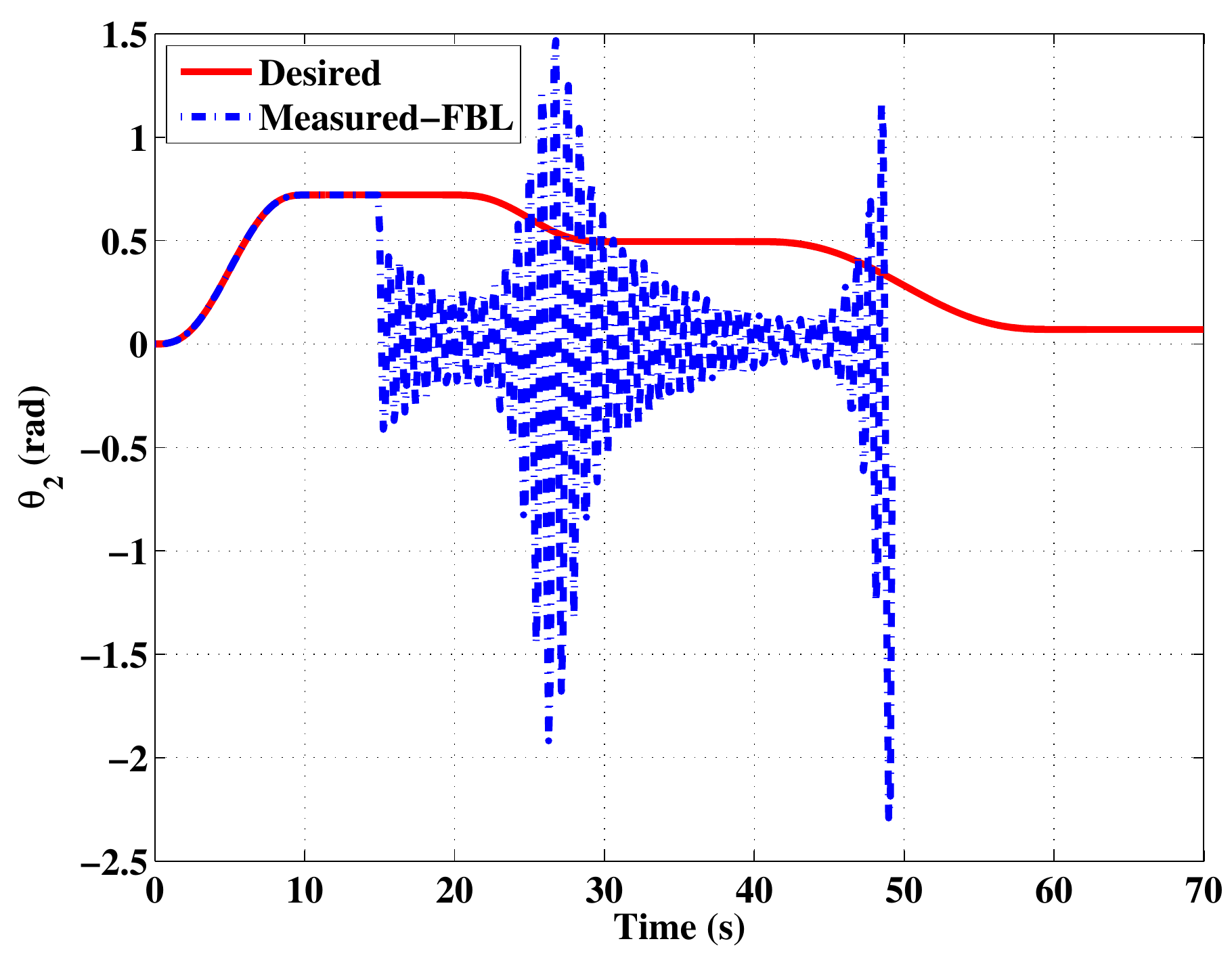}}\\
  \subfigure[$\phi$ Response]{\includegraphics [width=0.5\columnwidth, height=5cm]{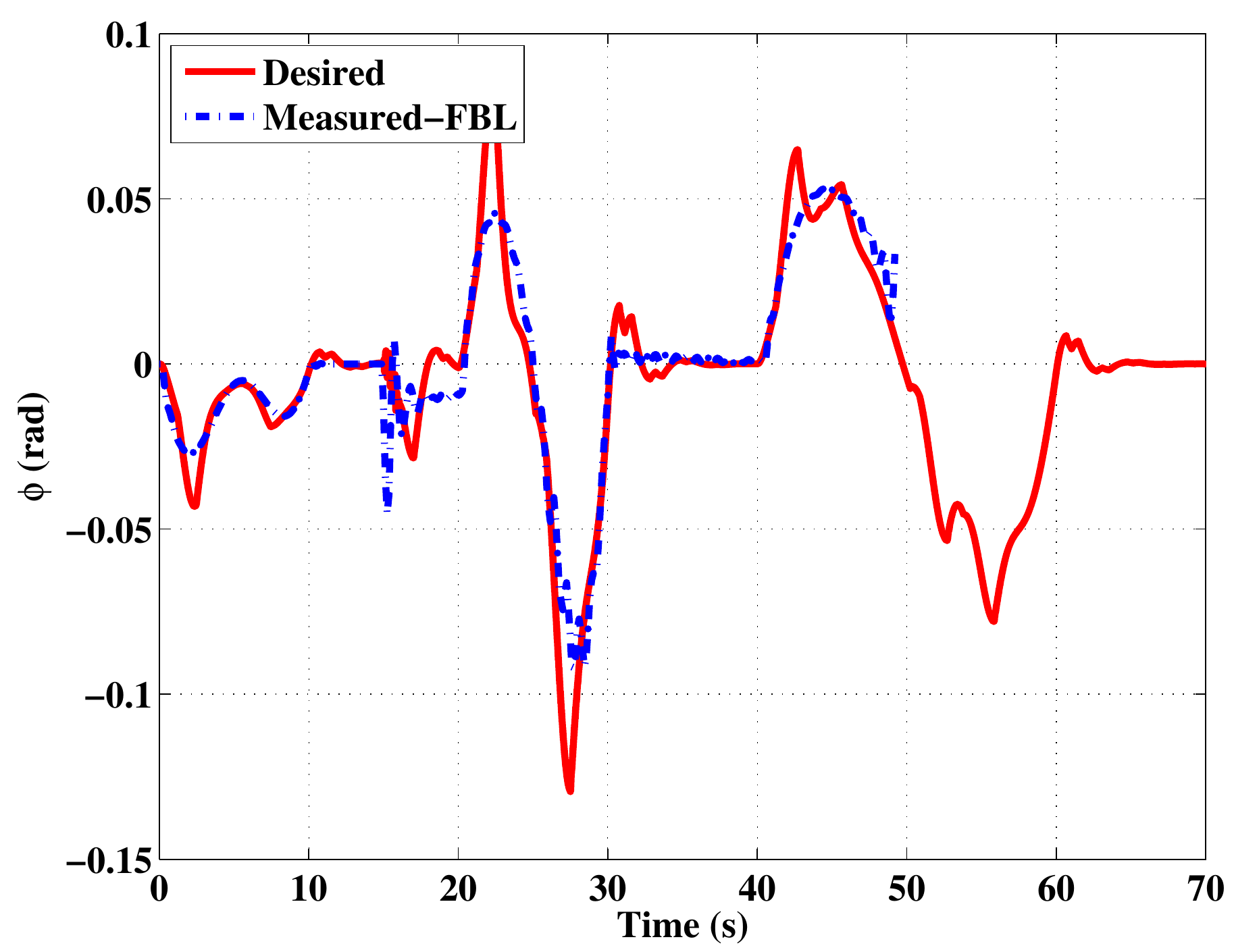}}&
 \subfigure[$\theta$ Response]{\includegraphics [width=0.5\columnwidth, height=5cm]{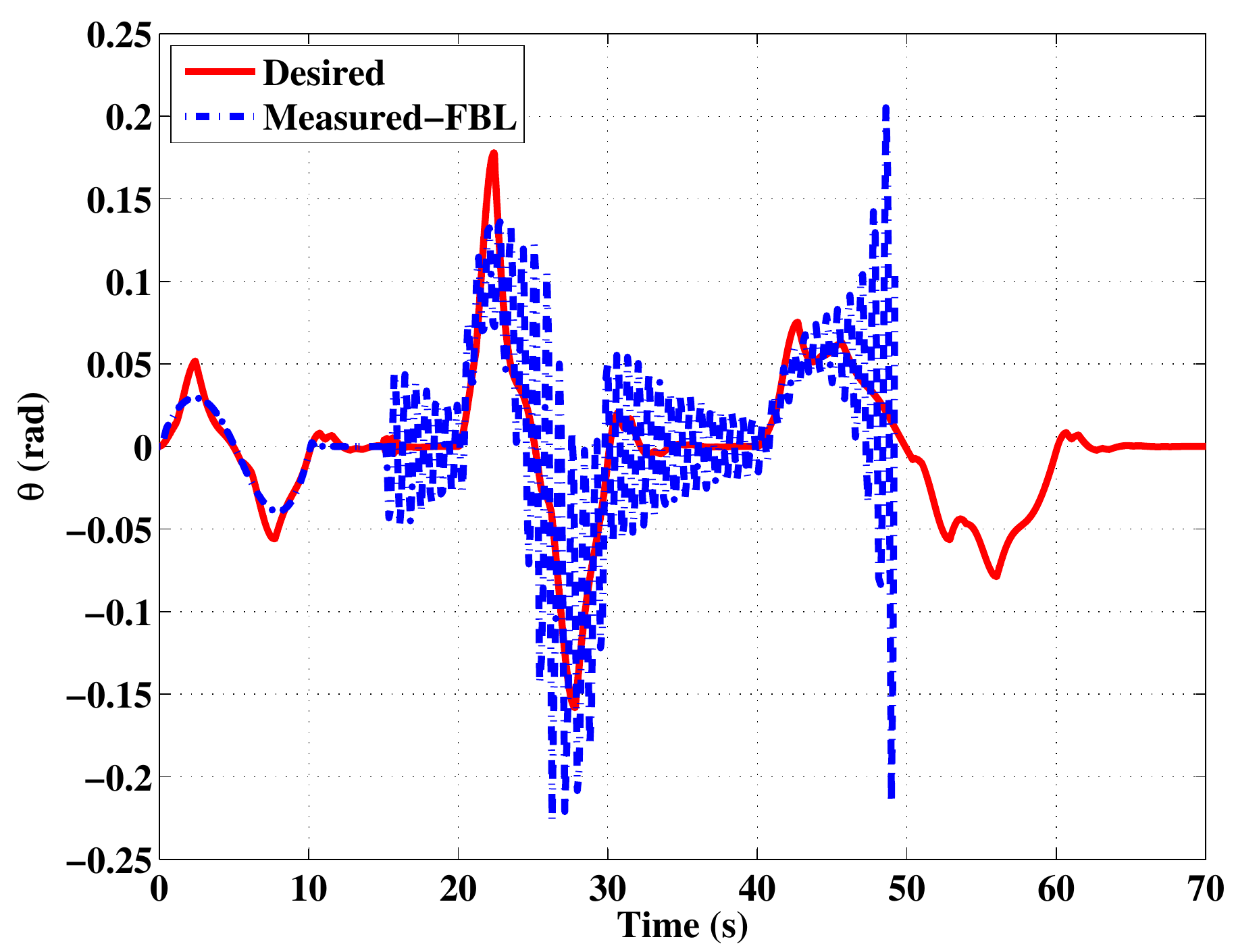}}
\end{tabular}
\caption{The Actual Response of Feedback Linearization Technique for the Quadrotor and Manipulator Variables: a) $X$, b) $Y$, c) $Z$, d) $\psi$, e) $\theta_1$, f) $\theta_2$, g) $\phi$, and h) $\theta$.}
\label {FBL_controller_Vechvariable}
\end{figure}

The controller are tested to stabilize and track the desired trajectories under the effect of picking a payload of value 150 g at instant 15 s and placing it at instant 65 s.
%===========================================================
\begin{table}
\caption{System Parameters}
\label{sys_par}
\begin{center}
\setlength{\tabcolsep}{4pt}
\begin{tabular}{|c||c||c||c||c||c|}
\hline
\hline
Par.	&Value	&Unit&	Par. &	Value&	Unit \\
\hline
$K_F$	& $1.667x10^{-5}$ & $Kg.m.rad^{-2}$ & $L_0$	& $30x10^{-3}$ & $m$ \\
\hline
$K_M$	&$3.965x10^{-7}$ &	$Kg.m^2.rad^{-2}$ &	$L_1$& $70x10^{-3}$&$m$ \\
\hline
$m$&	1 &	$kg$ &	$L_2$&	$85x10^{-3}$&	$m$ \\
\hline
$d$ &	$223.5 X 10^{-3}$ &	$m$ &$m_0$&	$30x10^{-3}$&	$kg$\\
\hline
$I_x$&	$13.215 X 10^{-3}$&	$N.m.s^2$	& $m_1$	&$55x10^{-3}$	&  $kg$\\
\hline
$I_y$	& $12.522 X 10^{-3}$ &	$N.m.s^2$ &	$m_2$ &	$112x10^{-3}$& $kg$\\
\hline
$I_z$&	$23.527 X 10^{-3}$&	$N.m.s^2$	& $I_r$	& $33.216 X 10^{-6}$& $N.m .s^2$\\
\hline
\hline
\end{tabular}
\end{center}
\end{table}
%===========================================================
\begin{table}
\caption{Feedback Linearization Controller Parameters}
\label{FBL parameters}
\begin{center}
\setlength{\tabcolsep}{4pt}
\begin{tabular}{|c||c||c||c|}
\hline
\hline
Par.	&Value	&	Par. &	Value \\
\hline
$[K_{p_z}  K_{d_z}  K_{i_z}]$ & $[16, 8, 0.01]$ &$[K_{p_\psi}  K_{d_\psi} K_{i_\psi}]$&$[16, 8, 0.01]$ \\
\hline
$[K_{p_\phi}  K_{d_\phi} K_{i_\phi}]$&	$[100, 8, 10]$&	$[K_{p_\theta1}  K_{d_\theta1} K_{i_\theta1}]$	& $[16, 8, 0.01]$\\
\hline
$[K_{p_\theta}  K_{d_\theta} K_{i_\theta}]$	& $[100, 8, 10]$ &	[ $[K_{p_\theta2}  K_{d_\theta2} K_{i_\theta2}]$	&  $[16, 8, 0.01]$ \\
\hline
\hline
\end{tabular}
\end{center}
\end{table}
%===========================================================
The simulation results are presented in Figure \ref{FBL_controller_Vechvariable}. These results show that the controller design based on feedback linearization can track the desired trajectories before picking the payload, but at the instant of picking and then holding the payload, it fails to track the desired trajectories and the system becomes unstable even if the payload is released.
%==============================================
%==============================================

\begin{figure}
\centering
\begin{tabular}{cc}
\subfigure[$x_{ee}$ Response]{\includegraphics[width=0.5\columnwidth]{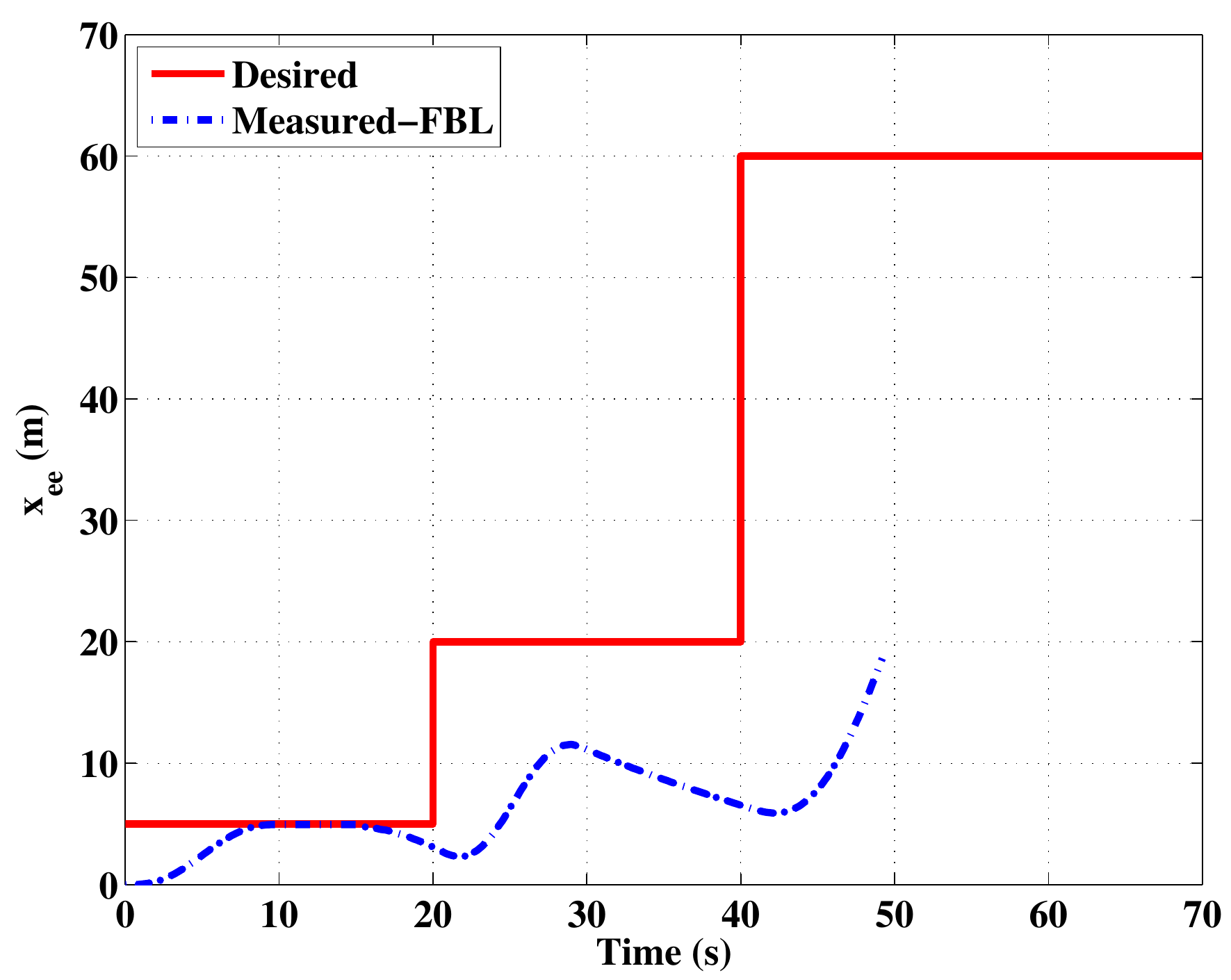}}&
\subfigure[$y_{ee}$ Response]{\includegraphics [width=0.5\columnwidth]{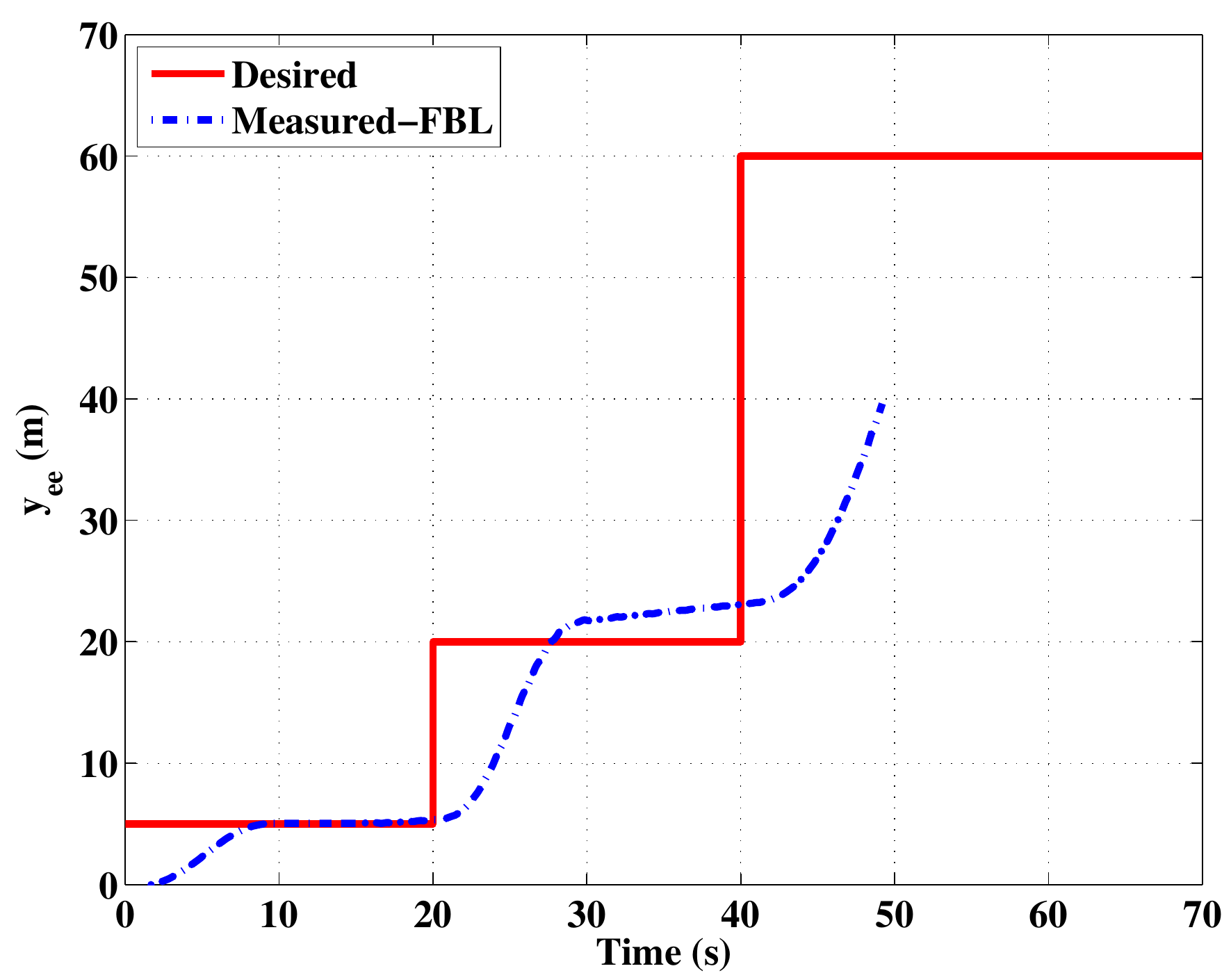}} \\
\subfigure[$z_{ee}$ Response]{\includegraphics [width=0.5\columnwidth]{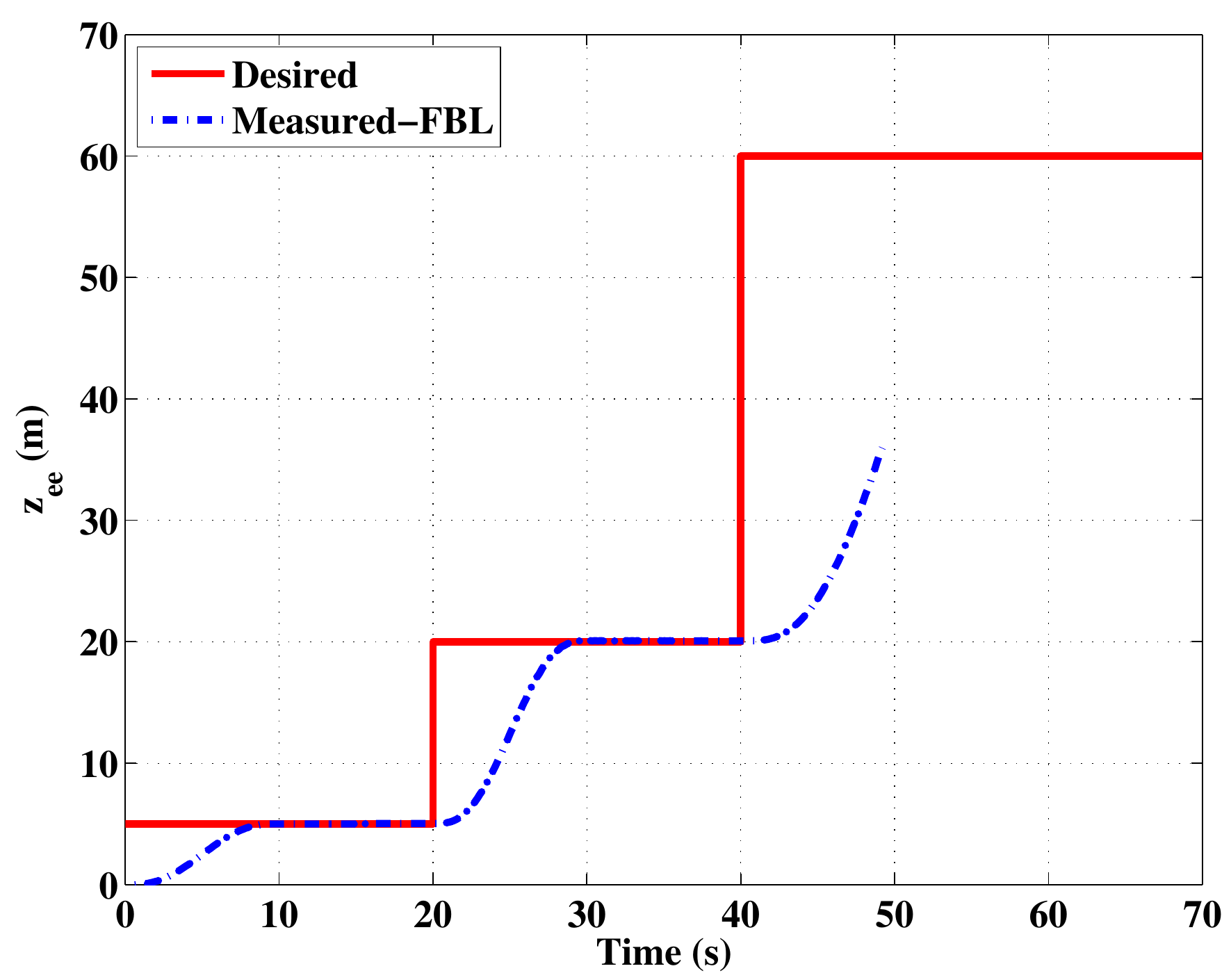}} &
\subfigure[$\phi_{ee}$ Response]{\includegraphics [width=0.5\columnwidth]{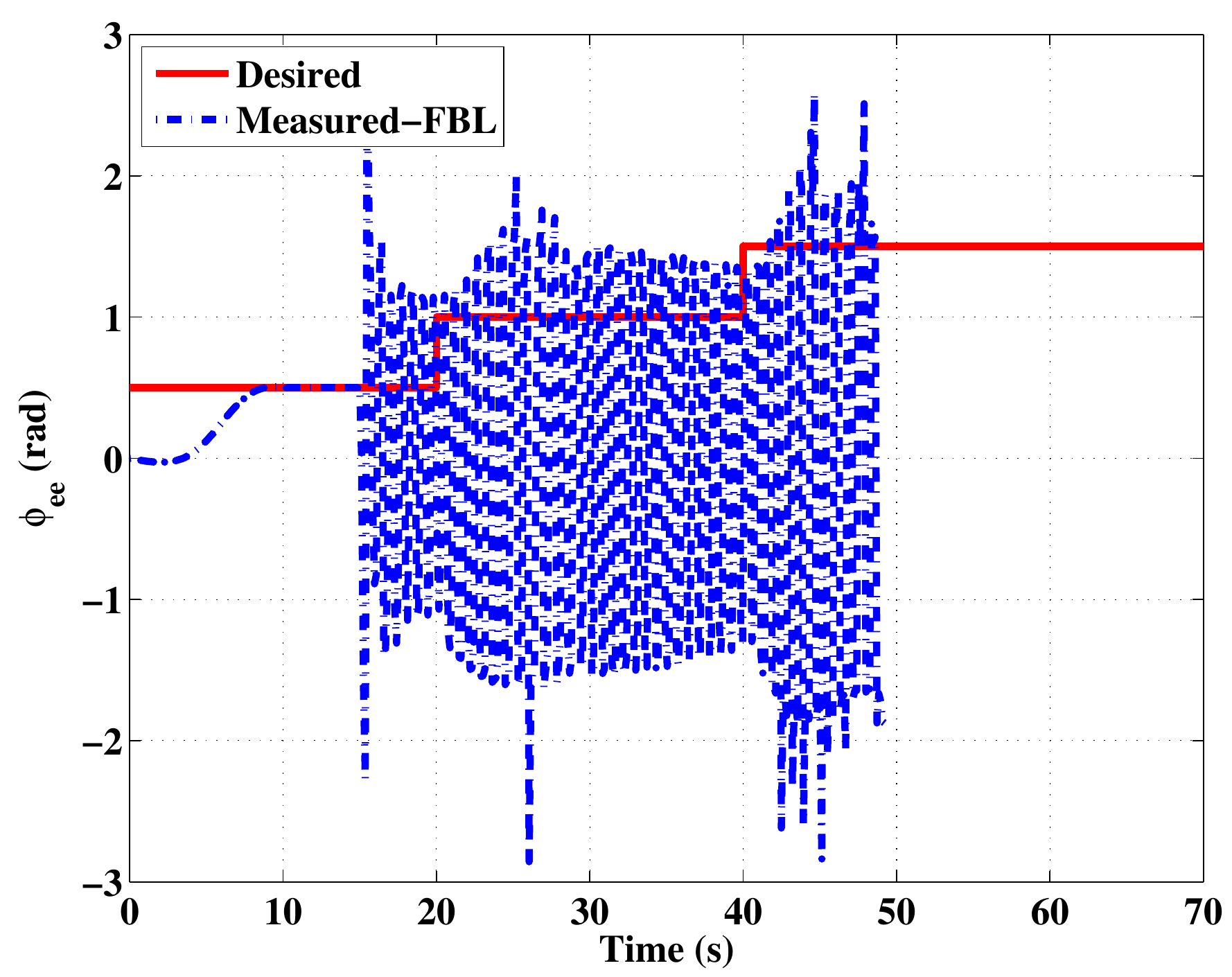}}\\
 \subfigure[$\theta_{ee}$ Response]{\includegraphics [width=0.5\columnwidth]{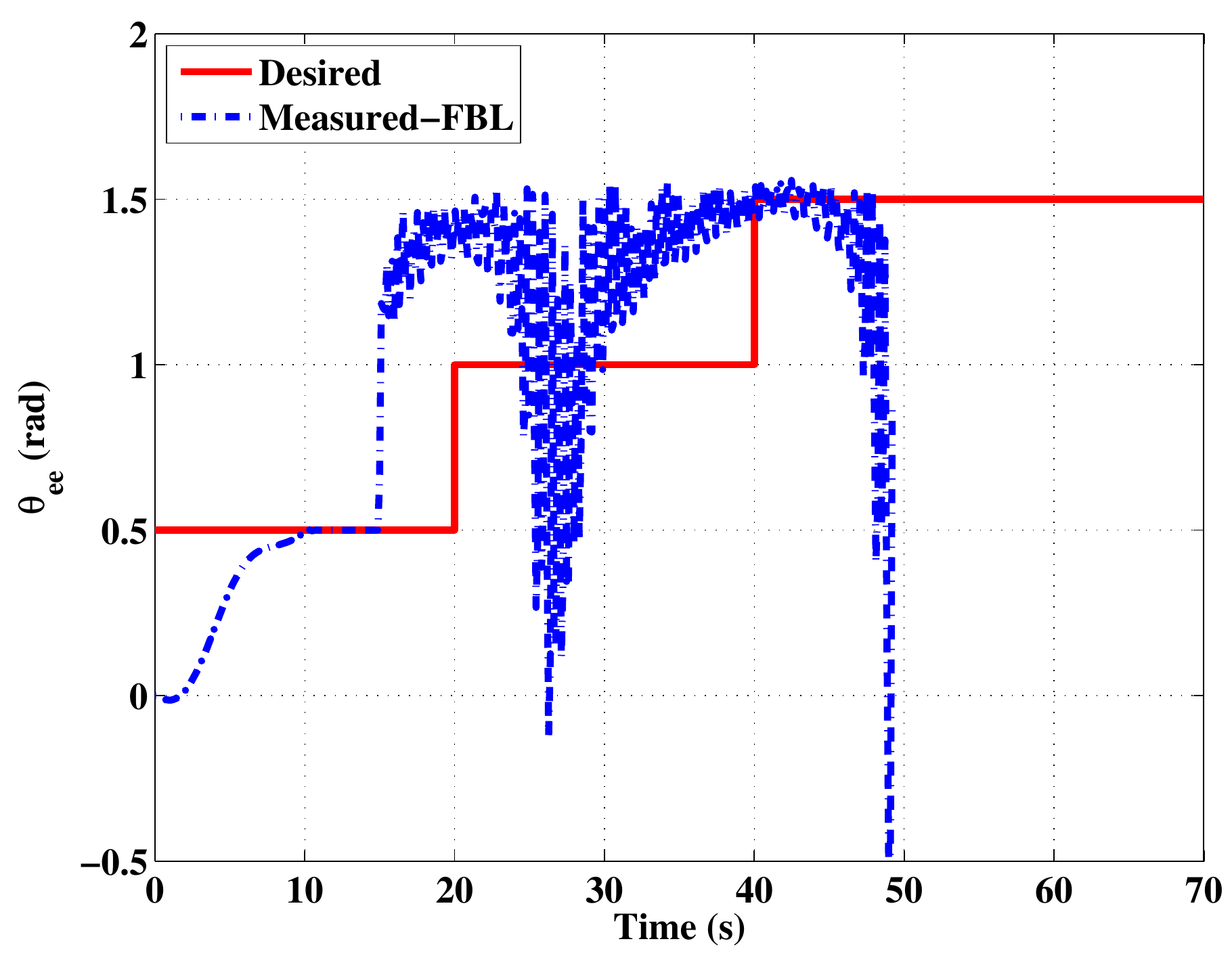}}&
 \subfigure[$\psi_{ee}$ Response]{\includegraphics [width=0.5\columnwidth]{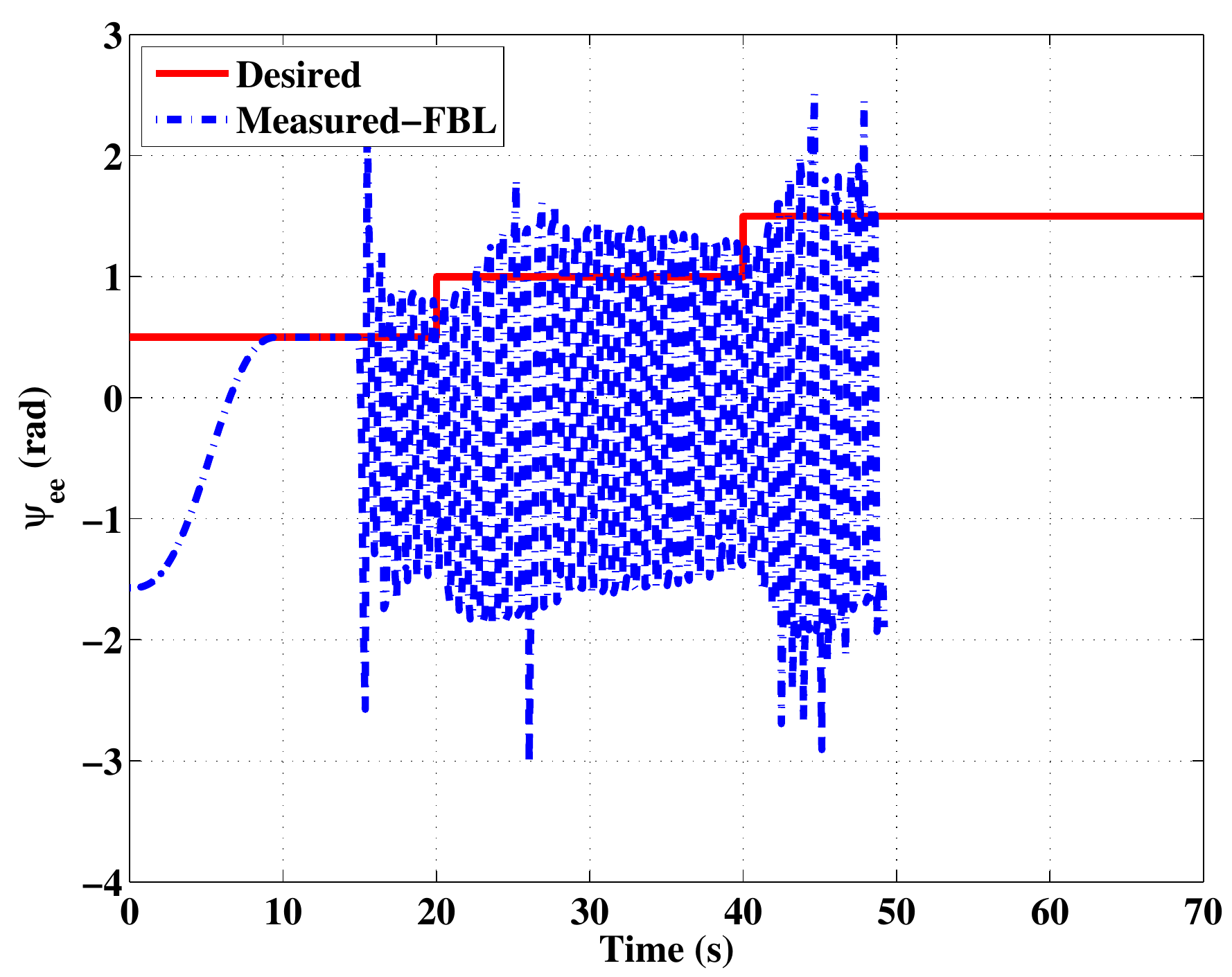}}
\end{tabular}
\caption{The Actual Response of Feedback Linearization Technique for the End Effector Position and Orientation: a) $x_{ee}$, b) $y_{ee}$, c) $z_{ee}$, d) $\phi_{ee}$, e) $\theta_{ee}$, and f) $\psi_{ee}$.}
\label {FBL_controller_endeffector}
\end{figure}

The end effector position and orientation can be found from the forward kinematics (see Figure \ref{FBL_controller_endeffector}).

The derived forward kinematics in Chapter \ref{ch:modeling} is used only at the end position where $\phi$ and $\theta$ are zero. If you want to use it at trajectory, you must modify it.
%==============================================
%==============================================

From the above discussion and results, the following items can be concluded about the performance of feedback linearization technique:
\begin{itemize}
  \item It provides a good trajectory tracking capabilities but it fail to make system stable against adding the payload.
  \item Due to the high nonlinearities and the complex dynamics in the system, the control laws are very complex and difficult to be implemented onboard (implementation in real time).
  \item Therefore, their is a need for an adaptive control technique to overcome the mentioned problems with lower complexity.
\end{itemize}
\section{Direct Fuzzy Logic Control\label{sec:DFLC}}
Recently, fuzzy logic control \cite{Passino_FLC_Book, FLC_Theory_app_Book} has become an alternative to conventional control algorithms to deal with complex processes and combine the advantages of classical controllers and human operator experience.

An intelligent controller, based on Direct Fuzzy Logic Control (DFLC), for a quadrotor was designed and presented in \cite{Fuzzy_Quad_only}. In this work, a modification of this technique is done and used to control the quadrotor-manipulator system to achieve the required objectives mentioned in Section \ref{sec:controldesign}.

\begin{figure}
      \centering
      \includegraphics[width=0.8\columnwidth, height=8cm]{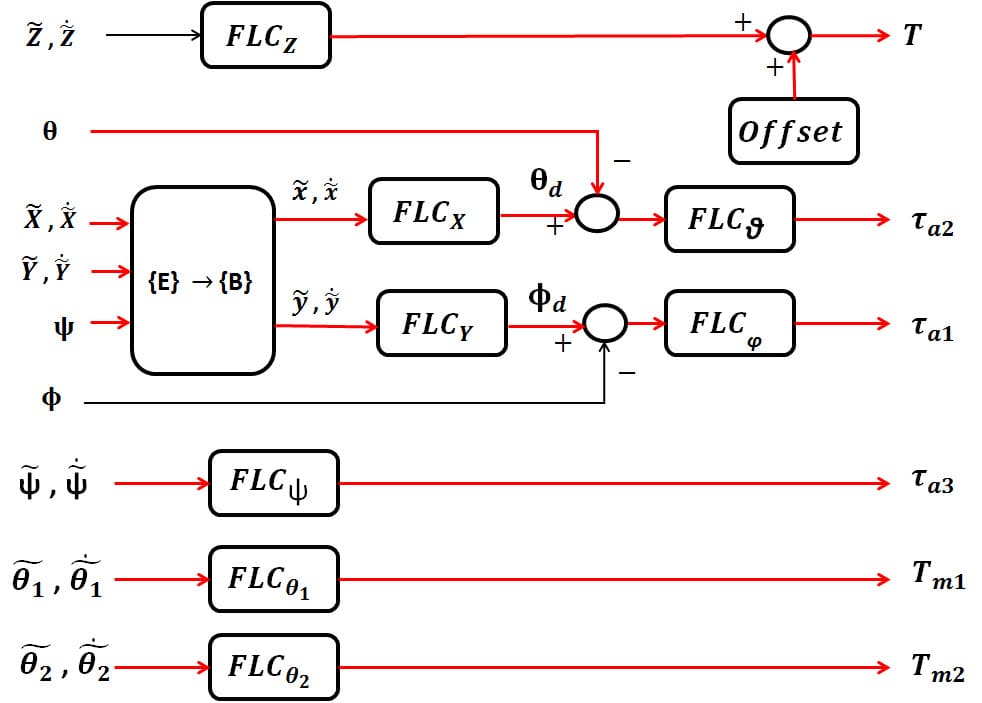}
      %\framebox{\parbox{1cm}{(1.a)}}
      \caption{Details of the Controller Block in Case of DFLC}
      \label{DFLC}
\end{figure}

In Figure \ref{DFLC} , three fuzzy controllers are designed to control the quadrotor's roll ($\phi$), pitch($\theta$) and yaw($\psi$) angles, denoted by  $FLC_\phi$, $FLC_\theta$, and  $FLC_\psi$ , respectively, with the  former two serving as attitude stabilizers. Three fuzzy  controllers, $FLC_x$, $FLC_y$ and $FLC_z$, are designed  to control the quadrotor's position. Also two fuzzy controllers $FLC_{\theta_1}$ and $FLC_{\theta_2}$ are designed to control the two joints' angles of the manipulator.

All eight fuzzy controllers have similar inputs that are:
\begin{itemize}
\item The error $e =(\tilde{.})= (.)_d - (.)$ , which  is  the  difference  between  the  desired  signal  $(.)_d$ and  its  measured value (.). This input is  normalized to the interval [-1, +1].
\item The error rate $c$, which is normalized to the interval [-3, +3].
\end{itemize}

\begin{figure}
      \centering
      \subfigure[Input Variable error $e$]{\includegraphics[width=0.5\columnwidth, height=4cm]
      {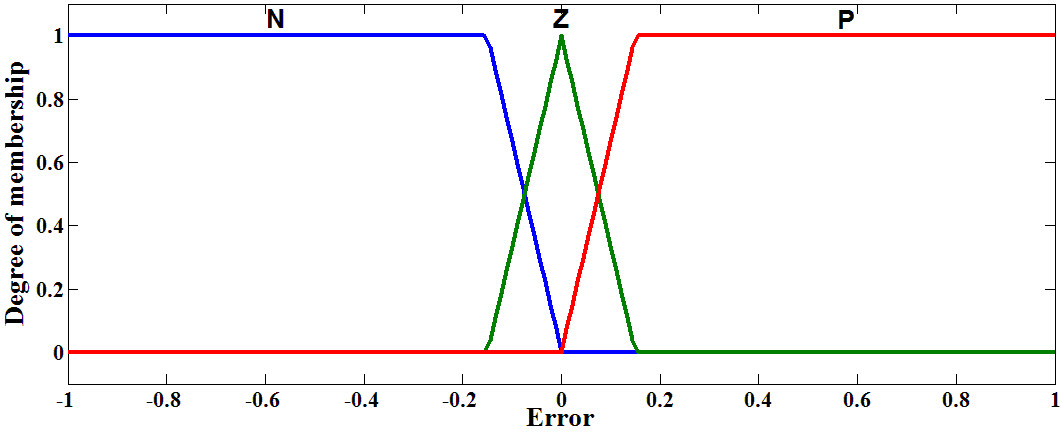}}\hspace{3cm}
      \subfigure[Input Variable error rate $c$]{\includegraphics[width=0.5\columnwidth, height=4cm]
      {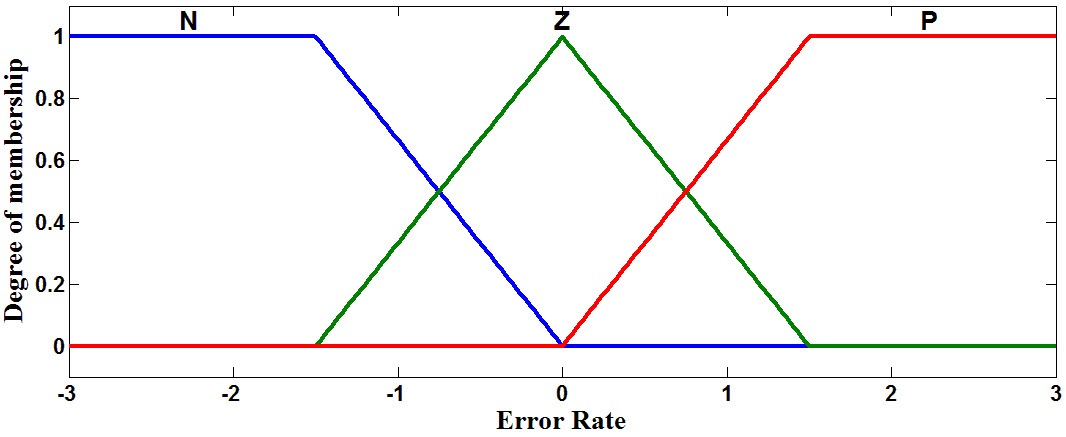}}\hspace{3cm}
      \subfigure[Output Variable $u$]{\includegraphics[width=0.5\columnwidth, height=4cm]{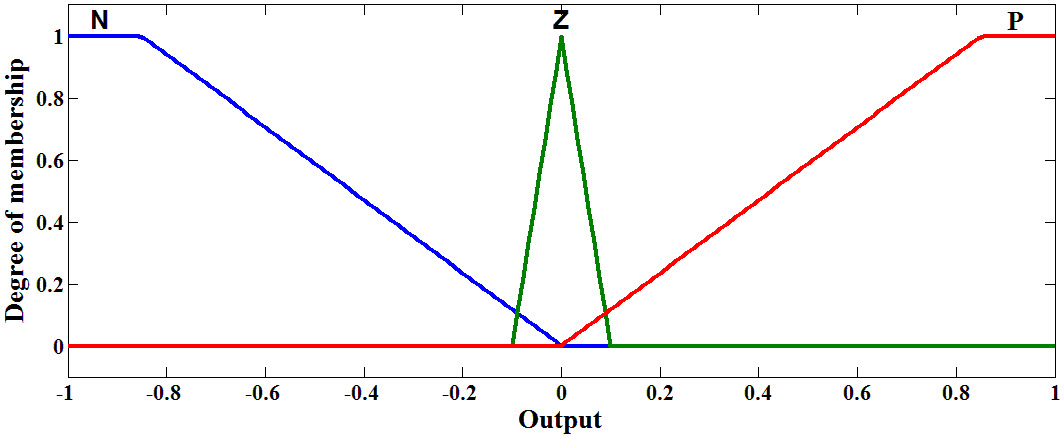}}
      \caption{Input and Output Membership Functions of DFLC: a) $Error$ , b) $Error$ $Rate$, and c) $Output$.}
      \label{DFLC_MSFs}
\end{figure}

In  this  control  strategy,  the  desired  pitch  and  roll  angles,  $\theta_d$  and  $\phi_d$ ,  are  not  explicitly provided  to  the  controller.  Instead,  they  are  continuously  calculated  by  controllers  $FLC_x$ and $FLC_y$  in  such  a  way  that  they  stabilize  the  quadrotor's  attitude. First, we convert the error and its rate of $X$ and $Y$ that is defined in the inertial frame into their corresponding values defined in the body frame. This conversion is done using the transformation matrix defined in (\ref{rot_mat}) assuming small angles ($\phi$ and $\theta$) as following:
\begin{equation}
\tilde{x} = \tilde{X}\cos(\psi) + \tilde{Y}\sin(\psi)
\label{x_error_body}
\end{equation}
\begin{equation}
\dot{\tilde{x}} = \dot{\tilde{X}}\cos(\psi) + \dot{\tilde{Y}}\sin(\psi)
\label{xd_error_body}
\end{equation}
\begin{equation}
\tilde{y} = \tilde{X}\sin(\psi) - \tilde{Y}\cos(\psi)
\label{y_error_body}
\end{equation}
\begin{equation}
\dot{\tilde{y}} = \dot{\tilde{X}}\sin(\psi) - \dot{\tilde{Y}}\cos(\psi)
\label{yd_error_body}
\end{equation}
%=============================================================
The input and output membership functions of each $FLC$ are tuned and chosen to be as shown in Figure \ref{DFLC_MSFs} with the linguistic values $N$ (Negative), $Z$ (Zero), and $P$ (Positive). Also the input and output scaling factors for the error, change of error, and fuzzy output ($K_{e_i}$, $K_{c_i}$, and $K_{u_i}$; $i$ = $x$, $y$, $z$, $\phi$, $\theta$, $\psi$, $\theta_1$, $\theta_2$) of each $FLC$ are tuned  such that required performance is obtained.

The rule base of each $FLC$ block is the same and is designed to provide a PD-like fuzzy controller. This rule base is given in Table \ref{rule base FLC}. A Mamdani  fuzzy inference method is used with a min-max operator for  the aggregation and the center of gravity method for defuzzification.

There is a need to add an 'Offset' value to the control signal from the $FLC_Z(T)$ in order to counter balance the weight of the quadrotor. This value is tuned.

It is important to note that the fuzzy controllers are designed in light of the knowledge acquired on the system's behavior and from its dynamic model. This property sets   the fuzzy controllers apart from conventional controllers which depend on the plant's mathematical model \cite{Fuzzy_Quad_only}.
% ===========================================
% ===========================================
\begin{table}
\caption{Rule Base of the DFLC}
\label{rule base FLC}
\begin{center}
\begin{tabular}{|c||c||c||c|}
\hline
\hline
e/c & N & Z & P \\
\hline
N & N & N & Z\\
\hline
Z & N & Z & P \\
\hline
P & Z & P& P \\
\hline
\hline
\end{tabular}
\end{center}
\end{table}
%================================================
\subsection{Simulation Results}
The system equations of motion and the control laws are simulated using MATLAB/SIMULINK program.

The controller parameters of the direct fuzzy logic controller are given in Table \ref{DFLC parameters}. Those parameters are tuned to get the required system performance.

The controller are tested to stabilize and track the desired trajectories under the effect of picking a payload of value 150 g at instant 15 s and placing it at instant 65 s.

\begin{figure}
\centering
\begin{tabular}{cc}
 \subfigure[$X$ Response]{\includegraphics[width=0.5\columnwidth, height=5cm]{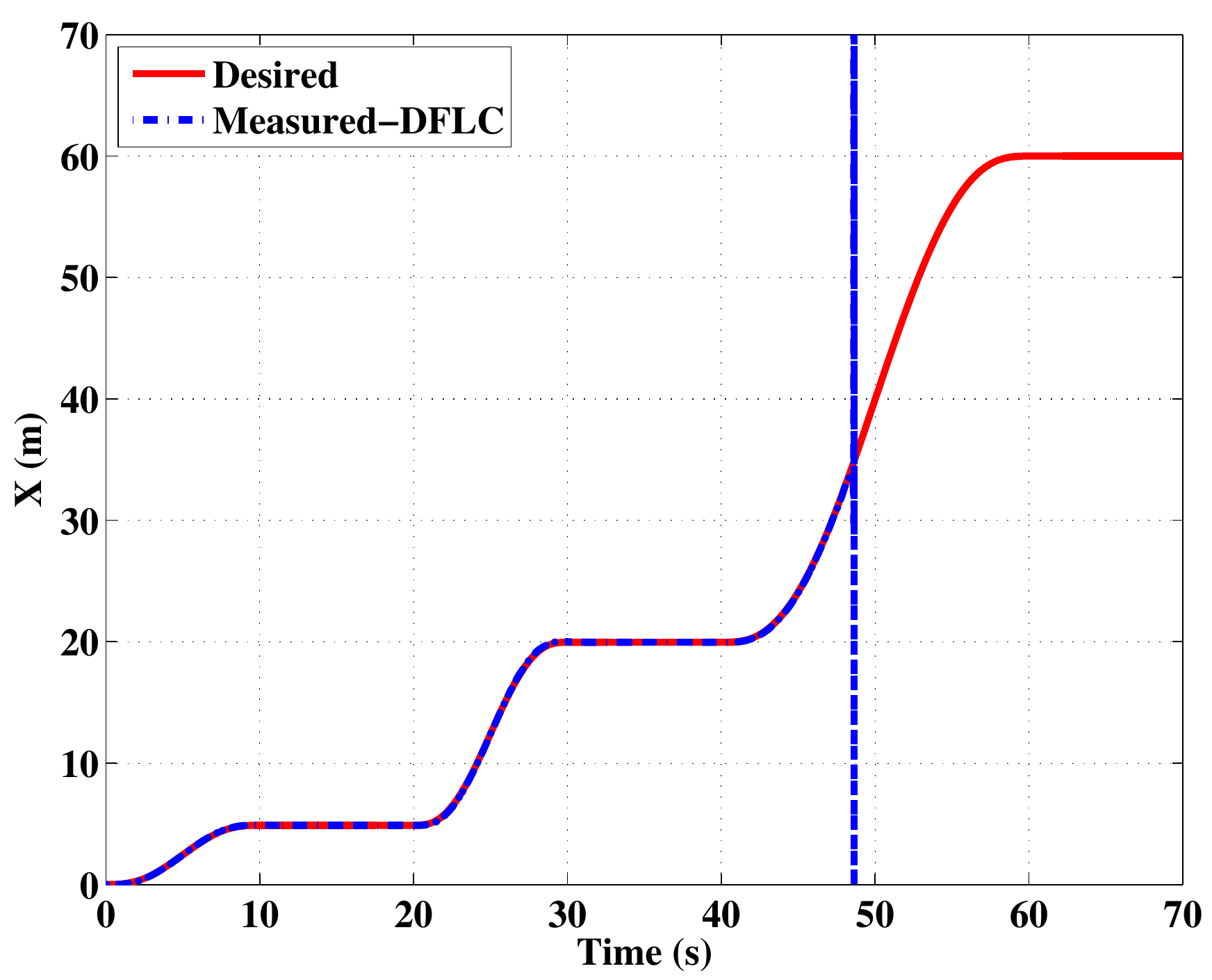}}&
 \subfigure[$Y$ Response]{\includegraphics [width=0.5\columnwidth, height=5cm]{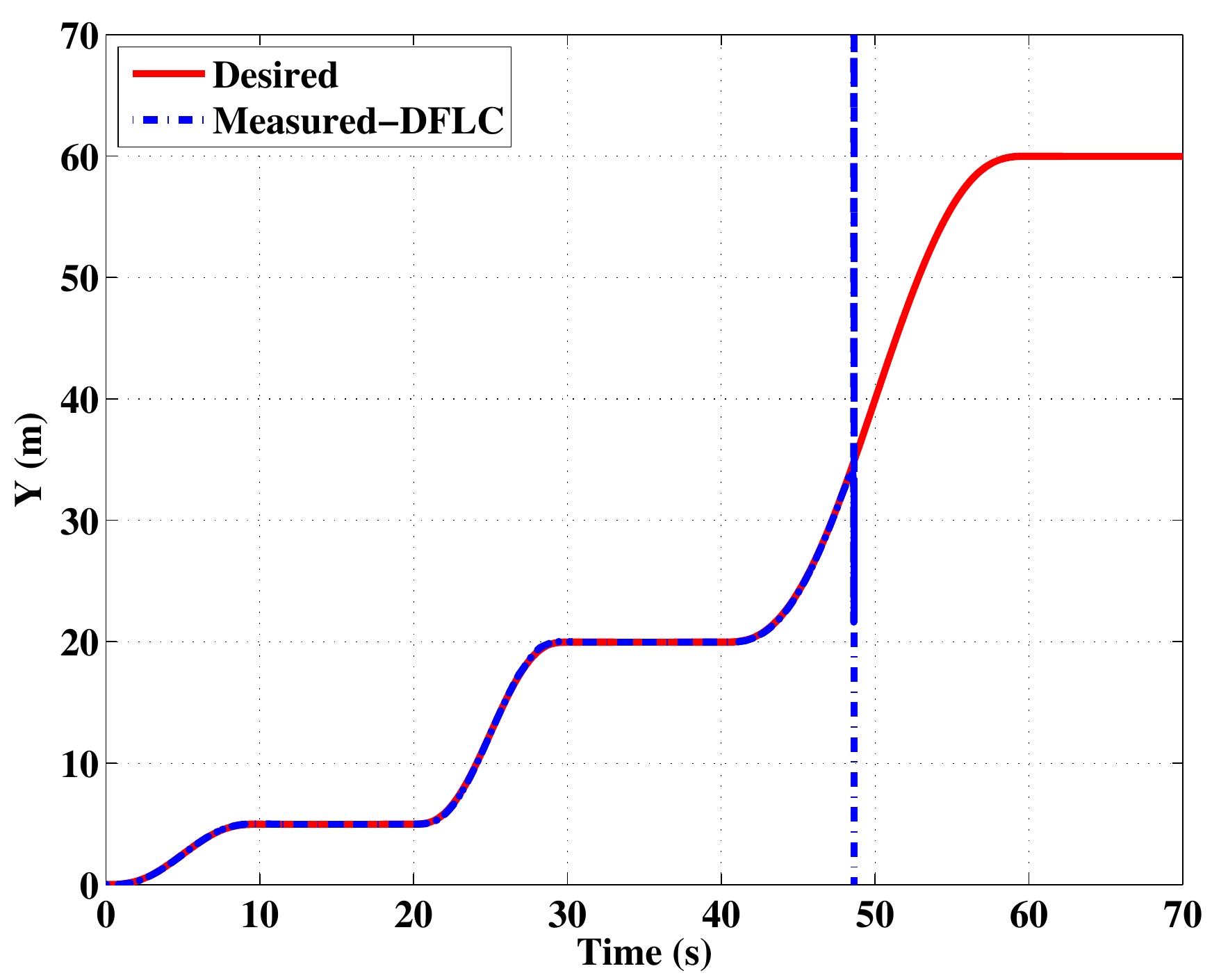}} \\
 \subfigure[$Z$ Response]{\includegraphics [width=0.5\columnwidth, height=5cm]{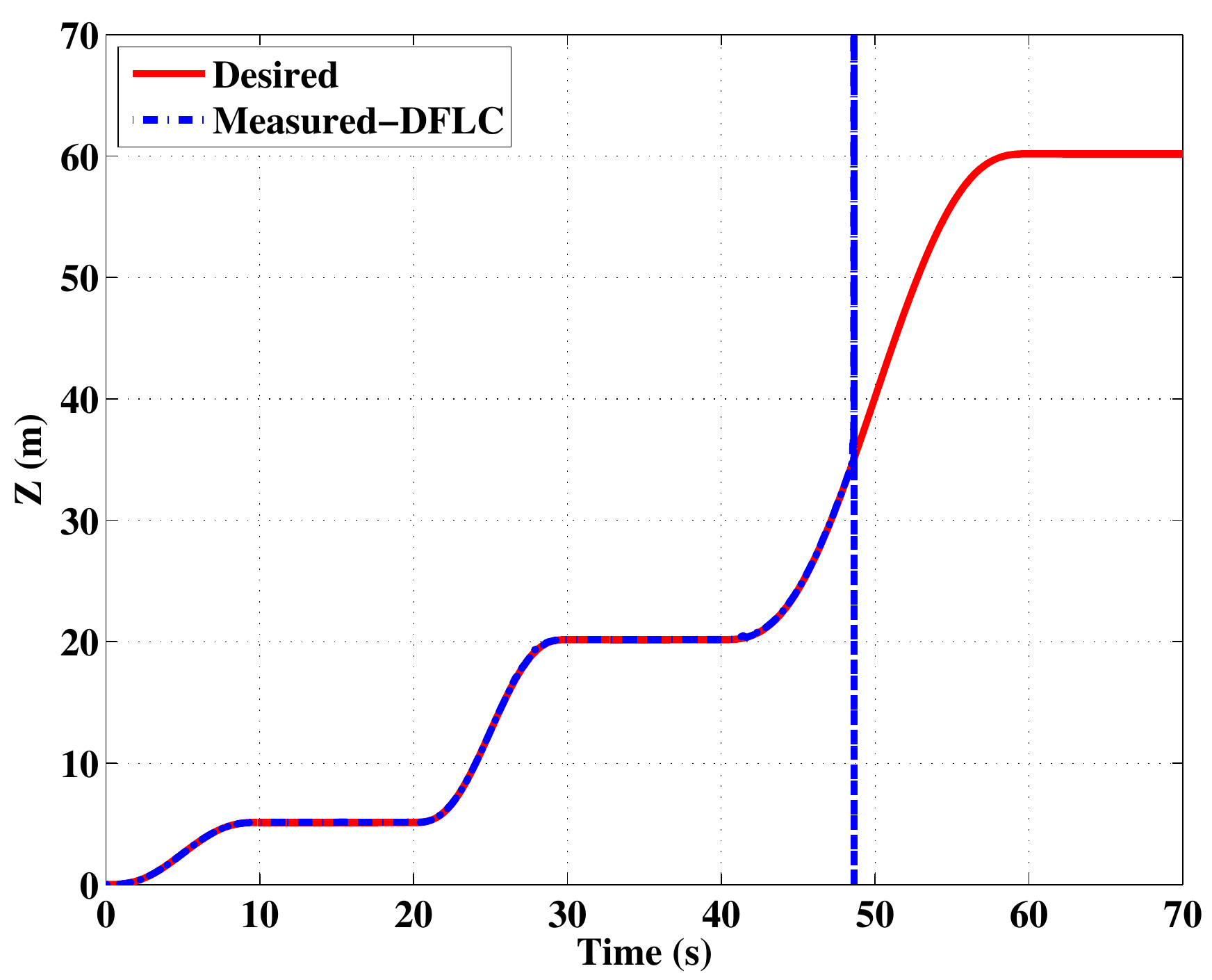}} &
 \subfigure[$\psi$ Response]{\includegraphics [width=0.5\columnwidth, height=5cm]{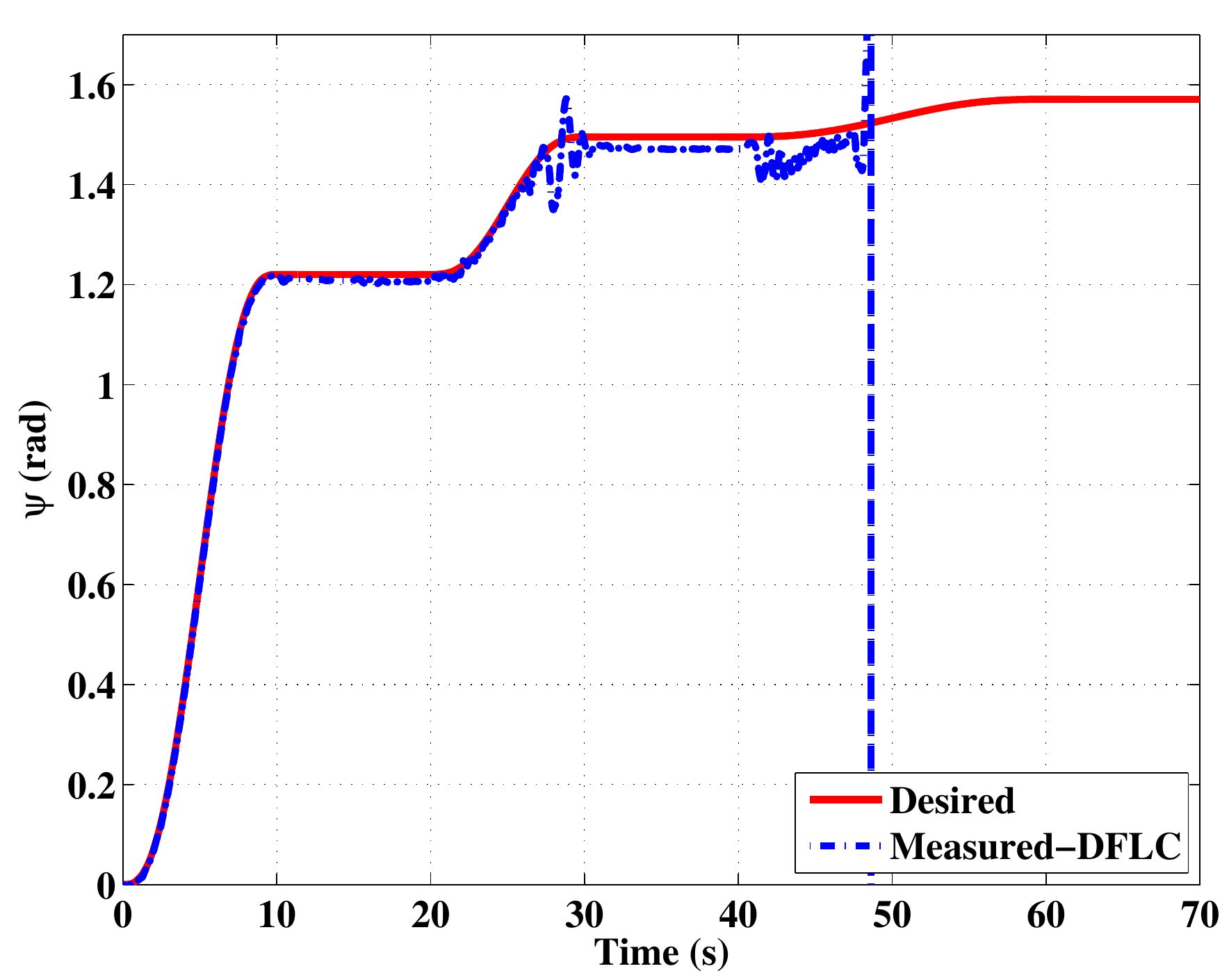}}\\
 \subfigure[$\theta_1$ Response]{\includegraphics [width=0.5\columnwidth, height=5cm]{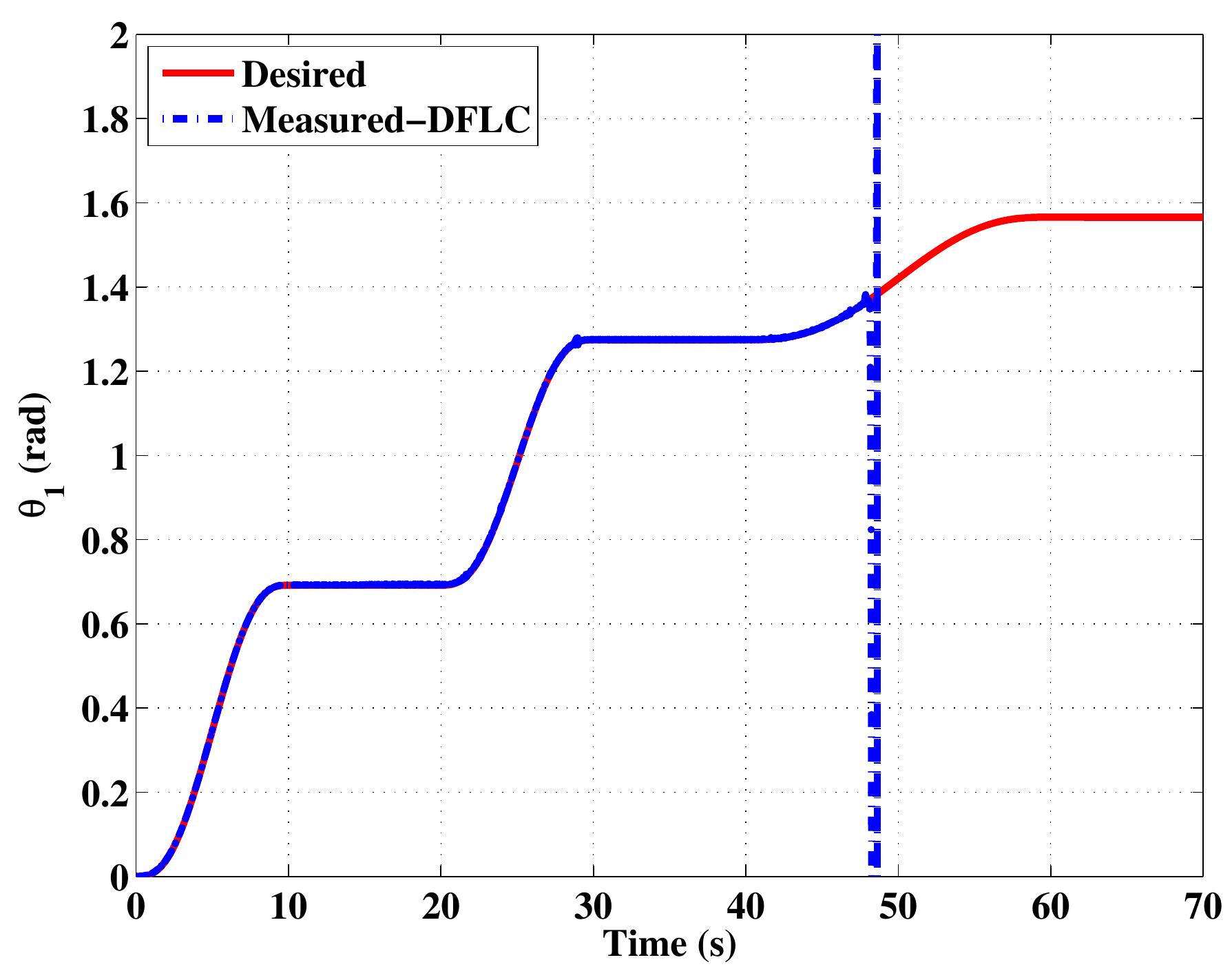}}&
 \subfigure[$\theta_2$ Response]{\includegraphics [width=0.5\columnwidth, height=5cm]{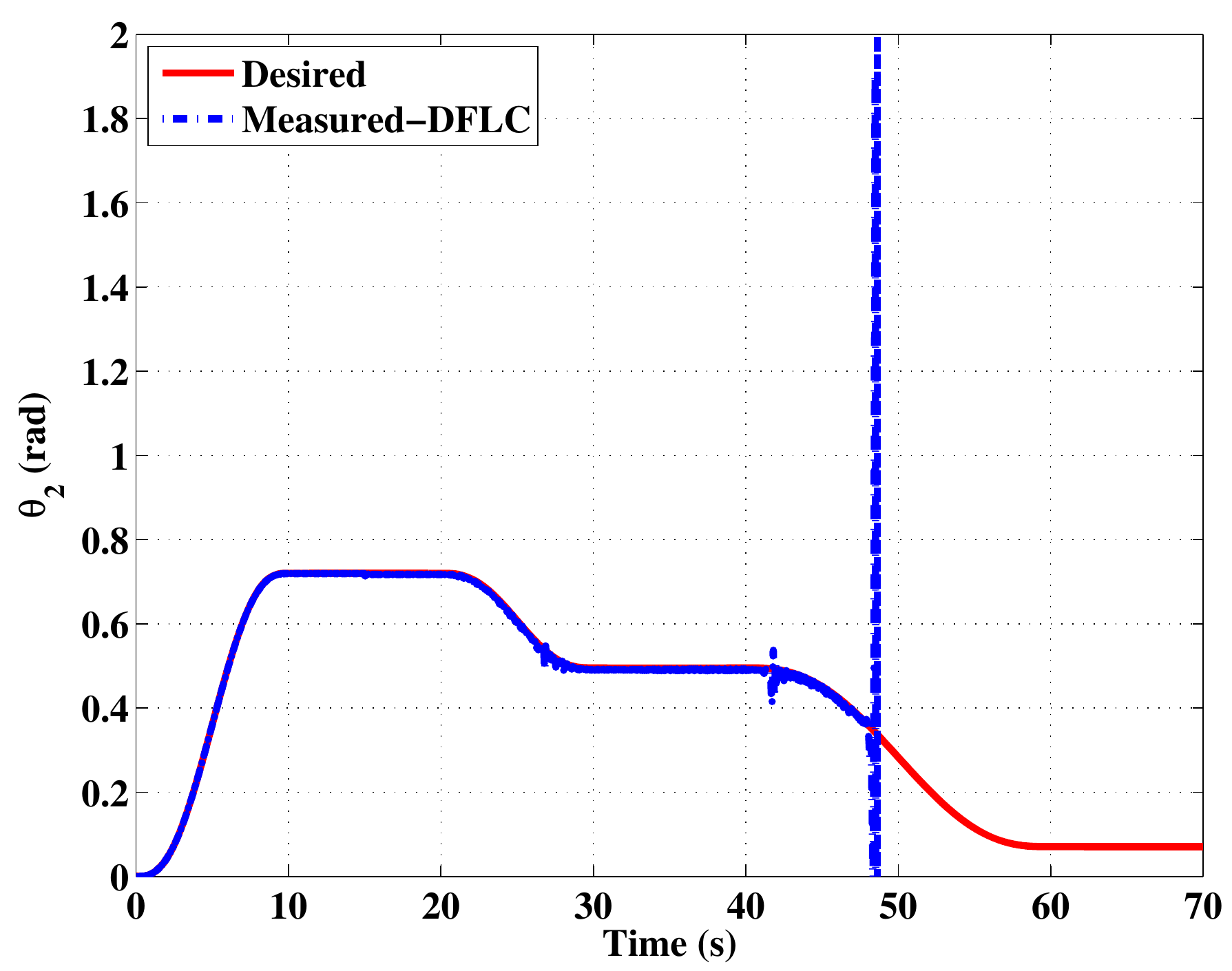}}\\
 \subfigure[$\phi$ Response]{\includegraphics [width=0.5\columnwidth, height=5cm]{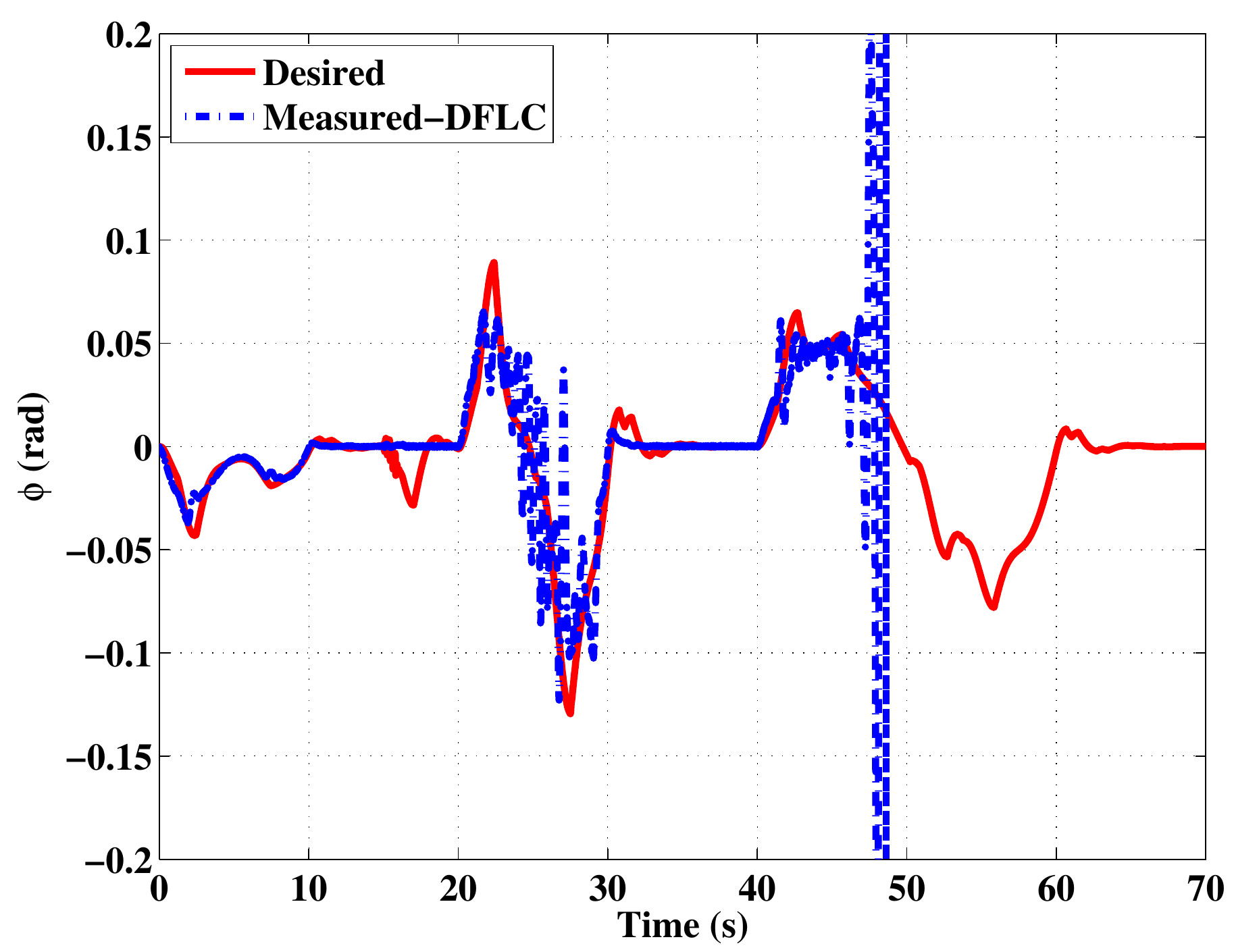}}&
 \subfigure[$\theta$ Response]{\includegraphics [width=0.5\columnwidth, height=5cm]{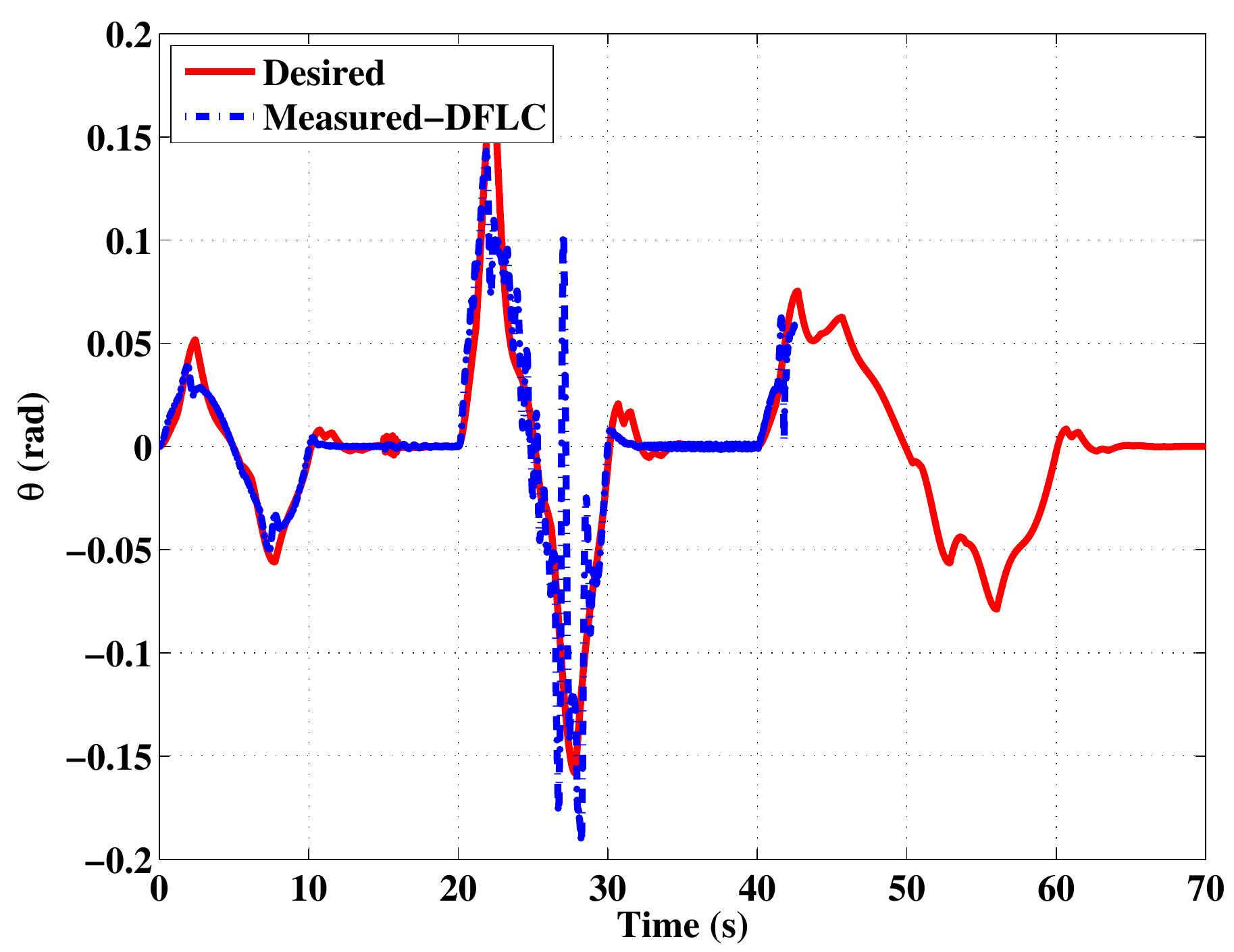}}
\end{tabular}
\caption{The Actual Response of DFLC Technique for the Quadrotor and Manipulator Variables: a) $X$, b) $Y$, c) $Z$, d) $\psi$, e) $\theta_1$, f) $\theta_2$, g) $\phi$, and h) $\theta$.}
\label {DFLC_controller_Vechvariable}
\end{figure}

The simulation results are presented in Figure \ref{DFLC_controller_Vechvariable}. These results show that DFLC  is able to track the desired trajectories before, during picking, and holding the payload. However, the DFLC fails to track the desired trajectories during changing the region of operation (operating point) because it need to retune its scaling factors.
%=================================================
\begin{table}
\caption{DFLC Parameters}
\label{DFLC parameters}
\begin{center}
\setlength{\tabcolsep}{2pt}
\begin{tabular}{|c||c||c||c|}
\hline
\hline
Par.	&Value	&	Par. &	Value \\
\hline
$[K_{e_x}  K_{c_x}  K_{u_x}]$ & $[.007,.05,5]$ &$[K_{e_y}  K_{c_y} K_{u_y}]$&$[.007,.05,5]$ \\
\hline
$[K_{e_z}  K_{c_z}  K_{u_z}]$ & $[1, .3, 16.5]$ &$[K_{e_\psi}  K_{c_\psi} K_{u_\psi}]$&$[1,.5,0.2]$ \\
\hline
$[K_{e_\phi}  K_{c_\phi} K_{u_\phi}]$&	$[.5,.5,9]$&	$[K_{e_{\theta_1}}  K_{c_{\theta_1}} K_{u_{\theta_1}}]$	& $[2, .05,4]$\\
\hline
$[K_{e_\theta}  K_{c_\theta} K_{u_\theta}]$	& $[.5,.5 , 10]$ &	[ $[K_{e_{\theta_2}}  K_{c_{\theta_2}} K_{u_{\theta_2}}]$	&  $[5, .3,0.3]$ \\
\hline
$Offset$& $7.85$ N &	 &  \\
\hline
\hline
\end{tabular}
\end{center}
\end{table}
%==============================================
%==============================================

\begin{figure}
\centering
\begin{tabular}{cc}
 \subfigure[$x_{ee}$ Response]{\includegraphics[width=0.5\columnwidth]{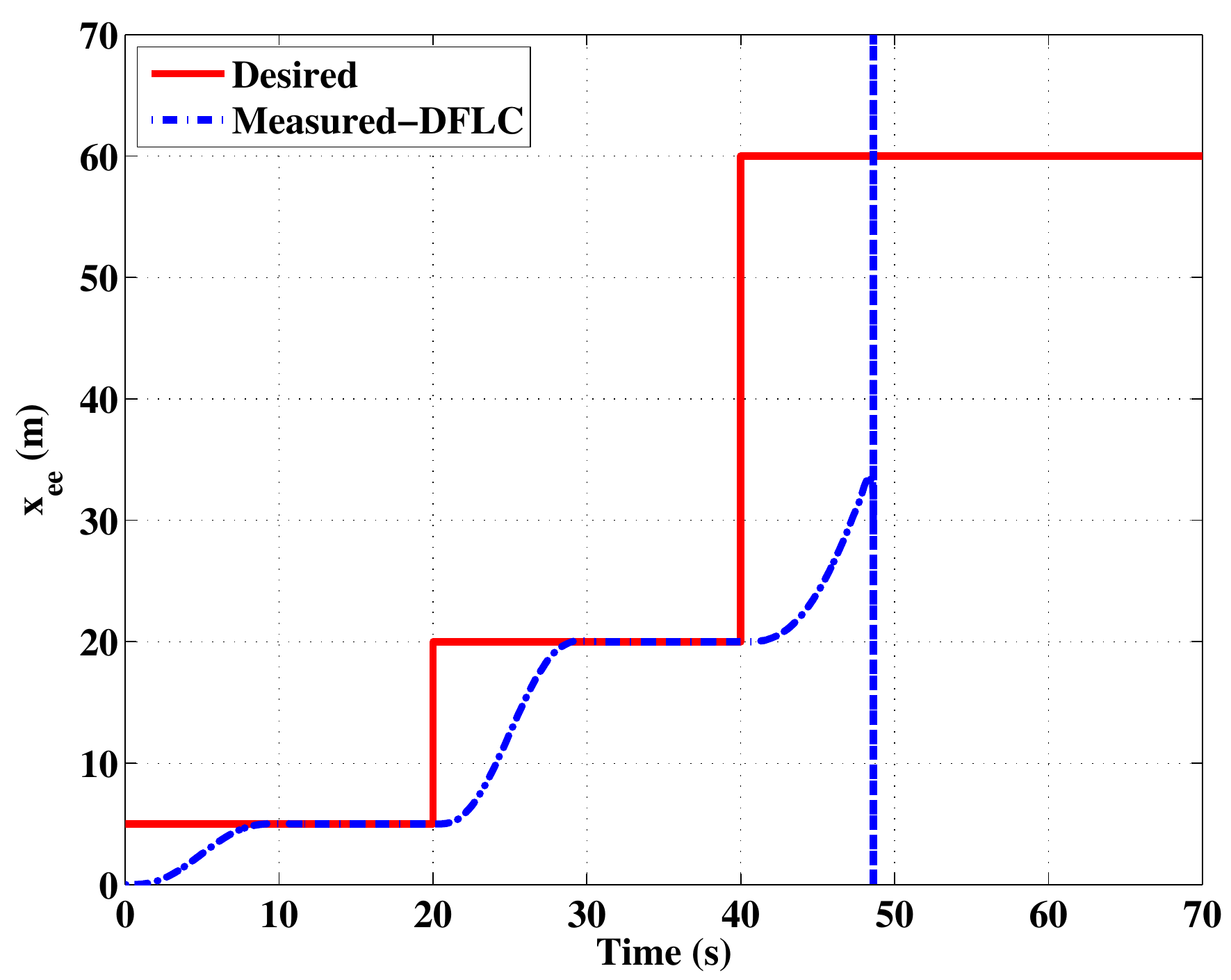}}&
 \subfigure[$y_{ee}$ Response]{\includegraphics [width=0.5\columnwidth]{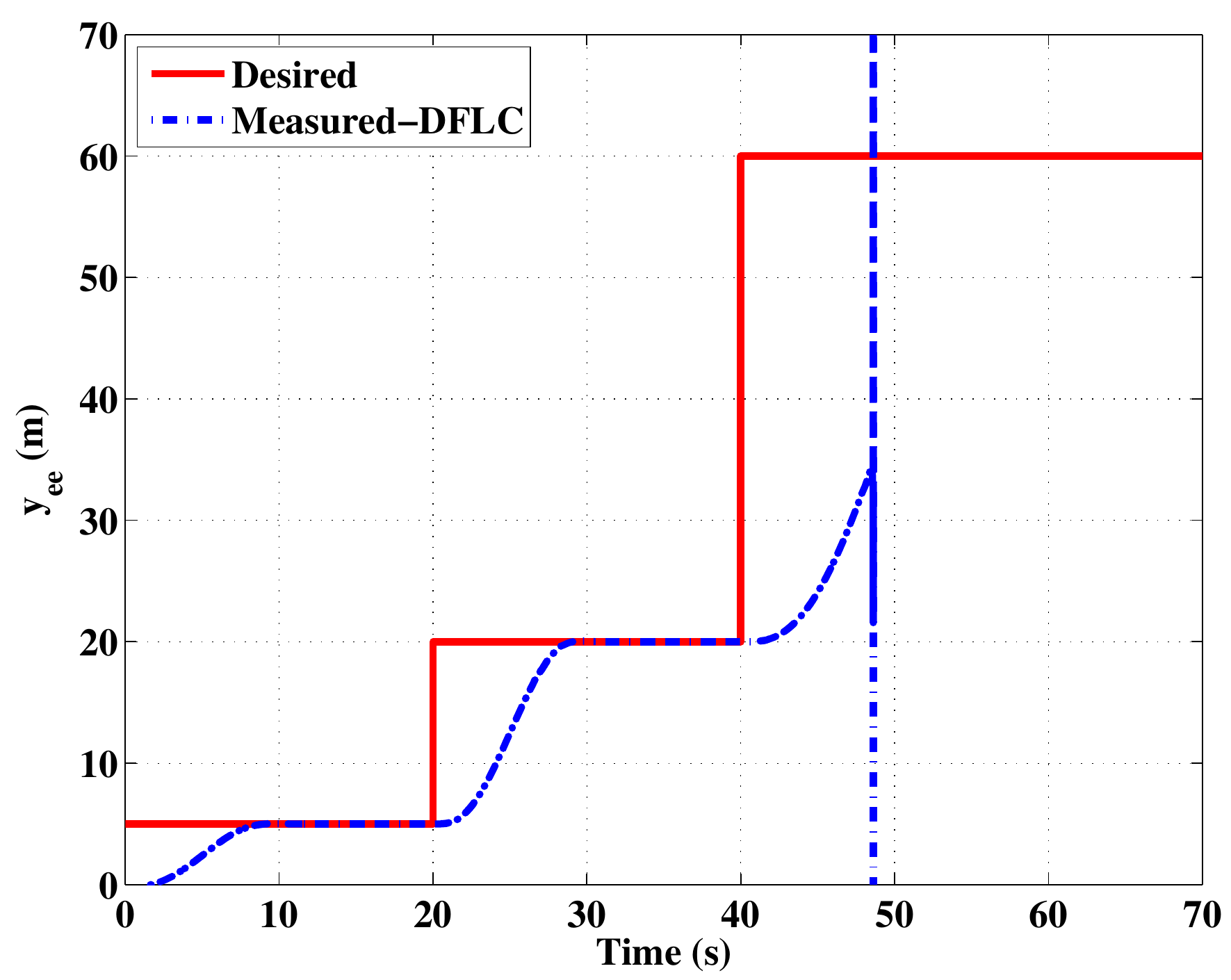}} \\
 \subfigure[$z_{ee}$ Response]{\includegraphics [width=0.5\columnwidth]{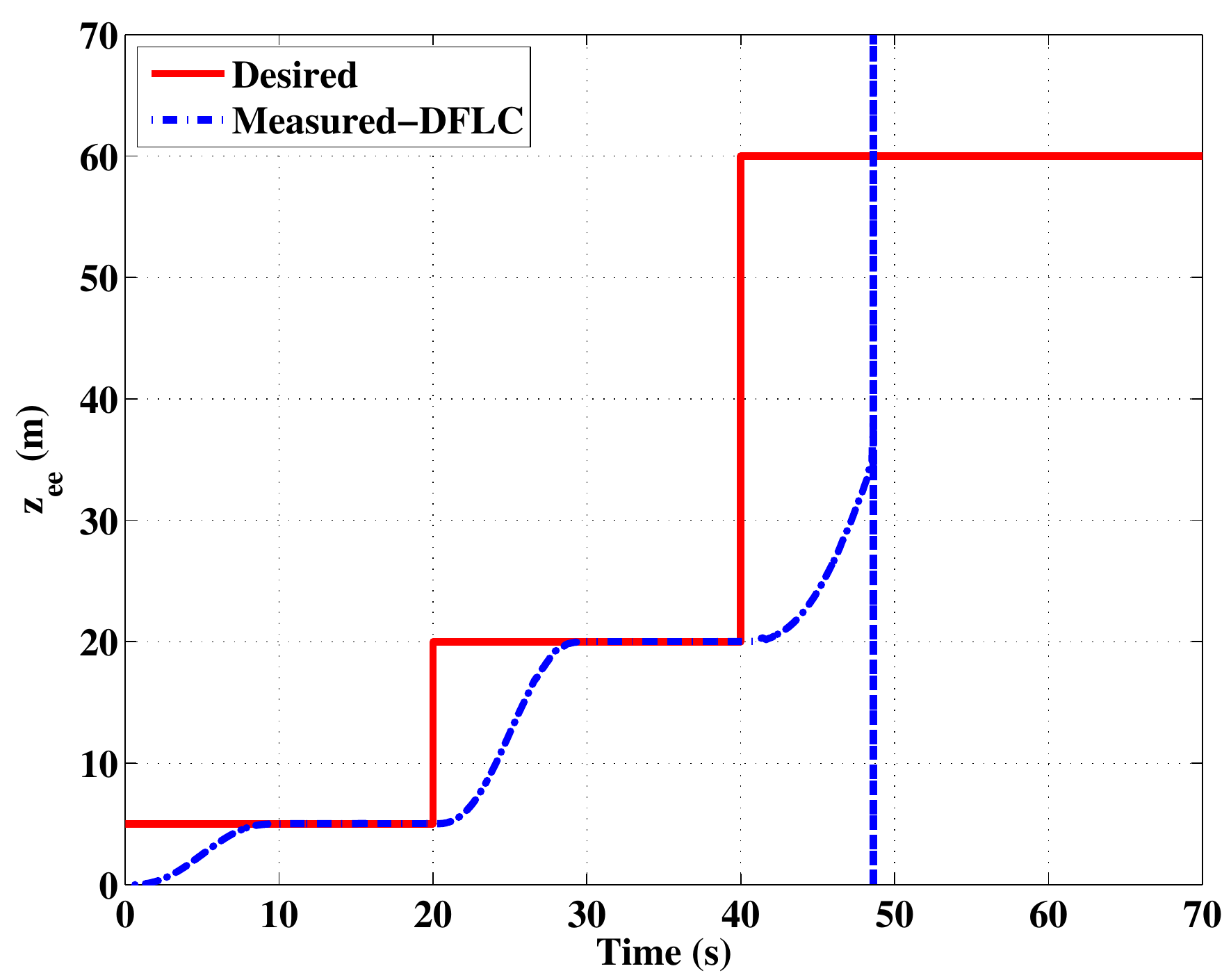}} &
 \subfigure[$\phi_{ee}$ Response]{\includegraphics [width=0.5\columnwidth]{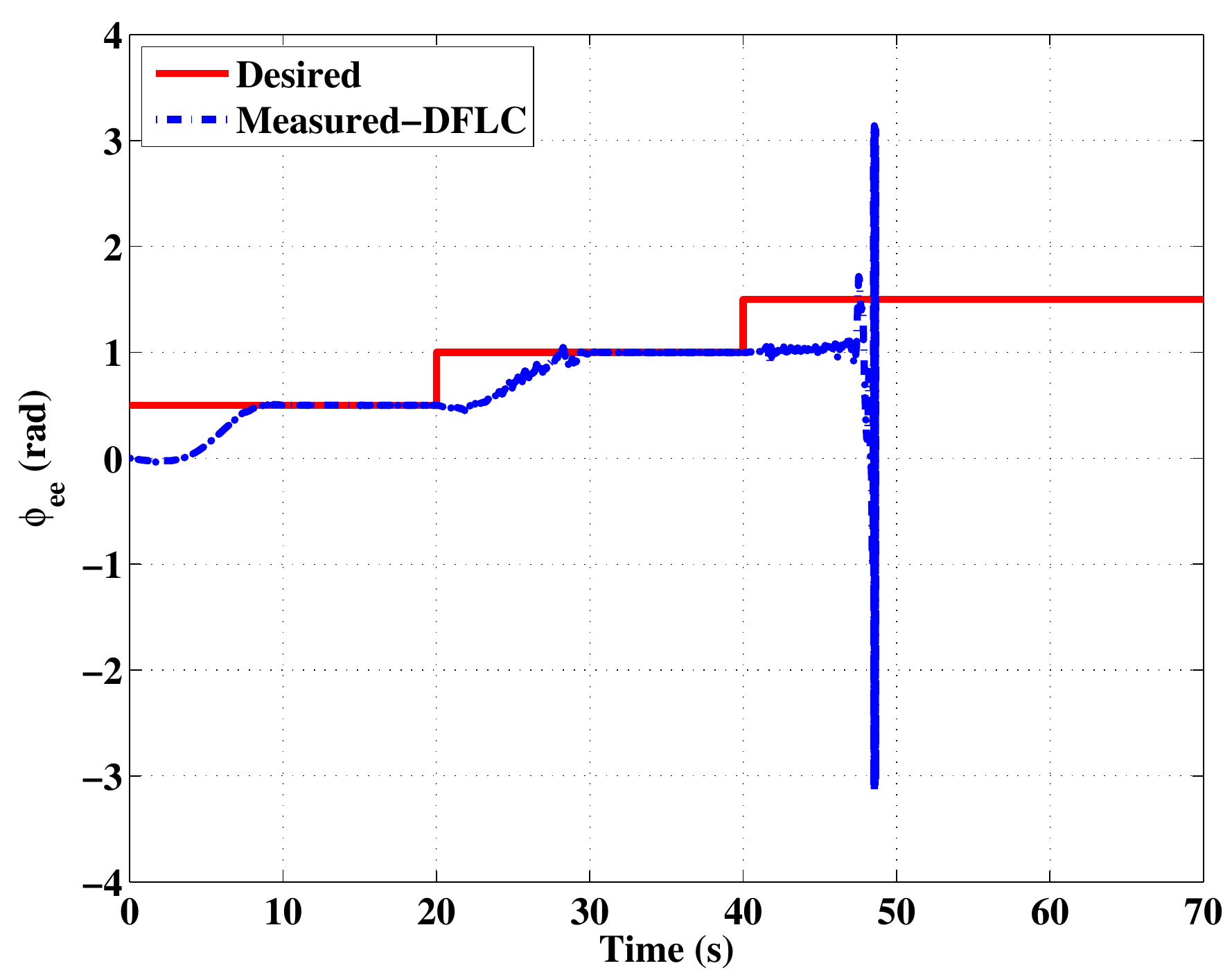}}\\
 \subfigure[$\theta_{ee}$ Response]{\includegraphics [width=0.5\columnwidth]{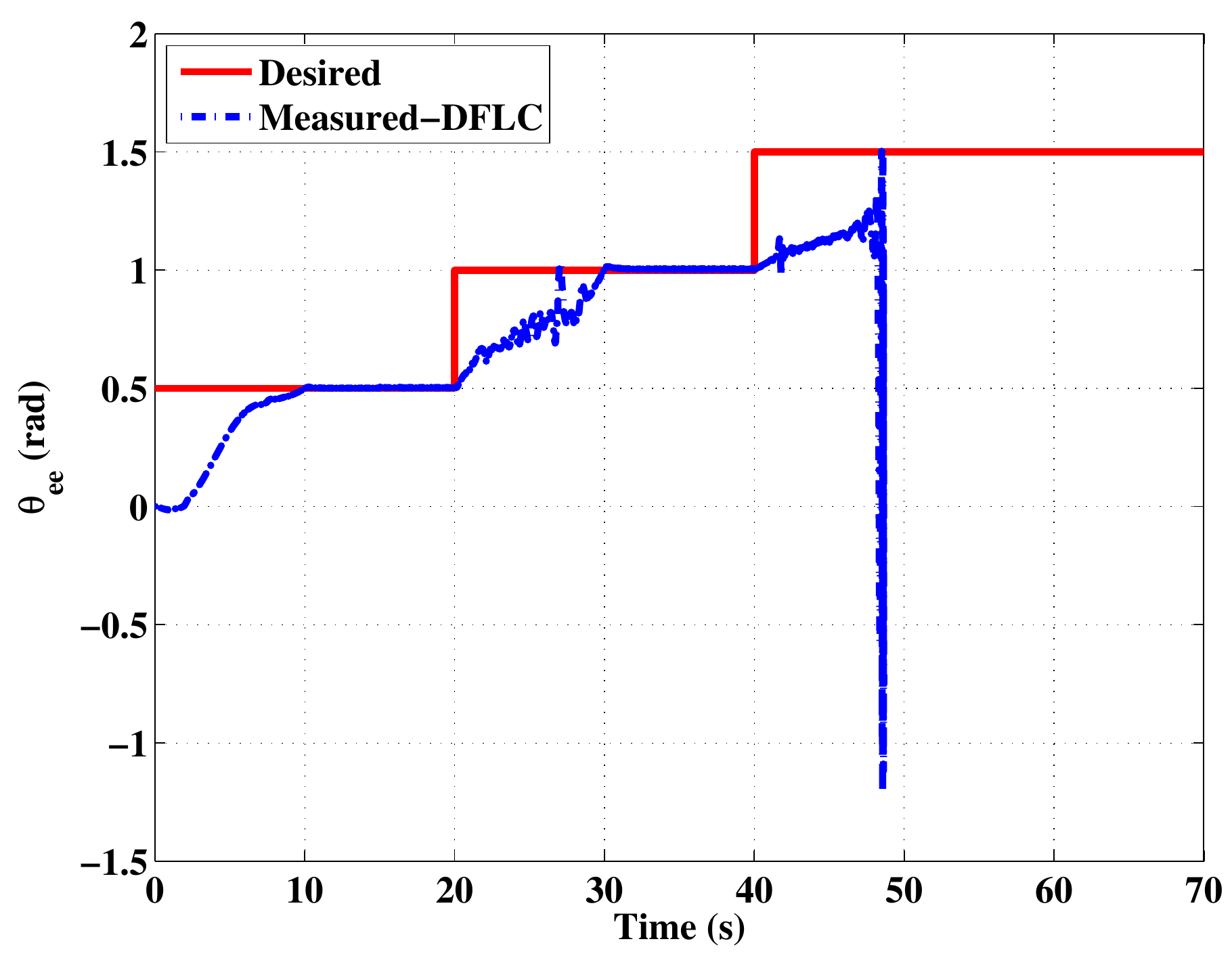}}&
 \subfigure[$\psi_{ee}$ Response]{\includegraphics [width=0.5\columnwidth]{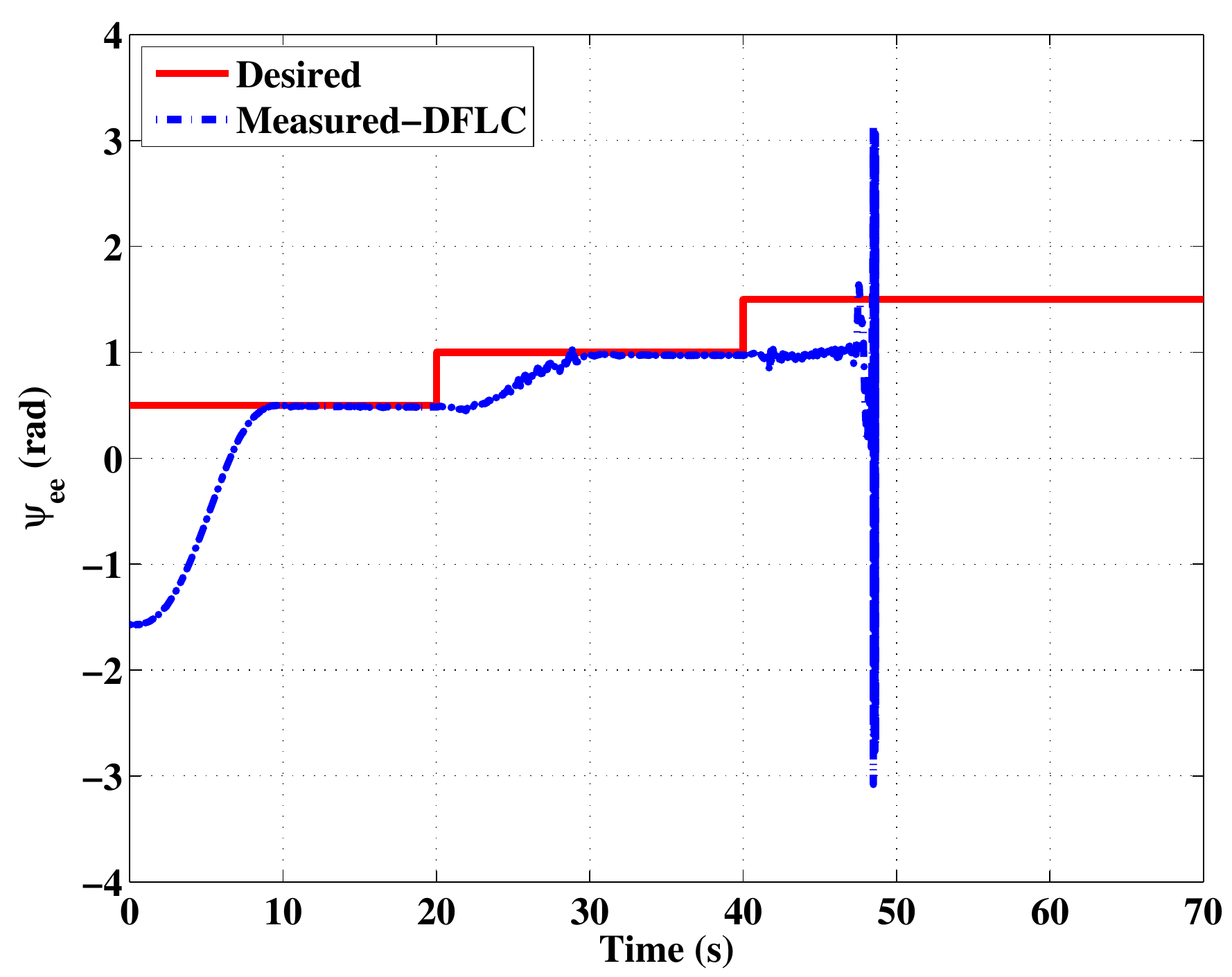}}
\end{tabular}
\caption{The Actual Response of DFLC Technique for the End Effector Position and Orientation: a) $x_{ee}$, b) $y_{ee}$, c) $z_{ee}$, d) $\phi_{ee}$, e) $\theta_{ee}$, and f) $\psi_{ee}$.}
\label {DFLC_controller_endeffector}
\end{figure}

The end effector position and orientation can be found from the forward kinematics (see Figure \ref{DFLC_controller_endeffector}).
%==============================================
%==============================================

From the above discussion and results, the following items can be concluded about the performance of DFLC technique:
\begin{itemize}
  \item DFLC technique succeeds to make system stable against adding the payload. However, it fails to provides a good trajectory tracking capabilities with different operation regions.
  \item DFLC suffers from the necessity of calibrating and determining the offset value which is affected by payload value and cannot be estimated accurately.
  \item Moreover, Considering the complexity of the controller implementation in real time, DFLC is fairly simple.
  \item Therefore, their is a need for high performance and more robust adaptive control technique to overcome these problems.
\end{itemize}
\section{Adaptive Fuzzy Logic Control}	
In this section, an adaptive fuzzy logic control based on "Fuzzy Model Reference Learning Controller" (FMRLC) is designed to control the proposed quadrotor manipulation system. This control technique is presented in details in \cite{Passino_FLC_Book, Passino_FMRLC_paper, Passino_DFLC_paper, Adapt_Fuz_globally, Adapt_Fuz_unknown}.

The main drawback of fuzzy controllers is the large amount of parameters to be tuned. Also, the DFLC designed in Section \ref{sec:DFLC} needs to retune its parameters in each operation region. Moreover, the fuzzy controller constructed for the nominal plant may later perform inadequately if significant and unpredictable plant parameter variations, or environmental disturbances occur \cite{Passino_FMRLC_paper}.

Many tuning methods are applied to improve the performance of fuzzy controllers such as fuzzy supervisors, genetic algorithms, and the ant colony algorithms. All these methods are capable of generating the optimum parameters to the control system but at the cost of computational time. \cite{fuzzy_tune_reinf}.

In this work, a learning control algorithm is used which helps to resolve some of these fuzzy controller design issues. This algorithm employs a reference model (a model of how you would like the plant to behave) to provide closed-loop performance feedback for tuning a fuzzy controller's knowledge-base.

\begin{figure}
      \centering
      \includegraphics[width=0.9\columnwidth, height=9cm]{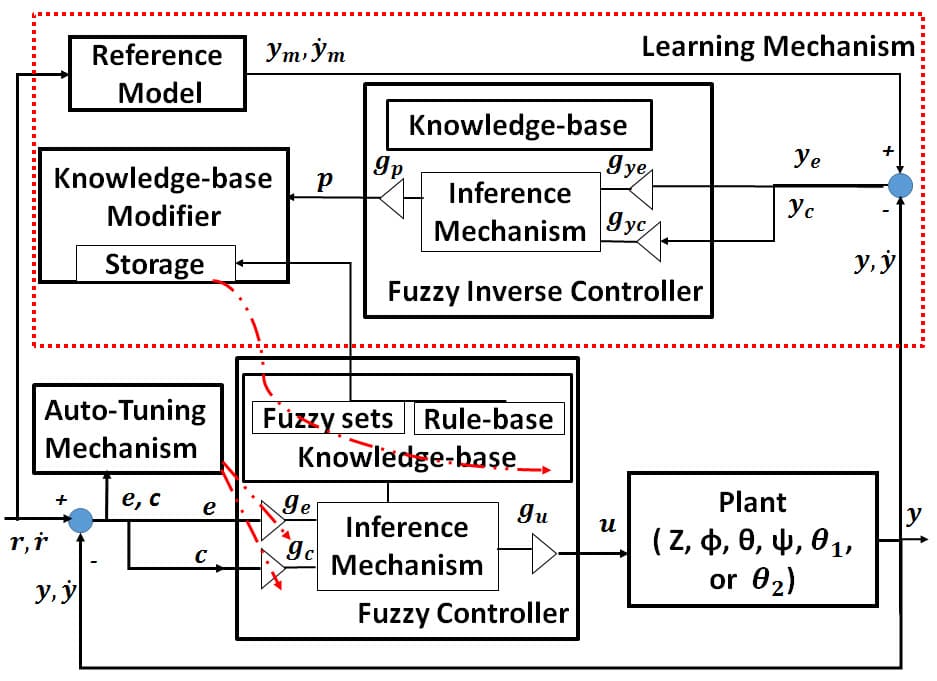}
      \caption{Functional Block Diagram for the FMRLC}
      \label{AdaptiveFLC}
   \end{figure}

The learning control technique, which is shown in Figure \ref{AdaptiveFLC}, operates as following: first, it observes data from a fuzzy control system, then it characterizes its current performance, and finally, adjusts the fuzzy controller such that some pre-determined performance objectives are met.  These performance objectives are characterized via the reference model.

The control system design is the same as in Figure \ref{DFLC} by replacing each of the $FLC_z$, $FLC_\phi$, $FLC_\theta$, $FLC_\psi$, $FLC_{\theta_1}$ and $FLC_{\theta_2}$ block with the block shown in Figure \ref{AdaptiveFLC}. However, there is no need for the offset value that is used in Figure \ref{DFLC} because the FMRLC can compensate the quadrotor weight. The blocks of $FLC_x$ and $FLC_y$ are still the same because there is no need for adaptation here, since these blocks are used to map the relation between the error in $X$ and $Y$ directions into the required roll and pitch motions based on the operation of the quadrotor.

The functional block diagram for the FMRLC is shown in Figure \ref{AdaptiveFLC}.
\subsection{The Fuzzy Controller}
The plant in Figure \ref{AdaptiveFLC} has output $y$ (which can be $Z$, $\phi$, $\theta$, $\psi$, $\theta_1$ or  $\theta_2$) ,and  an input $u$ (which can be $T$, $\tau_{a_1}$, $\tau_{a_2}$, $\tau_{a_3}$, $T_{m_1}$ or $T_{m_2}$). The scaling controller gains $g_e$, $g_c$, and $g_u$ for the error, $e$, change in error, $c$, and controller output, $u$, are used respectively, such that the universe of discourse of all inputs and outputs are the same and equal to [-1, 1]. These parameters are tuned to get required performance.

\begin{figure}
      \centering
      \subfigure[Input Variable error $e$]{\includegraphics[width=13cm]
      {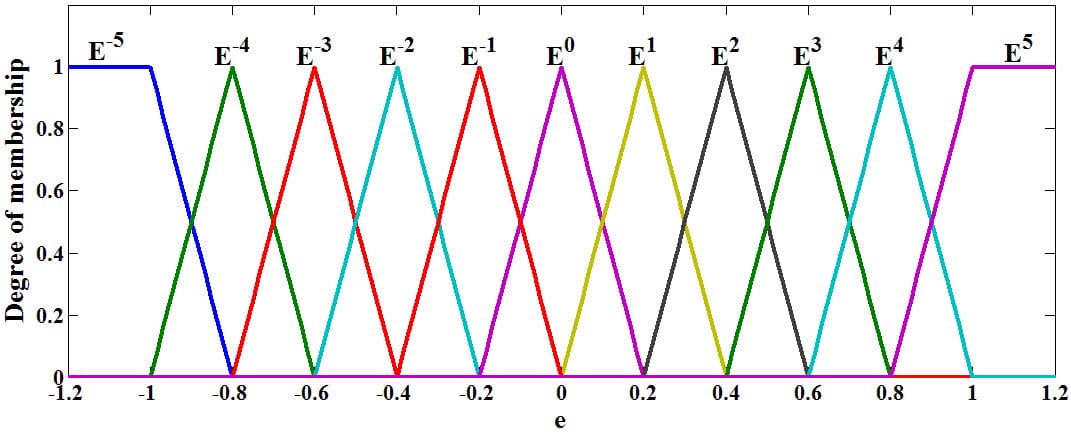}}\hspace{2cm}
      \subfigure[Input Variable error rate $c$]{\includegraphics[width=13cm]
      {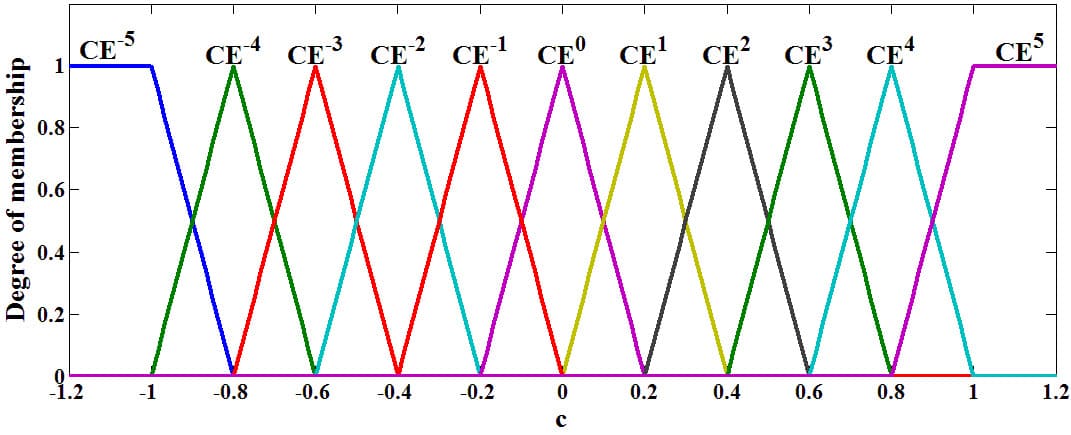}}
      \caption{Input Membership Functions of Fuzzy Controller of FMRLC: a) $Error$ ($e$) and b) $Error$ $Rate$ ($c$).}
      \label{FMRLC_FLC_MSFs}
\end{figure}

The membership functions are chosen to be 11 symmetric triangular-shaped functions for each controller input as shown in Figure \ref{FMRLC_FLC_MSFs}. The fuzzy sets for the fuzzy controller output are assumed to be symmetric and triangular-shaped with a base width of 0.4, and all centered at zero on the normalized universe of discourse. They are what the FMRLC will automatically tune through the learning mechanism. Thus, the initial rule base elements are set to zeros.

The rule-base for the fuzzy controller has rules of the form:
           $$IF \quad e \quad is \quad E^i \quad AND \quad \dot{e} \quad is \quad CE^j \quad THEN \quad u \quad is \quad U^m$$
where $E^i$ ($CE^j$) denotes the $i^{th}$ ($j^{th}$) linguistic value associated with $e$ ($c$), respectively. $U^m$ denotes the consequent linguistic value associated with $u$.

The centers of the input membership functions are tuned using the auto-tuning mechanism shown in figure \ref{AdaptiveFLC}.
Mamdani fuzzy inference method is used with a min-max operator for the aggregation. The standard center of gravity is used as a defuzzification technique.
%====================================================
\subsection{The Reference Model}
The reference model is used to quantify the desired performance.  A $1^{st}$ order model is chosen as the reference model:
\begin{equation}
\frac{y_{m}(s)}{r(s)} = \frac{1}{\tau_{c_i}S + 1}
\label{ref_model}
\end{equation}
where $y_{m}(s)$ is the output response of the reference system, and $r(s)$ is the desired value of the plant. $\tau_{c_i}$ ($i = z$, $\phi$, $\theta$, $\psi$, $\theta_1$, and $\theta_2$) is the time constant of the reference model. The value of this time constant is tuned such that it is suitable with the dynamics of the system.

The performance of the overall system is computed with respect to the reference model by the learning mechanism by generating an error signal:
\begin{equation}
y_{e} = y_{m} - y
\label{ref_error}
\end{equation}
\subsection{The Learning Mechanism}
The learning mechanism tunes the rule-base of the direct fuzzy controller so that the closed-loop system behaves like the reference model. Based on the performance of the reference model and the plant under control, the learning mechanism will take the required action. If the performance is met (i.e. $y_e$ is small), then the learning mechanism will not make significant modifications to the fuzzy controller. On the other hand, if $y_e$ is big, the desired performance is not achieved and the learning mechanism must adjust the fuzzy controller.

The learning mechanism consists of two parts, fuzzy inverse model and knowledge-base modifier.
\subsubsection{The Fuzzy Inverse Model}
The fuzzy inverse model performs the function of mapping $y_e$ (representing the deviation from the desired behavior), to changes in the process inputs $p$ that are necessary to force $y_e$ to be zero.

The input to the fuzzy inverse model, in Figure \ref{AdaptiveFLC}, includes the error ($y_e$) and change in error ($\dot{y_e}$ that is $y_c$) between the reference model ($y_m$) and the plant's output ($y$). Also, it has scaling gains, $g_{y_e}$, $g_{y_c}$ and $g_p$ for normalization of its universe of discourses of $y_e$, $y_c$ and $p$ respectively.

\begin{figure}
      \centering
      \subfigure[Input Variable error $y_e$]{\includegraphics[width=13cm]
      {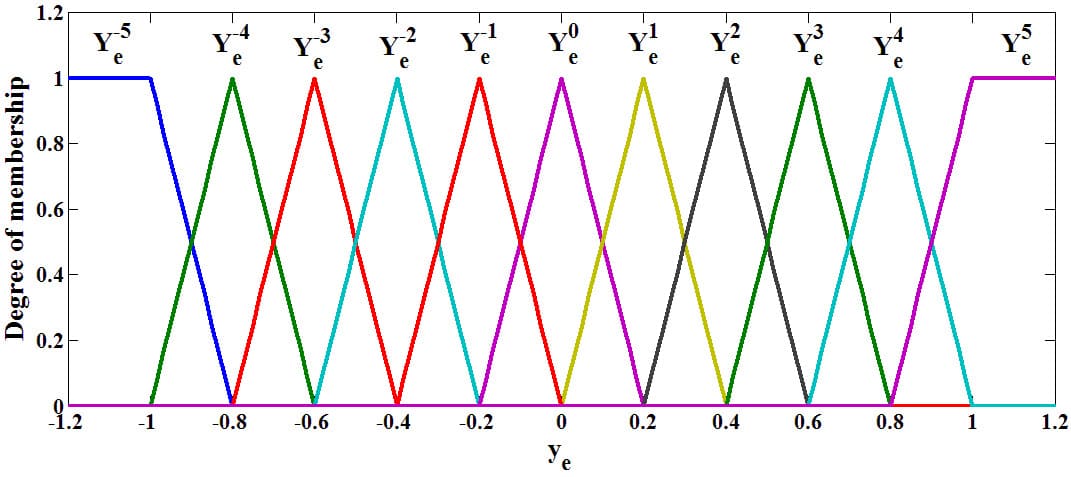}}\hspace{2cm}
      \subfigure[Input Variable error rate $y_c$]{\includegraphics[width=13cm]
      {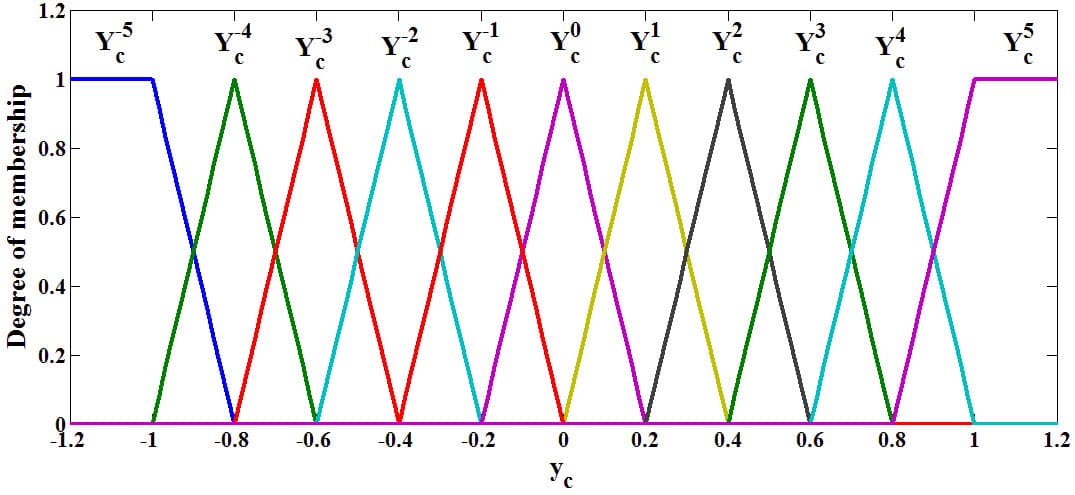}}
      \subfigure[Output Variable $p$]{\includegraphics[width=13cm]
      {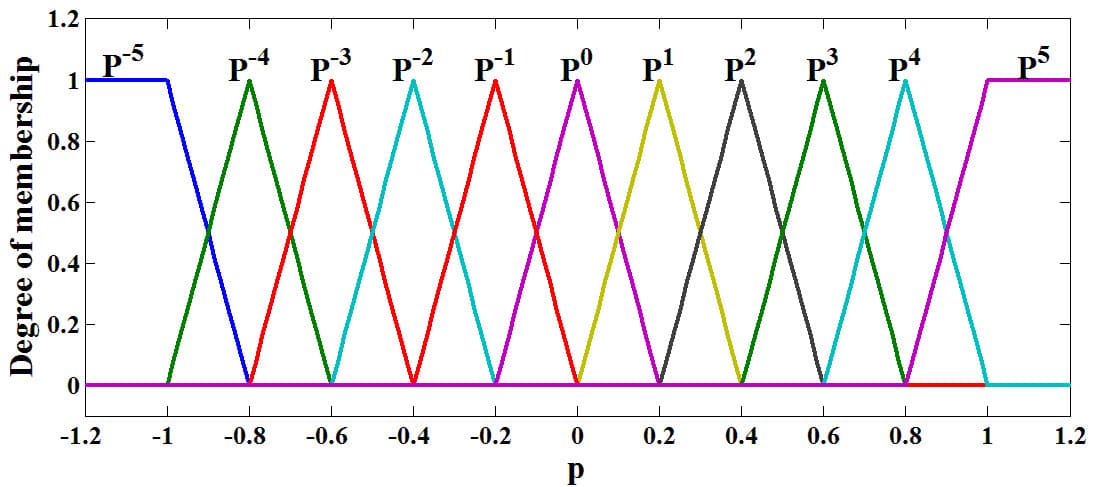}}
      \caption{Input Membership Functions of Fuzzy Inverse Model of FMRLC: a) $Error$($y_e$), b) $Error$ $Rate$ ($y_c$), and c) $Output$ ($p$).}
      \label{FMRLC_FIM_MSFs}
\end{figure}

For each of these inputs, 11 symmetric and triangular-shaped membership functions are used as shown in Figure \ref{FMRLC_FIM_MSFs}. Also, for the output universe of discourse, 11 symmetric triangular-shaped membership functions, with a base width of 0.4, are chosen.
The rule-base for the fuzzy controller has rules of the form:
           $$IF \quad y_e \quad is \quad Y_e^k \quad AND \quad y_c \quad is \quad Y_c^s \quad THEN \quad p \quad is \quad P^m$$
where $Y_e^k$ and  $Y_c^s$ denote the $k^{th}$ ($s^{th}$) linguistic value associated with $y_e (y_c)$, respectively, and $P^m$ denotes the consequent linguistic value associated with $p$.

Denoting the center of the output membership function for this rule $c_{k,s}$ that it is the center associated with the output membership function that has the $k^{th}$ membership function for the $y_e$ universe of discourse and the $s^{th}$ membership function for the $y_c$ universe of discourse. The rule base array shown in Table \ref{rulebasefuzzyinverse} is employed for the fuzzy inverse model. The entries of the table represent the center values of symmetric triangular-shaped membership functions $c_{k,s}$ with base widths 0.4 for output fuzzy sets $P^m$ on the normalized universe of discourse.

For example, if $k$ = $s$ = $0$, then we see from the table that $c_{k,s}$ = $c_{0,0}$ = $0$ (this is the center of the table). This cell in the table represents the rule that says "if $y_e$ = 0 and $y_c$ = 0 then $y$ is tracking $y_{m}$ perfectly", so there is no need to update the fuzzy controller. Thus, the output of the fuzzy inverse model will be zero. On the other hand, if $k$ = $2$ and $s$ = $1$, then $c_{k,s}$ = $c_{2,1}$ = $0.6$. This rule indicates that "if $y_e$ is positive (i.e., $y_m$ is greater than $y$) and $y_c$ is positive (i.e., $y_e$ is increasing), then increase value of $p$ such that increase value of $u$.

Mamdani fuzzy inference method is used with a min-max operator for the aggregation and the standard center of gravity is used as defuzzification technique.
%====================================================
%=====================================================
\begin{table}
\caption{Knowledge-Base Array for the Fuzzy Inverse Model}
\label{rulebasefuzzyinverse}
\begin{center}
\setlength{\tabcolsep}{7pt}
\begin{tabular}{|l|c|c|c|c|c|c|c|c|c|c|c|c|}
\hline
\multicolumn{2}{|c|}{} & \multicolumn{11}{c|}{$Y_{c}^{s}$} \\
\cline{3-13}
\multicolumn{2}{|c|}{$c_{k,s}$} & -5 & -4 & -3 & -2 & -1 & 0 & 1 & 2 & 3 & 4 & 5 \\
\hline
 & -5 & -1 & -1 & -1 & -1 & -1 & -1 & -0.8 & -0.6 & -0.4 & -0.2 & 0 \\
\cline{2-13}
 & -4 & -1 & -1 & -1 & -1 & -1 & -0.8 & -0.6 & -0.4 & -0.2 & 0 & 0.2 \\
\cline{2-13}
 & -3 & -1 & -1 & -1 & -1 & -0.8 & -0.6 & -0.4 & -0.2 & 0 & 0.2 & 0.4 \\
\cline{2-13}
 & -2 & -1 & -1 & -1 & -0.8 & -0.6 & -0.4 & -0.2 & 0 & 0.2 & 0.4 & 0.6 \\
\cline{2-13}
 & -1 & -1 & -1 & -0.8 & -0.6 & -0.4 & -0.2 & 0 & 0.2 & 0.4 & 0.6 & 0.8 \\
\cline{2-13}
$Y_{e}^{k}$ & 0 & -1 & -0.8 & -0.6 & -0.4 & -0.2 & 0 & 0.2 & 0.4 & 0.6 & 0.8 & 1 \\
\cline{2-13}
 & 1 & -0.8 & -0.6 & -0.4 & -0.2 & 0 & 0.2 & 0.4 & 0.6 & 0.8 & 1 & 1 \\
\cline{2-13}
 & 2 & -0.6 & -0.4 & -0.2 & 0 & 0.2 & 0.4 & 0.6 & 0.8 & 1 & 1 & 1 \\
\cline{2-13}
 & 3 & -0.4 & -0.2 & 0 & 0.2 & 0.4 & 0.6 & 0.8 & 1 & 1 & 1 & 1 \\
\cline{2-13}
 & 4 & -0.2 & 0 & 0.2 & 0.4 & 0.6 & 0.8 & 1 & 1 & 1 & 1 & 1 \\
\cline{2-13}
 & 5 & 0 & 0.2 & 0.4 & 0.6 & 0.8 & 1 & 1 & 1 & 1 & 1 & 1 \\
\hline
\end{tabular}
\end{center}
\end{table}
%===================================================================
\subsubsection{The Knowledge-Base Modifier}

Given the information about the necessary changes in the input, which are represented by $p$, to force the error $y_e$ to zero, the knowledge-base modifier changes the rule-base of the fuzzy controller so that the previously applied control action will be modified by the amount $p$.

Consider the previously computed control action $u(t-1)$, and assume that it contributed to the present good or bad system performance. Note that $e(t-1)$ and $c(t-1)$ would have been the error and change in error that were input to the fuzzy controller at that time. By modifying the fuzzy controller's knowledge-base, we may force the fuzzy controller to produce a desired output $u(t-1)$ + $p(t)$, which we should have put in at time $(t-1)$  to make $y_e$ smaller. Then, the next time we get similar values for the error and change in error, the input to the plant will be one that will reduce the error between the reference model and plant output.

Let $b_m$ denote the center of the membership function associated with $U^m$. Knowledge-base modification is performed by shifting centers $b_m$ of the membership functions of the output that are associated with
the fuzzy controller rules that contributed to the previous control action $u(t-1)$.

This knowledge-base modifier works as following:
\begin{itemize}
\item Define "active set" of rules at time $(t - 1)$ to be all the rules in the fuzzy controller whose membership value is:
    \begin{equation}
      \mu_i(e(t-1),c(t-1))>0
       \label{learn_mech_mu}
    \end{equation}
\item For all rules in the active set, use (\ref{learn_mech_bm}) to modify the output membership function centers.
      \begin{equation}
        b_{m}(t)= b_{m}(t-1) + p(t)
         \label{learn_mech_bm}
       \end{equation}
\end{itemize}
Rules that are not in the active set do not have their output membership functions modified.
\subsection{Auto-Tuning Mechanism}	
With the auto-tuning for the input scaling gains of the fuzzy controller, the centers of the input membership functions are tuned.

Note that the range covered by the linguistic term is reduced by increasing the scaling factor (decreasing the domain interval of the universe of discourse), and thus the true meanings of a membership function can be varied by the gains applied.

Tuning procedure that changes the gains $g_e$ and $g_c$ is altering the coverage of the control surface. Note that for a rule-base with a fixed number of rules, when the domain interval of the input universes of discourse are large (i.e., small $g_e$ and $g_c$ ), it represents a "coarse control" action; and when the input universes of discourse are small (i.e. large $g_e$ and $g_c$ ), it represents a "fine control" action. Hence, we can
vary the "granularity" of the rule-base near the point where the system is operating (e.g. the center region of the rule-base map by varying the gains $g_e$ and $g_c$.

The gains $g_e$ and $g_c$ are adjusted so that a smaller rule-base can be used on the input range needed the most. This is done by adjusting the meaning of the linguistic values based on the most recent input signals to the fuzzy controller so that the control surface is properly focused on the region that describes the system activity.

In the standard FMRLC design, the system performance is degraded with variation in the desired input value. An auto-tuning mechanism is used in \cite{Passino_DFLC_paper} to tune $g_e$ and $g_c$ gains online as following:

Let the maximum of each fuzzy controller inputs $(e, c)$ over a time interval of the last $T_a$ seconds be denoted by $max_{T_a}\{e\}$ and $max_{T_a}\{c\}$. Then this maximum value is defined as the gain of each input $e$ and $c$ so that,
\begin{equation}
g_e = \frac{1}{max_{T_a}\{e\}} \quad and \quad g_c = \frac{1}{max_{T_a}\{c\}}
\label{auto_tune}
\end{equation}

However, values of $g_e$ and $g_c$ must be limited to a maximum value; otherwise each input universe of discourse for the fuzzy controller may become zero that is when the these gains go to infinity.

The learning mechanism should operate at a higher rate than the auto-tuning mechanism in order to try to assure stability. If the auto-tuning mechanism is designed to be faster than the learning mechanism, the learning mechanism will not be able to catch up with the changes made by the auto-tuning mechanism so it will never be able to learn the rule-base correctly.
%
%This can be achieved by the following algorithm:
%\begin{itemize}
%  \item The input gains are updated every $n$ samples. Find the maximum $e$/$c$ over the most recent $n$ samples and denote it by $e_{max}$/$c_{max}$
%  \item Set the input gains;
%   $$ g_e =\frac{1}{|e_{max}|} \quad g_c =\frac{1}{|c_{max}|} $$
%\end{itemize}
%
%\subsection{Selection of the Scaling Gains}
%
%\begin{enumerate}
%  \item The initial and maximum value of the $g_e$/$g_c$ is chosen such that t normally encountered  values of $e$/$c$ will not result in saturation of the outermost input membership functions. Also, $g_u$ is chosen so that the range of outputs that are possible is the maximum one possible.
%  \item Begin with $g_p$ = $0$ (i.e., with the adaptation mechanism turned off) and simulate the system.
%  \item Choose the gains of the inverse model so that there is no saturation on its input universes of discourse.
%  \item  Increase $g_p$ slightly so that you just turn on the learning mechanism and it makes only small changes to the rule-base at each step. Perform any necessary tuning for the inverse model. If there exist unacceptable oscillations in the plant output response about the reference model response, then increase $g_{y_c}$ (we need additional derivative action in the learning mechanism to reduce the oscillations)
%  \item Continue to increase $g_p$ and subsequently tune the inverse model as needed. With $g_p$ large, you increase the adaptation rate, and hence if you increase it too much, you can get undesirable oscillations and sometimes instability.
%\end{enumerate}

\subsection{Simulation Results}
The system equations of motion and the control laws of the FMRLC are simulated using MATLAB/SIMULINK program.

The controller parameters of the FMRLC controller are given in Table \ref{FMRLC parameters}. Those parameters are tuned to get the required system performance.

The controller are tested to stabilize and track the desired trajectories under the effect of picking a payload of value 150 g at instant 15 s and placing it at instant 65 s.

\begin{figure}
\centering
\begin{tabular}{cc}
 \subfigure[$X$ Response]{\includegraphics[width=0.5\columnwidth, height=5cm]{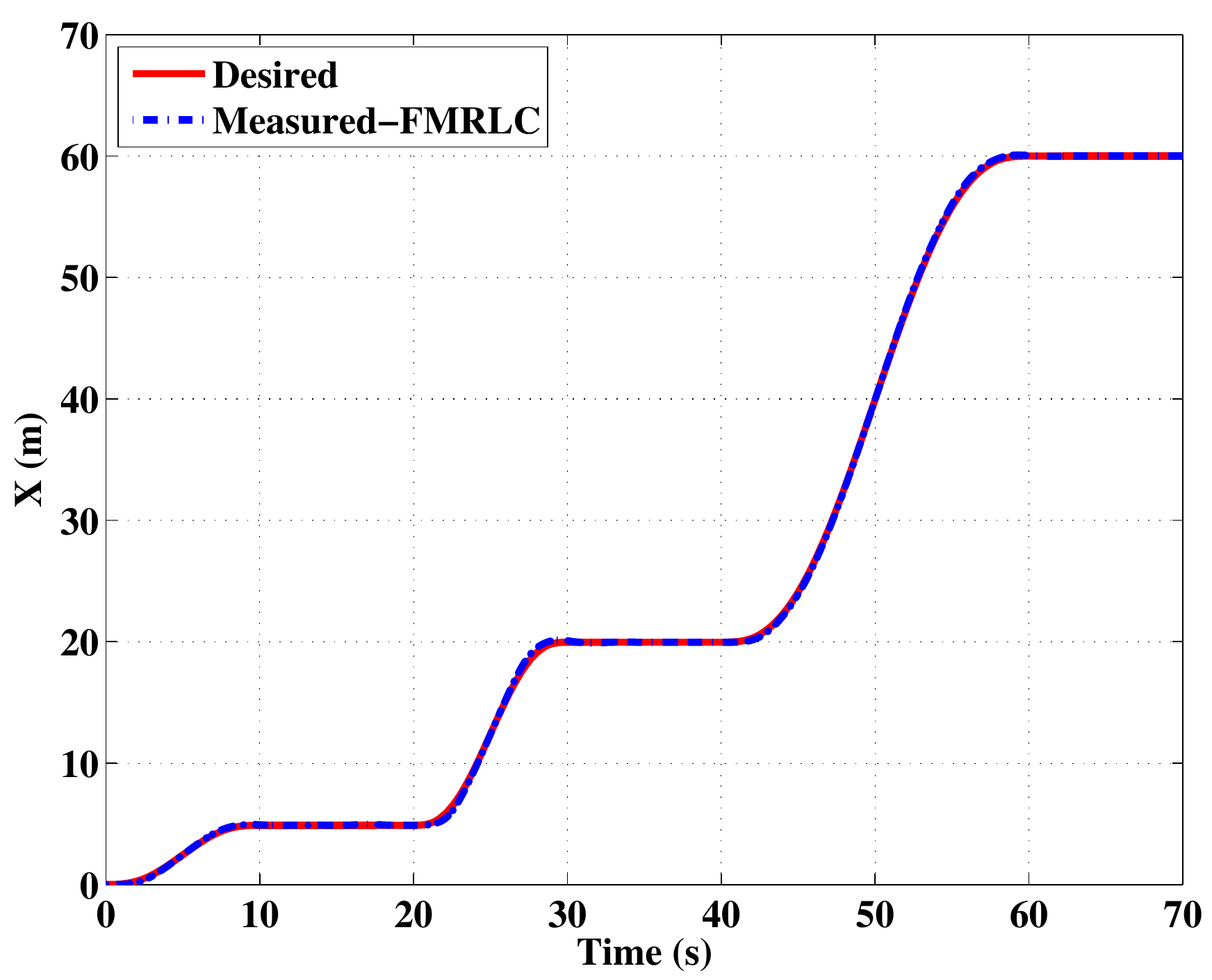}}&
 \subfigure[$Y$ Response]{\includegraphics [width=0.5\columnwidth, height=5cm]{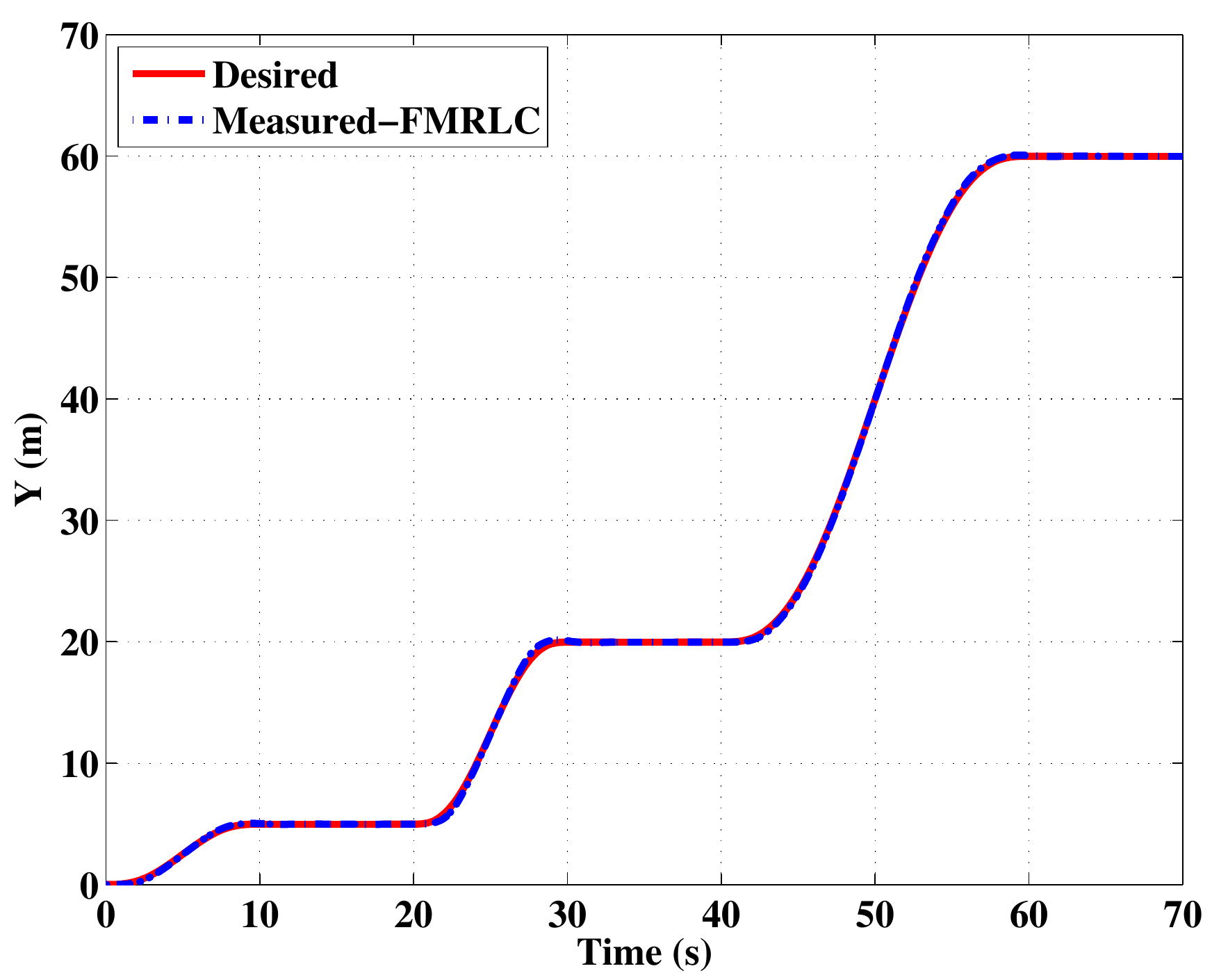}} \\
 \subfigure[$Z$ Response]{\includegraphics [width=0.5\columnwidth, height=5cm]{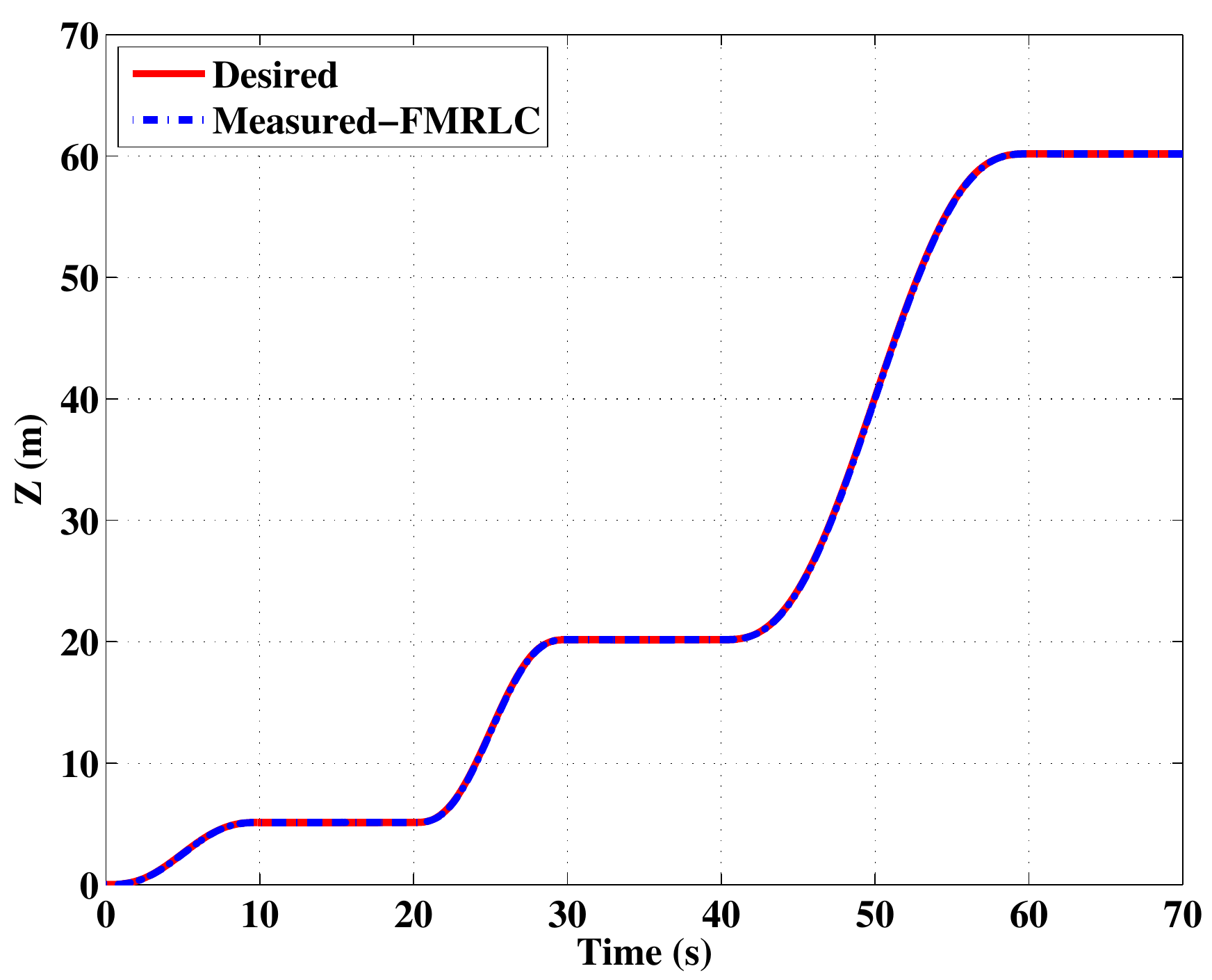}} &
 \subfigure[$\psi$ Response]{\includegraphics [width=0.5\columnwidth, height=5cm]{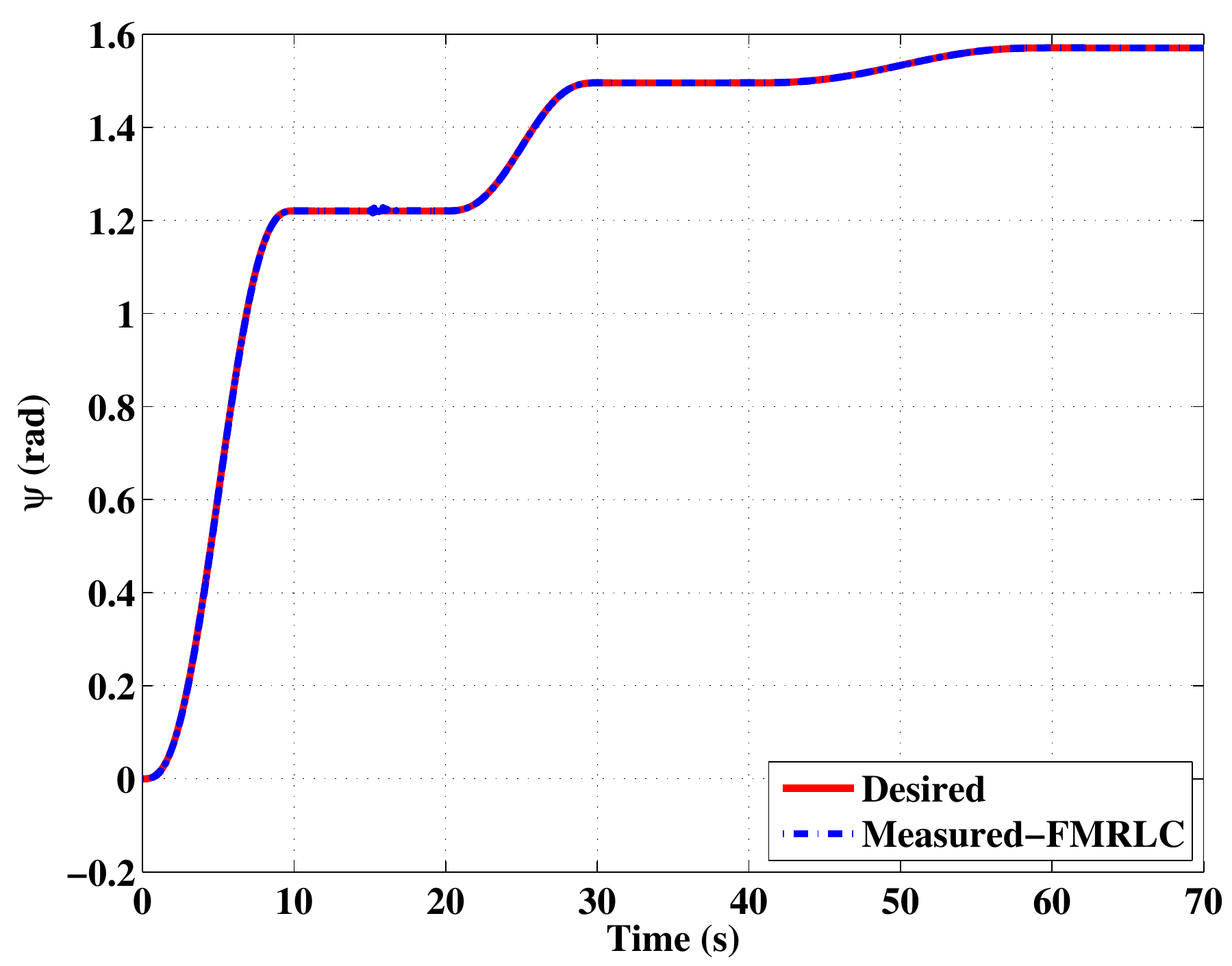}}\\
 \subfigure[$\theta_1$ Response]{\includegraphics [width=0.5\columnwidth, height=5cm]{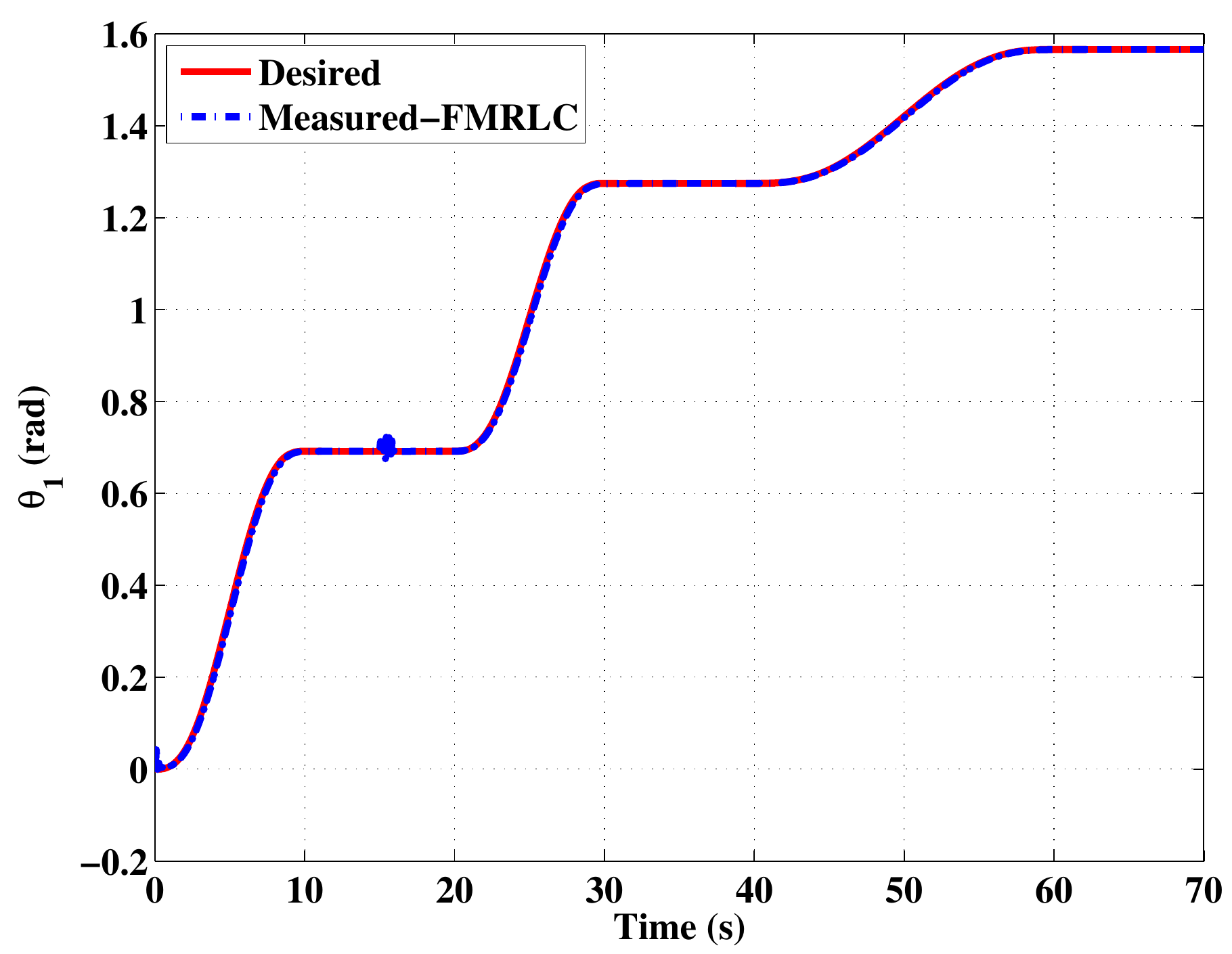}}&
 \subfigure[$\theta_2$ Response]{\includegraphics [width=0.5\columnwidth, height=5cm]{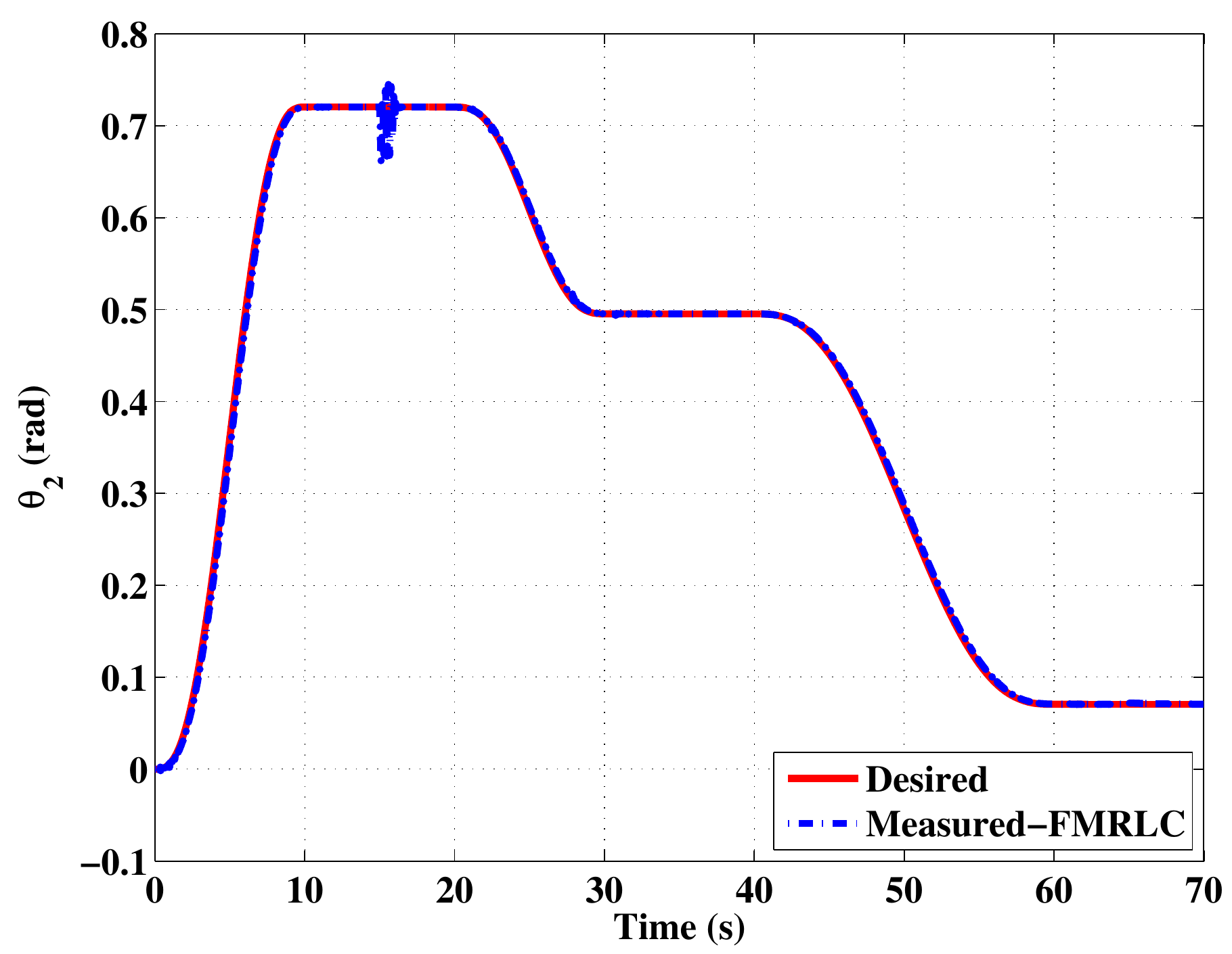}}\\
 \subfigure[$\phi$ Response]{\includegraphics [width=0.5\columnwidth, height=5cm]{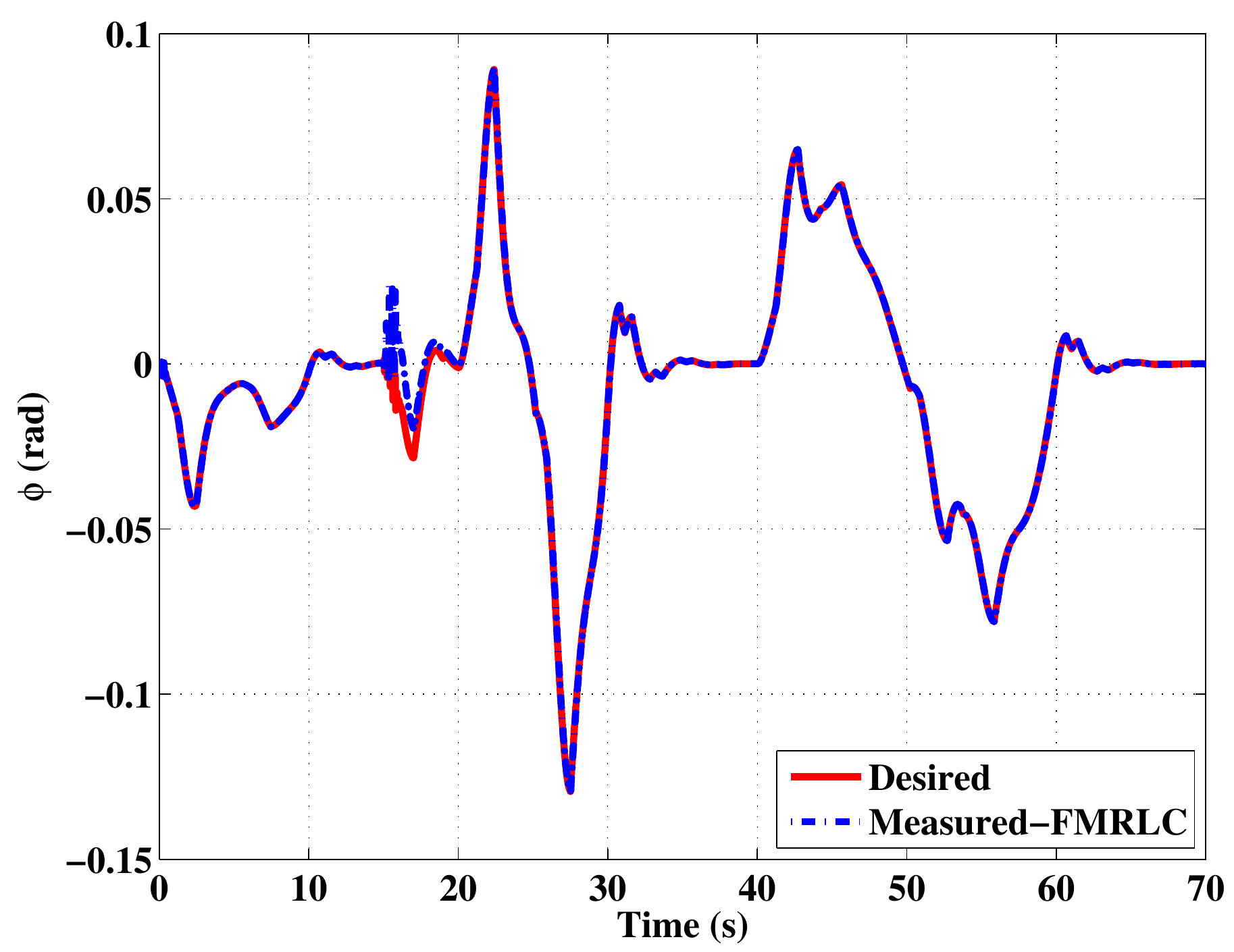}}&
 \subfigure[$\theta$ Response]{\includegraphics [width=0.5\columnwidth, height=5cm]{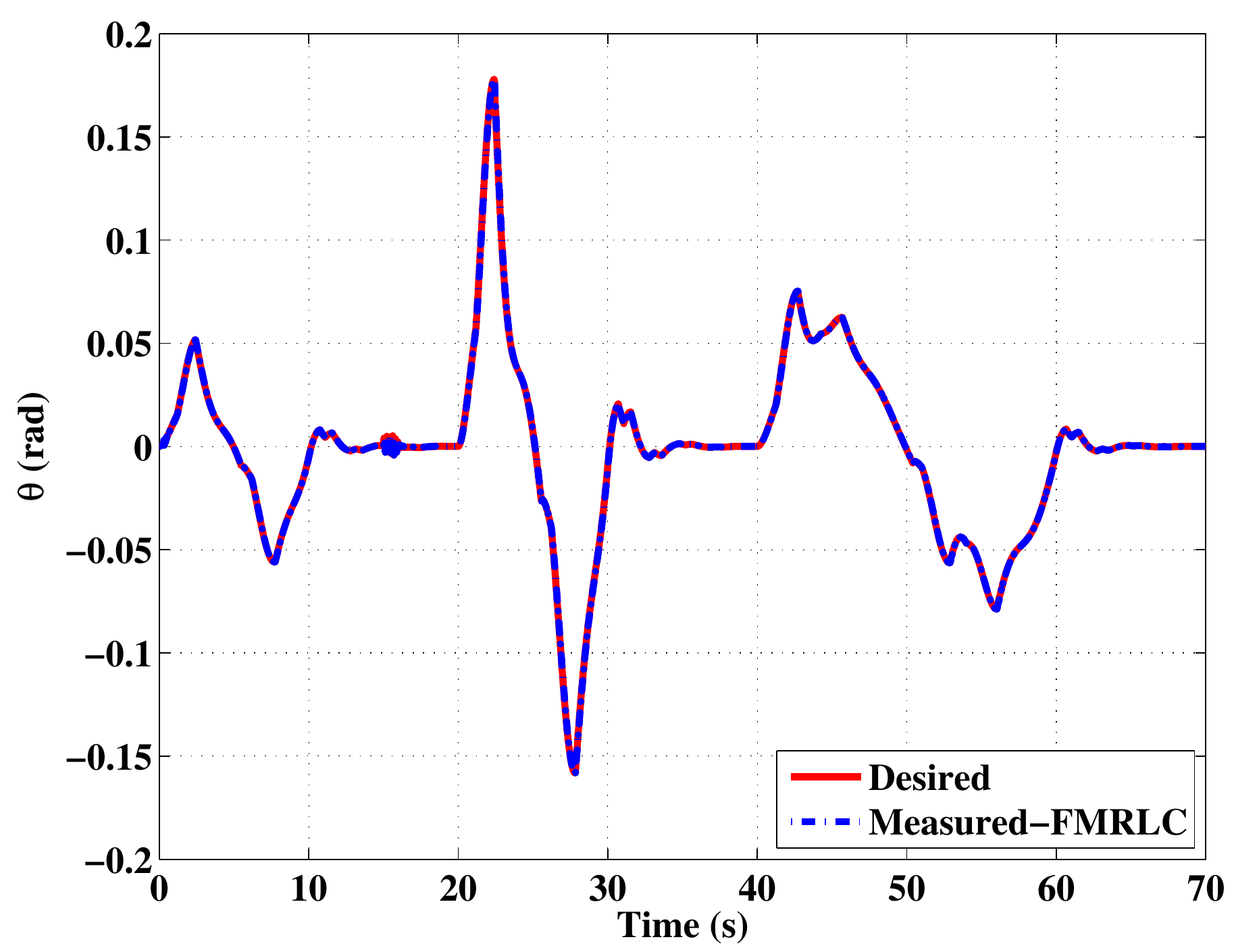}}
\end{tabular}
\caption{The Actual Response of FMRLC Technique for the Quadrotor and Manipulator Variables: a) $X$, b) $Y$, c) $Z$, d) $\psi$, e) $\theta_1$, f) $\theta_2$, g) $\phi$, and h) $\theta$.}
\label {FMRLC_controller_Vechvariable}
\end{figure}

The simulation results are presented in Figure \ref{FMRLC_controller_Vechvariable}. These results show that FMRLC  is able to track the desired trajectories (with different operating regions) before, during picking, holding, and placing the payload with zero tracking error.
\begin{table}
\caption{FMRLC Parameters}
\label{FMRLC parameters}
\begin{center}
\setlength{\tabcolsep}{5.5pt}
\begin{tabular}{|c||c||c||c||c||c||c|}
\hline
\hline
Par./Val. & Z & $\phi$ & $\theta$ & $\psi$ & $\theta_1$ & $\theta_2$\\
\hline
 $g_{e-initial}$ &	1/5&	2&	2&	1/3&	1/60	&1/60\\
\hline
 $g_{c-initial}$  &1/10&	1&	1&	1/30& 1/1000&   1/1000\\
\hline
$g_{u}$ &	16.5	&0.93&	0.93&	0.19&	0.63&	0.32\\
\hline
$g_{y_e}$	&	1/60&	1/.1	&1/.1&	1/.1	&1/2&	1/1.5\\
\hline
$g_{y_c}$	&1/15	&1/.1	&1/.1	&1/.1	&1/2&	1/1.5\\
\hline
$g_{p}$	&	3&	0.0029&	0.0029	&0.0019	&0.0063	&9.6e-4\\
\hline
$\tau_{c}(s)$ &	0.03	&0.01&	0.01	&0.01&	0.1&	0.1\\
\hline
$T_{a}(s)$	&0.1&	0.05&	0.05&	0.05&	0.1&	0.1\\
\hline
\hline
\end{tabular}
\end{center}
\end{table}
%==============================================
%==============================================

\begin{figure}
\centering
\begin{tabular}{cc}
 \subfigure[$x_{ee}$ Response]{\includegraphics[width=0.5\columnwidth]{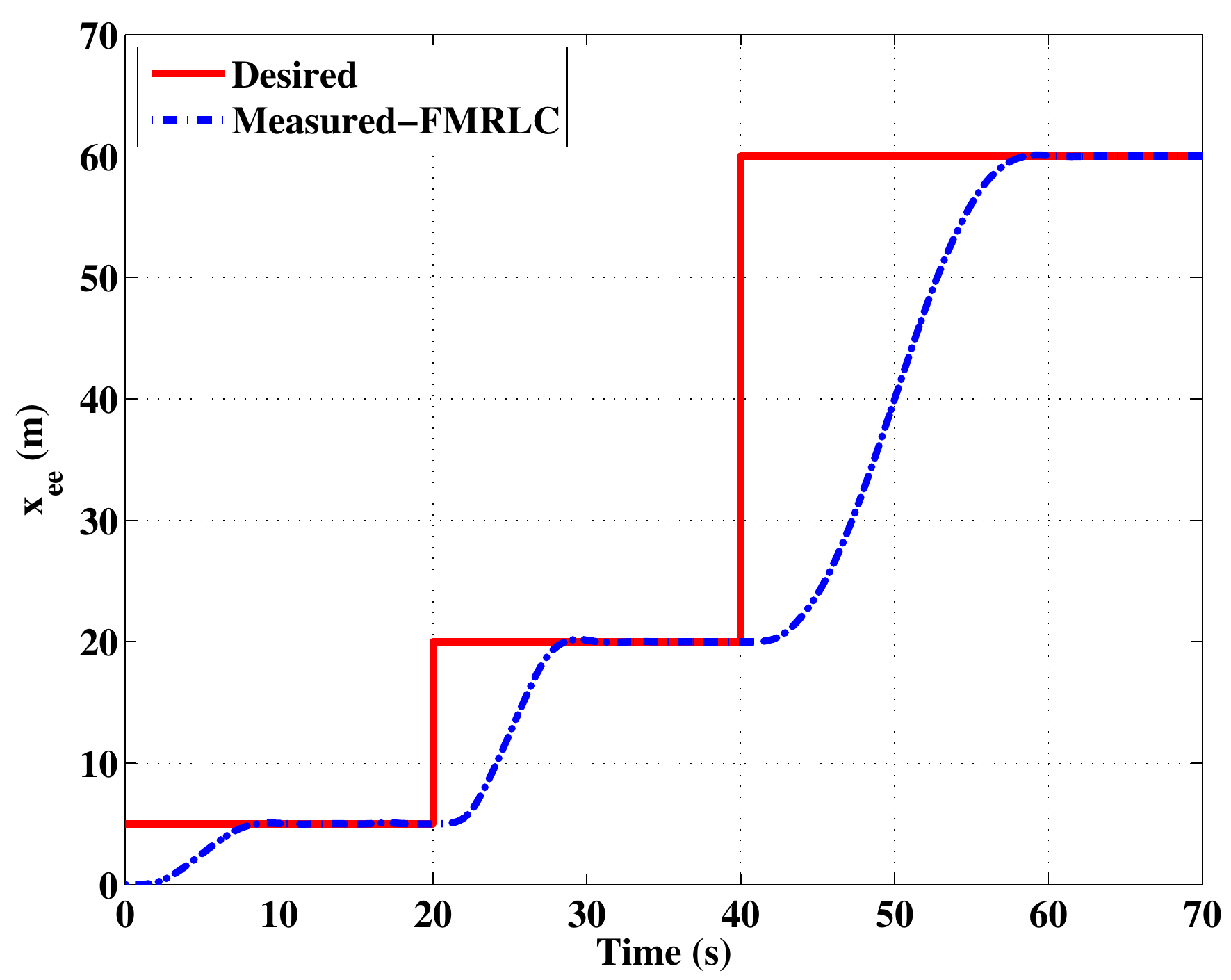}}&
 \subfigure[$y_{ee}$ Response]{\includegraphics [width=0.5\columnwidth]{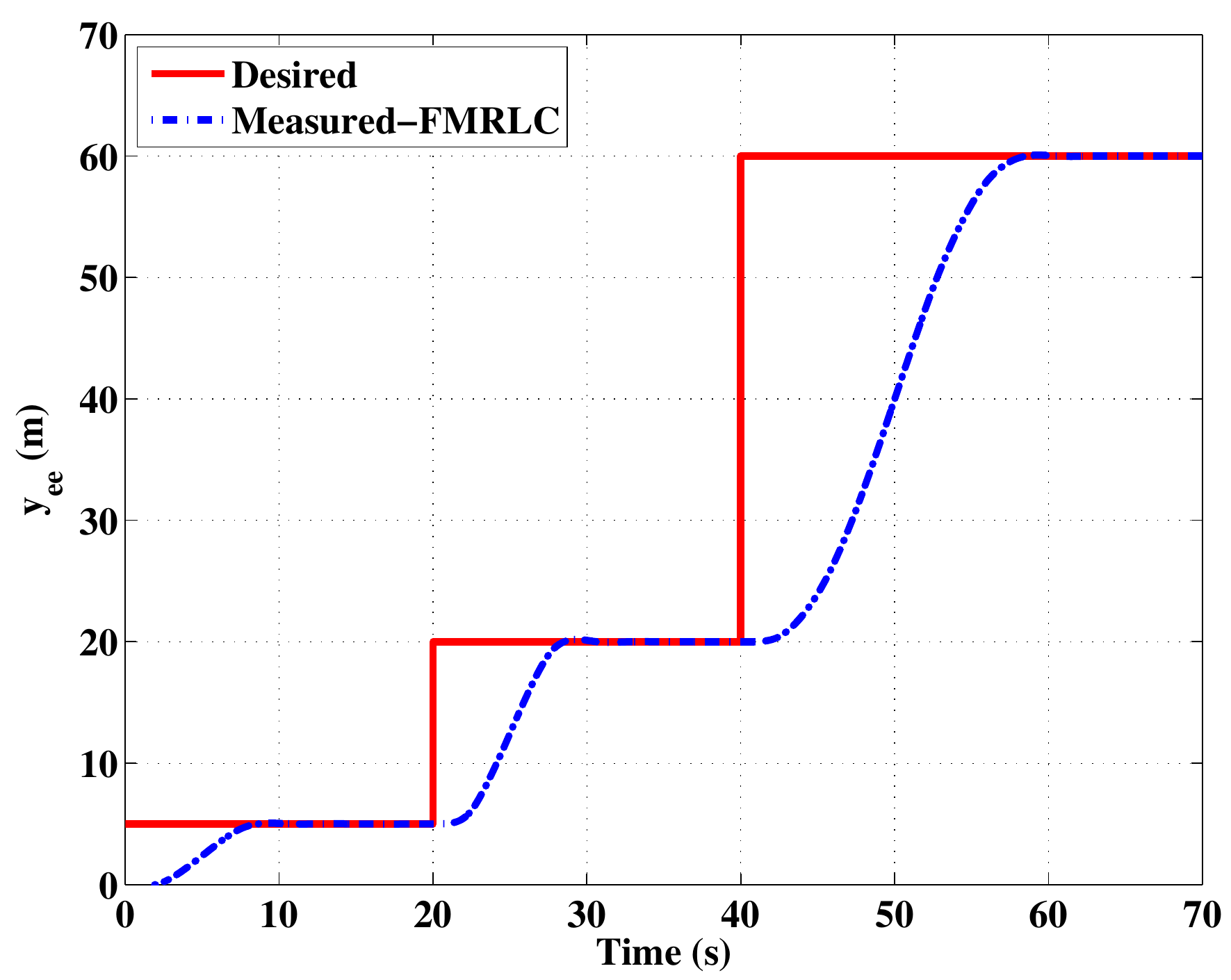}} \\
 \subfigure[$z_{ee}$ Response]{\includegraphics [width=0.5\columnwidth]{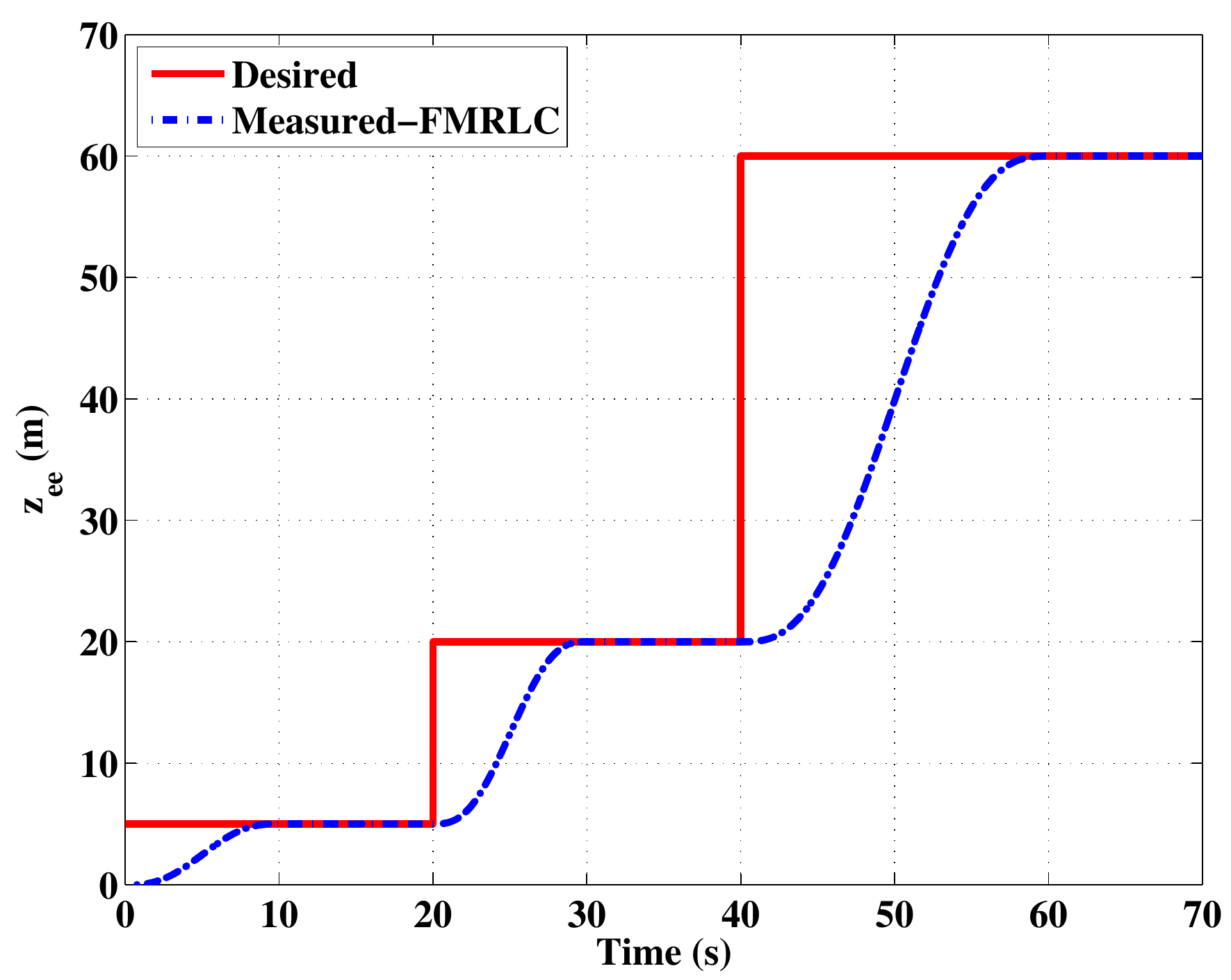}} &
 \subfigure[$\phi_{ee}$ Response]{\includegraphics [width=0.5\columnwidth]{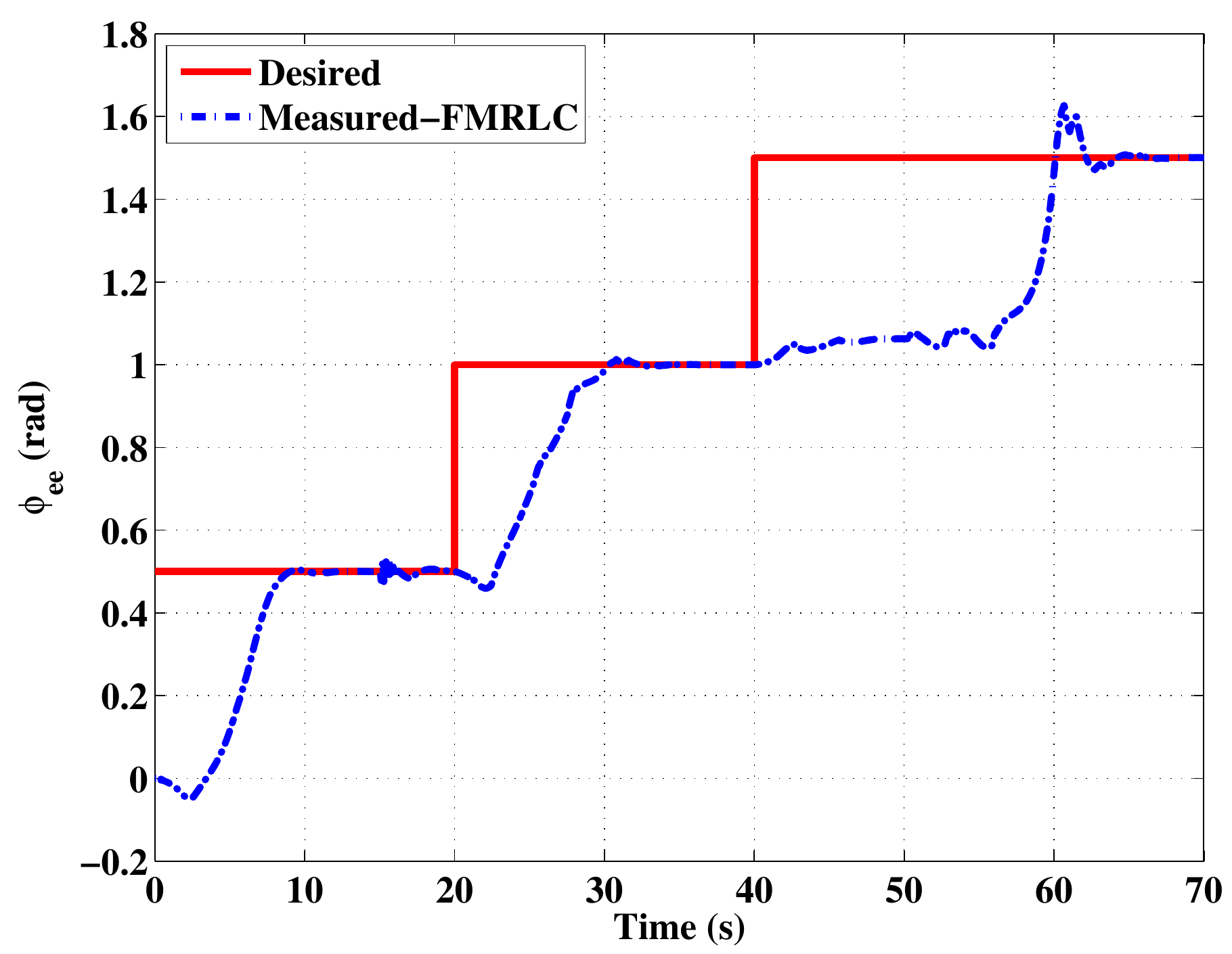}}\\
 \subfigure[$\theta_{ee}$ Response]{\includegraphics [width=0.5\columnwidth]{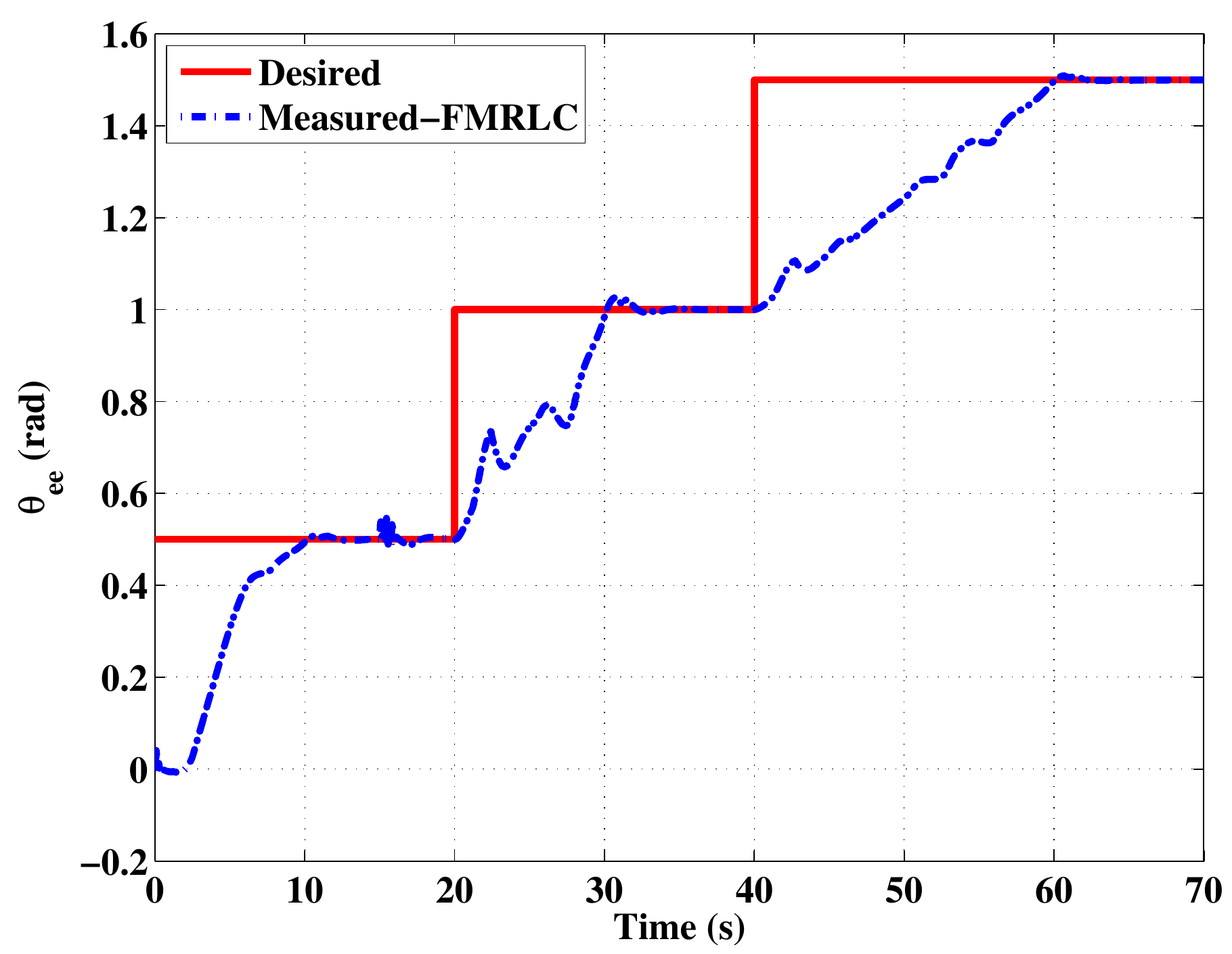}}&
 \subfigure[$\psi_{ee}$ Response]{\includegraphics [width=0.5\columnwidth]{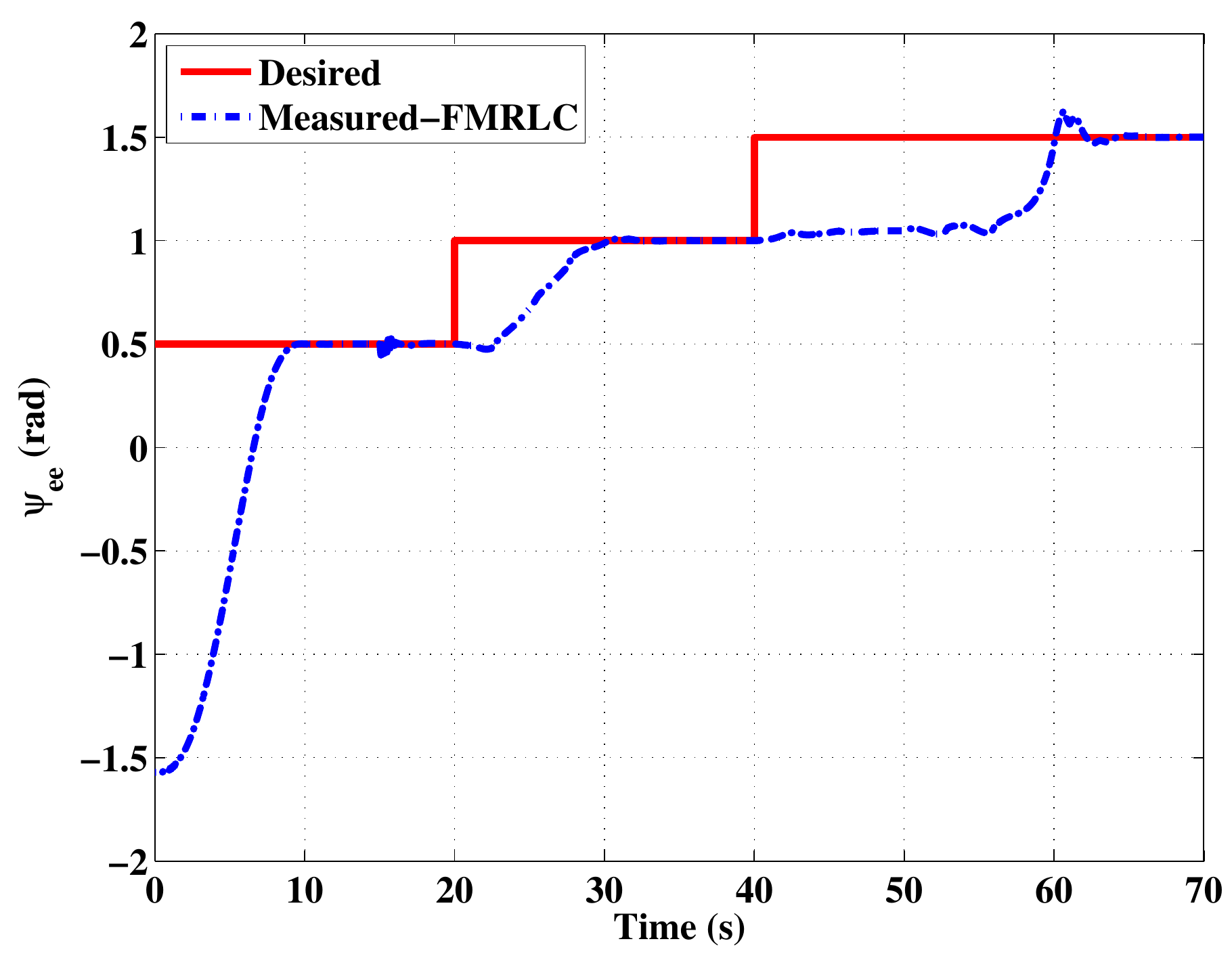}}
\end{tabular}
\caption{The Actual Response of FMRLC Technique for the End Effector Position and Orientation: a) $x_{ee}$, b) $y_{ee}$, c) $z_{ee}$, d) $\phi_{ee}$, e) $\theta_{ee}$, and f) $\psi_{ee}$.}
\label {FMRLC_controller_endeffector}
\end{figure}

The end effector position and orientation can be found from the forward kinematics (see Figure \ref{FMRLC_controller_endeffector}).
%==============================================
%==============================================

From the above discussion and results, the following items can be concluded about the performance of FMRLC technique:
\begin{itemize}
  \item FMRLC technique succeeds to make system stable against adding/releasing the payload with high accuracy, in addition to, provides a good trajectory tracking capabilities with different operation regions.
  \item Considering the complexity of the controller implementation in real time, FMRLC is moderate. It is more complicated than DFLC and simpler than feedback linearization.
  \item Therefore, FMRLC technique is able to achieve the performance objective of the system.
\end{itemize}

\chapter{\uppercase{Conclusion and Future Work }\label{ch:conclusion}}

\section{Conclusions}
\begin{itemize}
  \item Design, Kinematics, Dynamics and Control of a novel aerial manipulation system are discussed.
  \item The new proposed system adds new features to the current aerial manipulation systems (quadrotor with a gripper) such that the end-effector becomes capable of making arbitrary 6-DOF motion.
  \item Methodology to identify and verify the system parameters is implemented experimentally.
  \item A controller was designed based on Feedback Linearization, DFLC, and FMRLC techniques.
  \item The system equations of motion and the control laws are simulated using MATLAB/SIMULINK program.
  \item Feedback linearization fails to make system stable against adding the payload. Also, it has complex implementation in real time, because it puts the complex system dynamics in the control laws.
  \item DFLC technique succeeds to make system stable against adding/releasing the payload. However, it fails to provides a good trajectory tracking capabilities with different operation regions. In addition, it suffers from the necessity of calibrating and determining the offset value cannot be estimated accurately.
  \item FMRLC technique succeeds to make system stable against adding/releasing the payload with high accuracy. In addition, it provides a good trajectory tracking capabilities with different operation regions.
  \item Simulations results show that FMRLC has a superior performance compared with that of Feedback linearization and DFLC. Moreover, these results indicate the feasibility of the proposed quadrotor manipulation system.
\end{itemize}

\section{Future Work}
The final target is to build 6 DOF autonomous aerial manipulation system, and thus, there is a need to do the following steps as a future work:
\begin{itemize}
  \item Completing the experimental system.
  \item Adding a force sensor to the end effector in order to grasp and release the object accurately and without damage.
  \item Estimation of the linear position ($X$,$Y$) using a differential GPS.
  \item Testing the proposed control techniques experimentally.
  \item Implementing a vision-based object detection.
  \item Designing a navigation algorithm such that the system can be fully Autonomous.
\end{itemize} 
\begin{center}{
\huge \textbf{\uppercase{publications}}}
\end{center}
\addcontentsline{toc}{chapter}{\uppercase{publications}}
\begin{enumerate}
  \item A. Khalifa, M. Fanni, A. Ramadan, and A. Abo-Ismail, "Modeling and Control of a New Quadrotor Manipulation System", 2012 IEEE/RAS International Conference on Innovative Engineering Systems, IEEE ICIES 2012, Dec 7-9, 2012, pp.109–114, Alexandria, Egypt.
  \item M. Elsamanty, A. Khalifa, M. Fanni, A. Ramadan, and A. Abo-Ismail, "Methodology for Identifying Quadrotor Parameters, Attitude Estimation and Control", 2013 IEEE/ASME International Conference on Advanced Intelligent Mechatronics "AIM 2013", July 9-12, 2013, Wollongong, Australia.
  \item A. Khalifa, M. Fanni, A. Ramadan, and A.  Abo-Ismail, "Adaptive Intelligent Controller Design for a New Quadrotor Manipulation System", 2013 IEEE International Conference on Systems, Man, and Cybernetics "IEEE SMC 2013", Oct. 13-16, 2013, Manchester, United Kingdom. (To appear)
\end{enumerate}

\addcontentsline{toc}{chapter}{\uppercase{References}}
\addtocontents{toc}{\contentsline{chapter}{\uppercase{Arabic Summary}}{}}
\addtocontents{toc}{\contentsline{chapter}{\uppercase{Arabic Cover Page}}{}}
% Make the bibliography single spaced
\singlespacing
\bibliographystyle{plain}
\bibliography{thesis}
% add the Bibliography to the Table of Contents
%\cleardoublepage
\ifdefined\phantomsection
  \phantomsection  % makes hyperref recognize this section properly for pdf link
\else
\fi

%\blankpage
%\thispagestyle{empty}
%\includepdf[pages=-]{Arabic_Cover/Arabic_Summery.pdf} \blankpage
%\includepdf[pages=-]{Arabic_Cover/Arabic_Cover_2.pdf} \blankpage
%\includepdf[pages=-]{Arabic_Cover/Arabic_Cover_1.pdf} \blankpage

%\appendix % all chapters following will be labeled as appendices
%%% make sure the appendix will be one and half spacing
%\doublespacing
%\addcontentsline{toc}{chapter}{\uppercase{Appendices}}
%%\addcontentsline{toc}{chapter}{\uppercase{Appendices}}
%\addcontentsline{toc}{section}{\uppercase{[Appendix A] Android Phone Sensors}}
%\include{abstract}
%%\include{ch-appendicies/printing}
%% include your .bib file
\end{document}